Deep Learning Interviews is home to hundreds of fully-solved problems, from a wide range of key topics in AI. It is designed to both rehearse interview or exam-specific topics and provide machine learning M.Sc./Ph.D. students, and those awaiting an interview a well-organized overview of the field. The problems it poses are tough enough to cut your teeth on and to dramatically improve your skills-but they're framed within thought-provoking questions and engaging stories.

That is what makes the volume so specifically valuable to students and job seekers: it provides them with the ability to speak confidently and quickly on any relevant topic, to answer technical questions clearly and correctly, and to fully understand the purpose and meaning of interview questions and answers. These are powerful, indispensable advantages to have when walking into the interview room.

The book's contents is a large inventory of numerous topics relevant to DL job interviews and graduate-level exams. That places this work at the forefront of the growing trend in science to teach a core set of practical mathematical and computational skills. It is widely accepted that the training of every computer scientist must include the fundamental theorems of ML, and AI appears in the curriculum of nearly every university. This volume is designed as an excellent reference for graduates of such programs.

Shlomo Kashani, Author.    Amir Ivry, Chief Editor.

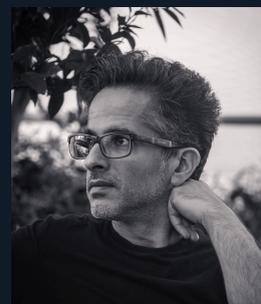
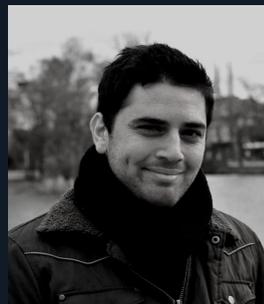

- Logistic Regression
- Information Theory
- Calculus
- Algorithmic Differentiation
- Bayesian Deep Learning
- Probabilistic Programming
- Ensemble Learning
- CNN Feature Extraction
- Deep Learning: Expanded Chapter

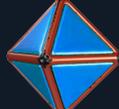

www.interviews.ai
DEEP LEARNING INTERVIEWS

**DEEP LEARNING INTERVIEWS**

SHLOMO KASHANI

# DEEP LEARNING INTERVIEWS

—REAL-WORLD DEEP LEARNING INTERVIEW PROBLEMS & SOLUTIONS—

*SECOND EDITION*

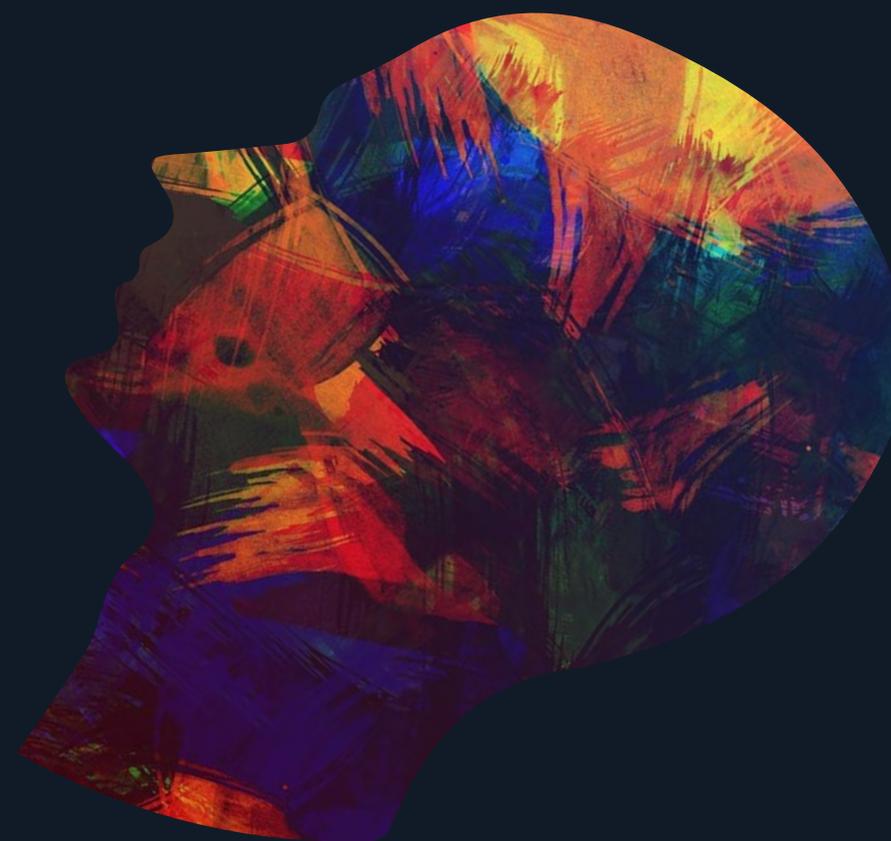

SHLOMO KASHANI

SHLOMO KASHANI

DEEP LEARNING INTERVIEWS

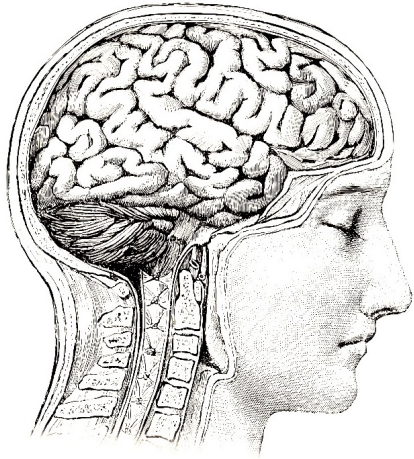

By Shlomo Kashani, M.Sc, QMUL, UK.

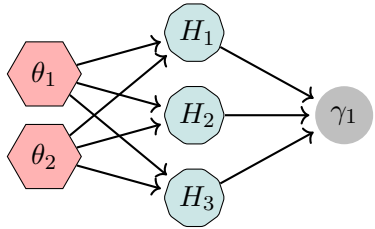

Published by Shlomo Kashani, Tel-Aviv, ISRAEL.
Visit: http://www.interviews.ai

# COPYRIGHT.

# FOREWORD.

*We will build a machine that will fly.*

— Joseph Michael Montgolfier, French Inventor/Aeronaut (1740-1810)

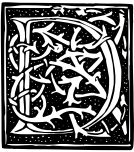 EEP learning interviews are technical, dense, and thanks to the fields competitiveness, often high-stakes. The prospect of preparing for one can be daunting, and the fear of failure can be paralyzing and many interviewees find their ideas slipping away alongside their confidence.

This book was written for you: an aspiring data scientist with a quantitative background, facing down the gauntlet of the interview process in an increasingly competitive field. For most of you, the interview process is the most significant hurdle between you and a dream job. Even though you have the ability, the background, and the motivation to excel in your target position, you might need some guidance on how to get your foot in the door.

Though this book is highly technical it is not too dense to work through quickly. It aims to be comprehensive, including many of the terms and topics involved in modern data science and deep learning. That thoroughness makes it unique; no other single work offers such breadth of learning targeted so specifically at the demands of the interview.

Most comparable information is available in a variety of formats, locations, structures, and resourcesblog posts, tech articles, and short books scattered across the internet. Those resources are simply not adequate to the demands of deep learning interview or exam preparation and were not assembled with this explicit purpose in mind. It is hoped that this book does not suffer the same shortcomings.

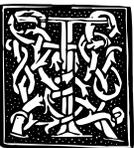 HIS books creation was guided by a few key principles: clarity and depth, thoroughness and precision, interest and accuracy. The volume was designed for use by job seekers in the fields of machine learning and deep learning whose abilities and background locate them firmly within STEM (science, technology, engineering, and mathematics). The book will still be of use to other readers, such as those still undergoing their initial education in a STEM field.

However, it is tailored most directly to the needs of **active job seekers and students attending M.Sc/Ph.D programmes in AI**. It is, in any case, a book for engineers, mathematicians, and computer scientists: nowhere does it include the kind of very basic background material that would allow it to be read by someone with no prior knowledge of quantitative and mathematical processes.

The books contents are a large inventory of numerous topics relevant to deep learning job interviews and graduate level exams. Ideas that are interesting or pertinent have been excluded if they are not valuable in that context. That places this work at the forefront of the growing trend in education and in business to emphasize a core set of practical mathematical and computational skills. It is now widely understood that the training of every computer scientist must include a course dealing with the fundamental theorems of machine learning in a rigorous manner; Deep Learning appears in the curriculum of nearly every university; and this volume is designed as a convenient ongoing reference for graduates of such courses and programs.

The book is grounded in both academic expertise and on-the-job experience and thus has two goals. First, it compresses all of the necessary information into a coherent package. And second, it renders that information accessible and makes it easy to navigate. As a result, the book helps the reader develop a thorough understanding of the principles and concepts underlying practical data science. None of the textbooks I read met all of those needs, which are:

1. **Appropriate presentation level**. I wanted a friendly introductory text accessible to graduate students who have not had extensive applied experience as data scientists.

2. **A text that is rigorous** and builds a solid understanding of the subject without getting bogged down in too many technicalities.

3. **Logical and notational consistency among topics**. There are intimate connections between calculus, logistic regression, entropy, and deep learning theory, which I feel need to be emphasized and elucidated if the reader is to fully understand the field. Differences in notation and presentation style in existing sources make it very difficult for students to appreciate these kinds of connections.

4. **Manageable size**. It is very useful to have a text compact enough that all of the material in it can be covered in few weeks or months of intensive review. Most candidates will have only that much time to prepare for an interview, so a longer text is of no use to them.

   The text that follows is an attempt to meet all of the above challenges. It will inevitably prove more successful at handling some of them than others, but it has at least made a sincere and devoted effort.

**A note about Bibliography**

The book provides a carefully curated bibliography to guide further study, whether for interview preparation or simply as a matter of interest or job-relevant research. A comprehensive bibliography would be far too long to include here, and would be of little immediate use, so the selections have been made with deliberate attention to the value of each included text.

Only the most important books and articles on each topic have been included, and only those written in English that I personally consulted. Each is given a brief annotation to indicate its scope and applicability. Many of the works cited will be found to include very full bibliographies of the particular subject treated, and I recommend turning there if you wish to dive deeper into a specific topic, method, or process.

We have a web page for this book, where we list errata, examples, and any additional information. You can access this page at: http://www.interviews.ai. To comment or ask technical questions about this book, send email to: entropy@interviews.ai.

**I would also like to solicit corrections, criticisms, and suggestions from students and other readers. Although I have tried to eliminate errors over the multi year process of writing and revising this text, a few undoubtedly remain. In particular, some typographical infelicities will no doubt find their way into the final version. I hope you will forgive them**.

# ACKNOWLEDGEMENTS.

The thanks and acknowledgements of the publisher are due to the following: My dear son, Amir Ivry, Matthew Isaac Harvey, Sandy Noymer, Steve foot and Velimir Gayevskiy.

# Author's Biography.

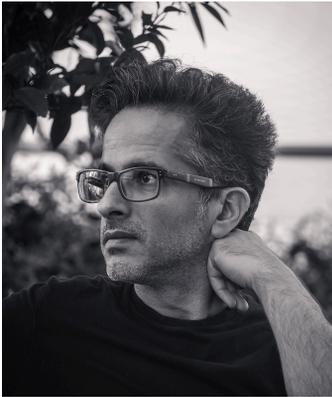

**When Shlomo** typed his book in LaTeX, he wanted it to reflect some of his passions: AI, design, typography, and most notably coding. On a typical day, his two halves - the scientist and the artist - spend hours meticulously designing AI systems, from epilepsy prediction and pulmonary nodule detection, to training a computer-vision model on a cluster.

Shlomo spends whole days in a lab full of GPUs working on his many interesting research projects. Though research satisfies his itch for discovery, his most important scientific contribution, he says, is helping other researchers.

And the results are evident in his publications. But, although theoretical studies are important, practical experience has many great virtues. As the Head of AI at DeepOncology, he developed uses of Deep Learning for precise tumour detection, expanding and refining what human experts are capable of. The work, which relies on CNN's, marks the culmination of a career spent applying AI techniques to problems in medical AI. Shlomo holds an MSc in Digital Signal Processing (Distinction) from the University of London.

**A PERSONAL NOTE:** In this first volume, I purposely present a coherent, cumulative, and content-specific **core curriculum** of the data science field, including topics such as information theory, Bayesian statistics, algorithmic differentiation, logistic regression, perceptrons, and convolutional neural networks.

I hope you will find this book stimulating. It is my belief that you **the postgraduate students and job-seekers** for whom the book is primarily meant will benefit from reading it; however, it is my hope that even the most experienced researchers will find it fascinating as well.

**SHLOMO KASHANI,TEL-AVIV,ISRAEL.**

# About the Chief Editor.

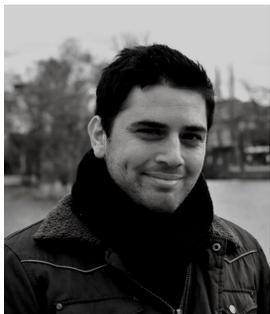 **Amir Ivry** has been an applied research scientist in the fields of deep learning and speech signal processing since 2015. A direct PhD candidate in the Electrical and Computer Engineering Faculty in the Technion - Israel Institute of Technology, Amir is the author of over a dozen academic papers in leading IEEE journals and top-tier conferences. For his contribution to the field of hands-free speech communication using deep neural networks, Amir has received more than a dozen awards and honors, including back-to-back Jacobs citations for research excellence, and most recently the international speech communication association grant. Being only 28 years old, he has been cemented as a popular lecturer in the machine learning community, and delivered technological sessions for MIT, Google for startups, Alibaba, and more. Amir is currently holding a position as an applied research intern in Microsoft Advanced Technology Labs.

# Contents

# IV Bachelors 183

## DEEP LEARNING: NN ENSEMBLES 185

## DEEP LEARNING: CNN FEATURE EXTRACTION 205

## VI  Volume two

**VOLUME TWO - PLAN**

# PART I
# RUSTY NAIL

# CHAPTER

<div style="background:maroon">1</div>

HOW-TO USE THIS BOOK

*The true logic of this world is in the **calculus** of probabilities.*

— James C. Maxwell

## Contents

## 1.1   Introduction

First of all, welcome to world of Deep Learning Interviews.

### 1.1.1   What makes this book so valuable

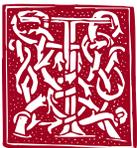ARGETED advertising. Deciphering dead languages. Detecting malignant tumours. Predicting natural disasters. Every year we see dozens of new uses for deep learning emerge from corporate R&R, academia, and plucky entrepreneurs. Increasingly, deep learning and artificial intelligence are ingrained in our cultural consciousness. Leading universities are dedicating programs to teaching them, and they make the headlines every few days.

That means jobs. It means intense demand and intense competition. It means a generation of data scientists and machine learning engineers making their way into the workforce and using deep learning to change how things work. This book is for them, and for you. It is aimed at current or aspiring experts and students in the field possessed of a strong grounding in mathematics, an active imagination, engaged creativity, and an appreciation for data. It is hand-tailored to give you the best possible preparation for deep learning job interviews by guiding you through hundreds of fully solved questions.

That is what makes the volume so specifically valuable to students and job seekers: it provides them with the ability to speak confidently and quickly on any relevant topic, to answer technical questions clearly and correctly, and to fully understand the purpose and meaning of interview questions and answers.

Those are powerful, indispensable advantages to have when walking into the interview room.

The questions and problems the book poses are tough enough to cut your teeth on-and to dramatically improve your skills but theyre framed within thought provoking questions, powerful and engaging stories, and cutting edge scientific information. What are bosons and fermions? What is choriionic villus? Where did the Ebola virus first appear, and how does it spread? Why is binary options trading so dangerous?

Your curiosity will pull you through the book's problem sets, formulas, and instructions, and as you progress, you'll deepen your understanding of deep learning. There are intricate connections between calculus, logistic regression, entropy, and deep learning theory; work through the book, and those connections will feel intuitive.

### 1.1.2 What will I learn

#### Starting Your Career

Are you actively pursuing a career in deep learning and data science, or hoping to do so? If so, you're in luck everything from deep learning to artificial intelligence is in extremely high demand in the contemporary workforce. Deep learning professionals are highly sought after and also find themselves among the highest-paid employee groups in companies around the world.

So your career choice is spot on, and the financial and intellectual benefits of landing a solid job are tremendous. But those positions have a high barrier to entry: the deep learning interview. These interviews have become their own tiny industry, with HR employees having to specialize in the relevant topics so as to distinguish well-prepared job candidates from those who simply have a loose working knowledge of the material. Outside the interview itself, the difference doesn't always feel import-ant. Deep learning libraries are so good that a machine learning pipeline can often be assembled with little high-skill input from the researcher themselves. But that level of ability won't cut it in the interview. You'll be asked practical questions, technical questions, and theoretical questions, and expected to answer them all confidently and fluently.

For unprepared candidates, that's the end of the road. Many give up after repeated post-interview rejections.

### Advancing Your Career

Some of you will be more confident. Those of you with years on the job will be highly motivated, exceptionally numerate, and prepared to take an active, hands-on role in deep learning projects. You probably already have extensive knowledge in applied mathematics, computer science, statistics, and economics. Those are all formidable advantages.

But at the same time, it's unlikely that you will have prepared for the interview itself. Deep learning interviews especially those for the most interesting, autonomous, and challenging positions demand that you not only know how to do your job but that you display that knowledge clearly, eloquently, and without hesitation. Some questions will be straightforward and familiar, but others might be farther afield or draw on areas you haven't encountered since college.

There is simply no reason to leave that kind of thing to chance. Make sure you're prepared. Confirm that you are up-to-date on terms, concepts, and algorithms. Refresh your memory of fundamentals, and how they inform contemporary research practices. And when the interview comes, walk into the room knowing that you're ready for what's coming your way.

### Diving Into Deep Learning

"Deep Learning Job Interviews" is organized into chapters that each consist of an Introduction to a topic, Problems illustrating core aspects of the topic, and complete Solutions. You can expect each question and problem in this volume to be clear, practical, and relevant to the subject. Problems fall into two groups, conceptual and application-based. Conceptual problems are aimed at testing and improving your knowledge of basic underlying concepts, while applications are targeted at practicing or applying what you've learned (most of these are relevant to Python and PyTorch). The chapters are followed by a reference list of relevant formulas and a selective bibliography for guide further reading.

### 1.1.3 How to Work Problems

In real life, like in exams, you will encounter problems of varying difficulty. A good skill to practice is recognizing the level of difficulty a problem poses. Job interviews will have some easy problems, some standard problems, and some much harder problems.

Each chapter of this book is usually organized into three sections: Introduction, Problems, and Solutions. As you are attempting to tackle problems, resist the temptation to prematurely peek at the solution; It is vital to allow yourself to struggle for a time with the material. Even professional data scientists do not always know right away how to resolve a problem. The art is in gathering your thoughts and figuring out a strategy to use what you know to find out what you don't.

**PRB-1  CH.PRB- 1.1.**

*Problems outlined in grey make up the* representative question set. *This set of problems is intended to cover the most essential ideas in each section. These problems are usually highly typical of what you'd see on an interview, although some of them are atypical but carry an important moral. If you find yourself unconfident with the idea behind one of these, it's probably a good idea to practice similar problems. This representative question set is our suggestion for a minimal selection of problems to work on. You are highly encouraged to work on more.*

**SOL-1  CH.SOL- 1.1.** *I am a solution.* ∎

If you find yourself at a real stand-off, go ahead and look for a clue in one of the recommended theory books. Think about it for a while, and don't be afraid to read back in the notes to look for a key idea that will help you proceed. If you still can't solve the problem, well, we included the Solutions section for a reason! As you're reading the solutions, try hard to understand why we took the steps we did, instead of memorizing step-by-step how to solve that one particular problem.

If you struggled with a question quite a lot, it's probably a good idea to return to it in a few days. That might have been enough time for you to internalize the necessary ideas, and you might find it easily conquerable. If you're still having troubles, read over the solution again, with an emphasis on understanding why each step makes sense. One of the reasons so many job candidates are required to demonstrate their ability to resolves data science problems on the board, is that it hiring managers assume it reflects their true problem-solving skills.

In this volume, you will learn lots of concepts, and be asked to apply them in a variety of situations. Often, this will involve answering one really big problem by breaking it up into manageable chunks, solving those chunks, then putting the pieces back together. When you see a particularly long question, remain calm and look for a way to break it into pieces you can handle.

### 1.1.4    Types of Problems

Two main types of problems are presented in this book.

CONCEPTUAL: The first category is meant to test and improve your understanding of basic underlying concepts. These often involve many mathematical calculations. They range in difficulty from very basic reviews of definitions to problems that require you to be thoughtful about the concepts covered in the section.

An example in Information Theory follows.

---

**PRB-2 ❷ CH.PRB- 1.2.**

*What is the distribution of maximum entropy, that is, the distribution which has the maximum entropy among all distributions on the bounded interval* $[a, b]$,$(-\infty, +\infty)$

---

**SOL-2 ✎ CH.SOL- 1.2.**

*The uniform distribution has the maximum entropy among all distributions on the bounded interval:* $[a, b]$,$(-\infty, +\infty)$.
*The variance of* $U(a, b)$ *is* $\sigma^2 = 1/12(b - a)^2$.
*Therefore the entropy is:*

$$1/2 \log 12 + \log \sigma. \tag{1.1}$$

---

APPLICATION: Problems in this category are for practicing skills. It's not enough to understand the philosophical grounding of an idea: you have to be able to apply it in appropriate situations. This takes practice! mostly in Python or in one of the available Deep Learning Libraries such as PyTorch.

An example in PyTorch follows.

**PRB-3**  **CH.PRB- 1.3.**

*Describe in your own words, what is the purpose of the following code in the context of training a Convolutional Neural Network.*

```python
self.transforms = []
if rotate:
    self.transforms.append(RandomRotate())
if flip:
    self.transforms.append(RandomFlip())
```

**SOL-3**  **CH.SOL- 1.3.**

*During the training of a Convolutional Neural Network, data augmentation, and to some extent dropout are used as core methods to decrease overfitting. Data augmentation is a regularization scheme that synthetically expands the data-set by utilizing label-preserving transformations to add more invariant examples of the same data samples. It is most commonly performed in real time on the CPU during the training phase whilst the actual training mode takes place on the GPU. This may consist for instance, random rotations, random flips, zooming, spatial translations etc.* ∎

# PART II
# KINDERGARTEN

# CHAPTER

## 2

<span style="color:red">LOGISTIC REGRESSION</span>

*You should call it entropy for two reasons. In the first place, your uncertainty function has been used in statistical mechanics under that name. In the second place, and more importantly, no one knows what entropy really is, so in a debate you will always have the advantage.*

— John von Neumann to Claude Shannon

## Contents

## 2.1   Introduction

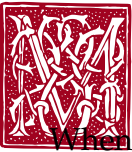Ultivariable methods are routinely utilized in statistical analyses across a wide range of domains. Logistic regression is the most frequently used method for modelling binary response data and binary classification. When the response variable is binary, it characteristically takes the form of 1/0, with 1 normally indicating a success and 0 a failure. Multivariable methods usually assume a relationship between two or more independent, predictor variables, and one dependent, response variable. The predicted value of a response variable may be expressed as a sum of products, wherein each product is formed by multiplying the value of the variable and its coefficient. How the coefficients are computed? from a respective data set. Logistic regression is heavily used in supervised machine learning and has become the workhorse for both binary and multiclass classification problems. Many of the questions introduced in this chapter are crucial for truly understanding the inner-workings of artificial neural networks.

## 2.2   Problems

### 2.2.1   General Concepts

**PRB-4 ❷ CH.PRB- 2.1.**
   *True or False: For a fixed number of observations in a data set, introducing more variables normally generates a model that has a better fit to the data. What may be the drawback of such a model fitting strategy?*

**PRB-5 ❷ CH.PRB- 2.2.**
   *Define the term **"odds of success"** both qualitatively and formally. Give a numerical example that stresses the relation between probability and odds of an event occurring.*

**PRB-6 ❷ CH.PRB- 2.3.**

   1. *Define what is meant by the term **"interaction"**, in the context of a logistic regression predictor variable.*

2. *What is the simplest form of an interaction? Write its formulae.*

3. *What statistical tests can be used to attest the significance of an interaction term?*

---

**PRB-7 ❷ CH.PRB- 2.4.**

**True or False:** *In machine learning terminology, unsupervised learning refers to the mapping of input covariates to a target response variable that is attempted at being predicted when the labels are known.*

---

**PRB-8 ❷ CH.PRB- 2.5.**

**Complete the following sentence:** *In the case of logistic regression, the response variable is the log of the odds of being classified in [...].*

---

**PRB-9 ❷ CH.PRB- 2.6.**

*Describe how in a logistic regression model, a transformation to the response variable is applied to yield a probability distribution. Why is it considered a more informative representation of the response?*

---

**PRB-10 ❷ CH.PRB- 2.7.**

**Complete the following sentence:** *Minimizing the negative log likelihood also means maximizing the [...] of selecting the [...] class.*

### 2.2.2   Odds, Log-odds

---

**PRB-11 ❷ CH.PRB- 2.8.**

*Assume the probability of an event occurring is $p = 0.1$.*

1. *What are the **odds** of the event occurring?.*

2. *What are the **log-odds** of the event occurring?.*

3. *Construct the **probability** of the event as a ratio that equals 0.1.*

---

**PRB-12 ❷ CH.PRB- 2.9.**

**True or False:** *If the odds of success in a binary response is* 4*, the corresponding probability of success is* 0.8.

---

**PRB-13 ❷ CH.PRB- 2.10.**

*Draw a graph of **odds to probabilities**, mapping the entire range of probabilities to their respective odds.*

---

**PRB-14 ❷ CH.PRB- 2.11.**

*The logistic regression model is a subset of a broader range of machine learning models known as generalized linear models (GLMs), which also include analysis of variance (ANOVA), vanilla linear regression, etc. There are three components to a GLM; **identify these three components for binary logistic regression.***

---

**PRB-15 ❷ CH.PRB- 2.12.**

*Let us consider the logit transformation, i.e., log-odds. Assume a scenario in which the logit forms the linear decision boundary:*

$$\log\left(\frac{\Pr(Y = 1|X)}{\Pr(Y = 0|X)}\right) = \theta_0 + \theta^T X, \tag{2.1}$$

*for a given vector of systematic components $X$ and predictor variables $\theta$. Write the mathematical expression for the hyperplane that describes the decision boundary.*

---

**PRB-16 ❷ CH.PRB- 2.13.**

**True or False:** *The logit function and the natural logistic (sigmoid) function are inverses of each other.*

### 2.2.3   The Sigmoid

The sigmoid (Fig. 2.1) also known as the logistic function, is widely used in binary classification and as a neuron activation function in artificial neural networks.

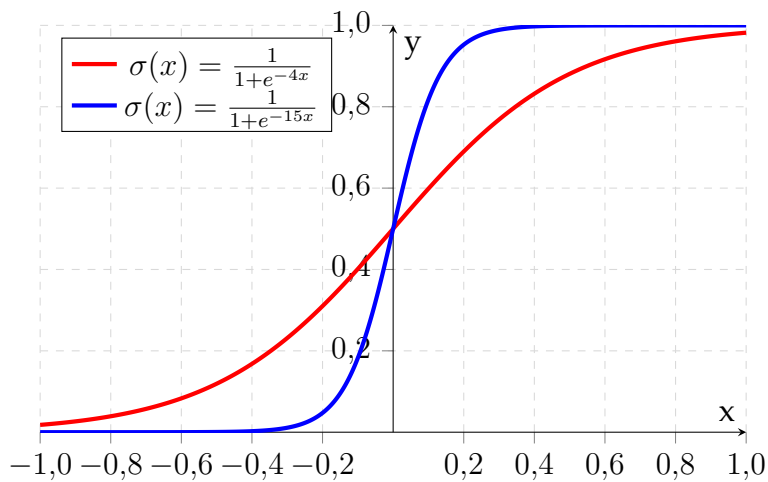

FIGURE 2.1: Examples of two sigmoid functions.

---

**PRB-17 ❷ CH.PRB- 2.14.**

*Compute the derivative of the natural sigmoid function:*

$$\sigma(x) = \frac{1}{1 + e^{-x}} \in (0, 1). \tag{2.2}$$

---

**PRB-18 ❷ CH.PRB- 2.15.**

*Remember that in logistic regression, the hypothesis function for some parameter vector $\beta$ and measurement vector $x$ is defined as:*

$$
\begin{aligned}
h_\beta(x) &= g(\beta^T x) = \frac{1}{1 + e^{-\beta^T x}} \\
&= P(y = 1 | x; \beta),
\end{aligned}
\tag{2.3}
$$

*where y holds the hypothesis value.*

*Suppose the coefficients of a logistic regression model with independent variables are as follows: $\beta_0 = -1.5$, $\beta_1 = 3$, $\beta_2 = -0.5$.*

*Assume additionally, that we have an observation with the following values for the dependent variables: $x_1 = 1$, $x_2 = 5$. As a result, the logit equation becomes:*

$$logit = \beta_0 + \beta_1 x_1 + \beta_2 x_2. \tag{2.4}$$

1. *What is the value of the **logit for this observation?***

2. *What is the value of the **odds for this observation?***

3. *What is the value of $P(y = 1)$ **for this observation?***

### 2.2.4 Truly Understanding Logistic Regression

**PRB-19 ❷ CH.PRB- 2.16.**

*Proton therapy (PT) [2] is a widely adopted form of treatment for many types of cancer including breast and lung cancer (Fig. 2.2).*

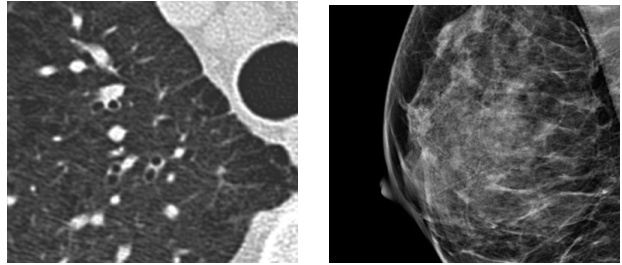

FIGURE 2.2: Pulmonary nodules (left) and breast cancer (right).

*A PT device which **was not properly calibrated** is used to simulate the treatment of cancer. As a result, the PT beam does not behave normally. A data scientist collects information relating to this simulation. The covariates presented in Table 2.1 are collected during* *the experiment. The columns **Yes** and **No** indicate if the tumour was eradicated or not, respectively.*

|              | Tumour eradication | |
| Cancer Type | Yes | No |
| --- | --- | --- |
| Breast | 560 | 260 |
| Lung | 69 | 36 |

TABLE 2.1: Tumour eradication statistics.

*Referring to Table 2.1:*

1. *What is the explanatory variable and what is the response variable?*

2. *Explain the use of relative risk and odds ratio for measuring association.*

3. *Are the two variables positively or negatively associated?*
   *Find the direction and strength of the association using both relative risk and odds ratio.*

4. *Compute a 95% confidence interval (CI) for the measure of association.*

5. *Interpret the results and explain their significance.*

---

**PRB-20 ❷ CH.PRB- 2.17.**

*Consider a system for radiation therapy planning (Fig. 2.3). Given a patient with a malignant tumour, the problem is to select the optimal radiation exposure time for that patient. A key element in this problem is estimating the probability that a given tumour will be eradicated given certain covariates. A data scientist collects information relating to this radiation therapy system.*

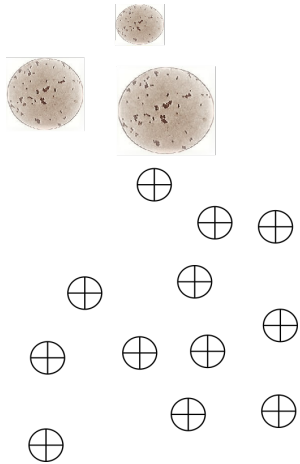

FIGURE 2.3: A multi-detector positron scanner used to locate tumours.

*The following covariates are collected; $X_1$ denotes time in milliseconds that a patient is irradiated with, $X_2 =$ holds the size of the tumour in centimeters, and $Y$ notates a binary response variable indicating if the tumour was eradicated. Assume that each response' variable $Y_i$ is a Bernoulli random variable with success parameter $p_i$, which holds:*

$$p_i = \frac{e^{\beta_0 + \beta_1 x_1 + \beta_2 x_2}}{1 + e^{\beta_0 + \beta_1 x_1 + \beta_2 x_2}}. \tag{2.5}$$

*The data scientist fits a logistic regression model to the dependent measurements and produces these estimated coefficients:*

$$\begin{aligned} \hat{\beta}_0 &= -6, \\ \hat{\beta}_1 &= 0.05, \\ \hat{\beta}_2 &= 1. \end{aligned} \tag{2.6}$$

1. *Estimate the probability that, given a patient who undergoes the treatment for 40 milliseconds and who is presented with a tumour sized 3.5 centimetres, the system eradicates the tumour.*

2. *How many milliseconds the patient in part (a) would need to be radiated with to have exactly a 50% chance of eradicating the tumour?*

**PRB-21 ❷ CH.PRB- 2.18.**

*Recent research [3] suggests that heating mercury containing dental amalgams may cause the release of toxic mercury fumes into the human airways. It is also presumed that drinking hot coffee, stimulates the release of mercury vapour from amalgam fillings (Fig. 2.4).*

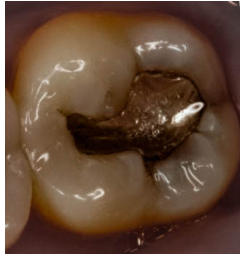

FIGURE 2.4: A dental amalgam.

*To study factors that affect migraines, and in particular, patients who have at least four dental amalgams in their mouth, a data scientist collects data from 200K users with and without dental amalgams. The data scientist then fits a logistic regression model with an indicator of a second migraine within a time frame of one hour after the onset of the first migraine, as the binary response variable (e.g., migraine=1, no migraine=0). The data scientist believes that the frequency of migraines may be related to the release of toxic mercury fumes.*

*There are two independent variables:*

1. *$X_1$ = 1 if the patient has at least four amalgams; 0 otherwise.*

2. *$X_2$ = coffee consumption (0 to 100 hot cups per month).*

*The output from training a logistic regression classifier is as follows:*

```
Analysis of LR Parameter Estimates
Parameter    Estimate  Std.Err   Z-val   Pr>|Z|

Intercept    -6.36347  3.21362  -1.980   0.0477
$X_1$        -1.02411  1.17101  -0.875   0.3818
$X_2$         0.11904  0.05497   2.165   0.0304
```

1. Using $X_1$ and $X_2$, express the **odds** of a patient having a migraine for a second time.

2. Calculate the **probability** of a second migraine for a patient that has at least four amalgams and drank 100 cups per month?

3. For users that have at least four amalgams, is **high** coffee intake associated with an **increased** probability of a second migraine?

4. Is there statistical evidence that having more than four amalgams is **directly** associated with a **reduction** in the probability of a second migraine?

---

**PRB-22 ❷ CH.PRB- 2.19.**

To study factors that affect Alzheimer's disease using logistic regression, a researcher considers the link between gum (periodontal) disease and Alzheimer as a plausible risk factor [1]. The predictor variable is a count of gum bacteria (Fig. 2.5) in the mouth.

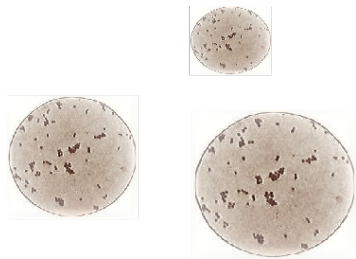

FIGURE 2.5: A chain of spherical bacteria.

The response variable, $Y$, measures whether the patient shows any remission (e.g. yes=1). The output from training a logistic regression classifier is as follows:

```
 Parameter   DF  Estimate     Std
  Intercept   1   -4.8792  1.2197
gum bacteria   1    0.0258  0.0194
```

1. Estimate the probability of improvement when the count of gum bacteria of a patient is 33.

2. *Find out the gum bacteria count at which the estimated probability of improvement is 0.5.*

3. *Find out the estimated odds ratio of improvement for an increase of **1** in the total gum bacteria count.*

4. *Obtain a 99% confidence interval for the true odds ratio of improvement increase of **1** in the total gum bacteria count. Remember that the most common confidence levels are 90%, 95%, 99%, and 99.9%. Table 9.1 lists the $z$ values for these levels.*

| Confidence Level | $z$ |
|:---:|:---:|
| 90% | 1.645 |
| 95% | 1.960 |
| 99% | **2.576** |
| 99.9% | 3.291 |

TABLE 2.2: Common confidence levels.

**PRB-23 ❷ CH.PRB- 2.20.**

   *Recent research [4] suggests that cannabis (Fig. 2.6) and cannabinoids administration in particular, may reduce the size of malignant tumours in rats.*

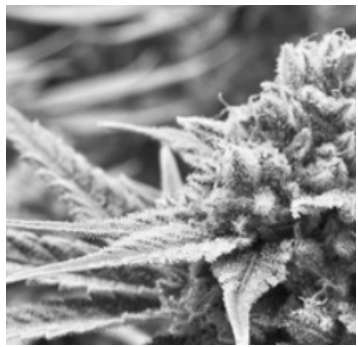

FIGURE 2.6: Cannabis.

To study factors affecting tumour shrinkage, a deep learning researcher collects data from two groups; one group is administered with placebo (a substance that is not medicine) and the other with cannabinoids. His main research revolves around studying the relationship (Table 2.3) between the anticancer properties of cannabinoids and tumour shrinkage:

| Group | Tumour Shrinkage In Rats | | |
| | Yes | No | Sum |
|---|---|---|---|
| Cannabinoids | 60 | 6833 | 6893 |
| Placebo | 130 | 6778 | 6909 |
| Sum | 190 | 13611 | 13801 |

TABLE 2.3: Tumour shrinkage in rats.

For the true odds ratio:

1. Find the sample odds ratio.

2. Find the sample log-odds ratio.

3. Compute a 95% confidence interval ($z_{0.95} = 1.645$; $z_{0.975} = 1.96$) for the true log odds ratio and true odds ratio.

### 2.2.5 The Logit Function and Entropy

**PRB-24 ❷ CH.PRB- 2.21.**
The entropy (see Chapter 4) of a single binary outcome with probability $p$ to receive 1 is defined as:

$$H(p) \equiv -p \log p - (1 - p) \log(1 - p). \tag{2.7}$$

1. At what $p$ does $H(p)$ attain its maximum value?

2. What is the relationship between the entropy $H(p)$ and the logit function, given $p$?

2.2.6    Python/PyTorch/CPP

---

**PRB-25 ❷ CH.PRB- 2.22.**

*The following C++ code (Fig. 2.7) is part of a (very basic) logistic regression implement-ation module. For a theoretical discussion underlying this question, refer to problem 2.17.*

```cpp
#include ...
std::vector<double> theta {-6,0.05,1.0};
double sigmoid(double x) {
 double tmp =1.0 / (1.0 + exp(-x));
 std::cout << "prob=" << tmp<<std::endl;
 return tmp;
}
double hypothesis(std::vector<double> x){
 double z;
 z=std::inner_product(std::begin(x), std::end(x),
   std::begin(theta), 0.0);
 std::cout << "inner_product=" << z<<std::endl;
 return sigmoid(z);
}
int classify(std::vector<double> x){
 int hypo=hypothesis(x) > 0.5f;
 std::cout << "hypo=" << hypo<<std::endl;
 return hypo;
}
int main() {
 std::vector<double> x1 {1,40,3.5};
 classify(x1);
}
```

FIGURE 2.7: Logistic regression in CPP

1. *Explain the purpose of line **10**, i.e., **inner_product**.*

2. *Explain the purpose of line **15**, i.e., **hypo(x) > 0.5f**.*

3. *What does θ **(theta)** stand for in line 2?*

4. *Compile and run the code, you can use:*
   *https://repl.it/languages/cpp11 to evaluate the code.*
   *What is the output?*

---

**PRB-26 ❷ CH.PRB- 2.23.**

*The following Python code (Fig. 2.8) runs a very simple linear model on a two-dimensional matrix.*

```
1  import torch
2  import torch.nn as nn
3
4  lin = nn.Linear(5, 7)
5  data = (torch.randn(3, 5))
6
7  print(lin(data).shape)
8  >?
```

FIGURE 2.8: A linear model in PyTorch

*Without actually running the code, determine what is the size of the matrix printed as a result of applying the linear model on the matrix.*

---

**PRB-27 ❷ CH.PRB- 2.24.**

*The following Python code snippet (Fig. 2.9) is part of a logistic regression implementation module in Python.*

```python
from scipy.special import expit
import numpy as np
import math

def Func001(x):
  e_x = np.exp(x - np.max(x))
  return e_x / e_x.sum()

def Func002(x):
  return 1 / (1 + math.exp(-x))

def Func003(x):
  return x * (1-x)
```

FIGURE 2.9: Logistic regression methods in Python.

Analyse the methods  Func001, Func002 and Func003 presented in Fig. 2.9, find their purposes and **name them**.

**PRB-28 ❷ CH.PRB- 2.25.**
The following Python code snippet (Fig. 2.10) is part of a machine learning module in Python.

```
1  ^^I^^I
2  from scipy.special import expit
3  import numpy as np
4  import math
5  ^^I^^I
6  def Func006(y_hat, y):
7   if y == 1:
8     return -np.log(y_hat)
9   else:
10     return -np.log(1 - y_hat)^^I
```

FIGURE 2.10: Logistic regression methods in Python.

*Analyse the method*  *presented in Fig. 2.10. What important concept in machine-learning does it implement?*

**PRB-29 ❷ CH.PRB- 2.26.**

*The following Python code snippet (Fig. 2.11) presents several different variations of the same function.*

```
1  ^^I^^I
2  from scipy.special import expit
3  import numpy as np
4  import math
5
6  def Ver001(x):
7   return 1 / (1 + math.exp(-x))
8
9  def Ver002(x):
10   return 1 / (1 + (np.exp(-x)))
11
12  WHO_AM_I = 709
13
14  def Ver003(x):
15   return 1 / (1 + np.exp(-(np.clip(x, -WHO_AM_I, None))))
```

FIGURE 2.11: Logistic regression methods in Python.

1. *Which mathematical function do these methods implement?*

2. *What is significant about the number $\boxed{709}$ in line 11?*

3. *Given a choice, which method would you use?*

## 2.3    Solutions

### 2.3.1    General Concepts

**SOL-4  CH.SOL- 2.1.**

*True. However, when an excessive and unnecessary number of variables is used in a logistic regression model, peculiarities (e.g., specific attributes) of the underlying data set disproportionately affect the coefficients in the model, a phenomena commonly referred to as "overfitting". Therefore, it is important that a logistic regression model does not start training with more variables than is justified for the given number of observations.* ■

## SOL-5  CH.SOL- 2.2.

*The odds of success are defined as the ratio between the probability of success $p \in [0, 1]$ and the probability of failure $1 - p$. Formally:*

$$Odds(p) \equiv \left( \frac{p}{1-p} \right).$$ 

(2.8)

*For instance, assuming the probability of success of an event is $p = 0.7$. Then, in our example, the odds of success are $7/3$, or $2.333$ to $1$. Naturally, in the case of equal probabilities where $p = 0.5$, the odds of success is $1$ to $1$.*

## SOL-6  CH.SOL- 2.3.

1. *An interaction is the product of two single predictor variables implying a **non-additive** effect.*

2. *The simplest interaction model includes a predictor variable formed by multiplying two ordinary predictors. Let us assume two variables $X$ and $Z$. Then, the logistic regression model that employs the simplest form of interaction follows:*

$$\beta_0 + \beta_1 X + \beta_2 Z + \beta_3 XZ,$$ 

(2.9)

*where the coefficient for the interaction term $XZ$ is represented by predictor $\beta_3$.*

3. *For testing the contribution of an interaction, two principal methods are commonly employed; the Wald chi-squared test or a likelihood ratio test between the model with and without the interaction term. Note: How does interaction relates to information theory? What added value does it employ to enhance model performance?*

## SOL-7  CH.SOL- 2.4.

***False.*** *This is exactly the definition of supervised learning; when labels are known then supervision guides the learning process.*

**SOL-8  CH.SOL- 2.5.**

*In the case of logistic regression, the response variable is the log of the odds of being classified in a group of binary or multi-class responses. This definition essentially demonstrates that odds can take the form of a vector.* ◼

**SOL-9  CH.SOL- 2.6.**

*When a transformation to the response variable is applied, it yields a probability distribution over the output classes, which is bounded between 0 and 1; this transformation can be employed in several ways, e.g., a softmax layer, the sigmoid function or classic normalization. This representation facilitates a soft-decision by the logistic regression model, which permits construction of probability-based processes over the predictions of the model. Note: What are the pros and cons of each of the three aforementioned transformations?* ◼

**SOL-10  CH.SOL- 2.7.**

*Minimizing the negative log likelihood also means maximizing the **likelihood** of selecting the **correct** class.* ◼

### 2.3.2    Odds, Log-odds

**SOL-11  CH.SOL- 2.8.**

1. *The odds of the event occurring are, by definition:*

$$odds = (\frac{0.1}{0.9}) = 0.11.$$ (2.10)

2. *The log-odds of the event occurring are simply taken as the log of the odds:*

$$log\text{-}odds = \ln(0.1/0.9) = -2.19685.$$ (2.11)

3. *The probability may be constructed by the following representation:*

$$probability = \frac{odds}{odds + 1} = \frac{0.11}{1.11} = 0.1,$$ (2.12)

*or, alternatively:*

$$p = \frac{\exp{(\ln odds)}}{\exp{(\ln odds)} + 1} = \frac{0.11}{1.11} = 0.1. \tag{2.13}$$

*Note: What is the intuition behind this representation?*

■

**SOL-12**  **CH.SOL- 2.9.**

**True**. *By definition of odds, it is easy to notice that $p = 0.8$ satisfies the following relation:*

$$odds = (\frac{0.8}{0.2}) = 4 \tag{2.14}$$

■

**SOL-13**  **CH.SOL- 2.10.**

*The graph of odds to probabilities is depicted in Figure 2.12.*

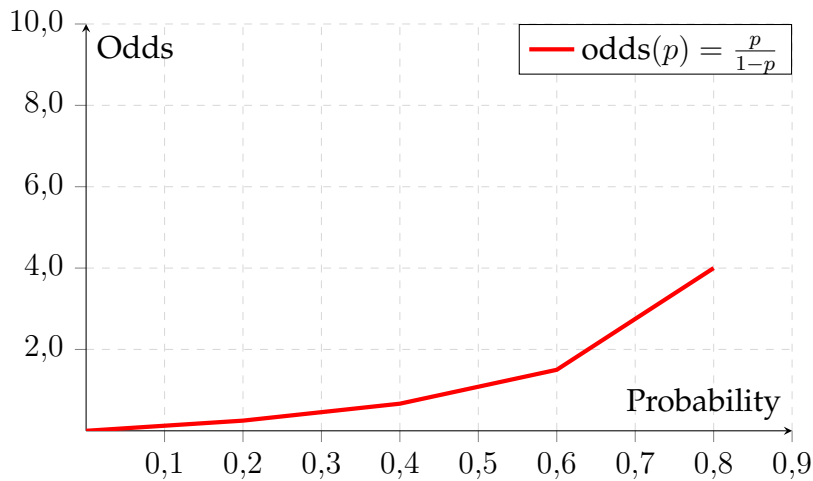

FIGURE 2.12: Odds vs. probability values.

■

**SOL-14**  **CH.SOL- 2.11.**

*A binary logistic regression GLM consists of there components:*

1. *Random component: refers to the probability distribution of the response variable* $(Y)$, *e.g., binomial distribution for* $Y$ *in the binary logistic regression, which takes on the values* $Y = 0$ *or* $Y = 1$.

2. *Systematic component: describes the explanatory variables:*
   $(X_1, X_2, ...)$ *as a combination of linear predictors. The binary case does not constrain these variables to any degree.*

3. *Link function: specifies the link between random and systematic components. It says how the expected value of the response relates to the linear predictor of explanatory variables.*

   *Note: Assume that* $Y$ *denotes whether a human voice activity was detected* $(Y = 1)$ *or not* $(Y = 0)$ *in a give time frame. Propose two systematic components and a link function adjusted for this task.*

■

**SOL-15**  **CH.SOL- 2.12.**

*The hyperplane is simply defined by:*

$$\theta_0 + \theta^T X = 0. \tag{2.15}$$

*Note: Recall the use of the logit function and derive this decision boundary rigorously.* ■

**SOL-16**  **CH.SOL- 2.13.**

***True.*** *The logit function is defined as:*

$$z(p) = logit(p) = log\left(\frac{p}{1-p}\right), \tag{2.16}$$

*for any $p \in [0,1]$. A simple set of algebraic equations yields the inverse relation:*

$$p(z) = \frac{\exp z}{1 + \exp z}, \tag{2.17}$$

*which exactly describes the relation between the output and input of the logistic function, also known as the sigmoid.* ∎

### 2.3.3 The Sigmoid

**SOL-17**  **CH.SOL- 2.14.**

There are various approaches to solve this problem, here we provide two; direct derivation or derivation via the softmax function.

1. *Direct derivation:*
   $\frac{d}{dx}\sigma(x) = \frac{d}{dx}((1 + e^{-x})^{-1}) = -((1 + e^{-x})^{(-2)})\frac{d}{dx}(1 + e^{-x}) = \frac{e^{-x}}{(1+e^{-x})^2}$.

2. *Softmax derivation:*
   *In a classification problem with mutually exclusive classes, where all of the values are positive and sum to one, a softmax activation function may be used. By definition, the softmax activation function consists of $n$ terms, such that $\forall i \in [1, n]$:*

$$f(\theta_i) = \frac{e^{\theta_i}}{\sum_k e^{v_k}} = \frac{1}{1 + e^{-\theta_i} \sum_{k \neq i} e^{\theta_k}}. \tag{2.18}$$

*To compute the partial derivative of 2.18, we treat all $\theta_k$ where $k \neq i$ as **constants** and then differentiate $\theta_i$ using regular differentiation rules. For a given $\theta_i$, let us define:*

$$\beta = \sum_{k \neq i} e^{\theta_k}, \tag{2.19}$$

*and*

$$f(\theta_i) = \frac{1}{1 + \beta e^{-\theta_i}} = (1 + \beta e^{-\theta_i})^{-1}. \tag{2.20}$$

*It can now be shown that the derivative with respect to $\theta_i$ holds:*

$$f'(\theta_i) = \left(1 + \beta e^{-\theta_i}\right)^{-2} \beta e^{-\theta_i}, \tag{2.21}$$

*which can take on the informative form of:*

$$f'(\theta_i) = f(\theta_i)(1 - f(\theta_i)). \tag{2.22}$$

*It should be noted that 2.21 holds for any constant $\beta$, and for $\beta = 1$ it clearly reduces to the sigmoid activation function.*

*Note: Characterize the sigmoid function when its argument approaches $0, \infty$ and $-\infty$. What undesired properties of the sigmoid function do this values entail when considered as an activation function?*

■

---

**SOL-18**  **CH.SOL- 2.15.**

1. *The logit value is simply obtained by substituting the values of the dependent variables and model coefficients into the linear logistic regression model, as follows:*

$$logit = \beta_0 + \beta_1 x_1 + \beta_2 x_2 = -1.5 + 3 \cdot 1 + -0.5 \cdot 5 = -1. \tag{2.23}$$

2. *According to the natural relation between the logit and the odds, the following holds:*

$$odds = e^{logit} = e^{\beta_0 + \beta_1 x_1 + \beta_2 x_2} = e^{-1} = 0.3678794. \tag{2.24}$$

3. *The **odds ratio** is, by definition:*

$$odds = \frac{P(y = 1)}{P(y = 0)}, \tag{2.25}$$

*so the logistic response function is:*

$$P(y = 1) = \frac{1}{1 + e^{-logit}} = \frac{1}{1 + e^1} = 0.2689414. \tag{2.26}$$

■

### 2.3.4   Truly Understanding Logistic Regression

**SOL-19**  **CH.SOL- 2.16.**

1. *Tumour eradication (Y) is the response variable and cancer type (X) is the explanatory variable.*

2. *Relative risk (RR) is the ratio of risk of an event in one group (e.g., exposed group) versus the risk of the event in the other group (e.g., non-exposed group). The odds ratio (OR) is the ratio of odds of an event in one group versus the odds of the event in the other group.*

3. *If we calculate odds ratio as a measure of association:*

$$\hat{\theta} = \frac{560 \times 36}{69 \times 260} = 1.23745. \tag{2.27}$$

*And the log-odds ratio is* $(\log(1.23745)) = 0.213052$*:*

*The odds ratio is **larger** than one, indicating that the odds for a breast cancer is more than the odds for a lung cancer to be eradicated. Notice however, that this result is too close to one, which prevents conclusive decision regarding the odds relation.*

*Additionally, if we calculate relative risk as a measure of association:*

$$RR = \frac{\frac{560}{560+260}}{\frac{69}{69+36}} = 1.0392. \tag{2.28}$$

4. *The* $95\%$ *confidence interval for the odds-ratio,* $\theta$ *is computed from the sample confidence interval for log odds ratio:*

$$\hat{\sigma}\left(\log(\hat{\theta})\right) = \sqrt{\frac{1}{560} + \frac{1}{260} + \frac{1}{69} + \frac{1}{36}} = 0.21886. \tag{2.29}$$

*Therefore, the 95% CI for* $\log(\theta)$ *is:*

$$0.213052 \pm 1.95 \times 0.21886 = (0.6398298, -0.2137241). \tag{2.30}$$

*Therefore, the 95% CI for θ is:*

$$(e^{-0.210}, e^{0.647}) = (0.810, 1.909). \tag{2.31}$$

5. *The CI* $(0.810, 1.909)$ *contains 1, which indicates that the true odds ratio is not significantly different from 1 and* **there is not enough evidence** *that tumour eradication is dependent on cancer type.*

■

**SOL-20**  **CH.SOL- 2.17.**

1. *By using the defined values for $X_1$ and $X_2$, and the known logistic regression model, substitution yields:*

$$\hat{p}(X) = \frac{e^{-6+0.05X_1+X_2}}{(1 + e^{-6+0.05X_1+X_2})} = 0.3775. \tag{2.32}$$

2. *The equation for the predicted probability tells us that:*

$$\frac{e^{-6+0.05X_1+3.5}}{(1 + e^{-6+0.05X_1+3.5})} = 0.5, \tag{2.33}$$

*which is equivalent to constraining:*

$$e^{-6+0.05X_1+3.5} = 1. \tag{2.34}$$

*By taking the logarithm of both sides, we get that the number of milliseconds needed is:*

$$X_1 = \frac{2.5}{0.05} = 50. \tag{2.35}$$

■

**SOL-21**  **CH.SOL- 2.18.**

*For the purpose of this exercise, it is instructive to pre-define $z$ as:*

$$z\left(X_1, X_2\right) = -6.36 - 1.02 \times X_1 + 0.12 \times X_2. \qquad (2.36)$$

1. *By employing the classic logistic regression model:*

$$odds = \exp(z\left(X_1, X_2\right)). \qquad (2.37)$$

2. *By substituting the given values of $X_1, X_2$ into $z\left(X_1, X_2\right)$, the probability holds:*

$$p = exp(z\left(1, 100\right))/(1 + exp(z\left(1, 100\right))) = 0.99. \qquad (2.38)$$

3. *Yes. The coefficient for coffee consumption is positive (0.119) and the p-value is less than 0.05 (0.0304).*
   *Note: Can you describe the relation between these numerical relations and the positive conclusion?*

4. *No. The p-value for this predictor is 0.3818 > 0.05.*
   *Note: Can you explain why this inequality implicates a lack of statistical evidence?*

## SOL-22  CH.SOL- 2.19.

1. *The estimated probability of improvement is:*

$$\hat{\pi}(gum\ bacteria) =$$
$$\frac{\exp(-4.8792 + 0.0258 \times gum\ bacteria)}{1 + \exp(-4.8792 + 0.0258 \times gum\ bacteria)}.$$

*Hence, $\hat{\pi}(33) = 0.01748$.*

2. *For* $\hat{\pi}(gum\ bacteria) = 0.5$ *we know that:*

$$\hat{\pi}(gum) = \frac{\exp(\hat{\alpha} + \hat{\beta}x)}{1 + \exp(\hat{\alpha} + \hat{\beta}x)} = 0.5 \qquad (2.39)$$

$$gum\ bacteria = -\hat{\alpha}/\hat{\beta} = 4.8792/0.0258 = 189.116. \qquad (2.40)$$

3. *The estimated odds ratio are given by:*

$$\exp(\hat{\beta}) = \exp(0.0258) = 1.0504. \qquad (2.41)$$

4. *A* 99% *confidence interval for* $\beta$ *is calculated as follows:*

$$\hat{\beta} \pm z_{0.005} \times ASE(\hat{\beta}) = \qquad (2.42)$$
$$0.0258 \pm 2.576 \times 0.0194 \qquad (2.43)$$
$$= (-0.00077, 0.9917). \qquad (2.44)$$

*Therefore, a* 99% *confidence interval for the true odds ratio* $\exp(\beta)$ *is given by:*

$$(\exp(-0.00077), \exp(0.9917)) = (0.99923, 2.6958). \qquad (2.45)$$

■

## SOL-23  CH.SOL- 2.20.

1. *The sample odds ratio is:*

$$\hat{\theta} = \frac{130 \times 6833}{60 \times 6778} = 2.1842. \qquad (2.46)$$

2. *The estimated standard error for $\log\left(\hat{\theta}\right)$ is:*

$$\hat{\sigma}\left(\log\hat{\theta}\right) = \sqrt{\frac{1}{60} + \frac{1}{6833} + \frac{1}{130} + \frac{1}{6778}} = 0.1570. \tag{2.47}$$

3. *According to previous sections, the 95% CI for the true log odds ratio is:*

$$0.7812 \pm 1.96 \times 0.1570 = (0.4734, 1.0889). \tag{2.48}$$

*Correspondingly, the 95% CI for the true odds ratio is:*

$$(e^{0.4734}, e^{1.0889}) = (1.6060, 2.9710). \tag{2.49}$$

### 2.3.5 The Logit Function and Entropy

**SOL-24**  **CH.SOL- 2.21.**

1. *The entropy (Fig. 2.13) has a maximum value of $\log_2(2)$ for probability $p = 1/2$, which is the most chaotic distribution. A lower entropy is a more predictable outcome, with zero providing full certainty.*

2. *The derivative of the entropy with respect to $p$ yields **the negative of the logit** function:*

$$\frac{dH(p)}{dp} = -logit(p). \tag{2.50}$$

*Note: The curious reader is encouraged to rigorously prove this claim.*

### 2.3.6 Python, PyTorch, CPP

**SOL-25**  **CH.SOL- 2.22.**

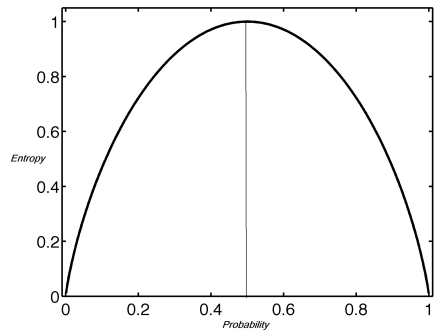

FIGURE 2.13: Binary entropy.

1. *During inference, the purpose of **inner_product** is to multiply the vector of logistic regression coefficients with the vector of the input which we like to evaluate, e.g., calculate the probability and binary class.*

2. *The line **hypo(x) > 0.5f** is commonly used for the evaluation of binary classification wherein probability values above 0.5 (i.e., a threshold) are regarded as TRUE whereas values below 0.5 are regarded as FALSE.*

3. *The term $\theta$ **(theta)** stands for the logistic regression coefficients which were evaluated during training.*

4. *The output is as follows:*

```
> inner_product=-0.5
> prob=0.377541
> hypo=0
```

FIGURE 2.14: Logistic regression in C++

■

**SOL-26 ✏ CH.SOL- 2.23.**

*Because the second dimension of  is 7, and the first dimension of* <span>data</span> *is 3, the resulting matrix has a shape of* <span>torch.Size([3, 7])</span>.

■

## SOL-27 ✏ CH.SOL- 2.24.

*Ideally, you should be able to recognize these functions immediately upon a request from the interviewer.*

1. *A softmax function.*

2. *A sigmoid function.*

3. *A derivative of a sigmoid function.*

■

## SOL-28 ✏ CH.SOL- 2.25.

*The function implemented in Fig. 2.10 is the*  *cross-entropy function.*  ■

## SOL-29 ✏ CH.SOL- 2.26.

1. *All the methods are variations of the sigmoid function.*

2. *In Python, approximately* $1.797e + 308$ *holds the largest possible valve for a floating point variable. The logarithm of which is evaluated at* $709.78$. *If you try to execute the following expression in Python, it will result in* <span>inf</span>: $np.log(1.8e + 308)$.

3. *I would use*  *because of its stability. Note: Can you entail why is this method more stable than the others?*

■

# CHAPTER

<span style="color:#b01030">PROBABILISTIC PROGRAMMING & BAYESIAN DL</span>

*Anyone who considers arithmetical methods of producing random digits is, of course, in a state of sin.*

— John von Neumann (1903-1957)

## Contents

## 3.1 Introduction

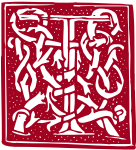 HE Bayesian school of thought has permeated fields such as mechanical statistics, classical probability, and financial mathematics [13]. In tandem, the subject matter itself has gained attraction, particularly in the field of BML. It is not surprising then, that several new Python based probabilistic programming libraries such as PyMc3 and Stan [11] have emerged and have become widely adopted by the machine learning community.

This chapter aims to introduce the Bayesian paradigm and apply Bayesian inferences in a variety of problems. In particular, the reader will be introduced with real-life examples of conditional probability and also discover one of the most important results in Bayesian statistics: that the family of beta distributions is **conjugate to a binomial likelihood**. It should be stressed that Bayesian inference is a subject matter that students evidently find hard to grasp, since it heavily relies on rigorous probabilistic interpretations of data. Specifically, several obstacles hamper with the prospect of learning Bayesian statistics:

1. Students typically undergo merely basic introduction to classical probability and statistics. Nonetheless, what follows requires a very solid grounding in these fields.

2. Many courses and resources that address Bayesian learning do not cover essential concepts.

3. A strong comprehension of Bayesian methods involves numerical training and sophistication levels that go beyond first year calculus.

Conclusively, this chapter may be much harder to understand than other chapters. Thus, we strongly urge the readers to thoroughly solve the following questions and verify their grasp of the mathematical concepts in the basis of the solutions [8].

## 3.2 Problems

### 3.2.1 Expectation and Variance

**PRB-30 ❷ CH.PRB- 3.1.**
*Define what is meant by a **Bernoulli trial**.*

**PRB-31 ❷ CH.PRB- 3.2.**

*The binomial distribution is often used to model the probability that $k$ out of a group of $n$ objects bare a specific characteristic. Define what is meant by a **binomial random variable** $X$.*

**PRB-32 ❷ CH.PRB- 3.3.**

*What does the following shorthand stand for?*

$$X \sim \text{Binomial}(n, p) \tag{3.1}$$

**PRB-33 ❷ CH.PRB- 3.4.**

*Find the probability mass function (PMF) of the following random variable:*

$$X \sim \text{Binomial}(n, p) \tag{3.2}$$

**PRB-34 ❷ CH.PRB- 3.5.**

*Answer the following questions:*

1. *Define what is meant by (mathematical) expectation.*

2. *Define what is meant by variance.*

3. *Derive the expectation and variance of a the binomial random variable $X \sim \text{Binomial}(n, p)$ in terms of $p$ and $n$.*

**PRB-35 ❷ CH.PRB- 3.6.**

*Proton therapy (PT) is a widely adopted form of treatment for many types of cancer [6]. A PT device which **was not properly calibrated** is used to treat a patient with pancreatic cancer (Fig. 3.1). As a result, a PT beam randomly shoots 200 particles independently and **correctly** hits cancerous cells with a probability of 0.1.*

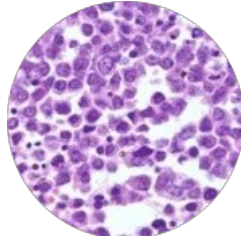

FIGURE 3.1: Histopathology for pancreatic cancer cells.

1. *Find the **statistical distribution** of the number of correct hits on cancerous cells in the described experiment. What are the expectation and variance of the corresponding random variable?*

2. *A radiologist using the device claims he was able to hit exactly 60 cancerous cells. How likely is it that he is **wrong**?*

### 3.2.2 Conditional Probability

**PRB-36 ❷ CH.PRB- 3.7.**
 *Given two events $A$ and $B$ in probability space $H$, which occur with probabilities $P(A)$ and $P(B)$, respectively:*

1. *Define the conditional probability of $A$ given $B$. Mind singular cases.*

2. *Annotate each part of the conditional probability formulae.*

3. *Draw an instance of Venn diagram, depicting the intersection of the events $A$ and $B$. Assume that $A \cup B = H$.*

**PRB-37 ❷ CH.PRB- 3.8.**
 *Bayesian inference amalgamates data information in the likelihood function with known **prior** information. This is done by conditioning the prior on the likelihood using the Bayes formulae. Assume two events $A$ and $B$ in probability space $H$, which occur with probabilities* $P(A)$ and $P(B)$, respectively. Given that $A \bigcup B = H$, state the Bayes formulae for this case, interpret its components and annotate them.

**PRB-38 ❷ CH.PRB- 3.9.**

Define the terms **likelihood** and **log-likelihood** of a discrete random variable $X$ given a fixed parameter of interest $\gamma$. Give a practical example of such scenario and derive its likelihood and log-likelihood.

**PRB-39 ❷ CH.PRB- 3.10.**

Define the term **prior distribution** of a likelihood parameter $\gamma$ in the continuous case.

**PRB-40 ❷ CH.PRB- 3.11.**

Show the **relationship** between the prior, posterior and likelihood probabilities.

**PRB-41 ❷ CH.PRB- 3.12.**

In a Bayesian context, if a first experiment is conducted, and then another experiment is followed, what does the **posterior** become for the **next** experiment?

**PRB-42 ❷ CH.PRB- 3.13.**

What is the condition under which two events $A$ and $B$ are said to be **statistically independent**?

### 3.2.3   Bayes Rule

**PRB-43 ❷ CH.PRB- 3.14.**

In an experiment conducted in the field of particle physics (Fig. 3.2), a certain particle may be in two distinct **equally probable** quantum states: integer spin or half-integer spin. It is well-known that particles with integer spin are bosons, while particles with half-integer spin are fermions [4].

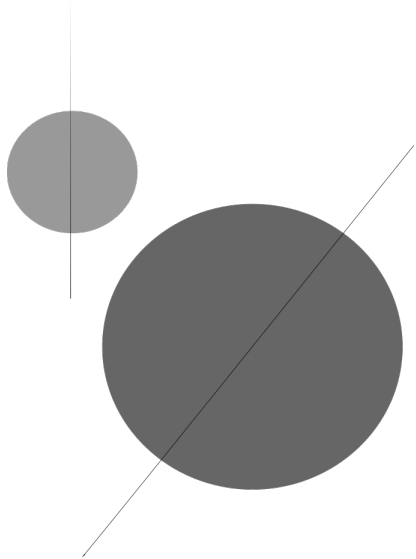

FIGURE 3.2: Bosons and fermions: particles with half-integer spin are fermions.

*A physicist is observing two such particles, while at least one of which is in a half-integer state. What is the probability that **both** particles are fermions?*

**PRB-44 ❷ CH.PRB- 3.15.**

*During pregnancy, the Placenta Chorion Test [1] is commonly used for the diagnosis of hereditary diseases (Fig. 3.3). The test has a probability of 0.95 of being correct **whether or not** a hereditary disease is present.*

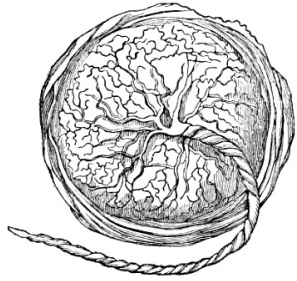

FIGURE 3.3: Foetal surface of the placenta

*It is known that 1% of pregnancies result in hereditary diseases. Calculate the probability of a test indicating that a hereditary disease is present.*

**PRB-45 ❷ CH.PRB- 3.16.**

*The Dercum disease [3] is an extremely rare disorder of multiple painful tissue growths. In a population in which the ratio of females to males is equal, 5% of females and 0.25% of males have the Dercum disease (Fig. 3.4).*

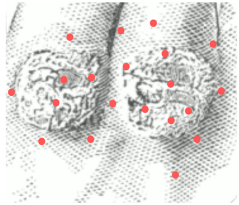

FIGURE 3.4: The Dercum disease

*A person is chosen at random and that person has the Dercum disease. Calculate the probability that the person is female.*

**PRB-46 ❷ CH.PRB- 3.17.**

*There are numerous fraudulent binary options websites scattered around the Internet, and for every site that shuts down, new ones are sprouted like mushrooms. A **fraudulent** AI*

*based stock-market prediction algorithm utilized at the New York Stock Exchange, (Fig. 3.6)*
*can correctly predict if a certain binary option [7] shifts states from 0 to 1 or the other way*
*around, with* 85% *certainty.*

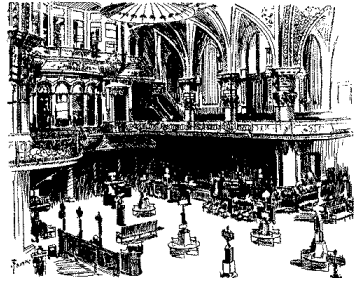

FIGURE 3.5: The New York Stock Exchange.

*A financial engineer has created a portfolio consisting twice as many state-1 options then*
*state-0 options. A stock option is selected at random and is determined by said algorithm to*
*be in the state of 1. What is the probability that the prediction made by the AI is correct?*

### PRB-47 ❷ CH.PRB- 3.18.

*In an experiment conducted by a hedge fund to determine if monkeys (Fig. 3.6) can*
*outperform humans in selecting better stock market portfolios, 0.05 of humans and 1 out of*
*15 monkeys could correctly predict stock market trends correctly.*

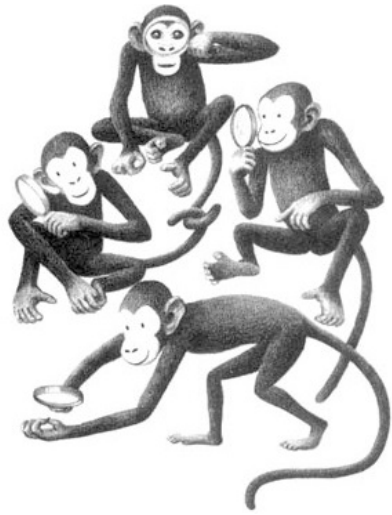

FIGURE 3.6: Hedge funds and monkeys.

*From an equally probable pool of humans and monkeys an "expert" is chosen at random. When tested, that expert was correct in predicting the stock market shift. What is the probability that the expert is a **human**?*

---

**PRB-48 ❷ CH.PRB- 3.19.**

*During the cold war, the U.S.A developed a speech to text (STT) algorithm that could theoretically detect the hidden dialects of Russian sleeper agents. These agents (Fig. 3.7), were trained to speak English in Russia and subsequently sent to the US to gather intelligence. The FBI was able to apprehend ten such hidden Russian spies [9] and accused them of being "sleeper" agents.*

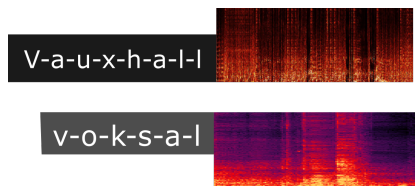

FIGURE 3.7: Dialect detection.

*The Algorithm relied on the acoustic properties of Russian pronunciation of the word (v-o-k-s-a-l) which was borrowed from English V-a-u-x-h-a-l-l. It was alleged that it is **impossible for Russians to completely hide** their accent and hence when a Russian would say V-a-u-x-h-a-l-l, the algorithm would yield the text "**v-o-k-s-a-l**". To test the algorithm at a diplomatic gathering where 20% of participants are Sleeper agents and the rest Americans, a data scientist randomly chooses a person and asks him to say V-a-u-x-h-a-l-l. A single letter is then chosen randomly from the word that was generated by the algorithm, which is observed to be an "l". What is the probability that the person is indeed a Russian sleeper agent?*

**PRB-49 ❷ CH.PRB- 3.20.**

*During World War II, forces on both sides of the war relied on encrypted communications. The main encryption scheme used by the German military was an Enigma machine [5], which was employed extensively by Nazi Germany. Statistically, the Enigma machine sent the symbols X and Z Fig. (3.8) according to the following probabilities:*

$$P(X) = \frac{2}{9} \tag{3.3}$$

$$P(Z) = \frac{7}{9} \tag{3.4}$$

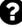

FIGURE 3.8: The Morse telegraph code.

*In one incident, the German military sent encoded messages while the British army used countermeasures to deliberately tamper with the transmission. Assume that as a result of the British countermeasures, an X is erroneously received as a Z (and mutatis mutandis) with a*

*probability $\frac{1}{7}$. If a recipient in the German military received a Z, what is the probability that a Z was actually transmitted by the sender?*

### 3.2.4    Maximum Likelihood Estimation

**PRB-50 ❷ CH.PRB- 3.21.**

*What is **likelihood function** of the independent identically distributed (i.i.d) random variables:*
$X_1, \cdots, X_n$ *where* $X_i \sim \text{binomial}(n, p)$, $\forall i \in [1, n]$,
*and where $p$ is the parameter of interest?*

**PRB-51 ❷ CH.PRB- 3.22.**

*How can we derive the **maximum likelihood estimator (MLE)** of the i.i.d samples* $X_1, \cdots, X_n$ *introduced in Q. 3.21?*

**PRB-52 ❷ CH.PRB- 3.23.**

*What is the relationship between the likelihood function and the log-likelihood function?*

**PRB-53 ❷ CH.PRB- 3.24.**

*Describe how to analytically find the MLE of a likelihood function?*

**PRB-54 ❷ CH.PRB- 3.25.**

*What is the term used to describe **the first derivative** of the log-likelihood function?*

**PRB-55 ❷ CH.PRB- 3.26.**

*Define the term **Fisher information**.*

### 3.2.5    Fisher Information

**PRB-56 ❷ CH.PRB- 3.27.**

*The 2014 west African Ebola (Fig. 9.10) epidemic has become the largest and fastest-spreading outbreak of the disease in modern history [2] with a death tool far exceeding all past outbreaks combined. Ebola (named after the Ebola River in Zaire) first emerged in 1976 in Sudan and Zaire and infected over 284 people with a mortality rate of 53%.*

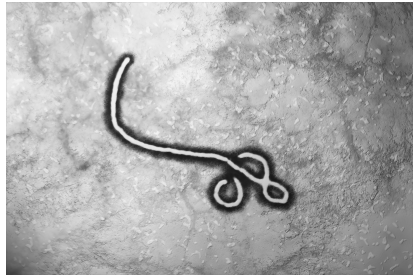

FIGURE 3.9: The Ebola virus.

*This rare outbreak, underlined the challenge medical teams are facing in containing epidemics. A junior data scientist at the center for disease control (CDC) models the possible spread and containment of the Ebola virus using a numerical simulation. He knows that out of a population of $k$ humans (the number of trials), $x$ are carriers of the virus (success in statistical jargon). He believes the sample likelihood of the virus in the population, follows a Binomial distribution:*

$$L(\gamma \mid y) = \begin{pmatrix} n \\ y \end{pmatrix} \gamma^y (1-\gamma)^{n-y}, \quad \gamma \in [0,1], \quad y = 1, 2, \ldots, n \quad (3.5)$$

*As the senior researcher in the team, you guide him that his parameter of interest is $\gamma$, the proportion of infected humans in the entire population. The expectation and variance of the binomial distribution are:*

$$E(y|\gamma, n) = n\gamma, \quad V(y|\gamma, n) = n\gamma(1-\gamma) \quad (3.6)$$

*Answer the following; for the likelihood function of the form $L_x(\gamma)$:*

1. *Find the log-likelihood function $l_x(\gamma) = \ln L_x(\gamma)$.*

2. *Find the gradient of $l_x(\gamma)$.*

3. *Find the Hessian matrix $H(\gamma)$.*

4. *Find the Fisher information $I(\gamma)$.*

5. *In a population spanning 10,000 individuals, 300 were infected by Ebola. Find the MLE for $\gamma$ and the standard error associated with it.*

---

**PRB-57**  **CH.PRB- 3.28.**

*In this question, you are going to derive the Fisher information function for several distributions. Given a probability density function (PDF) $f(X|\gamma)$, you are provided with the following definitions:*

1. *The natural logarithm of the PDF $\ln f(X|\gamma) = \Phi(X|\gamma)$.*

2. *The first partial derivative $\Phi'(X|\gamma)$.*

3. *The second partial derivative $\Phi''(X|\gamma)$.*

4. *The Fisher Information for a continuous random variable:*

$$I(\gamma) = -\mathrm{E}_\gamma \left[ \Phi'(X|\gamma)^2 \right].\qquad(3.7)$$

*Find the Fisher Information $I(\gamma)$ for the following distributions:*

1. *The Bernoulli Distribution $X \sim B(1, \gamma)$.*

2. *The Poisson Distribution $X \sim Poiss(\theta)$.*

---

**PRB-58**  **CH.PRB- 3.29.**

1. ***True or False:*** *The Fisher Information is used to compute the Cramer-Rao bound on the variance of any unbiased maximum likelihood estimator.*

2. ***True or False:*** *The Fisher Information matrix is also the Hessian of the symmetrized KL divergence.*

### 3.2.6   Posterior & prior predictive distributions

**PRB-59 ❷ CH.PRB- 3.30.**

    *In chapter 3 we discussed the notion of a prior and a posterior distribution.*

1. *Define the term **posterior distribution**.*

2. *Define the term **prior predictive distribution**.*

**PRB-60 ❷ CH.PRB- 3.31.**

    *Let $y$ be the number of successes in 5 independent trials, where the probability of success is $\theta$ in each trial. Suppose your prior distribution for $\theta$ is as follows: $P(\theta = 1/2) = 0.25$, $P(\theta = 1/6) = 0.5$, and $P(\theta = 1/4) = 0.25$.*

1. *Derive the **posterior distribution** $p(\theta|y)$ **after** observing $y$.*

2. *Derive the **prior predictive distribution** for $y$.*

### 3.2.7   Conjugate priors

**PRB-61 ❷ CH.PRB- 3.32.**

    *In chapter 3 we discussed the notion of a prior and a posterior.*

1. *Define the term **conjugate prior**.*

2. *Define the term **non-informative prior**.*

The Beta-Binomial distribution

**PRB-62 ❷ CH.PRB- 3.33.**

    *The Binomial distribution was discussed extensively in chapter 3. Here, we are going to show one of the most important results in Bayesian machine learning. Prove that the family of beta distributions is **conjugate to a binomial likelihood**, so that if a prior is in that*

*family then so is the posterior. That is, show that:*

$$x \sim \text{Ber}(\gamma), \quad \gamma \sim \mathcal{B}(\alpha, \beta) \quad \Rightarrow \quad \gamma | x \sim \mathcal{B}(\alpha', \beta') \tag{3.8}$$

*For instance, for $h$ heads and $t$ tails, the posterior is:*

$$\mathcal{B}(h + \alpha, t + \beta) \tag{3.9}$$

### 3.2.8    Bayesian Deep Learning

**PRB-63 ❷ CH.PRB- 3.34.**

*A recently published paper presents a new layer for a new Bayesian neural network (BNN). The layer behaves as follows. During the feed-forward operation, each of the hidden neurons $H_n$, $n \in 1, 2$ in the neural network (Fig. 3.10) **may, or may not** fire independently of each other according to a known prior distribution.*

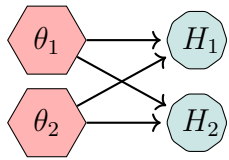

FIGURE 3.10: Likelihood in a BNN model.

*The chance of firing, $\gamma$, is the same for each hidden neuron. Using the formal definition, calculate the likelihood function of each of the following cases:*

1. *The hidden neuron is distributed according to $X \sim \text{binomial}(n, \gamma)$ random variable and fires with a probability of $\gamma$. There are 100 neurons and only 20 are **fired**.*

2. *The hidden neuron is distributed according to $X \sim Uniform(0, \gamma)$ random variable and fires with a probability of $\gamma$.*

**PRB-64 ❷ CH.PRB- 3.35.**

*Your colleague, a veteran of the Deep Learning industry, comes up with an idea for for*

*a BNN layer entitled **OnOffLayer**. He suggests that each neuron will stay **on** (the other state is off) following the distribution $f(x) = e^{-x}$ for $x > 0$ and $f(x) = 0$ otherwise (Fig. 3.11). X indicates the **time in** seconds the neuron stays on. In a BNN, 200 such neurons are activated independently in said OnOffLayer. The OnOffLayer is set to off (e.g. not active) **only if at least 150 of the neurons are shut down**. Find the probability that the OnOffLayer will be active for at least 20 seconds without being shut down.*

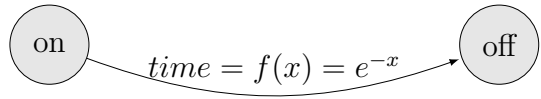

FIGURE 3.11: OnOffLayer in a BNN model.

**PRB-65 ❷ CH.PRB- 3.36.**

*A Dropout layer [12] (Fig. 3.12) is commonly used to regularize a neural network model by randomly equating several outputs (the crossed-out **hidden** node H) to 0.*

Dropout

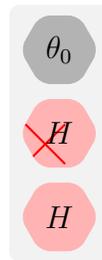

FIGURE 3.12: A Dropout layer (simplified form).

*For instance, in PyTorch [10], a Dropout layer is declared as follows (3.1):*

```
1 import torch
2 import torch.nn as nn
3 nn.Dropout(0.2)
```

CODE 3.1: Dropout in PyTorch

Where $nn.Dropout(0.2)$ *(Line #3 in 3.1) indicates that the probability of zeroing an element is* $0.2$.

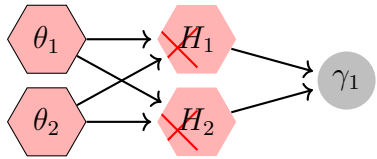

FIGURE 3.13: A Bayesian Neural Network Model

*A new data scientist in your team suggests the following procedure for a Dropout layer which is based on Bayesian principles. Each of the neurons* $\theta_n$ *in the neural network in (Fig. 8.33) may drop (or not) independently of each other exactly like a Bernoulli trial.*

*During the training of a neural network, the Dropout layer randomly drops out outputs of the previous layer, as indicated in (Fig. 3.12). Here, for illustration purposes, all two neurons are dropped as depicted by the crossed-out **hidden** nodes* $H_n$.

*You are interested in the proportion* $\theta$ *of dropped-out neurons. Assume that the chance of drop-out,* $\theta$, *is **the same** for each neuron (e.g. a **uniform prior** for* $\theta$*). Compute the **posterior** of* $\theta$.

**PRB-66 ❷ CH.PRB- 3.37.**

*A new data scientist in your team, who was formerly a Quantum Physicist, suggests the following procedure for a Dropout layer entitled **QuantumDrop** which is based on Quantum principles and the **Maxwell Boltzmann distribution**. In the Maxwell-Boltzmann*

*distribution, the likelihood of finding a particle with a particular velocity $v$ is provided by:*

$$n(v)dv = \frac{4\pi N}{V}\left(\frac{m}{2\pi kT}\right)^{3/2}v^2 e^{-\frac{mv^2}{2kT}}\,dv \tag{3.10}$$

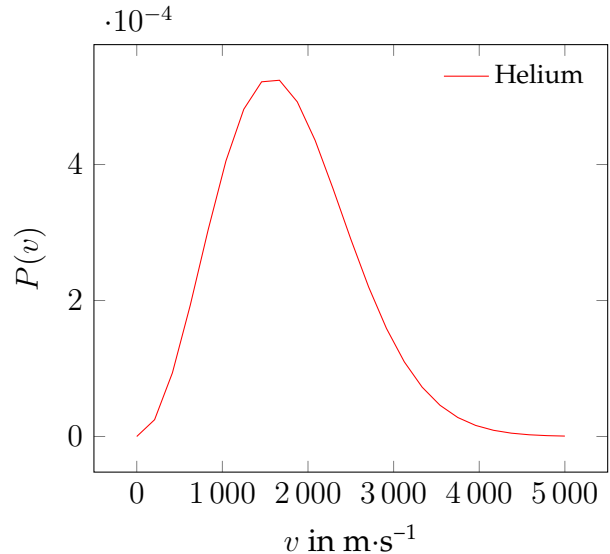

FIGURE 3.14: The Maxwell-Boltzmann distribution.

*In the suggested QuantumDrop layer (3.15), each of the neurons behaves like a molecule and is distributed according to the Maxwell-Boltzmann distribution and **fires only when the most probable speed is reached**. This speed is the velocity associated with the highest point in the Maxwell distribution (3.14). Using calculus, brain power and some mathematical manipulation, find the **most likely value (speed) at which the neuron will fire**.*

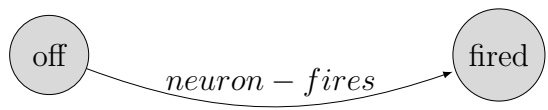

FIGURE 3.15: A QuantumDrop layer.

## 3.3   Solutions

### 3.3.1   Expectation and Variance

**SOL-30 ✏ CH.SOL- 3.1.**

*The notion of a Bernoulli trial refers to an experiment with two dichotomous binary outcomes; success (x = 1), and failure (x = 0).*  ▪

**SOL-31 ✏ CH.SOL- 3.2.**

*A binomial random variable $X = k$ represents $k$ successes in $n$ mutually independent Bernoulli trials.*  ▪

**SOL-32 ✏ CH.SOL- 3.3.**

*The shorthand $X \sim \text{Binomial}(n, p)$ indicates that the random variable $X$ has the binomial distribution (Fig. 3.16). The positive integer parameter $n$ indicates the number of Bernoulli trials and the real parameter $p$, $0 < p < 1$ holds the probability of success in each of these trials.*

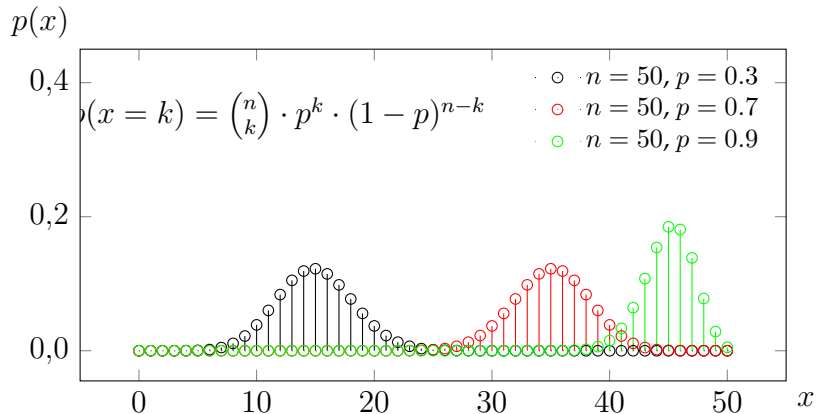

FIGURE 3.16: The binomial distribution.

▪

**SOL-33 ✏ CH.SOL- 3.4.**

*The random variable $X \sim \text{Binomial}(n,\, p)$ has the following PMF:*

$$P(X = k) = \binom{n}{k} p^k \left(1-p\right)^{n-k};\; k = 0,\, 1,\, 2,\, \ldots,\, n. \tag{3.11}$$

---

### SOL-34  CH.SOL- 3.5.

*The answers below regard a discrete random variable. The curious reader is encouraged to expend them to the continuous case.*

1. *For a random variable $X$ with probability mass function $P(X = k)$ and a set of outcomes $K$, the* **expected value** *of $X$ is defined as:*

$$E[X] := \sum_{k \in K} kP(X = k). \tag{3.12}$$

   *Note: The expectation of $X$ may also be denoted by $\mu_X$.*

2. *The* **variance** *of $X$ is defined as:*

$$\text{Var}[X] := E\big[(X - E[X])^2\big]. \tag{3.13}$$

   *Note: The variance of $X$ may also be denoted by $\sigma_X^2$, while $\sigma_X$ itself denotes the* **standard deviation** *of $X$.*

3. *The population mean and variance of a binomial random variable with parameters $n$ and $p$ are:*

$$E[X] = np \qquad V[X] = np(1 - p) \tag{3.14}$$

   *Note: Why is this solution intuitive? What information theory-related phenomenon occurs when $p = 1/2$?*

---

### SOL-35  CH.SOL- 3.6.

1. *This scenario describes an experiment that is repeated 200 times independently with a success probability of 0.1. Thus, if the random variable $X$ denotes the number of times success was obtained, then it is best characterized by the binomial distribution with parameters $n = 200$ and $p = 0.1$. Formally:*

$$X \sim \text{Binomial}(200, \, 0.1). \tag{3.15}$$

*The expectation of $X$ is given by:*

$$x = E(x) = 200 \times 0.1 = 20, \tag{3.16}$$

*and its respective variance is:*

$$Var = 200 \times 0.10(1 - 0.10) = 18.0. \tag{3.17}$$

2. *Here we propose two distinguished methods to answer the question.*
   *Primarily, the straightforward solution is to employ the definition of the binomial distribution and substitute the value of $X$ in it. Namely:*

$$
\begin{aligned}
P(X = 60; n &= 200, p = 0.1) \\
&= \binom{200}{60} 0.1^{60} \left(1 - 0.1\right)^{200-60} \\
&= \approx 2.7 \times e^{-15}.
\end{aligned}
\tag{3.18}
$$

*This leads to an extremely high probability that the radiologist is mistaken.*

*The following approach is longer and more advanced, but grants the reader with insights and intuition regarding the results. To derive how wrong the radiologist is, we can employ an approximation by considering the standard normal distribution. In statistics, the **Z-score** allows us to understand how far from the mean is a data point in units of standard deviation, thus revealing how likely it is to occur (Fig. 3.17).*

$$z = \frac{x - \mu}{\sigma}. \qquad (3.19)$$

FIGURE 3.17: Z-score

*Therefore, the probability of correctly hitting 60 cells is:*

$$P(X \geq 60) = P(Z \geq \frac{60 - 20}{\sqrt{18.0}}) = P(Z \geq 9.428) \approx 0. \qquad (3.20)$$

*Again, the outcome shows the likelihood that the radiologist was wrong approaches 1. Note: Why is the relation depicted in Fig. 3.17 deduces that Z is a standard Gaussian? Under what terms is this conclusion valid? Why does eq. (3.20) employs the cumulative distribution function and not the probability mass function?*

### 3.3.2 Conditional Probability

**SOL-36 ✎ CH.SOL- 3.7.**

1. *For two events A and B with $P(B) > 0$, the* **conditional probability** *of A given that B has occurred is defined as:*

$$P(A|B) = \frac{P(A \cap B)}{P(B)}. \qquad (3.21)$$

*It is easy to note that if $P(B) = 0$, this relation is not defined mathematically. In this case, $P(A|B) = P(A \cap B) = P(A)$.*

2. *The annotated probabilities are displayed in Fig. 3.18:*

A given B

A and B

$$P(A|B) = \frac{P(A \cap B)}{P(B)}. \qquad (3.22)$$

B only

FIGURE 3.18: Conditional probability

3. *An example of a diagram depicting the intersected events $A$ and $B$ is displayed in Fig. 3.19:*

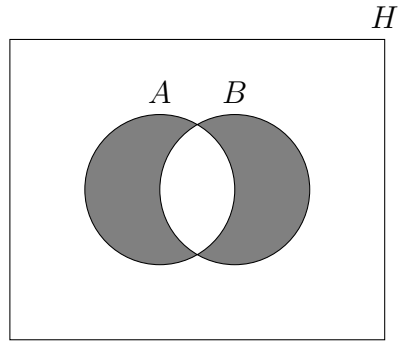

FIGURE 3.19: Venn diagram of the intersected events A and B in probability space H

**SOL-37 ✎ CH.SOL- 3.8.**

*The Bayes formulae reads:*

$$P(A|B) = \frac{P(B|A)P(A)}{P(B|A)P(A) + P(B|A^c)P(A^c)}, \qquad (3.23)$$

*where $P(A^c)$ is the complementary probability of $P(A)$. The interpretation of the elements in*

*Bayes formulae is as follows:*

$$posterior\ probability\ = \frac{likelihood\ of\ the\ data \times prior\ probability}{normalization\ constant}. \tag{3.24}$$

*Note: What is the important role of the normalization constant? Analyze the cases where $P(B) \to 0$ and $P(B) \to 1$. The annotated probabilities are displayed in (Fig. 3.20):*

$$P(A|B) = \frac{P(B|A)P(A)}{P(B|A)P(A) + P(B|A^c)P(A^c)}. \tag{3.25}$$

FIGURE 3.20: Annotated components of the Bayes formula (eq. 3.23)

---

### SOL-38 ✏ CH.SOL- 3.9.

*Given $X$ as a discrete randomly distributed variable and given $\gamma$ as the parameter of interest, the likelihood and the log-likelihood of $X$ given $\gamma$ follows respectively:*

$$\mathcal{L}_\gamma(X = x) = p(X = x|\gamma) \tag{3.26}$$

$$\ell_\gamma(X = x) = \ln\left(p(X = x|\gamma)\right) \tag{3.27}$$

*The term likelihood can be intuitively understood from this definition; it deduces how likely is to obtain a value $x$ when a prior information is given regarding its distribution, namely the parameter $\gamma$. For example, let us consider a biased coin toss with $p_h = \gamma$. Then:*

$$\mathcal{L}_\gamma(X = "h") = p(X = "h"|\gamma) = \gamma. \tag{3.28}$$

$$\ell_\gamma(X = "h") = \ln\left(p(X = "h"|\gamma)\right) = \ln\left(\gamma\right). \tag{3.29}$$

*Note: The likelihood function may also follow continuous distributions such as the normal distribution. In the latter, it is recommended and often obligatory to employ the log-likelihood. Why? We encourage the reader to modify the above to the continuous case of normal distribution and derive the answer.* ■

**SOL-39**  **CH.SOL- 3.10.**

*The continuous prior distribution, $f(\Gamma = \gamma)$ represents what is known about the probability of the value $\gamma$ before the experiment has commenced. It is termed as being **subjective**, and therefore may vary considerably between researchers. By proceeding the previous example, $f(\Gamma = 0.8)$ holds the probability of randomly flipping a coin that yields "heads" with chance of 80% of times.* ■

**SOL-40**  **CH.SOL- 3.11.**

*The essence of Bayesian analysis is to draw inference of unknown quantities or quantiles from the posterior distribution $p(\Gamma = \gamma | X = x)$, which is traditionally derived from prior beliefs and data information. Bayesian statistical conclusions about chances to obtain the parameter $\Gamma = \gamma$ or unobserved values of random variable $X = x$, are made in terms of* prob-ability *statements. These probability statements are conditional on the observed values of $X$, which is denoted as $p(\Gamma = \gamma | X = x)$, called posterior distributions of parameter $\gamma$. Bayesian analysis is a practical method for making inferences from data and prior beliefs using probability models for quantities we observe and for quantities which we wish to learn. Bayes rule provides a relationship of this form:*

$$posterior \ \propto p(\mathbf{x}|\boldsymbol{\gamma})p(\boldsymbol{\gamma}) \propto \ data \ given \ prior \times \ chance \ of \ prior. \qquad (3.30)$$

■

**SOL-41**  **CH.SOL- 3.12.**

*The posterior density summarizes what is known about the parameter of interest $\gamma$ after the data **is observed**. In Bayesian statistics, the posterior density $p(\Gamma = \gamma | X = x)$ becomes the **prior** for this next experiment. This is part of the well-known Bayesian updating mechanism wherein we update our knowledge to reflect the actual distribution of data that we observed. To summarize, from the perspective of Bayes Theorem, we update the **prior distribution** to a **posterior distribution** after seeing the data.* ■

**SOL-42**  **CH.SOL- 3.13.**

*Two events $A$ and $B$ are statistically independent if (and only if):*

$$P(A \cap B) = P(A)P(B). \tag{3.31}$$

*Note: Use conditional probability and rationalize this outcome. How does this property become extremely useful in practical researches that consider likelihood of normally distributed features?* ◼

### 3.3.3 Bayes Rule

**SOL-43**  **CH.SOL- 3.14.**

*Let $\gamma$ stand for the number of half-integer spin states, and given the prior knowledge that both states are equally probable:*

$$P(\gamma = 2 | \gamma \geq 1) \tag{3.32}$$

$$= \frac{P(\gamma = 2, \gamma \geq 1)}{P(\gamma \geq 1)} \tag{3.33}$$

$$= \frac{P(\gamma = 2)}{1 - P(\gamma = 0)} = \frac{1/4}{1 - 1/4} = \frac{1}{3} \tag{3.34}$$

*Note: Under what statistical property do the above relations hold?* ◼

**SOL-44**  **CH.SOL- 3.15.**

*Let event A indicate present hereditary-disease and let event B to hold a positive test result. The calculated probabilities are presented in Table 3.1. We were asked to find the probability of a test indicating that hereditary-disease is present, namely $P(B)$. According to the law of total probability:*

$$P(B) = P(B|A) * P(A) + P(B|\overline{A}) * P(\overline{A})$$
$$= [0.95 * 0.01] + [0.05 * 0.99] = 0.059 \tag{3.35}$$

*Note: In terms of performance evaluation, $P(B|A)$ is often referred to as the probability of*

| PROBABILITY | EXPLANATION |
|---|---|
| P(A)= 0.01 | The probability of hereditary-disease. |
| P($\overline{A}$)=1-0.01=.99 | The probability of no hereditary-disease. |
| P($\overline{B}$ \| $\overline{A}$)=0.95 | The probability that the test will yield a negative result [B̃] if hereditary-disease is NOT present [Ã]. |
| P(B \| $\overline{B}$)=1-0.95=.05 | The probability that the test will yield a positive result [B] if hereditary-disease is NOT present [Ã] (probability of false alarm). |
| P(B \| A)=0.95 | The probability that the test will yield a positive result [B] if hereditary-disease is present [A] (probability of detection). |
| P($\overline{B}$ \| A)=1-0.95=.05 | The probability that the test will yield a negative result [B̃] if hereditary-disease is present [A]. |

TABLE 3.1: Probability values of hereditary-disease detection.

*detection and $P(B|\overline{A})$ is considered the probability of false alarm. Notice that these measures do not, neither logically nor mathematically, combine to probability of 1.*  ∎

**SOL-45**  **CH.SOL- 3.16.**
  *We first enumerate the probabilities one by one:*

$$P(Dercum|female) = 0.05, \tag{3.36}$$

$$P(Dercum|male) = 0.0025, \tag{3.37}$$

$$P(male) = P(female) = 0.5. \tag{3.38}$$

*We are asked to find $P(female|Dercum)$. Using Bayes Rule:*

$$P(female|Dercum) = \frac{P(Dercum|female)P(female)}{P(Dercum)}. \tag{3.39}$$

*However we are missing the term $P(Dercum)$. To find it, we apply the Law of Total Probability:*

$$P(Dercum) = P(Dercum|female)P(female)$$
$$+P(Dercum|male)P(male)$$
$$=$$
$$0.05 \cdot 0.5 + 0.0025 \cdot 0.5 = 0.02625.$$

*And finally, returning to eq.*  *(3.39):*

$$P(female|Dercum) = \frac{0.05 \cdot 0.5}{0.02625} \approx 0.9524 \qquad (3.40)$$

*Note: How could this result be reached with one mathematical equation?* ■

---

**SOL-46** ✏ **CH.SOL- 3.17.**

*In order to solve this problem, we introduce the following events:*

1. *$AI$: the AI predicts that the state of the stock option is 1.*

2. *$State1$: the state of the stock option is 1.*

3. *$State0$: the state of the stock option is 0.*

*A direct application of Bayes formulae yields:*

$$P(State1|AI) = \qquad (3.41)$$
$$\frac{P(AI|State1)P(State1)}{P(AI|State1)P(State1)+P(AI|State0)P(State0)} \qquad (3.42)$$
$$= \frac{0.85 \cdot 2/3}{0.85 \cdot 2/3 + 0.15 \cdot 1/3} \approx 0.9189.$$

■

---

**SOL-47** ✏ **CH.SOL- 3.18.** *In order to solve this problem, we introduce the following events:*

1. *$H$: a human.*

2. $M$: a monkey.

3. $C$: a correct prediction.

By employing Bayes theorem and the Law of Total probability:

$$
\begin{aligned}
P(H|C) &= \frac{P(H \cap C)}{P(C)} \\
&= \frac{P(C|H)P(H)}{P(C|H)P(H) + P(C|M)P(M)} \\
&= \frac{\frac{1}{20} \cdot \frac{1}{2}}{\frac{1}{20} \cdot \frac{1}{2} + \frac{1}{15} \cdot \frac{1}{2}} \\
&\approx 0.42.
\end{aligned}
\tag{3.43}
$$

Note: If something seems off in this outcome, do not worry - it is a positive sign for understanding of conditional probability. ■

**SOL-48**  **CH.SOL- 3.19.**

In order to solve this problem, we introduce the following events:

1. $RUS$: a Russian sleeper agent is speaking.

2. $AM$: an American is speaking.

3. $L$: the TTS system generates an "l".

We are asked to find the value of $P(RUS|L)$. Using Bayes Theorem we can write:

$$
P(RUS|L) = \frac{P(L|RUS)P(RUS)}{P(L)}.
\tag{3.44}
$$

We were told that the Russians consist 1/5 of the attendees at the gathering, therefore:

$$
P(RUS) = \frac{1}{5}.
\tag{3.45}
$$

*Additionally, because "v-o-k-s-a-l" has a single l out of a total of six letters:*

$$P(L|RUS) = \frac{1}{6}.$$  (3.46)

*Additionally, because "V-a-u-x-h-a-l-l" has two l's out of a total of eight letters:*

$$P(L|AM) = \frac{2}{8}.$$  (3.47)

*An application of the Law of Total Probability yields:*

$$P(L) = P(AM)P(L|AM) + P(RUS)P(L|RUS)$$
$$= \left(\frac{4}{5}\right)\left(\frac{2}{8}\right) + \left(\frac{1}{5}\right)\left(\frac{1}{6}\right) = \frac{7}{30}.$$  (3.48)

*Using Bayes Theorem we can write:*

$$P(RUS|L) = \frac{\frac{1}{5}\left(\frac{1}{6}\right)}{\frac{7}{30}} = \frac{1}{7}.$$  (3.49)

*Note: What is the letter by which the algorithm is most likely to discover a Russian sleeper agent?* ■

---

**SOL-49**  **CH.SOL- 3.20.**

*We are given that:*
$P(X \text{ is erroneously received as a } Z) = 1/7.$ *Using Bayes Theorem we can write:*

$$P(Z \text{ } trans|Z \text{ } received) =$$
$$= \frac{P(Z \text{ } received|Z \text{ } trans)P(Z \text{ } trans)}{P(Z \text{ } received)}.$$  (3.50)

*An application of the Law of Total Probability yields:*

$$P(Z\ received) =$$
$$P(Z\ received|Z\ trans)P(Z\ trans)$$
$$+P(Z\ received|X\ trans)P(X\ trans)$$
$$= \frac{6}{7} \cdot \frac{7}{9} + \frac{1}{7} \cdot \frac{2}{9}$$
$$= \frac{44}{63}.$$

*So, using Bayes Rule, we have that*

$$P(Z\ trans|Z\ received)$$
$$= \frac{P(Z\ received|Z\ trans)P(Z\ trans)}{P(Z\ received)}$$
$$= \frac{\frac{6}{7}\frac{7}{9}}{\frac{44}{63}}$$ \hfill (3.51)
$$= \frac{44}{63} = 0.95.$$

### 3.3.4   Maximum Likelihood Estimation

**SOL-50**  **CH.SOL- 3.21.**

*For the set of i.i.d samples $X_1, \cdots, X_n$, the* **likelihood** **function** *is the product of the probability functions:*

$$L(p) = p(X_1 = x_1; p)p(X_2 = x_2; p) \cdots p(X_n = x_n; p)$$
$$= \prod_{i=1}^{n} \binom{n}{x_i} p^{x_i}(1-p)^{n-x_i}.$$ \hfill (3.52)

*Note: What is the distribution of $X^n$ when $X$ is a Bernoulli distributed random variable?*

## SOL-51  CH.SOL- 3.22.

*The* maximum likelihood estimator (MLE) *of $p$ is the value of all possible $p$ values that maximizes $L(p)$. Namely, the $p$ value that renders the set of measurements $X_1, \cdots, X_n$ as the* most likely. *Formally:*

$$\hat{p} = arg\, max_{0 \leq p \leq 1} L(p) \tag{3.53}$$

*Note: The curious student is highly encouraged to derive $\hat{p}$ from $L(p)$. Notice that $L(p)$ can be extremely simplified.* ∎

## SOL-52  CH.SOL- 3.23.

*The* log-likelihood *is the logarithm of the* likelihood *function. Intuitively, maximizing the likelihood function $L(\gamma)$ is equivalent to maximizing $\ln L(\gamma)$ in terms of finding the MLE $\hat{\gamma}$, since $\ln$ is a monotonically increasing function. Often, we maximize $\ln(f(\gamma))$ instead of the $f(\gamma)$. A common example is when $L(\gamma)$ is comprised of normally distribution random variables.*

*Formally, if $X_1, \cdots, X_n$ are i.i.d, each with probability mass function (PMF) of $f_{X_i}(x_i \mid \gamma)$, then*

$$f(\gamma) = \prod_{i=1}^{n} f_{X_i}(x_i \mid \gamma), \tag{3.54}$$

$$\ln(f(\gamma)) = \sum_{i=1}^{n} \ln f_{X_i}(x_i \mid \gamma). \tag{3.55}$$

∎

## SOL-53  CH.SOL- 3.24.

*The general procedure for finding the MLE, given that the likelihood function is differentiable, is as follows:*

1. *Start by differentiating the log-likelihood function $\ln(L(\gamma))$ with respect to a parameter of interest $\gamma$.*

2. *Equate the result to zero.*

3. *Solve the equation to find $\hat{\gamma}$ that holds:*

$$\frac{\partial \ln L(\hat{\gamma} \mid x_1, \cdots x_n)}{\partial \gamma} = 0 \tag{3.56}$$

4. *Compute the second derivative to verify that you indeed have a maximum rather than a minimum.*

■

**SOL-54**  **CH.SOL- 3.25.**

   *The first derivative of the log-likelihood function is commonly known as the **Fisher score function**, and is defined as:*

$$\mathbf{u}(\gamma) = \frac{\partial \ln L(\gamma \mid x_1, \cdots x_n)}{\partial \gamma} \tag{3.57}$$

■

**SOL-55**  **CH.SOL- 3.26.**

   ***Fisher information**, is the term used to describe the expected value of the second derivatives (the curvature) of the log-likelihood function, and is defined by:*

$$\mathbf{I}(\gamma) = -E\left[\frac{\partial^2 \ln L(\gamma \mid x_1, \cdots x_n)}{\partial \gamma^2}\right] \tag{3.58}$$

■

### 3.3.5   Fisher Information

**SOL-56**  **CH.SOL- 3.27.**

1. *Given $L(\gamma)$:*

$$\ln L(\gamma) = \ln \left( \; ny \; \right) + y * \ln(\gamma) + (n - y)\ln(1 - \gamma). \tag{3.59}$$

2. *To find the gradient, we differentiate once:*

$$g(\gamma) = y\gamma^{-1} - (n-y)(1-\gamma)^{-1} = \\ (\gamma(1-\gamma))^{-1}y - n(1-\gamma)^{-1}. \tag{3.60}$$

3. *The Hessian is generated by differentiating $g(\gamma)$:*

$$H(\gamma) = -y\gamma^{-2} - (n-y)(1-\gamma)^{-2} \tag{3.61}$$

4. *The Fisher information is calculated as follows:*

$$I(\gamma) = -E(H(\gamma)) = \frac{n}{\gamma(1-\gamma)}, \tag{3.62}$$

*since:*

$$E(y|\gamma, n) = n * \gamma \tag{3.63}$$

5. *Equating the gradient to zero and solving for our parameter $\gamma$, we get:*

$$\hat{\gamma} = \frac{y}{n} \tag{3.64}$$

*In our case this equates to: $300/10000 = 0.03$. Regarding the error, there is a close relationship between the variance of $\gamma$ and the Fisher information, as the former is the inverse of the latter:*

$$\text{var}(\gamma) = [I(\gamma)]^{-1} \\ V(\gamma) = \frac{\gamma(1-\gamma)}{n} \tag{3.65}$$

*Plugging the numbers from our question:*

$$\hat{V}(\hat{\gamma}) = \frac{0.03(1-0.03)}{10000} = 2.9 \times 10^{-7}. \tag{3.66}$$

*Statistically, the standard error that we are asked to find is the square root of eq. 3.66 which equals $5.3 \times 10^{-4}$. Note: What desired property is revealed in this experiment? At was cost could we ensure a low standard error?*

■

**SOL-57**  **CH.SOL- 3.28.**

*The Fisher Information for the distributions is as follows:*

1. *Bernoulli:*

$$\Phi(x|\gamma) = x \log \gamma + (1-x) \log(1-\gamma), \qquad (3.67)$$

$$\Phi'(x|\gamma) = \frac{x}{\gamma} - \frac{1-x}{1-\gamma}, \qquad (3.68)$$

$$\Phi''(x|\gamma) = -\frac{x}{\gamma^2} - \frac{1-x}{(1-\gamma)^2}, \qquad (3.69)$$

$$I(\gamma) = -\mathrm{E}_\gamma\left[\frac{X(1-\gamma)^2 + (1-X)\gamma^2}{\gamma^2(1-\gamma)^2}\right] = \frac{1}{\gamma(1-\gamma)}. \qquad (3.70)$$

2. *Poisson:*

$$
\begin{aligned}
\lambda(x|\theta) &= x \log \theta - \log x! - \theta, \\
\lambda'(x|\theta) &= \frac{x-\theta}{\theta}, \\
\lambda''(x|\theta) &= -\frac{x}{\theta^2}, \\
I(\theta) &= -\mathrm{E}_\theta\left[\frac{(X-\theta)^2}{\theta^2}\right] = \frac{1}{\theta}.
\end{aligned}
\qquad (3.71)
$$

■

**SOL-58**  **CH.SOL- 3.29.**

1. *True.*

2. *True.*

■

### 3.3.6  Posterior & prior predictive distributions

**SOL-59**  **CH.SOL- 3.30.**

1. *Given a sample of the form $\underline{x} = (x_1, \cdots, x_n)$ drawn from a density $p(\theta; \underline{x})$ and $\theta$ is randomly generated according to a prior density of $p(\theta)$. Then the posterior density is defined by:*

$$p(\theta|\underline{x}) = \frac{p(\theta; \underline{x})p(\theta)}{p(\underline{x})}. \tag{3.72}$$

2. *The prior predictive density is:*

$$p(\underline{x}) = \int_{\theta \in \Theta} p(\theta; \underline{x})p(\theta)d\theta \tag{3.73}$$

■

**SOL-60**  **CH.SOL- 3.31.**

1. *The posterior $p(\theta|y) \propto p(y|\theta)p(\theta)$ is:*

$$\begin{cases} \binom{5}{y}(1/2)^y(1/2)^{5-y}0.25, & \theta = 1/2 \\ \binom{5}{y}(1/6)^y(5/6)^{5-y}0.5, & \theta = 1/6 \\ \binom{5}{y}(1/4)^y(3/4)^{5-y}0.25, & \theta = 1/4 \\ 0, & otherwise \end{cases}$$

2. *The prior predictive distribution $p(y)$:*

$$\binom{5}{y}((1/2)^y(1/2)^{5-y}0.25 \tag{3.74}$$

$+$

$$(1/6)^y(5/6)^{5-y}0.5 + (1/4)^y(3/4)^{5-y}0.25). \tag{3.75}$$

■

### 3.3.7   Conjugate priors

**SOL-61**  **CH.SOL- 3.32.**

1. *A class $\mathcal{F}$ of prior distributions is said to form a conjugate family if the posterior density is in $\mathcal{F}$ for all each sample, whenever the prior density is in $\mathcal{F}$.*

2. *Often we would like a prior that favours no particular values of the parameter over others. Bayesian analysis requires prior information, however sometimes there is no particularly useful information before data is collected. In these situations, priors with "no information" are expected. Such priors are called* non-informative priors.

■

**SOL-62**  **CH.SOL- 3.33.**

*If $x \sim \mathrm{B}(n, \gamma)$ so*
   $p(x|\gamma) \propto \gamma^x(1-\gamma)^{n-x}$
*and the prior for $\gamma$ is $\mathcal{B}(\alpha, \beta)$ so*
   $p(\gamma) \propto \gamma^{\alpha-1}(1-\gamma)^{\beta-1}$
*then the posterior is*
   $\gamma|x \sim \mathcal{B}(\alpha + x, \beta + n - x)$
*It is immediately clear the family of beta distributions is conjugate to a binomial likelihood.*

■

### 3.3.8   Bayesian Deep Learning

**SOL-63**  **CH.SOL- 3.34.**

1. *The hidden neuron is distributed according to:*
   $X \sim \text{binomial}(n, \gamma)$ *random variable and fires with a probability of* $\gamma$. *There are 100 neurons and only 20 are* **fired**.

$$P(x = 20|\theta) = \left( \begin{array}{c} 100 \\ 20 \end{array} \right) \theta^{20}(1 - \theta)^{80} \tag{3.76}$$

2. *The hidden neuron is distributed according to:*
   $X \, uniform(0, \gamma)$ *random variable and fires with a probability of* $\gamma$.

   *The uniform distribution is, of course, a very simple case:*

$$f(x; a, b) = \frac{1}{b - a} \quad for \quad a \leq x \leq b \tag{3.77}$$

   *Therefore:*

$$f(x|\gamma) = \left\{ \begin{array}{ll} 0 & \textit{if } \gamma < x \textit{ or } x < 0 \\ 1/\gamma & \textit{if } 0 \leq x \leq \theta \end{array} \right. \tag{3.78}$$

■

**SOL-64**  **CH.SOL- 3.35.**

*The provided distribution is from the exponential family. Therefore, a single neuron becomes inactive with a probability of:*

$$p = P(X < 20) = \int_0^{20} e^{-x} \, dx = 1 - e^{-20}. \tag{3.79}$$

*The OnOffLayer is off only if at least 150 out of 200 neurons are off. Therefore, this may be represented as a Binomial distribution and the probability for the layer to be off is:*

$$V = \sum_{n \geq 150} \left( \begin{array}{c} 200 \\ n \end{array} \right) \tilde{p}^n (1 - \tilde{p})^{200-n} \tag{3.80}$$

*Hence, the probability of the layer being active for at least 20 seconds is 1 minus this value:*

$$[1 - V]. \tag{3.81}$$

■

**SOL-65**  **CH.SOL- 3.36.**

*The observed data, e.g the dropped neurons are distributed according to:*

$$(x_1, \ldots, x_n)|\theta \overset{iid}{\sim} \text{Bern}(\theta) \tag{3.82}$$

*Denoting $s$ and $f$ as success and failure respectively, we know that the likelihood is:*

$$p(x_1, \ldots, x_n|\theta) = \theta^s (1-\theta)^f \tag{3.83}$$

*With the following parameters $\alpha = \beta = 1$ the beta distribution acts like Uniform prior:*

$$\theta \sim \text{Beta}(\alpha, \beta), \text{ given } \alpha = \beta = 1 \tag{3.84}$$

*Hence, the **prior** density is:*

$$p(\theta) = \frac{1}{B(\alpha, \beta)} \theta^{\alpha-1} (1-\theta)^{\beta-1} \tag{3.85}$$

*Therefore the **posterior** is:*

$$\begin{aligned} p(\theta|x_1, \ldots, x_n) &\propto p(x_1, \ldots, x_n|\theta)\, p(\theta) \\ &\propto \theta^S (1-\theta)^f \theta^{\alpha-1} (1-\theta)^{\beta-1} \\ &= \theta^{\alpha+s-1} (1-\theta)^{\beta+f-1} \end{aligned} \tag{3.86}$$

■

**SOL-66**  **CH.SOL- 3.37.**

*Neurons are dropped whenever their value (or the equivalent quantum term- speed) reach*

*the most likely value:*

$$n(v)dv = \frac{4\pi N}{V}\left(\frac{m}{2\pi kT}\right)^{3/2} v^2 e^{-\frac{mv^2}{2kT}}\, dv \qquad (3.87)$$

*From calculus, we know that in order to maximize a function, we have to equate its first derivative to zero:*

$$\frac{d}{dv}n(v) = 0 \qquad (3.88)$$

*The constants can be taken out as follows:*

$$\frac{d}{dv}v^2 e^{-\frac{mv^2}{2kT}} = 0 \qquad (3.89)$$

*Applying the chain rule from calculus:*

$$2ve^{-\frac{mv^2}{2kT}} + v^2\left(-\frac{m}{2kT}2v\right)e^{-\frac{mv^2}{2kT}} = 0 \qquad (3.90)$$

*We notice that several terms cancel out:*

$$v^2\frac{m}{2kT} = 1 \qquad (3.91)$$

*Now the quadratic equation can be solved yielding:*

$$v_{most\_probable} = \sqrt{\frac{2kT}{m}} \qquad (3.92)$$

*Therefore, this is the most probable value at which the dropout layer will fire.*

■

# PART III
# HIGH SCHOOL

# CHAPTER

*A basic idea in information theory is that information can be treated very much like a physical quantity, such as mass or energy.*

— Claude Shannon, 1985

## Contents

## 4.1 Introduction

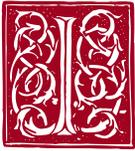 NDUCTIVE inference, is the problem of reasoning under conditions of in-complete information, or **uncertainty**. According to Shannon's theory [2], information and uncertainty are two sides of the same coin: the more uncertainty there is, the more information we gain by removing the uncertainty. Entropy plays central roles in many scientific realms ranging from physics and statistics to data science and economics. A basic problem in information theory is encoding large quantities of information [2].

Shannon's discovery of the fundamental laws of data compression and transmission marked the birth of information theory. In his fundamental paper of 1948, "*A Mathematical Theory of Communication*" [4], Shannon proposed a measure of the uncertainty associated with a random memory-less source, called *Entropy*.

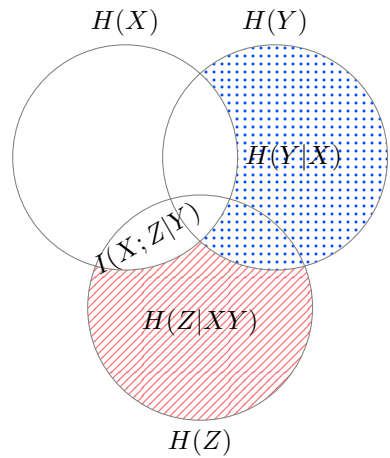

FIGURE 4.1: Mutual information

Entropy first emerged in thermodynamics in the $18^{th}$ century by Carnot, [1] in his pioneering work on steam entitled "*Reflection on the Motive Power of Fire*" (Fig. 4.2). Subsequently it appeared in statistical mechanics where it was viewed as a measure of *disorder*. However, it was Boltzmann (4.30) who found the connection between entropy and probability, and the notion of information as used by Shannon is a generalization of the notion of entropy. Shannon's entropy shares some instinct with Boltzmann's entropy, and likewise the mathematics developed in information theory is highly relevant in statistical mechanics.

REFLECTIONS

ON THE

·MOTIVE POWER OF HEAT.

*FROM THE ORIGINAL FRENCH OF*
N.-L.-S. ÇARNOT,
*Graduate of the Polytechnic School.*

ACCOMPANIED BY

AN ACCOUNT OF CARNOT'S THEORY.
BY SIR WILLIAM THOMSON (LORD KELVIN).

EDITED BY

R. H. THURSTON, M.A., LL.D., Dr.Eng'g ;
*Director of Sibley College, Cornell University ;*
*" Officier de l'Instruction Publique de France,"*
*etc., etc., etc.*

FIGURE 4.2: Reflection on the motive power of fire.

The **majority** of candidates I interview fail to come up with an answer to the following question: *what is the entropy of tossing a non-biased coin?* Surprisingly, even after I explicitly provide them with Shannon's formulae for calculating entropy (4.4), many are still unable to calculate simple logarithms. The purpose of this chapter is to present the aspiring data scientist with some of the most significant notions of entropy and to elucidate its relationship to probability. Therefore, it is primarily focused on basic quantities in information theory such as entropy, cross-entropy, conditional entropy, mutual information and Kullback-Leibler divergence, also known as relative entropy. It does not however, discuss more advanced topics such as the concept of 'active information' introduced by Bohm and Hiley [3].

## 4.2   Problems

### 4.2.1   Logarithms in Information Theory

It is important to note that all numerical calculations in this chapter use the binary logarithm $\log_2$. This specific logarithm produces units of bits, the commonly used units of information in the field on information theory.

**PRB-67 ❷ CH.PRB- 4.1.**

*Run the following Python code (4.3) in a Python interpreter. What are the results?*

```python
import math
import numpy
print (math.log(1.0/0.98)) # Natural log (ln)
print (numpy.log(1.0/0.02)) # Natural log (ln)

print (math.log10(1.0/0.98)) # Common log (base 10)
print (numpy.log10(1.0/0.02)) # Common log (base 10)

print (math.log2(1.0/0.98)) # Binary log (base 2)
print (numpy.log2(1.0/0.02)) # Binary log (base 2)
```

FIGURE 4.3: Natural (ln), binary ($\log_2$) and common ($\log_{10}$) logarithms.

**PRB-68 ❷ CH.PRB- 4.2.**

*The three basic laws of logarithms:*

1. *First law*

$$\log A + \log B = \log AB. \tag{4.1}$$

   *Compute the following expression:*

$$\log_{10} 3 + \log_{10} 4.$$

2. *Second law*

$$\log A^n = n \log A. \tag{4.2}$$

*Compute the following expression:*

$$\log_2 4^6.$$

3. ***Third law***

$$\log A - \log B = \log \frac{A}{B}. \tag{4.3}$$

*Therefore, subtracting $\log B$ from $\log A$ results in $\log \frac{A}{B}$.*

*Compute the following expression:*

$$\log_e 15 - \log_e 3.$$

### 4.2.2   Shannon's Entropy

**PRB-69 ❷ CH.PRB- 4.3.**
*Write Shannon's famous general formulae for **uncertainty**.*

**PRB-70 ❷ CH.PRB- 4.4.**
*Choose exactly **one, and only one** answer.*

1. *For an event which is **certain to happen**, what is the entropy?*

    *(a) 1.0*

    *(b) 0.0*

    *(c) The entropy is undefined*

    *(d) −1*

    *(e) 0.5*

    *(f) $\log_2(N)$, N being the number of possible events*

2. For $N$ **equiprobable events**, what is the entropy?

   (a) 1.0

   (b) 0.0

   (c) The entropy is undefined

   (d) $-1$

   (e) 0.5

   (f) $\log_2(N)$

---

## PRB-71 ❷ CH.PRB- 4.5.

   Shannon found that entropy was the only function satisfying **three natural properties**. Enumerate these properties.

---

## PRB-72 ❷ CH.PRB- 4.6.

   In information theory, minus the logarithm of the probability of a symbol (essentially the number of bits required to represent it efficiently in a binary code) is defined to be the **information** conveyed by transmitting that symbol. In this context, the entropy can be interpreted as the expected information conveyed by transmitting a single symbol from an alphabet in which the symbols occur with the probabilities $\pi_k$.
   **Mark the correct answer**: Information is a/an **[decrease|increase]** in uncertainty.

---

## PRB-73 ❷ CH.PRB- 4.7.

   Claud Shannon's paper "A mathematical theory of communication" [4], marked the birth of information theory. Published in 1948, it has become since the Magna Carta of the information age. Describe in your own words what is meant by the term **Shannon bit**.

---

## PRB-74 ❷ CH.PRB- 4.8.

   With respect to the notion of **surprise** in the context of information theory:

   1. Define what it actually meant by being **surprised**.

2. *Describe how it is related to the likelihood of an event happening.*

3. **True or False:** *The less likely the occurrence of an event, the smaller information it conveys.*

---

**PRB-75 ❷ CH.PRB- 4.9.**

   *Assume a source of signals that transmits a given message a with probability $P_a$. Assume further that the message is encoded into an ordered series of ones and zeros (a bit string) and that a receiver has a decoder that converts the bit string back into its respective message. Shannon devised a formulae that describes the **size** that the mean length of the bit string can be **compressed to**. Write the formulae.*

---

**PRB-76 ❷ CH.PRB- 4.10.**

   *Answer the following questions:*

1. *Assume a source that provides a constant stream of $N$ **equally likely** symbols $\{x_1, x_2, \ldots, x_N\}$. What does Shannon's formulae (4.4) reduce to in this particular case?*

2. *Assume that each equiprobable pixel in a monochrome image that is fed to a DL classification pipeline, can have values ranging from 0 to 255. Find the entropy in **bits**.*

---

**PRB-77 ❷ CH.PRB- 4.11.**

   *Given Shannon's famous general formulae for uncertainty (4.4):*

$$H = -\sum_{a=1}^{N} P_a \log_2 P_a \quad \text{(bits per symbol)}.$$   (4.4)

1. *Plot a graph of the curve of probability vs. uncertainty.*

2. **Complete the sentence:** *The curve is [**symmetrical/asymmetrical**].*

3. **Complete the sentence:** *The curve rises to a [**minimum/maximum**] when the two symbols are equally likely ($P_a = 0.5$).*

---

### PRB-78 ❷ CH.PRB- 4.12.

Assume we are provided with biased coin for which the event 'heads' is assigned probability $p$, and 'tails' - a probability of $1 - p$. Using (4.4), the respective entropy is:

$$H(p) = -p \log p - (1-p) \log (1-p).$$ (4.5)

Therefore, $H \geq 0$ and the maximum possible uncertainty is attained when $p = 1/2$, is $H_{\max} = \log_2 2$.

Given the above formulation, describe a helpful property of the entropy that **follows from the concavity of the logarithmic function.**

---

### PRB-79 ❷ CH.PRB- 4.13.

**True or False:** *Given random variables $X, Y$ and $Z$ where $Y = X + Z$ then:*

$$H(X,Y) = H(X,Z).$$ (4.6)

---

### PRB-80 ❷ CH.PRB- 4.14.

What is the entropy of a **biased coin?** Suppose a coin is biased such that the probability of 'heads' is $p(x_h) = 0.98$.

1. **Complete the sentence:** *We can predict 'heads' for each flip with an accuracy of [__-_]%.*

2. **Complete the sentence:** *If the result of the coin toss is 'heads', the amount of Shannon information gained is [____] bits.*

3. **Complete the sentence:** *If the result of the coin toss is 'tails', the amount of Shannon information gained is [____] bits.*

4. **Complete the sentence:** *It is always true that the more information is associated with an outcome, the [**more/less**] surprising it is.*

5. *Provided that the ratio of tosses resulting in 'heads' is $p(x_h)$, and the ratio of tosses resulting in 'tails' is $p(x_t)$, and also provided that $p(x_h) + p(x_t) = 1$, what is formulae for the **average surprise**?*

6. *What is the value of the **average surprise** in bits?*

### 4.2.3    Kullback-Leibler Divergence (KLD)

**PRB-81 ❷ CH.PRB- 4.15.**
   *Write the formulae for the Kullback-Leibler divergence between two discrete probability density functions $P$ and $Q$.*

**PRB-82 ❷ CH.PRB- 4.16.**
   *Describe one intuitive interpretation of the KL-divergence with respect to bits.*

**PRB-83 ❷ CH.PRB- 4.17.**

1. ***True or False:*** *The KL-divergence is not a symmetric measure of similarity, i.e.:*

$$\mathrm{D}_{KL}(P\|Q) \neq \mathrm{D}_{KL}(Q\|P).$$

2. ***True or False:*** *The KL-divergence satisfies the triangle inequality.*

3. ***True or False:*** *The KL-divergence is not a distance metric.*

4. ***True or False:*** *In information theory, KLD is regarded as a measure of the information gained when probability distribution $Q$ is used to approximate a true probability distribution $P$.*

5. ***True or False:*** *The units of KL-divergence are units of information.*

6. ***True or False:*** *The KLD is always non-negative, namely:*

$$\mathrm{D}_{KL}(P\|Q) \geq 0.$$

.

7. **True or False:** *In a decision tree, **high** information gain indicates that adding a split to the decision tree results in a **less** accurate model.*

---

**PRB-84 ❷ CH.PRB- 4.18.**

Given two distributions $f_1$ and $f_2$ and their respective joint distribution $f$, write the formulae for the mutual information of $f_1$ and $f_2$.

---

**PRB-85 ❷ CH.PRB- 4.19.**

*The question was commented out but remained here for the consistency of the numbering system.*

---

### 4.2.4  Classification and Information Gain

---

**PRB-86 ❷ CH.PRB- 4.20.**

*There are several measures by which one can determine how to optimally split attributes in a decision tree. List the three most commonly used measures and write their formulae.*

---

**PRB-87 ❷ CH.PRB- 4.21.**

**Complete the sentence:** *In a decision tree, the attribute by which we choose to split is the one with [**minimum/maximum**] information gain.*

---

**PRB-88 ❷ CH.PRB- 4.22.**

*To study factors affecting the decision of a frog to jump (or not), a deep learning researcher from a Brazilian rain-forest, collects data pertaining to several independent binary co-variates.*

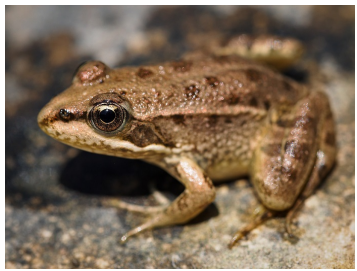

FIGURE 4.4: A Frog in its natural habitat. Photo taken by my son.

*The binary response variable **Jump** indicates whether a jump was observed. Referring to Table (4.1), each row indicates the observed values, columns denote features and rows denote labelled instances while class label (**Jump**) denotes whether the frog had jumped.*

| Observation | Green | Rain | Jump |
|:-----------:|:-----:|:----:|:----:|
| $x1$ | 1 | 0 | + |
| $x2$ | 1 | 1 | + |
| $x3$ | 1 | 0 | + |
| $x4$ | 1 | 1 | + |
| $x5$ | 1 | 0 | + |
| $x6$ | 0 | 1 | + |
| $x7$ | 0 | 0 | − |
| $x8$ | 0 | 1 | − |

TABLE 4.1: Decision trees and frogs.

*Without explicitly determining the information gain values for each of the three attributes, which attribute should be chosen as the attribute by which the decision tree should be first partitioned? e.g which attribute has the highest predictive power regarding the decision of the frog (Fig. 4.4) to jump.*

**PRB-89 ❷ CH.PRB- 4.23.**

*This question discusses the link between binary classification, information gain and decision trees. Recent research [5] suggests that Cannabis (Fig. 4.5), and Cannabinoids administration in particular may reduce the size of malignant tumours in rodents. The data (Table 9.2) comprises a training set of feature vectors with corresponding class labels which a researcher intents classifying using a decision tree.*

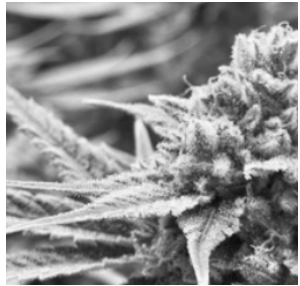

FIGURE 4.5: Cannabis

*To study factors affecting tumour shrinkage, the deep learning researcher collects data regrading two independent binary variables; $\theta_1$ (T/F) indicating whether the rodent is a female, and $\theta_2$ (T/F) indicating whether the rodent was administrated with Cannabinoids. The binary response variable, $\gamma$, indicates whether tumour shrinkage was observed (e.g. shrinkage=+, no shrinkage=-). Referring to Table (9.2), each row indicates the observed values, columns ($\theta_i$) denote features and class label ($\gamma$) denotes whether shrinkage was observed.*

| $\gamma$ | $\theta_1$ | $\theta_2$ |
|---|---|---|
| + | T | T |
| - | T | F |
| + | T | F |
| + | T | T |
| - | F | T |

TABLE 4.2: Decision trees and Cannabinoids administration

1. Describe what is meant by **information gain**.

2. Describe in your own words how does a decision tree work.

3. Using $log_2$, and the provided dataset, calculate the sample entropy $H(\gamma)$.

4. What is the information gain $IG(X_1) \equiv H(\gamma) - H(|\theta_1)$ for the provided training corpus?

---

**PRB-90 ❷ CH.PRB- 4.24.**

To study factors affecting the expansion of stars, a physicist is provided with data regrading two independent variables; $\theta_1$ (T/F) indicating whether a star is dense, and $\theta_2$ (T/F) indicating whether a star is adjacent to a black-hole. He is told that the binary response variable, $\gamma$, indicates whether expansion was observed.

e.g.:
expansion=+, no expansion=-. Referring to table (4.3), each row indicates the observed values, columns ($\theta_i$) denote features and class label ($\gamma$) denotes whether expansion was observed.

| $\gamma$ (expansion) | $\theta_1$ (dense) | $\theta_2$ (black-hole) |
|:---:|:---:|:---:|
| + | F | T |
| + | T | T |
| + | T | T |
| - | F | T |
| + | T | F |
| - | F | F |
| - | F | F |

TABLE 4.3: Decision trees and star expansion.

1. Using $log_2$ and the provided dataset, calculate the sample entropy $H(\gamma)$ (expansion) before splitting.

2. Using $log_2$ and the provided dataset, calculate the **information gain** of $H(\gamma|\theta_1)$.

3. *Using $log_2$ and the provided dataset, calculate the **information gain** of $H(\gamma|\theta_2)$.*

---

**PRB-91 ❷ CH.PRB- 4.25.**

*To study factors affecting tumour shrinkage in humans, a deep learning researcher is provided with data regrading two independent variables; $\theta_1$ (S/M/L) indicating whether the tumour is small(S), medium(M) or large(L), and $\theta_2$ (T/F) indicating whether the tumour has undergone radiation therapy. He is told that the binary response variable, $\gamma$, indicates whether tumour shrinkage was observed (e.g. shrinkage=+, no shrinkage=-).*

*Referring to table (4.4), each row indicates the observed values, columns ($\theta_i$) denote features and class label ($\gamma$) denotes whether shrinkage was observed.*

| $\gamma$ (shrinkage) | $\theta_1$ | $\theta_2$ |
|:---:|:---:|:---:|
| - | S | F |
| + | S | T |
| - | M | F |
| + | M | T |
| + | H | F |
| + | H | T |

TABLE 4.4: Decision trees and radiation therapy.

1. *Using $log_2$ and the provided dataset, calculate the sample entropy $H(\gamma)$ (shrinkage).*

2. *Using $log_2$ and the provided dataset, calculate the entropy of $H(\gamma|\theta_1)$.*

3. *Using $log_2$ and the provided dataset, calculate the entropy of $H(\gamma|\theta_2)$.*

4. ***True or false:** We should split on a specific variable that minimizes the information gain, therefore we should split on $\theta_2$ (radiation therapy).*

### 4.2.5 Mutual Information

---

**PRB-92 ❷ CH.PRB- 4.26.**

*Shannon described a communications system consisting five elements (4.6), two of which are the source $S$ and the destination $D$.*

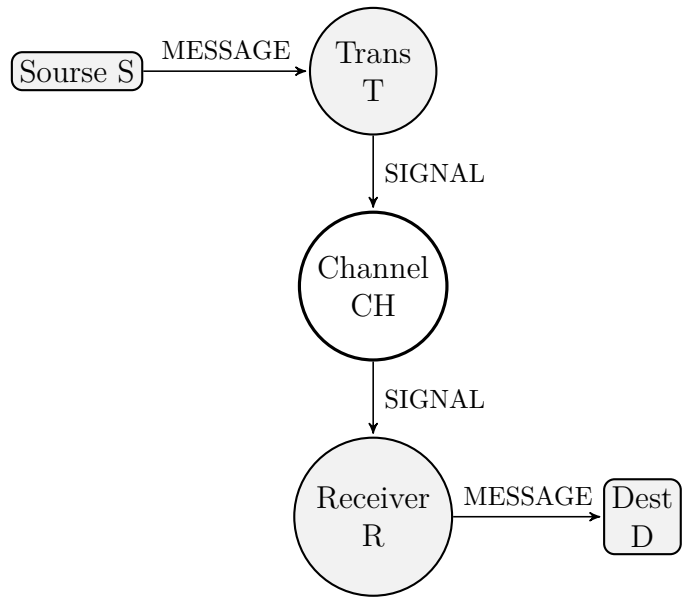

FIGURE 4.6: Shannon's five element communications system.

1. *Draw a Venn diagram depicting the relationship between the entropies of the source $H(S)$ and of the destination $H(D)$.*

2. *Annotate the part termed **equivocation**.*

3. *Annotate the part termed **noise**.*

4. *Annotate the part termed **mutual information**.*

5. *Write the formulae for **mutual information**.*

**PRB-93 ❷ CH.PRB- 4.27.**
   ***Complete the sentence:*** *The relative entropy $D(p||q)$ is the measure of (a) [___] between*

*two distributions. It can also be expressed as a measure of the (b)[___] of assuming that the distribution is q when the (c)[___] distribution is p.*

---

**PRB-94 ❷ CH.PRB- 4.28.**

**Complete the sentence:** *Mutual information is a Shannon entropy-based measure of dependence between random variables. The mutual information between $X$ and $Z$ can be understood as the (a) [___] of the (b) [___] in $X$ given $Z$:*

$$I(X; Z) := H(X) - H(X \mid Z), \tag{4.7}$$

*where $H$ is the Shannon entropy, and $H(X \mid Z)$ is the conditional entropy of $Z$ given $X$.*

---

### 4.2.6   Mechanical Statistics

Some books have a tendency of sweeping "unseen" problems under the rug. We will not do that here. This subsection may look intimidating and for a good reason; it involves equations that, unless you are a physicists, you have probably never encountered before. Nevertheless, the ability to cope with new concepts lies at the heart of every job interview.

For some of the questions, you may need these constants:

**PHYSICAL CONSTANTS**

| | | |
|---|---|---|
| **k** | Boltzmanns  constant | $1.381 \times 10^{-23}$ J K$^{-1}$ |
| **c** | Speed  of  light in vacuum | $2.998 \times 10^{8}$ m s$^{-1}$ |
| **h** | Planck's  constant | $6.626 \times 10^{-34}$ J s |

---

**PRB-95 ❷ CH.PRB- 4.29.**

*What is the expression for the Boltzmann probability distribution?*

---

**PRB-96 ❷ CH.PRB- 4.30.**

*Information theory, quantum physics and thermodynamics are closely interconnected. There are several equivalent formulations for the second law of thermodynamics. One approach to describing uncertainty stems from Boltzmanns fundamental work on* entropy *in*

*statistical mechanics. Describe what is meant by **Boltzmanns entropy.***

---

**PRB-97 ❷ CH.PRB- 4.31.**

   *From Boltzmanns perspective, what is the entropy of an octahedral dice (4.7)?*

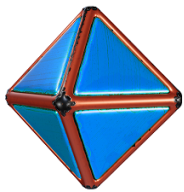

FIGURE 4.7: An octahedral dice.

---

### 4.2.7   Jensen's inequality

---

**PRB-98 ❷ CH.PRB- 4.32.**

1. *Define the term concave function.*

2. *Define the term convex function.*

3. *State Jensen's inequality and its implications.*

---

**PRB-99 ❷ CH.PRB- 4.33.**

   ***True or False:*** *Using Jensen's inequality, it is possible to show that the KL divergence is always greater or equal to zero.*

---

## 4.3   Solutions

### 4.3.1   Logarithms in Information Theory

**SOL-67**  **CH.SOL- 4.1.**

*Numerical results (4.8) are provided using Python interpreter version 3.6.*

```python
import math
import numpy
print (math.log(1.0/0.98)) # Natural log (ln)
> 0.02020270731751947
print (numpy.log(1.0/0.02)) # Natural log (ln)
> 3.912023005428146
print (math.log10(1.0/0.98)) # Common log (base 10)
> 0.008773924307505152
print (numpy.log10(1.0/0.02)) # Common log (base 10)
> 1.6989700043360187
print (math.log2(1.0/0.98)) # Binary log (base 2)
> 0.02914634565951651
print (numpy.log2(1.0/0.02)) # Binary log (base 2)
> 5.643856189774724
```

FIGURE 4.8: Logarithms in information theory.

■

**SOL-68**  **CH.SOL- 4.2.**

*The logarithm base is explicitly written in each solution.*

*1.*
$$\log_{10} 3 + \log_{10} 4 = \log_{10}(3 \times 4) = \log_{10} 12.$$

*2.*
$$\log_2 4^6 = 6\log_2 4.$$

*3.*
$$\log_e 15 - \log_e 3 = \log_e \tfrac{15}{3} = \log_e 5.$$

### 4.3.2    Shannon's Entropy

**SOL-69**  **CH.SOL- 4.3.**

*Shannons famous general formulae for uncertainty is:*

$$H = -\sum_{a=1}^{N} P_a \log_2 P_a \quad \text{(bits per symbol).}$$   (4.8)

**SOL-70**  **CH.SOL- 4.4.**

1. *No information is conveyed by an event which is a-priori known to occur for certain* ($P_a = 1$), *therefore the entropy is* 0.

2. *Equiprobable events mean that* $P_i = 1/N \ \forall i \in [1, N]$. *Therefore for* $N$ *equally-likely events, the entropy is* $\log_2(N)$.

**SOL-71**  **CH.SOL- 4.5.**

*The three properties are as follows:*

1. $H(X)$ *is always non-negative, since information cannot be lost.*

2. *The uniform distribution maximizes* $H(X)$, *since it also maximizes uncertainty.*

3. *The additivity property which relates the sum of entropies of two independent events. For instance, in thermodynamics, the total entropy of two isolated systems which co-exist in equilibrium is the sum of the entropies of each system in isolation.*

**SOL-72**  **CH.SOL- 4.6.**

*Information is an [**increase**] in uncertainty.* ■

**SOL-73**  **CH.SOL- 4.7.**

*The Shannon bit has two distinctive states; it is either 0 or 1, but never both at the same time. Shannon devised an experiment in which there is a question whose only two possible answers were **equally likely to happen**.*

*He then defined **one bit** as the amount of information gained (or alternatively, the amount of entropy removed) once an answer to the question has been learned. He then continued to state that when the a-priori probability of any one possible answer is higher than the other, the answer would have conveyed less than one bit of information.* ■

**SOL-74**  **CH.SOL- 4.8.**

*The notion of surprise is directly related to the likelihood of an event happening. Mathematically is it inversely proportional to the probability of that event.*

*Accordingly, learning that a high-probability event has taken place, for instance the sun rising, is **much less of a surprise** and gives less information than learning that a low-probability event, for instance, rain in a hot summer day, has taken place. Therefore, the less likely the occurrence of an event, the greater information it conveys.*

*In the case where an event is a-priori known to occur for certain ($P_a = 1$), then no information is conveyed by it. On the other hand, an extremely intermittent event conveys a lot of information as it **surprises** us and informs us that a very improbable state exists. Therefore, the statement in part 3 is false.*

■

**SOL-75**  **CH.SOL- 4.9.**

*This quantity $I_{Sh}$, represented in the formulae is called the **Shannon information of the source**:*

$$I_{Sh} = - \sum_a p_a \log_2 p_a. \tag{4.9}$$

*It refers to the mean length in bits, per message, into which the messages can be compressed*

*to. It is then possible for a communications channel to transmit $I_{Sh}$ bits per message with a capacity of $I_{Sh}$.*   ■

---

**SOL-76**  **CH.SOL- 4.10.**

1. *For N equiprobable events it holds that $P_i = 1/N$, $\forall i \in [1, N]$. Therefore if we substitute this into Shannon's equation we get:*

$$H_{equiprobable} = -\sum_{i=1}^{N} \frac{1}{N} \log_2 \frac{1}{N}. \tag{4.10}$$

*Since N does not depend on $i$, we can pull it out of the sum:*

$$
\begin{aligned}
H_{equiprobable} &= -\left(\frac{1}{N} \log_2 \frac{1}{N}\right) \sum_{i=1}^{N} 1 \tag{4.11} \\
&= -\left(\frac{1}{N} \log_2 \frac{1}{N}\right) N \\
&= -\log_2 \frac{1}{N} \tag{4.12} \\
&= \log_2 N.
\end{aligned}
$$

*It can be shown that for a given number of symbols (i.e., N is fixed) the uncertainty H has its largest value only when the symbols are equally probable.*

2. *The probability for each pixel to be assigned a value in the given range is:*

$$p_i = 1/256. \tag{4.13}$$

*Therefore the entropy is:*

$$H = -(256)(1/256)(-8) = 8 \text{ [bits/symbol]}. \tag{4.14}$$

■

---

**SOL-77**  **CH.SOL- 4.11.**

*Refer to Fig. 4.9 for the corresponding illustration of the graph, where information is shown as a function of $p$. It is equal to 0 for $p = 0$ and for $p = 1$. This is reasonable because for such values of $p$ the outcome is certain, so no information is gained by learning the outcome. The entropy in maximal uncertainty equals to 1 bit for $p = 0.5$. Thus, the information gain is maximal when the probabilities of two possible events are equal. Furthermore, for the entire range of probabilities between $p = 0.4$ and $p = 0.6$ the information is close to 1 bit.* ∎

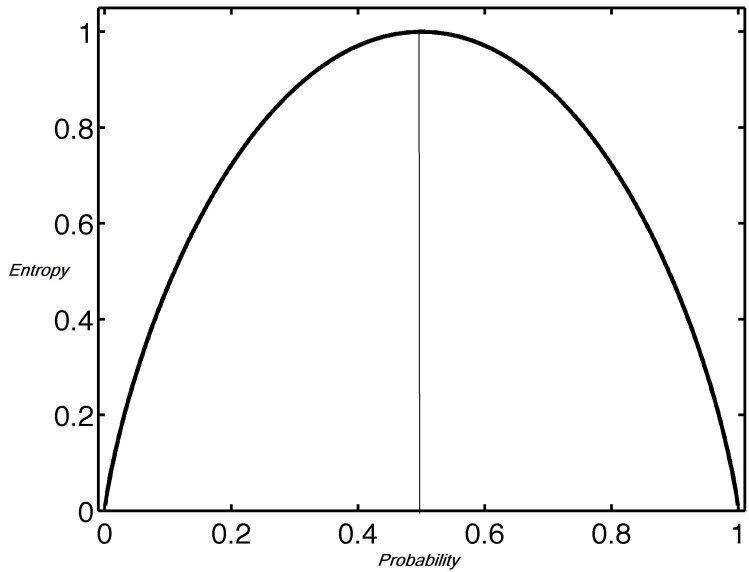

FIGURE 4.9: H vs. Probability

**SOL-78** ✏ **CH.SOL- 4.12.**

*An important set of properties of the entropy follows from the concavity of the entropy, which follows from the concavity of the logarithm. Suppose that in an experiment, we cannot decide whether the actual probability of 'heads' is $p_1$ or $p_2$. We may decide to assign probability $q$ to the first alternative and probability $1 - q$ to the second. The actual probability of 'heads' then is the mixture $qp_1 + (1 - q)p_2$. The corresponding entropies satisfy the inequality:*

$$S(qp_1 + (1-q)p_2) \geq qS(p_1) + (1-q)S(p_2), \tag{4.15}$$

*These probabilities, are equal in the extreme cases where $p_1 = p_2$, or $q = 0$, or $q = 1$.*  ■

---

**SOL-79**  **CH.SOL- 4.13.**

Given $(X, Y)$, we can determine $X$ and $Z = Y - X$. Conversely, given $(X, Z)$, we can determine $X$ and $Y = X + Z$. Hence, $H(X, Y) = H(X, Z)$ due to the existence of this bijection.  ■

---

**SOL-80**  **CH.SOL- 4.14.**

The solution and numerical calculations are provided using $\log_2$.

1. *We can predict 'heads' for each flip with an accuracy of $p(x_h) = 98$%.*

2. *According to Fig. (4.10), if the result of the coin toss is 'heads', the amount of Shannon information gained is $\log_2(1/0.98)$ [bits].*

```python
import  math
import numpy
print (math.log2(1.0/0.98)) # Binary log (base 2)
> 0.02914634565951651
print (numpy.log2(1.0/0.02)) # Binary log (base 2)
> 5.643856189774724
```

FIGURE 4.10: Shannon information gain for a biased coin toss.

3. *Likewise, if the result of the coin toss is 'tails', the amount of Shannon information gained is $\log_2(1/0.02)$ [bits].*

4. *It is always true that the **more** information is associated with an outcome, the **more** surprising it is.*

5. *The formulae for the **average surprise** is:*

$$H(x) \;=\; p(x_h) \log \frac{1}{p(x_h)} + p(x_t) \log \frac{1}{p(x_t)}. \tag{4.16}$$

6. *The value of the **average surprise** in bits is (4.11):*

$$H(x) = [0.98 \times 0.0291] + [0.02 \times 5.643] \qquad (4.17)$$
$$= 0.1414 \text{ [bits]}.$$

```
1  import autograd.numpy as np
2  def binaryEntropy (p):
3    return -p*np.log2(p) -(1-p)*np.log2(1-p)
4    print ("binaryEntropy(p) is:{}
       ↪ bits".format(binaryEntropy(0.98)))
5  > binaryEntropy(p) is:0.1414 bits
```

FIGURE 4.11: Average surprise

### 4.3.3 Kullback-Leibler Divergence

**SOL-81**  **CH.SOL- 4.15.**

*For discrete probability distributions $P$ and $Q$, the Kullback-Leibler divergence **from** $P$ **to** $Q$, the KLD is defined as:*

$$D(P \parallel Q) = \sum_x P(x) \log \frac{P(x)}{Q(x)} \qquad (4.18)$$
$$= \mathrm{E}_P \left[ \log \frac{1}{Q(x)} - \log \frac{1}{P(x)} \right]$$
$$= \underbrace{H_P(Q)}_{\text{Cross Entropy}} - \underbrace{H(P)}_{\text{Entropy}}.$$

**SOL-82**  **CH.SOL- 4.16.**

*One interpretation is the following: the KL-divergence indicates the average number of **additional bits** required for transmission of values $x \in X$ which are distributed according to $P(x)$, but we erroneously encoded them according to distribution $Q(x)$. This makes sense since you have to "pay" for additional bits to compensate for not knowing the true distribution, thus using a code that was optimized according to other distribution. This is one of the reason that the KL-divergence is also known as relative entropy. Formally, the cross entropy has an information interpretation quantifying how many bits are wasted by using the wrong code:*

$$H_P(Q) = \sum_x \underbrace{P(x)}_{Sending\ P} \overbrace{\log \frac{1}{Q(x)}}^{code\ for\ Q}. \tag{4.19}$$

---

**SOL-83**  **CH.SOL- 4.17.**

1. **True** *KLD is a non-symmetric measure, i.e.* $D(P \parallel Q) \neq D(Q \parallel P)$.

2. **False** *KLD does not satisfy the triangle inequality.*

3. **True** *KLD is not a distance metric.*

4. **True** *KLD is regarded as a measure of the information **gain**. Notice that, however, KLD is the **amount** of information **lost**.*

5. **True** *The units of KL divergence are units of information (bits, nats, etc.).*

6. **True** *KLD is a non-negative measure.*

7. **True** *Performing splitting based on highly informative event usually leads to low model generalization and a less accurate one as well.*

---

**SOL-84**  **CH.SOL- 4.18.**

*Formally, mutual information attempts to measure how correlated two variables are with each other:*

$$
\begin{aligned}
I(X;Y) &= \sum_{x,y} P(x,y) \log \frac{P(x,y)}{P(x)P(y)} \\
&= \mathrm{E}\left[ \log \frac{1}{P(x)} + \log \frac{1}{P(y)} - \log \frac{1}{P(x,y)} \right] \\
&= H(X) + H(Y) - H(X,Y).
\end{aligned}
\tag{4.20}
$$

*Regarding the question at hand, given two distributions $f_1$ and $f_2$ and their joint distribution $f$, the* mutual information *of $f_1$ and $f_2$ is defined as $I(f_1, f_2) = H(f, f_1 f_2)$. If the two distributions are independent, i.e. $f = f_1 \cdot f_2$, the mutual information will vanish. This concept has been widely used as a similarity measure in image analysis.* ∎

**SOL-85**  **CH.SOL- 4.19.**

*The question was commented out but remained here for the consistency of the numbering system.* ∎

### 4.3.4 Classification and Information Gain

**SOL-86**  **CH.SOL- 4.20.**

*The three most widely used methods are:*

*1.*

$$
\text{Entropy } (t) = -\sum_{i=0}^{c-1} p(i) \log_2 p(i).
\tag{4.21}
$$

*2.*

$$
1 - \sum_{i=0}^{c-1} [p(i)]^2
\tag{4.22}
$$

*3.*

$$\text{Classification error } (t) = 1 - \max_i [p(i)]. \tag{4.23}$$

**SOL-87 ☑ CH.SOL- 4.21.**

*In a decision tree, the attribute by which we choose to split is the one with **[maximum]** information gain.*

**SOL-88 ☑ CH.SOL- 4.22.**

*It is clear that the entropy will be **decreased more** by first splitting on Green rather than on Rain.*

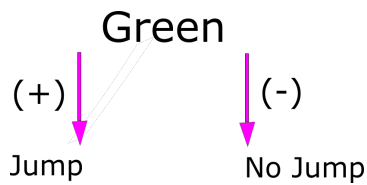

FIGURE 4.12: First split.

**SOL-89 ☑ CH.SOL- 4.23.**

1. *Information gain is the expected reduction in entropy caused by partitioning values in a dataset according to a given attribute.*

2. *A decision tree learning algorithm chooses the next attribute to partition the currently selected node, by first computing the information gain from the entropy, for instance, as a splitting criterion.*

3. *There are 3 positive examples corresponding to Shrinkage=+, and 2 negative examples*

*corresponding to Shrinkage=-. Using the formulae:*

$$H(Y) = -\sum_{i=1}^{k} P(Y = y_i) \log_2 P(Y = y_i) \tag{4.24}$$

*and the probabilities:*

$$P(\gamma = +) = \frac{3}{5}, \tag{4.25}$$

$$P(\gamma = -) = \frac{2}{5}, \tag{4.26}$$

*the overall entropy before splitting is (4.13):*

$$\begin{aligned} E_{orig} &= -(3/5)\log(3/5) - (2/5)\log(2/5) \\ &= H(\gamma) \approx 0.97095[bits/symbol]. \end{aligned} \tag{4.27}$$

```python
import autograd.numpy as np
def binaryEntropy (p):
  return -p*np.log2(p) -(1-p)*np.log2(1-p)

print ("binaryEntropy(p) is:{} bits".format(binaryEntropy(4/7)))
> binaryEntropy(p) is: 0.97095 bits
```

FIGURE 4.13: Entropy before splitting.

4. *If we split on $\theta_1$, (4.5) the relative shrinkage frequency is:*

| Total | $\theta_1 = T$ | $\theta_1 = F$ |
|:---:|:---:|:---:|
| ⊕ | 3 | 0 |
| ⊖ | 1 | 1 |

TABLE 4.5: Splitting on $\theta_1$.

*To compute the information gain (IG) based on feature $\theta_1$, we must first compute the entropy of $\gamma$ after a split based on $\theta_1$, $H(\gamma|\theta_1)$:*

$$H(\gamma|\theta_1)$$
$$= -\sum_{j=1}^{v}\left[\sum_{i=1}^{k} P(\gamma = \gamma_i|\theta_1 = \theta_j)\log_2 P(\gamma = \gamma_i|\theta_1 = \theta_j)\right]$$
$$P(\theta_1 = \theta_j).$$

*Therefore, using the data for the the relative shrinkage frequency (4.5), the information gain after splitting on $\theta_1$ is:*

$$E_{\theta_1=T} = -\frac{3}{4}\log\frac{3}{4} - \frac{1}{4}\log\frac{1}{4} = 0.8112,$$
$$E_{\theta_1=F} = -\frac{0}{1}\log\frac{0}{1} - \frac{1}{1}\log\frac{1}{1} = 0.0. \tag{4.28}$$

*Now we know that $P(\theta_1 = T) = \boxed{4/5}$ and $P(\theta_1 = F) = \boxed{1/5}$, therefore:*

$$\Delta = E_{orig} - \boxed{(4/5)}E_{\theta_1=T} - \boxed{(1/5)}E_{\theta_1=F}$$
$$= 0.97095 - (4/5)*0.8112 - (1/5)*(0.0) \tag{4.29}$$
$$= \approx 0.32198 \ [bits/symbol].$$

■

**SOL-90**  **CH.SOL- 4.24.**
*There are 4 positive examples corresponding to Expansion=+, and 3 negative examples*

*corresponding to Expansion=-.*

1. *The overall entropy before splitting is (4.14):*

$$E_{orig} = -(4/7)\log(4/7) - (3/7)\log(3/7)$$
$$= 0.9852281 \text{ [bits/symbol].}$$

(4.30)

```
1  import autograd.numpy as np
2  def binaryEntropy (p):
3    return -p*np.log2(p) -(1-p)*np.log2(1-p)
4
5  print ("binaryEntropy(p) is:{} bits".format(binaryEntropy(4/7)))
6  > binaryEntropy(p) is:0.9852281 bits
```

FIGURE 4.14: Entropy before splitting.

2. *If we split on $\theta_1$, (4.6) the relative star expansion frequency is:*

| Total | $\theta_1 = T$ | $\theta_1 = F$ |
|:-----:|:--------------:|:--------------:|
| ⊕ | 3 | 1 |
| ⊖ | 0 | 3 |

TABLE 4.6: Splitting on $\theta_1$.

*Therefore, the information gain after splitting on A is:*

$$E_{\theta_1=T} = -\frac{3}{3}\log\frac{3}{3} - \frac{0}{3}\log\frac{0}{3} = 0.0,$$
$$E_{\theta_1=F} = -\frac{3}{4}\log\frac{3}{4} - \frac{1}{4}\log\frac{1}{4} = 0.81127.$$

(4.31)

*Now we know that $P(\theta_1 = T) = \boxed{3/7}$ and $P(\theta_1 = F) = \boxed{4/7}$, therefore:*

$$\begin{aligned}
\Delta &= E_{orig} - \boxed{(3/7)}E_{\theta_1=T} - \boxed{(4/7)}E_{\theta_1=F} \\
&= 0.98522 - (3/7) * 0.0 - (4/7) * (0.81127) \\
&= 0.52163 \text{ [bits/symbol].}
\end{aligned} \tag{4.32}$$

3. *If we split on $\theta_2$, (4.7) the relative star expansion frequency is:*

| Total | $\theta_2 = T$ | $\theta_2 = F$ |
|:-----:|:--------------:|:--------------:|
| + | 3 | 1 |
| - | 1 | 2 |

TABLE 4.7: Splitting on $\theta_2$.

*The information gain after splitting on B is:*

$$\begin{aligned}
E_{\theta_2=T} &= -\frac{3}{4}\log\frac{3}{4} - \frac{1}{4}\log\frac{1}{4} = 0.0.8112, \\
E_{\theta_2=F} &= -\frac{1}{3}\log\frac{1}{3} - \frac{2}{3}\log\frac{2}{3} = 0.9182.
\end{aligned} \tag{4.33}$$

*Now we know that $P(\theta_2 = T) = \boxed{4/7}$ and $P(\theta_2 = F) = \boxed{3/7}$, therefore:*

$$\begin{aligned}
\Delta &= E_{orig} - \boxed{(4/7)}E_{\theta_2=T} - \boxed{(3/7)}E_{\theta_2=F} \\
&= 0.98522 - (4/7) * 0.8122 - (3/7) * (0.9182) \\
&0.1275 \text{ [bits/symbol].}
\end{aligned}$$

$$\begin{aligned}
\Delta &= 0.98522 - (4/7) * 0.8122 - (3/7) * (0.9182) \\
&0.1275 \text{ [bits/symbol].}
\end{aligned} \tag{4.34}$$

---

**SOL-91**  **CH.SOL- 4.25.**

*1.*

$$\begin{aligned}
H(\gamma) &= -\left(\frac{2}{6}\log_2\frac{2}{6} + \frac{4}{6}\log_2\frac{4}{6}\right) \\
H(\gamma) &= -\left(\frac{1}{3}\log_2\frac{1}{3} + \frac{2}{3}\log_2\frac{2}{3}\right) \\
&\approx 0.92 \text{ [bits/symbol].}
\end{aligned} \tag{4.35}$$

*2.*

$$\begin{aligned}
H(\gamma|\theta_1) &= -\frac{1}{3}\left(\frac{1}{2}\log_2\frac{1}{2} + \frac{1}{2}\log_2\frac{1}{2}\right) - \\
\frac{1}{3}\left(\frac{1}{2}\log_2\frac{1}{2} + \frac{1}{2}\log_2\frac{1}{2}\right) &- \frac{1}{3}\left(1\log_2 1\right). \\
H(\gamma|\theta_1) &= \frac{1}{3}(1) + \frac{1}{3}(1) + \frac{1}{3}(0). \\
H(\gamma|\theta_1) &= \frac{2}{3} \approx 0.66 \text{[bits/symbol].}
\end{aligned} \tag{4.36}$$

*3.*

$$\begin{aligned}
H(\gamma|\theta_2) &= -\frac{1}{2}\left(\frac{1}{3}\log_2\frac{1}{3} + \frac{2}{3}\log_2\frac{2}{3}\right) - \frac{1}{2}\left(1\log_2 1\right). \\
H(\gamma|\theta_2) &= \frac{1}{2}\left(\log_2 3 - \frac{2}{3}\right). \\
H(\gamma|\theta_2) &= \frac{1}{2}\log_2 3 - \frac{1}{3} \approx 0.46 \text{ [bits/symbol].}
\end{aligned} \tag{4.37}$$

*4. **False.***

■

### 4.3.5  Mutual Information

---

**SOL-92**  **CH.SOL- 4.26.**

*1. The diagram is depicted in Fig. 4.15.*

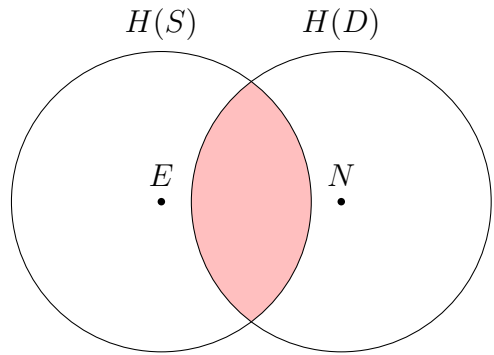

FIGURE 4.15: Mutual Information between $H(S)$ & $H(D)$.

2. ***Equivocation*** *is annotated by* $E$.

3. ***Noise*** *is annotated by* $N$.

4. *The intersection (shaded area) in (4.15) corresponds to mutual information of the source* $H(S)$ *and of the destination* $H(D)$.

5. *The formulae for mutual information is:*

$$H(S; D) = H(S) - E = H(D) - N. \qquad (4.38)$$

**SOL-93**  **CH.SOL- 4.27.**

*The relative entropy* $D(p||q)$ *is the measure of* ***difference*** *between two distributions. It can also be expressed like a measure of the* ***inefficiency*** *of assuming that the distribution is* $q$ *when the* ***true*** *distribution is* $p$.

**SOL-94**  **CH.SOL- 4.28.**

*Mutual information is a Shannon entropy-based measure of dependence between random variables. The mutual information between* $X$ *and* $Z$ *can be understood as the* ***reduction*** *of*

*the **uncertainty** in X given Z:*

$$I(X; Z) := H(X) - H(X \mid Z), \tag{4.39}$$

*where H is the Shannon entropy, and $H(X \mid Z)$ is the conditional entropy of Z given X.* ∎

### 4.3.6 Mechanical Statistics

**SOL-95  CH.SOL- 4.29.**

*Is this question valuable?* ∎

**SOL-96  CH.SOL- 4.30.**

*Boltzmann related the degree of disorder of the state of a physical system to the logarithm of its probability. If, for example, the system has $n$ non-interacting and identical particles, each capable of existing in each of $K$ equally likely states, the leading term in the logarithm of the probability of finding the system in a configuration with $n_1$ particles in state 1, $n_2$ in state 2, etc, is given by the* Boltzmann entropy *$\mathcal{H}_\pi = -\sum_1^K \pi_i \log(\pi_i)$, where $\pi_i = n_i/n$.* ∎

**SOL-97  CH.SOL- 4.31.**

*There are $8$ equiprobable events in each roll of the dice, therefore:*

$$H = -\sum_{i=1}^{8} \frac{1}{8} \log_2 \frac{1}{8} = 3 \ [bits] . \tag{4.40}$$

∎

### 4.3.7 Jensen's inequality

**SOL-98  CH.SOL- 4.32.**

1. *A function $f$ is concave in the range $[a, b]$ if $f\phi 2$ is negative in the range $[a, b]$.*

2. *A function $f$ is convex in the range $[a, b]$ if $f\phi 2$ is positive in the range $[a, b]$.*

3. *The following inequality was published by J.L. Jensen in 1906:*

   **(Jensen's Inequality)** *Let $f$ be a function convex up on $(a, b)$. Then for any $n \geq 2$ numbers $x_i \in (a, b)$:*
   $$f\left(\frac{\sum_{i=1}^{n} x_i}{n}\right) \leq \frac{\sum_{i=1}^{n} f(x_i)}{n},$$

   *and that the equality is attained if and only if $f$ is linear or all $x_i$ are equal.*

   *For a convex down function, the sign of the inequality changes to $\geq$.*

   *Jensen's inequality states that if $f$ is convex in the range $[a, b]$, then:*

   $$\frac{f(a) + f(b)}{2} \geq f\left(\frac{a+b}{2}\right).$$

   *Equality holds if and only if $a = b$. Jensen's inequality states that if $f$ is concave in the range $[a, b]$, then:*
   $$\frac{f(a) + f(b)}{2} \leq f\left(\frac{a+b}{2}\right).$$

   *Equality holds if and only if $a = b$.*

   ■

**SOL-99  CH.SOL- 4.33.**
   **True** *The non-negativity of KLD can be proved using Jensen's inequality.*    ■

# Chapter

## 5

*The true logic of this world is in the **calculus** of probabilities.*

— James C. Maxwell

## Contents

## 5.1   Introduction

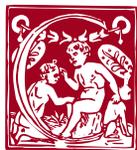 ALCULUS is the mathematics of change; the differentiation of a function is key to almost every domain in the scientific and engineering realms and calculus is also very much *central* to DL. A standard curriculum of *first year calculus* includes topics such as limits, differentiation, the derivative, Taylor series, integration, and the integral. Many aspiring data scientists who lack a relevant mathematical background and are shifting careers, hope to easily enter the field but frequently encounter a mental barricade.

| $f(x)$ | $f'(x)$ |
|--------|---------|
| $\sin(x)$ | $\cos(x)$ |
| $\cos(x)$ | $-\sin(x)$ |
| $\log(x)$ | $\frac{1}{x}$ |
| $e^x$ | $e^x$ |

Thanks to the rapid advances in processing power and the proliferation of GPUs, it is possible to lend the burden of computation to a computer with high efficiency and precision. For instance, extremely fast implementations of backpropagation, the gradient descent algorithm, and *automatic differentiation* (AD) [5] brought artificial intelligence from a mere concept to reality.

Calculus is frequently taught in a way that is very burdensome to the student, therefore I tried incorporating the writing of Python code snippets into the learning process and the usage of:

**DAGs** (**D**irected **A**cyclic **G**raphs). Gradient descent is the essence of optimization in deep learning, which requires efficient access to first and second order derivatives that AD frameworks provide. While older AD frameworks were written in C++ ([4]), the newer ones are Python-based such as Autograd ([10]) and JAX ([3], [1]).

Derivatives are also crucial in graphics applications. For example, in a rendering technique entitled *global illumination*, photons bounce in a synthetically generated scene while their direction and colour has to be determined using derivatives based on the specific material each photon hits. In ray tracing algorithms, the colour of the pixels is determined by tracing the trajectory the photons travel from the eye of the observer through a synthetic 3D scene.

A function is usually represented by a **DAG**. For instance, one commonly used form is to represent intermediate values as nodes and operations as arcs (5.2). One other commonly used form is to represent not only the values but also the operations as nodes (5.11).

The first representation of a function by a DAG goes back to [7].

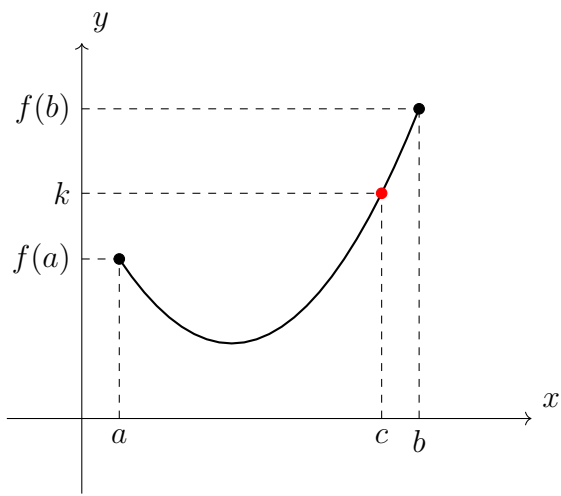

FIGURE 5.1: Intermediate value theorem

Manual differentiation is tedious and error-prone and practically unusable for real-time graphics applications wherein numerous successive derivatives have to be repeatedly calculated. Symbolic differentiation on the other hand, is a computer based method that uses a collection of differentiation rules to analytically calculate an exact derivative of a function resulting in a purely symbolic derivatives. Many symbolic differentiation libraries utilize what is known as *operator-overloading* ([9]) for both the forward and reverse forms of differentiation, albeit they are not quite as fast as AD.

## 5.2  Problems

### 5.2.1  AD, Gradient descent & Backpropagation

AD [5] is the application of the chain rule to functions by computers in order to automatically compute derivatives. AD plays a significant role in training deep learning algorithms and in order to understand AD you need a solid grounding in Calculus. As opposed to numerical differentiation, AD is a procedure for establishing **exact** derivatives without any truncation errors. AD breaks a computer program into a series of fundamental mathematical operations, and the gradient or Hessian of the computer program is found by successive application of the chain rule (5.1) to it's elementary constituents.

For instance, in the C++ programming language, two techniques ([4]) are commonly utilized in transforming a program that calculates numerical values of a function into a program which calculates numerical values for derivatives of that function; (1) an operator overloading approach and (2) systematic source code transformation.

$$\frac{\partial}{\partial t} f(g(t))\bigg|_{t=t_0} = \left(\frac{\partial}{\partial s} f(s)\bigg|_{s=g(t_0)}\right)\left(\frac{\partial}{\partial t} g(t)\bigg|_{t=t_0}\right) \tag{5.1}$$

One notable feature of AD is that the values of the derivatives produced by applying AD, as opposed to numerical differentiation (finite difference formulas), are **exact and accurate**. Two variants of AD are widely adopted by the scientific community: the forward mode or the reverse mode where the underlying distinction between them is the order in which the chain rule is being utilized. The forward mode, also entitled tangent mode, propagates derivatives from the dependent towards the independent variables, whereas the reverse or adjoint mode does exactly the opposite. AD makes heavy use of a concept known as dual numbers (DN) first introduced by Clifford ([2]).

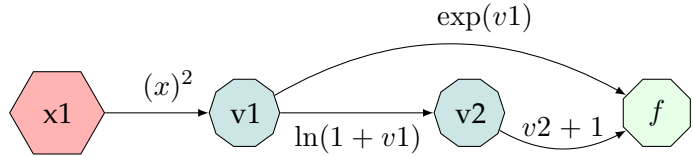

FIGURE 5.2: A Computation graph with intermediate values as nodes and operations as arcs.

**PRB-100 ❷ CH.PRB- 5.1.**

1. *Write the formulae for the **finite difference rule** used in numerical differentiation.*

2. *What is the main problem with this formulae?*

3. *Indicate one problem with software tools which utilize numerical differentiation and successive operations on floating point numbers.*

---

**PRB-101 ❷ CH.PRB- 5.2.**

1. *Given a function $f(x)$ and a point $a$, define the **instantaneous rate of change** of $f(x)$ at $a$.*

2. *What other commonly used alternative name does the instantaneous rate of change have?*

3. *Given a function $f(x)$ and a point $a$, define the **tangent line** of $f(x)$ at $a$.*

---

### 5.2.3 Directed Acyclic Graphs

There are two possible ways to traverse a **DAG** (**D**irected **A**cyclic **G**raph). One method is simple. Start at the bottom and go through all nodes to the top of the computational tree. That is nothing else than passing the corresponding computation sequence top down. Based on this method, the so called forward mode or of AD was developed [8]. In contrast to this forward mode the reverse mode was first used by Speelpenning [13] who passed the underlying graph top down and propagated the gradient backwards.

---

**PRB-102 ❷ CH.PRB- 5.3.**

1. *State the definition of the derivative $f(c)$ of a function $f(x)$ at $x = c$.*

2. *With respect to the **DAG** depicted in 5.3:*

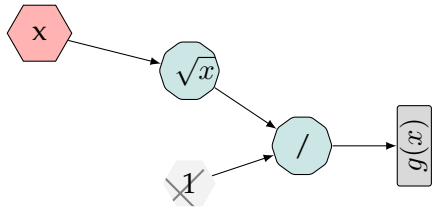

FIGURE 5.3: An expression graph for $g(x)$. Constants are shown in gray, crossed-out since derivatives should not be propagated to constant operands.

(a) *Traverse the graph 5.3 and find the function $g(x)$ it represents.*

(b) *Using the definition of the derivative, find $g'(9)$.*

**PRB-103 ❷ CH.PRB- 5.4.**

1. *With respect to the expression graph depicted in 5.4, traverse the graph and find the function $g(x)$ it represents.*

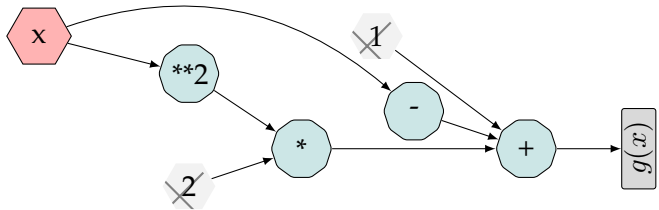

FIGURE 5.4: An expression graph for $g(x)$. Constants are shown in gray, crossed-out since derivatives should not be propagated to constant operands.

2. *Using the definition of the derivative find the derivative of $g(x)$.*

5.2.4    The chain rule

**PRB-104 ❷ CH.PRB- 5.5.**

1. *The **chain rule** is key concept in differentiation. Define it.*

2. *Elaborate how the chain rule is utilized in the context of neural networks.*

### 5.2.5   Taylor series expansion

The idea behind a Taylor series is that if you know a function and all its derivatives at one point $x = a$, you can approximate the function at other points near $a$. As an example, take $f(x) = \sqrt{x}$. You can use Taylor series to approximate $\sqrt{10}$ by knowing $f(9)$ and all the derivatives $f'(9)$, $f''(9)$.

The MacLaurin series (5.2) is a special case of Taylor series when $f(0)$, $f'(0)$ are known:

$$f(x) = f(0) + xf'(0) + \frac{x^2}{2!}f''(0) + \frac{x^3}{3!}f'''(0) + \cdots = \\ \sum_{p=0}^{\infty} \frac{x^p}{p!}f^{(p)}(0) \tag{5.2}$$

For instance, the Maclaurin expansion of $cos(x)$ is:

$$f(x) = \cos x, \quad f'(x) = -\sin x, \\ f''(x) = -\cos x, \quad f'''(x) = \sin x \tag{5.3}$$

When evaluated at 0 results in:

$$\cos x = 1 - \frac{x^2}{2!} + \frac{x^4}{4!} - \frac{x^6}{6!} + \cdots \tag{5.4}$$

---

### PRB-105 ❷ CH.PRB- 5.6.

*Find the Taylor series expansion for:*

*1.*

$$\frac{1}{1-x} \tag{5.5}$$

*2.*

$$e^x \tag{5.6}$$

*3.*

$$\sin(x) \tag{5.7}$$

*4.*

$$\cos(x) \tag{5.8}$$

**PRB-106 ❷ CH.PRB- 5.7.**
*Find the Taylor series expansion for:*

$$\log(x) \tag{5.9}$$

**PRB-107 ❷ CH.PRB- 5.8.**
*Find the Taylor series expansion centered at $x = -3$ for:*

$$f(x) = 5x^2 - 11x + 1 \tag{5.10}$$

**PRB-108 ❷ CH.PRB- 5.9.**
*Find the $101$th degree Taylor polynomial centered at $x = 0$ for:*

$$f(x) = \cos(x) \tag{5.11}$$

**PRB-109 ❷ CH.PRB- 5.10.**
*At $x = 1$, compute the first 7 terms of the Taylor series expansion of:*

$$f(x) = \ln 3x. \tag{5.12}$$

### 5.2.6 Limits and continuity

**Theorem 1** (L'Hopital's rule).

$$[\lim_{x \to a} \frac{f(x)}{g(x)} = \lim_{x \to a} \frac{f'(x)}{g'(x)}]. \tag{5.13}$$

**PRB-110 ❷ CH.PRB- 5.11.**
   *Find the following limits:*

   1. $\lim_{x \to 3} \dfrac{e^{x^3} - e^{27}}{3x - 9}$

   2. $\lim_{x \to 0} \dfrac{e^{x^2} - x - 1}{3 \cos x - x - 3}$

   3. $\lim_{x \to \infty} \dfrac{x - \ln x}{\sqrt[100]{x} + 4}$

### 5.2.7 Partial derivatives

**PRB-111 ❷ CH.PRB- 5.12.**

   1. **True or false**: *When applying a partial derivative, there are two variables considered constants - the dependent and independent variable.*

   2. *Given $g(x, y)$, find its partial derivative with respect to $x$:*

$$g(x, y) = x^2 y + yx + 8y. \tag{5.14}$$

**PRB-112 ❷ CH.PRB- 5.13.**

*The gradient of a two-dimensional function is given by*

$$\nabla f(x,y) = \frac{\partial f}{\partial x}i + \frac{\partial f}{\partial y}j \tag{5.15}$$

1. *Find the gradient of the function:*

$$f(x,y) = xy^2 - y^2 + x^3 \tag{5.16}$$

2. *Given the function:*

$$g(x,y) = x^2 y = xy^2 - y - 1, \tag{5.17}$$

   *evaluate it at* $(-1,0)$, *directed at* $(1,1)$.

---

**PRB-113 ❷ CH.PRB- 5.14.**
  *Find the partial derivatives of:*

$$f(x,y) = 3\sin^2(x-y) \tag{5.18}$$

---

**PRB-114 ❷ CH.PRB- 5.15.**
  *Find the partial derivatives of:*

$$z = 2\sin(x)\sin(y) \tag{5.19}$$

5.2.8    Optimization

---

**PRB-115 ❷ CH.PRB- 5.16.**
  *Consider* $f(x) = \dfrac{x^2 + 1}{(x+2)^2}$.

1. *Where is* $f(x)$ *well defined?*

2. *Where is $f(x)$ increasing and decreasing?*

3. *Where is $f(x)$ reaching minimum and maximum values.*

**PRB-116 ❷ CH.PRB- 5.17.**
  *Consider $f(x) = 2x^3 - x$.*

1. *Derive $f(x)$ and conclude on its behavior.*

2. *Derive once again and discuss the concavity of the function $f(x)$.*

**PRB-117 ❷ CH.PRB- 5.18.**
*Consider the function*
$$f(x, y) = 2x^2 - xy + y^2,$$
*and find maximum, minimum, and saddle points.*

5.2.9  The Gradient descent algorithm

**PRB-118 ❷ CH.PRB- 5.19.**
  *The gradient descent algorithm can be utilized for the minimization of convex functions. Stationary points are required in order to minimize a convex function. A very simple approach for finding stationary points is to start at an arbitrary point, and move along the gradient at that point towards the next point, and repeat until converging to a stationary point.*

1. *What is the term used to describe the vector of all partial derivatives for a function $f(x)$?*

2. *Complete the sentence: when searching for a minima, if the derivative is positive, the function is **increasing/decreasing**.*

3. The function $x^2$ as depicted in 5.5, has a derivative of $f'(x) = 2x$. Evaluated at $x = -1$, the derivative equals $f'(x = -1) = -2$. At $x = -1$, the function is **decreasing** as $x$ gets larger. We will happen if we wish to find a **minima** using gradient descent, and increase (decrease) $x$ by **the size of the gradient**, and then again repeatedly keep jumping?

4. How this phenomena can be alleviated?

5. **True or False:** The gradient descent algorithm is guaranteed to find a local minimum if the learning rate is correctly decreased and a finite local minimum exists.

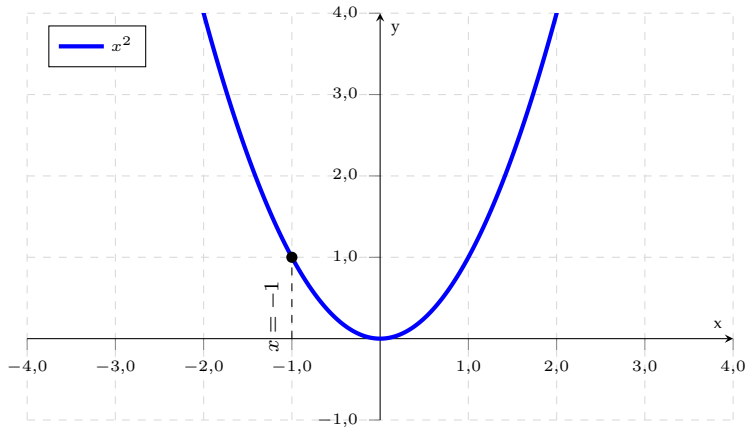

FIGURE 5.5: $x^2$ Function

---

**PRB-119 ❷ CH.PRB- 5.20.**

1. Is the data linearly separable?

| $X_1$ | $X_2$ | $Y$ |
|-------|-------|-----|
| 1 | 1 | $+$ |
| 12 | 12 | $-$ |
| 4 | 5 | $-$ |
| 12 | 12 | $+$ |

(5.20)

2. *What is loss function for linear regression?*

3. *What is the gradient descent algorithm to minimize a function $f(x)$?*

### 5.2.10   The Backpropagation algorithm

The most important, expensive and hard to implement part of any hardware realization of ANNs is the non-linear activation function of a neuron. Commonly applied activation functions are the sigmoid and the hyperbolic tangent. In the most used learning algorithm in present day applications, back-propagation, the derivatives of the sigmoid function are needed when back propagating the errors.

The backpropagation algorithm looks for the minimum of the error function in weight space using the method of gradient descent.

---

**PRB-120 ❷ CH.PRB- 5.21.**

1. *During the training of an ANN, a sigmoid layer applies the sigmoid function to every element in the forward pass, while in the backward pass the chain rule is being utilized as part of the backpropagation algorithm. With respect to the backpropagation algorithm, given a sigmoid $\sigma(x) = \frac{e^x}{1+e^x}$ activation function, and a $J$ as the cost function, annotate each part of equation*  *(5.21):*

$$dZ = \frac{dJ}{d\sigma(x)}\frac{d\sigma(x)}{dx} = dA \cdot \sigma(x) \cdot \Big(1 - \sigma(x)\Big) \tag{5.21}$$

2. *Code snippet 5.6 provides a pure Python-based (e.g. not using Autograd) implementation of the forward pass for the sigmoid function. Complete the backward pass that directly computes the analytical gradients.*

```
1  class Sigmoid:
2    def forward(self,x):
3      self.x = x
4      return 1/(1+np.exp(-x))
5    def backward(self, grad):
6      grad_input = [???]
7      return grad_input
```

FIGURE 5.6: Forward pass for the sigmoid function.

**PRB-121 ❷ CH.PRB- 5.22.**
   *This question deals with the effect of customized transfer functions. Consider a neural network with hidden units that use $x^3$ and output units that use $\sin(2x)$ as transfer functions. Using the chain rule, starting from $\partial E/\partial y_k$, derive the formulas for the weight updates $\Delta w_{jk}$ and $\Delta w_{ij}$. Notice - do not include partial derivatives in your final answer.*

### 5.2.11   Feed forward neural networks

Understanding the inner-workings of Feed Forward Neural Networks (FFNN) is crucial to the understanding of other, more advanced Neural Networks such as CNN's.

   A Neural Network (NN) is an interconnected assembly of simple processing elements, *units* or *nodes*, whose functionality is loosely based on the animal neuron. The processing ability of the network is stored in the inter-unit connection strengths, or *weights*, obtained by a process of adaptation to, or *learning* from, a set of training patterns. [6]

The *Backpropagation Algorithm* is the most widely used learning algorithm for FFNN. Backpropagation is a training method that uses the *Generalized Delta Rule*. Its basic idea is to perform a gradient descent on the total squared error of the network output, considered as a function of the weights. It was first described by *Werbos* and made popular by *Rumelhart*'s, *Hinton*'s and *Williams*' paper [12].

### 5.2.12 Activation functions, Autograd/JAX

Activation functions, and most commonly the sigmoid activation function, are heavily used for the construction of NNs. We utilize Autograd ([10]) and the recently published JAX ([1]) library to learn about the relationship between activation functions and the Backpropagation algorithm.

Using a logistic, or sigmoid, activation function has some benefits in being able to easily take derivatives and then interpret them using a logistic regression model. Autograd is a core module in PyTorch ([11]) and adds inherit support for automatic differentiation for all operations on tensors and functions. Moreover, one can implement his own custom Autograd function by sub classing the autograd *Function* and implementing the forward and backward passes which operate on PyTorch tensors. PyTorch provides a simple syntax (5.7) which is transparent to both CPU/GPU support.

```python
import torch
from torch.autograd import Function
class DLFunction(Function):
 @staticmethod
 def forward(ctx, input):
  ...

 @staticmethod
 def backward(ctx, grad_output):
 ...
```

FIGURE 5.7: PyTorch syntax for autograd.

**PRB-122 ❷ CH.PRB- 5.23.**

1. ***True or false:*** *In Autograd, if any input tensor of an operation has requires_grad=True, the computation will be tracked. After computing the backward pass, a gradient w.r.t. this tensor is accumulated into .grad attribute*

2. **True or false:** *In Autograd, multiple calls to backward will sum up previously computed gradients if they are not zeroed.*

---

### PRB-123 ❷ CH.PRB- 5.24.

*Your friend, a veteran of the DL community wants to use logistic regression and implement custom activation functions using Autograd. Logistic regression is used when the variable y that we want to predict can only take on discrete values (i.e. classification). Considering a binary classification problem (y = 0 or y = 1) (5.8), the hypothesis function could be defined so that it is bounded between [0, 1] in which we use some form of logistic function, such as the **sigmoid function**. Other, more efficient functions exist such as the **ReLU** (Rectified Linear Unit) which we discussed later. Note: The weights in (5.8) are only meant for illustration purposes and are not part of the solution.*

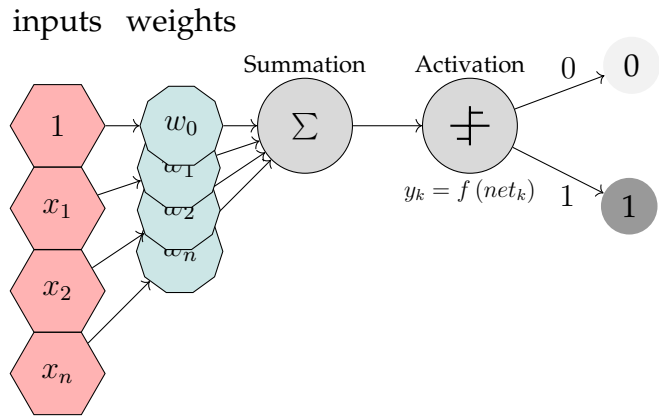

FIGURE 5.8: A typical binary classification problem.

1. *Given the sigmoid function:* $g(x) = \frac{1}{1+e^{-z}}$ *what is the expression for the corresponding hypothesis in logistic regression?*

2. *What is the decision boundary?*

3. *What does* $h_{\Theta}(x) = 0.8$ *mean?*

4. *Using an Autograd based Python program, implement both the forward and backward pass for the sigmoid activation function and evaluate it's derivative at* $x = 1$

5. *Using an Autograd based Python program, implement both the forward and backward pass for the ReLU activation function and evaluate it's derivative at $x = 1$*

**PRB-124 ❷ CH.PRB- 5.25.**

*For real values, $-1 < x < 1$ the hyperbolic tangent function is defined as:*

$$\tanh^{-1} x = \frac{1}{2}[\ln(1 + x) - \ln(1 - x)] \tag{5.22}$$

*On the other hand, the artanh function, which returns the **inverse** hyperbolic tangent of its argument x, is implemented in numpy as arctanh().*

*Its derivative is given by:*

$$(arctanh(x))' = \frac{1}{1 - x^2} \tag{5.23}$$

*Your friend, a veteran of the DL community wants to implement a custom activation function for the arctanh function using Autograd. Help him in realize the method.*

1. *Use this numpy array as an input $[[0.37, 0.192, 0.571]]$ and evaluate the result using pure Python.*

2. *Use the PyTorch based $torch.autograd.Function$ class to implement a custom Function that implements the forward pass for the arctanh function in Python.*

3. *Use the PyTorch based $torch.autograd.Function$ class to implement a custom Function that implements the backward pass for the arctanh function in Python.*

4. *Name the class ArtanhFunction, and using the gradcheck method from torch.autograd, verify that your numerical values equate the analytical values calculated by gradcheck. Remember you must implement a method entitled .apply(x) so that the function can be invoked by Autograd.*

Dual numbers (DN) are analogous to complex numbers and augment real numbers with a dual element by adjoining an infinitesimal element $\mathbf{d}$, for which $\mathbf{d}^2 = 0$.

---

**PRB-125 ❷ CH.PRB- 5.26.**

1. *Explain how AD uses floating point numerical rather than symbolic expressions.*

2. *Explain the notion of DN as introduced by ([2]).*

3. *What arithmetic operations are possible on DN?.*

4. *Explain the relationship between a Taylor series and DN.*

---

**PRB-126 ❷ CH.PRB- 5.27.**

1. *Expand the following function using DN:*

$$\sin(x + \dot{x}\mathbf{d}) \tag{5.24}$$

2. *With respect to the expression graph depicted in 5.9:*

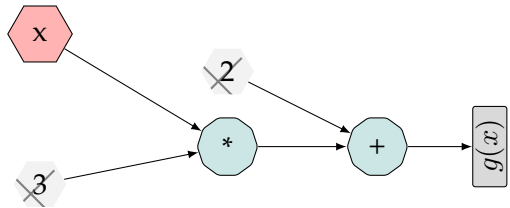

FIGURE 5.9: An expression graph for $g(x)$. Constants are shown in gray, crossed-out since derivatives should not be propagated to constant operands.

   (a) *Traverse the graph 5.9 and find the function $g(x)$ it represents.*

   (b) *Expand the function $g(x)$ using DN.*

3. *Show that the* ⸢general identity⸣:

$$g(x + \dot{x}\mathbf{d}) = g(x) + g'(x)\dot{x}\mathbf{d} \tag{5.25}$$

*holds in this particular case too.*

4. *Using the derived DN, evaluate the function $g(x)$ at $x = 2$.*

5. *Using an Autograd based Python program implement the function and evaluate it's derivative at $x = 2$.*

---

**PRB-127 ❷ CH.PRB- 5.28.**

*With respect to the expression graph depicted in 5.10:*

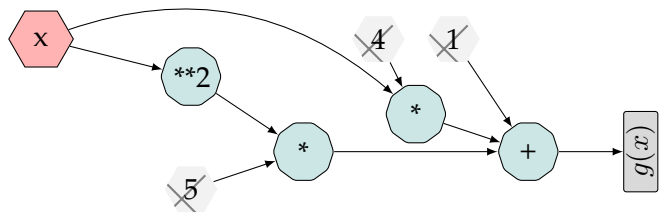

FIGURE 5.10: An expression graph for $g(x)$. Constants are shown in gray, crossed-out since derivatives should not be propagated to constant operands.

1. *Traverse the graph 5.10 and find the function $g(x)$ it represents.*

2. *Expand the function $g(x)$ using DN.*

3. *Using the derived DN, evaluate the function $g(x)$ at $x = 5$.*

4. *Using an AutoGrad based Python program implement the function and evaluate it's derivative at $x = 5$.*

5.2.14   Forward mode AD

---

**PRB-128 ❷ CH.PRB- 5.29.**

*When differentiating a function using forward-mode AD, the computation of such an expression can be computed from its corresponding directed a-cyclical graph by propagating the numerical values.*

1. *Find the function, $g(A, B, C)$ represented by the expression graph in* 5.11.

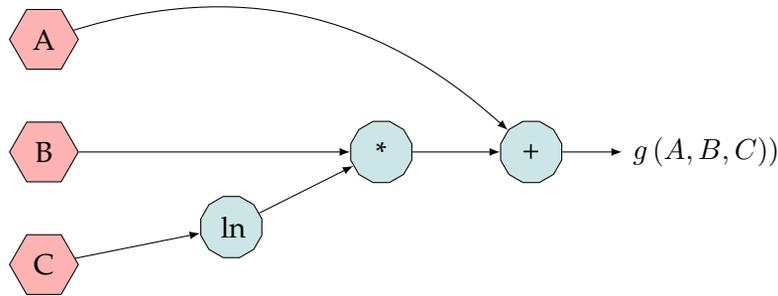

FIGURE 5.11: A computation graph for $g(x)$

2. *Find the partial derivatives for the function $g(x)$.*

---

**PRB-129 ❷ CH.PRB- 5.30.**

*Answer the following given that a computational graph of a function has N inputs and M outputs.*

1. **True or False?:**

   (a) *Forward and reverse mode AD always yield the same result.*

   (b) *In reverse mode AD there are fewer operations (time) and less space for intermediates (memory).*

   (c) *The cost for forward mode grows with N.*

   (d) *The cost for reverse mode grows with M.*

---

**PRB-130 ❷ CH.PRB- 5.31.**

1. *Transform the source code in code snippet 5.1 into a function $g(x_1, x_2)$.*

   CODE 5.1: A function, $g(x_1, x_2)$ in the C programming language.

```c
float g( float x1 , float x2) {
 float v1, v2, v3 , v4 , v5;
 v1=x1;
 v2=x2;
 v3 = v1 * v2;
 v4 = ln (v1 );
 v5 = v3 + v4;
 return v5;
}
```

2. *Transform the function $g(x_1, x_2)$ into an expression graph.*

3. *Find the partial derivatives for the function $g(x_1, x_2)$.*

### 5.2.15   Forward mode AD table construction

**PRB-131 ❷ CH.PRB- 5.32.**

1. *Given the function:*

$$f(x_1, x_2) = x_1 x_2 + ln(x_1) \tag{5.26}$$

   *and the graph 5.1, annotate each vertex (edge) of the graph with the partial derivatives that would be propagated in forward mode AD.*

2. *Transform the graph into a **table** that computes the **function**:*
   *$g(x_1, x_2)$ evaluated at $(x1; x2) = (e^2; \pi)$ using forward-mode AD.*

3. *Write and run a Python code snippet to prove your results are correct.*

4. *Describe the role of **seed values** in forward-mode AD.*

5. *Transform the graph into a **table** that computes the **derivative** of $g(x_1, x_2)$ evaluated at $(x1; x2) = (e^2; \pi)$ using forward-mode AD for $x_1$ as the chosen independent variable.*

6. *Write and run a Python code snippet to prove your results are correct.*

### 5.2.16   Symbolic differentiation

In this section, we introduce the basic functionality of the SymPy (SYMbolic Python) library commonly used for symbolic mathematics as a means to deepen your understanding in both Python and calculus. If you are using Sympy in a Jupyter notebook in Google Colab (e.g. https://colab.research.google.com/) then rendering sympy equations requires MathJax to be available within each cell output. The following is a hook function that will make this possible:

CODE 5.2: Sympy in Google Colab

```python
from IPython.display import Math, HTML
def enable_sympy_in_cell():
display(HTML("<script
    src='https://cdnjs.cloudflare.com/ajax/libs/"
"mathjax/2.7.3/latest.js?config=default'>
</script>"))
get_ipython().events.register('pre_run_cell',
    enable_sympy_in_cell)
```

After successfully registering this hook, SymPy rendering (5.3) will work correctly:

CODE 5.3: Rendering Sympy in Google Colab

```python
import sympy
from sympy import *
init_printing()
x, y, z = symbols('x y z')
Integral(sqrt(1/x), (x, 0, oo))
```

It is also recommended to use the latest version of Sympy:

```
> pip install --upgrade sympy
```

### 5.2.17   Simple differentiation

**PRB-132 ❷ CH.PRB- 5.33.**
 *Answer the following questions:*

1. *Which differentiation method is inherently prone to rounding errors?*

2. *Define the term symbolic differentiation.*

**PRB-133 ❷ CH.PRB- 5.34.**
 *Answer the following questions:*

1. *Implement the sigmoid function $\sigma(x) = \frac{1}{1+e^{-x}}$ symbolically using a Python based SymPy program.*

2. *Differentiate the sigmoid function using SymPy and compare it with the analytical derivation $\sigma'(x) = \sigma(x)(1 - \sigma(x))$.*

3. *Using SymPy, evaluate the gradient of the sigmoid function at $x = 0$.*

4. *Using SymPy, plot the resulting gradient of the sigmoid function.*

### 5.2.18   The Beta-Binomial model

**PRB-134 ❷ CH.PRB- 5.35.**

*You will most likely not be given such a long programming task during a face-to-face interview. Nevertheless, an extensive home programming assignment is typically given at many of the start-ups I am familiar with. You should allocate around approximately four to six hours to completely answer all questions in this problem.*

*We discussed the Beta-Binomial model extensively in chapter 3. Recall that the Beta-Binomial distribution is frequently used in Bayesian statistics to model the number of successes in $n$ trials. We now employ SymPy to do the same; demonstrate computationally how a prior distribution is updated to develop into a posterior distribution after observing the data via the relationship of the Beta-Binomial distribution.*

*Provided the probability of success, the number of successes after n trials follows a binomial distribution. Note that the beta distribution is a conjugate prior for the parameter of the binomial distribution. In this case, the likelihood function is binomial, and a beta prior distribution yields a beta posterior distribution.*

*Recall that for the Beta-Binomial distribution the following relationships exist:*

| | | |
|---|---|---|
| *Prior of θ* | *Beta(a,b)* | |
| *Likelihood* | *binomial $(n, \theta)$* | (5.27) |
| *Posterior of θ* | *Beta $(a + x, b + n - x)$* | |
| *Posterior Mean* | $(a + x)/(a + b + n - x)$ | |

1. *Likelihood: The starting point for our inference problem is the Likelihood, the probability of the **observed** data. Find the Likelihood function symbolically using sympy. Convert the SymPy representation to a purely Numpy based callable function with a Lambda expression. Evaluate the Likelihood function at $\theta = 0.5$ with 50 successful trials out of 100.*

2. *Prior: The Beta Distribution. Define the Beta distribution which will act as our **prior** distribution symbolically using sympy. Convert the SymPy representation to a purely Numpy based callable function. Evaluate the Beta Distribution at $\theta : 0.5, a : 2, b : 7$*

3. *Plot the Beta distribution, using the Numpy based function.*

4. *Posterior: Find the **posterior** distribution by multiplying our Beta prior by the Binomial Likelihood symbolically using sympy. Convert the SymPy representation to* *a purely Numpy based callable function. Evaluate the Posterior Distribution at* $\theta$ : $0.5, a : 2, b : 7$

5. *Plot the posterior distribution, using the Numpy based function.*

6. *Show that the posterior distribution has the same functional dependence on* $\theta$ *as the prior, and it is just* **another Beta distribution**.

7. *Given:*
   **Prior** : $\text{Beta}(\theta|a = 2, b = 7) = 56\theta\left(-\theta + 1\right)^6$ *and:*
   **Likelihood** : $\text{Bin}(r = 3|n = 6, \theta) = 19600\theta^3\left(-\theta + 1\right)^{47}$ *find the resulting posterior distribution and plot it.*

## 5.3   Solutions

### 5.3.1   Algorithmic differentiation, Gradient descent

### 5.3.2   Numerical differentiation

**SOL-100**  **CH.SOL- 5.1.**

1. *The formulae is:*

$$f'(x) \approx \frac{f(x + h) - f(x)}{h}. \tag{5.28}$$

2. *The main problem with this formulae is that it suffers from numerical instability for small values of* $h$.

3. *In some numerical software systems, the number* $\sqrt{2}$ *may be represented as the a floating point number* $\approx 1.414213562$. *Therefore, the result of:*
   $float\left(\sqrt{(2)}\right) * float\left(\sqrt{(2)}\right)$ *may equal* $\approx 2.000000446$.

■

**SOL-101**  **CH.SOL- 5.2.**

1. *The instantaneous rate of change equals:*

$$\lim_{h \to 0} \frac{f(a+h) - f(a)}{a+h-a}. \tag{5.29}$$

2. *The instantaneous rate of change of $f(x)$ at $a$ is also commonly known as the tangent line of $f(x)$ at $a$.*

3. *Given a function $f(x)$ and a point $a$, the tangent (Fig. 5.12) line of $f(x)$ at $a$ is a line that touches $f(a)$ but does not cross $f(x)$ (sufficiently close to $a$).*

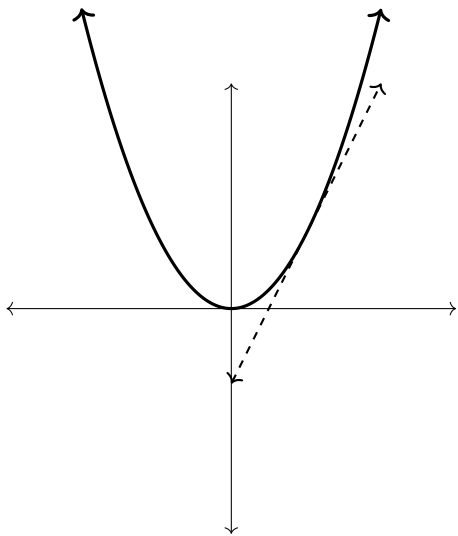

FIGURE 5.12: A Tangent line

### 5.3.3   Directed Acyclic Graphs

**SOL-102**  **CH.SOL- 5.3.**

1. *The definition is:*

$$f'(c) = \lim_{h \to 0} \frac{f(c+h) - f(c)}{h}.$$

2. *If we traverse the graph 5.3 from left to right we derive the following function:*

$$g(x) = \frac{1}{\sqrt{x}}. \tag{5.30}$$

$$
\begin{aligned}
f'(9) &= \lim_{h \to 0} \frac{1/\sqrt{9+h} - 1/\sqrt{9}}{h} \\
&= \lim_{h \to 0} \frac{\sqrt{9} - \sqrt{9+h}}{\sqrt{9} \cdot \sqrt{9+h} \cdot h} \\
&= \lim_{h \to 0} \frac{(3 - \sqrt{9+h})(3 + \sqrt{9+h})}{3\sqrt{9+h} \cdot (3 + \sqrt{9+h}) \cdot h} \\
&= \lim_{h \to 0} \frac{9 - (9+h)}{9\sqrt{9+h} \cdot h + 3 \cdot (9+h) \cdot h} \\
&= -\frac{1}{9 \cdot 3 + 3 \cdot 9} \\
&= -\frac{1}{54}
\end{aligned}
$$

■

## SOL-103 ✎ CH.SOL- 5.4.

1. *The function $g(x) = 2x^2 - x + 1$ represents the expression graph depicted in 5.4.* 

2. *By the definition:*

$$\begin{aligned}
f'(x) &= \lim_{h \to 0} \frac{f(x+h) - f(x)}{x + h - x} \\
&= \lim_{h \to 0} \frac{2(x+h)^2 - (x+h) + 1 - 2x^2 + x - 1}{h} \\
&= \lim_{h \to 0} \frac{2(x^2 + 2xh + h^2) - x - h + 1 - 2x^2 + x - 1}{h} \\
&= \lim_{h \to 0} \frac{2x^2 + 4xh + 2h^2 - x - h + 1 - 2x^2 + x - 1}{h} \\
&= \lim_{h \to 0} \frac{4xh + 2h^2 - h}{h} \\
&= \lim_{h \to 0} 4x + 2h - 1 \\
&= 4x - 1.
\end{aligned}$$
(5.31)

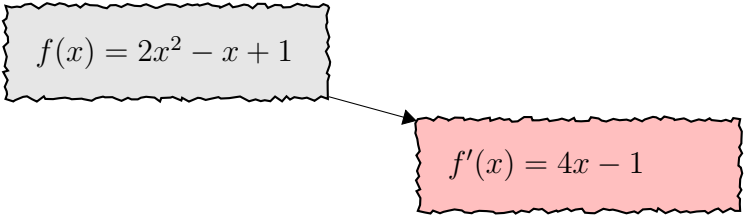

$$f(x) = 2x^2 - x + 1$$

$$f'(x) = 4x - 1$$

### 5.3.4    The chain rule

**SOL-104 ✏ CH.SOL- 5.5.**

1. *The chain rule states that the partial derivative of $E = E(x, y)$ with respect to $x$ can be calculated via another variable $y = y(x)$, as follows:*

$$\frac{\partial E}{\partial x} = \frac{\partial E}{\partial y} \cdot \frac{\partial y}{\partial x}$$
(5.32)

2. *For instance, the chain rule [8] is applied in neural networks to calculate the change in*

*its weights resulting from tuning the cost function. This derivative is calculated via a chain of partial derivatives (e.g. of the activation functions).*

### 5.3.5 Taylor series expansion

**SOL-105**  **CH.SOL- 5.6.**

*1.*

$$\frac{1}{1-x} \ = \ \sum_{n=0}^{\infty} x^n = 1 + x + x^2 + x^3$$

*(when $-1 < x < 1$)* (5.33)

*2.*

$$e^x \ = \ \sum_{n=0}^{\infty} \frac{x^n}{n!} = 1 + x + \frac{x^2}{2!} + \frac{x^3}{3!} + \cdots$$ (5.34)

*3.*

$$\sin x \ = \ \sum_{n=0}^{\infty} \frac{(-1)^n}{(2n+1)!} x^{2n+1} \ = x - \frac{x^3}{3!} + \frac{x^5}{5!} - \cdots$$ (5.35)

*4.*

$$\cos x \ = \ \sum_{n=0}^{\infty} \frac{(-1)^n}{(2n)!} x^{2n} \ = 1 - \frac{x^2}{2!} + \frac{x^4}{4!} - \cdots$$ (5.36)

**SOL-106**  **CH.SOL- 5.7.**

$$\log x = \sum_{n=1}^{\infty} \frac{(-1)^{n+1}(x-1)^n}{n} = (x-1) - \frac{(x-1)^2}{2} +$$
$$\frac{(x-1)^3}{3} - \frac{(x-1)^4}{4} + \cdots \tag{5.37}$$

**SOL-107**  **CH.SOL- 5.8.**

*In this case, all derivatives can be computed:*

$$f^0(x) = 5x^2 - 11x + 1,$$
$$f^0(-3) = 79,$$
$$f^1(x) = 10x - 11,$$
$$f^1(-3) = -41, \tag{5.38}$$
$$f^2(x) = 10,$$
$$f^2(-3) = 10,$$
$$f^n(x) = 0, \forall n \geq 3.$$

**SOL-108**  **CH.SOL- 5.9.**

*The immediate answer is 1. Refer to eq. 5.36 to verify this logical consequence.*

**SOL-109**  **CH.SOL- 5.10.**

*By employing eq. 5.37, one can substitute $x$ by $3 - x$ and generate the first 7 terms of the $x$-dependable outcome before assigning the point $x = 1$.*

### 5.3.6    Limits and continuity

**SOL-110**  **CH.SOL- 5.11.**

1. With an indeterminate form $0/0$, L'Hopital's rule holds. We look at

$$\lim_{x \to 3} \frac{3x^2 e^{x^3}}{3} = 9e^{27},$$

which equals to the original limit.

2. Again, we yield $0/0$ at interim, so we look at the first order derivative

$$\lim_{x \to 0} \frac{2xe^x - 1}{-3\sin x - 1} = 1.$$

The original limit is also equal to $1$.

3. This time, the intermediate form is of $\infty/\infty$ and L'Hopital applies as well. The quotient of the derivatives is

$$\frac{1 - \frac{1}{x}}{0.01x^{-99/100}} = 100(x-1)x^{1/99}$$

As $x \to \infty$, this goes to $\infty$, so the original limit is equal to $\infty$ also.

■

### 5.3.7   Partial derivatives

**SOL-111**  **CH.SOL- 5.12.**

1. True.

2. By treating $y$ as constant, one can derive that

$$\frac{\partial g}{\partial x} = 2xy + y. \tag{5.39}$$

■

**SOL-112**  **CH.SOL- 5.13.**

1.

$$\begin{aligned}
\boldsymbol{\nabla} f(x,y) &= \frac{\partial f}{\partial x}\boldsymbol{i} + \frac{\partial f}{\partial y}\boldsymbol{j} \\
&= \left(y^2 + 3x^2\right)\boldsymbol{i} + (2xy - 2y)\boldsymbol{j}
\end{aligned} \tag{5.40}$$

2. *It can be shown that* $\nabla g(x,y) = \left(2xy + y^2\right)\boldsymbol{i} + \left(x^2 + 2xy - 1\right)\boldsymbol{j}$ *at* $(-1,0)$ *equals* $(0,0)$. *According to the definition of directional derivative:*

$$\frac{(0,0) \cdot (1,1)}{|(1,1)|} = 0 \tag{5.41}$$

**SOL-113** 🖉 **CH.SOL- 5.14.**

$$\begin{aligned}
\frac{\partial f}{\partial x} &= 6\sin(x-y)\cos(x-y) \\
\frac{\partial f}{\partial y} &= -6\sin(x-y)\cos(x-y)
\end{aligned} \tag{5.42}$$

**SOL-114** 🖉 **CH.SOL- 5.15.**

$$\begin{aligned}
\frac{\partial z}{\partial x} &= 2\cos x \sin y \\
\frac{\partial z}{\partial y} &= 2\sin x \cos y
\end{aligned} \tag{5.43}$$

### 5.3.8   Optimization

**SOL-115** 🖉 **CH.SOL- 5.16.**

1. *The function is only defined where $x \neq -2$, in the domain of:*
   $(-\infty, -2) \cup (-2, +\infty)$.

2. *By a simple quotient-based derivation:*

$$f'(x) = \frac{2(x+2)(2x-1)}{(x+2)^4}. \tag{5.44}$$

   *Namely, expect for the ill-defined $x = -2$, the critical point of $x = 0.5$ should be considered. For $x > 0.5$, the derivative is positive and the function increases, in contrast to $x < 0.5$.*

3. *The requested coordinate is $(0.5, 0.2)$.*

■

---

## SOL-116  CH.SOL- 5.17.

1. $f'(x) = 6x^2 - 1$, *which entails the behavior of the function changes around the points $x = \pm \frac{1}{\sqrt{6}}$. The derivative is negative between $x = -\frac{1}{\sqrt{6}}$ and $x = \frac{1}{\sqrt{6}}$, i.e., it decreases in the domain, and increases otherwise.*

2. *The second derivative is $f''(x) = 12x$, which means the function is concave for negative $x$ values and convex otherwise.*

■

---

## SOL-117  CH.SOL- 5.18.

*The function should be derived according to each variable separately and be equated to 0, as follows:*

$$f_x(x, y) = 4x - y = 0, \quad f_y(x, y) = -y + 2y = 0.$$

*So, the solution to these equations yield the coordinate $(0, 0)$, and $f(0, 0) = 0$.*

*Let us derive the second order derivative, as follows:*

$$\frac{\partial^2 f}{\partial x^2}(x, y) = 4, \ \frac{\partial^2 f}{\partial y^2}(x, y) = 2, \ \frac{\partial^2 f}{\partial x \partial y}(x, y) = -1,$$

*Also, the following relation exists:*

$$D(x,y) = \frac{\partial^2 f}{\partial x^2}\frac{\partial^2 f}{\partial y^2} - \left(\frac{\partial^2 f}{\partial x \partial y}\right)^2 = 7\,,$$

*Thus, the critical point $(0,0)$ is a minimum.*                                    ■

### 5.3.9    The Gradient descent algorithm

**SOL-118**  **CH.SOL- 5.19.**

1. *It is the gradient of a function which is mathematically represented by:*

$$\nabla f(x,y) = \begin{pmatrix} \frac{\partial f(x,y)}{\partial x} \\ \frac{\partial f(x,y)}{\partial y} \end{pmatrix} \qquad (5.45)$$

2. *Increasing.*

3. *We will keep jumping between the same two points without ever reaching a minima.*

4. *This phenomena can be alleviated by using a **learning rate or step size**. For instance, $x+=2*\eta$ where $\eta$ is a learning rate with small value such as $\eta = 0.25$.*

5. *True.*

                                                                                    ■

**SOL-119**  **CH.SOL- 5.20.**

1. *The point (12,12) has two classes, so the classes cannot be separated by any line.*

2.

$$J(\theta) = \frac{1}{2m}\sum_{i=1}^{m}\left(\hat{y}_i - y_i\right)^2 \qquad (5.46)$$

3. *Simple but fundamental algorithm for minimizing $f$. Just repeatedly move in the direction of the negative gradient*

  (a) *Start with initial guess $\theta^{(0)}$, step size $\eta$*

  (b) *For $k = 1, 2, 3, \ldots$:*

    i. *Compute the gradient $\nabla f(\theta^{(k-1)})$*

    ii. *Check if gradient is close to zero; is so stop, otherwise continue*

    iii. *Update $\theta^{(k)} = \theta^{(k-1)} - \eta \nabla f(\theta^{(k-1)})$*

  (c) *Return final $\theta^{(k)}$ as approximate solution $\theta^*$*

### 5.3.10   The Backpropagation algorithm

**SOL-120**  **CH.SOL- 5.21.**

1. *The annotated parts of equation (5.21) appear in (5.47):*

$$\sigma(x) = \frac{e^x}{1 + e^x} = \textit{The Sigmoid activation function}$$

$$\sigma(x) \cdot \Big(1 - \sigma(x)\Big)$$

*The deriviative of the Sigmoid activation function =*

$$1Z = \textit{The input}$$

$$dZ = \textit{The error introduced by input } Z.$$

$$A = \textit{The output}$$

$$dA = \textit{The error introduced by output } A.$$

(5.47)

2. *Code snippet 5.13 provides an implementation of both the forward and backward passes for the sigmoid function.*

```
1  class Sigmoid:
2   def forward(self,x):
3    self.x = x
4    return 1/(1+np.exp(-x))
5
6   def backward(self, grad):
7    grad_input = self.x*(1-self.x) * grad
8    return grad_input
```

FIGURE 5.13: Forward and backward passes for the sigmoid activation function in pure Python.

■

### SOL-121  CH.SOL- 5.22.

*The key concept in this question is merely understanding that the transfer function and **its derivatives** are changing compared to traditional activation functions, namely:*

$$\frac{\partial E}{\partial y_k} = (y_k - d_k) \tag{5.48}$$

$$\frac{\partial E}{\partial net_k} = \frac{\partial E}{\partial y_k} \cdot \frac{\partial y_k}{\partial net_k} = (y_k - d_k) \cdot 2\cos(2net_k) \tag{5.49}$$

$$\Delta w_{jk} = -\eta \frac{\partial E}{\partial w_{jk}} = -\eta \frac{\partial E}{\partial net_k} \cdot \frac{\partial net_k}{\partial w_{jk}} = -\eta \cdot (y_k - d_k) \cdot 2\cos(2net_k) \cdot y_j \tag{5.50}$$

$$\frac{\partial E}{\partial y_j} = \sum_k \left( \frac{\partial E}{\partial net_k} \cdot \frac{\partial net_k}{\partial y_j} \right) = \sum_k \left( \frac{\partial E}{\partial net_k} w_{jk} \right) \tag{5.51}$$

$$\frac{\partial E}{\partial net_j} = \frac{\partial E}{\partial y_j} \cdot \frac{\partial y_j}{\partial net_j} = \frac{\partial E}{\partial y_j} \cdot 3net_j^2 \tag{5.52}$$

$$\begin{aligned}
\Delta w_{ij} &= -\eta \frac{\partial E}{\partial w_{ij}} \\
&= -\eta \frac{\partial E}{\partial net_j} \cdot \frac{\partial net_j}{\partial w_{ij}} \\
&= -\eta \cdot \left( \sum_k \left[ (y_k - d_k) \cdot 2\cos(2net_k) \cdot w_{jk} \right] \right) \cdot 3net_j^2 \cdot y_i
\end{aligned}$$ (5.53)

$\blacksquare$

### 5.3.11 Feed forward neural networks

### 5.3.12 Activation functions, Autograd/JAX

**SOL-122 ☑ CH.SOL- 5.23.**

1. *True.*

2. *True.*

$\blacksquare$

**SOL-123 ☑ CH.SOL- 5.24.**

*The answers are as follows:*

1. $h_\Theta(x) = g(\Theta^T x) = \frac{1}{1+e^{-\Theta^T x}}$.

2. *The decision boundary for the logistic sigmoid function is where $h_\Theta(x) = 0.5$ (values less than 0.5 mean false, values equal to or more than 0.5 mean true).*

3. *That there is a 80% chance that the instance is of the corresponding class, therefore:*

   - $h_\Theta(x) = g(\Theta_0 + \Theta_1 x_1 + \Theta_2 x_2)$. *We can predict $y = 1$ if $x_0 + x_1 + x_2 \geq 0$.*

4. *The code snippet in 5.14 implements the function using Autograd.*

```python
from torch.autograd import Function
class Sigmoid(Function):
 @staticmethod
 def forward(ctx, x):
  output = 1 / (1 + torch.exp(-x))
  ctx.save_for_backward(output)
  return output

 @staticmethod
 def backward(ctx, grad_output):
  output, = ctx.saved_tensors
  grad_x = output * (1 - output) * grad_output
  return grad_x
```

FIGURE 5.14: Forward and backward for the sigmoid function in Autograd.

5. The code snippet in 5.15 implements the function using Autograd.

```
1  from torch.autograd import Function
2  class ReLU(torch.autograd.Function):
3   @staticmethod
4   def forward(ctx, input):
5    ctx.save_for_backward(input)
6    return input.clamp(min=0)
7
8   @staticmethod
9   def backward(ctx, grad_output):
10   input, = ctx.saved_tensors
11   grad_input = grad_output.clone()
12   grad_input[input < 0] = 0
13   return grad_input
```

FIGURE 5.15: Forward and backward for the ReLU function in Autograd.

**SOL-124**  **CH.SOL- 5.25.** *The answers are as follows:*

1. *Code snippet 5.16 implements the forward pass using pure Python.*

```python
import numpy as np
xT = torch.abs(torch.tensor([[0.37,0.192,0.571]],
requires_grad=True)).type(torch.DoubleTensor)
xT_np=xT.detach().cpu().numpy()
print ("Input: \n",xT_np)
arctanh_values = np.arctanh(xT_np)
print ("Numpy:", arctanh_values)
> Numpy: [[0.38842311 0.1944129  0.64900533]]
```

FIGURE 5.16: Forward pass for equation (5.23) using pure Python.

2. *Code snippet 5.17 implements the forward pass using Autograd.*

```python
import torch
from torch.autograd import Function
class ArtanhFunction(Function):
  @staticmethod
  def forward(ctx, x):
   ctx.save_for_backward(x)
   r = (torch.log_(1 + x).sub_(torch.log_(1 - x))).mul_(0.5)
   return r
```

FIGURE 5.17: Forward pass for equation (5.23).

3. *Code snippet 5.18 implements the backward pass using Autograd.*

```
1  from torch.autograd import Function
2  class ArtanhFunction(Function):
3  @staticmethod
4   input, = ctx.saved_tensors
5   out= grad_output / (1 - input ** 2)
6   print ("backward:{}".format(out))
7   return out
```

FIGURE 5.18: Backward pass for equation (5.23).

4. *Code snippet 5.19 verifies the correctness of the implementation using gradcheck.*

```
1  import numpy as np
2
3  xT =
   ↪  torch.abs(torch.tensor([[0.11,0.19,0.57]],requires_grad=True))
4  .type(torch.DoubleTensor)
5  arctanh_values_torch = arctanhPyTorch(xT)
6  print ("Torch:", arctanh_values_torch)
7  from torch.autograd import gradcheck, Variable
8  f = ArtanhFunction.apply
9  test=gradcheck(lambda t: f(t), xT)
10 print(test)
11
12 > PyTorch version: 1.7.0
13 > Torch: tensor([[0.3884, 0.1944, 0.6490]], dtype=torch.float64,
14 >     grad_fn=<ArtanhFunctionBackward>)
15 > backward:tensor([[1.1586, 1.0383,1.4838]], dtype=torch.float64,
16       grad_fn=<CopyBackwards>)
```

FIGURE 5.19: Invoking arctanh using gradcheck

### 5.3.13    Dual numbers in AD

**SOL-125**  **CH.SOL- 5.26.**

*The answers are as follows:*

1. *The procedure of AD is to use verbatim text of a computer program which calculates a numerical value and to transform it into the text of a computer program called the transformed program which calculates the desired derivative values. The transformed computer program carries out these derivative calculations by repeated use of the chain rule however applied to **actual floating point values** rather than to a symbolic representation.*

2. *Dual numbers extend all numbers by adding a second component $x \mapsto x + \dot{x}\mathbf{d}$ where $x + \dot{x}$ is the **dual part**.*

3. *The following arithmetic operations are possible on DN:*

   (a) $\mathbf{d}^2 = 0$

   (b) $(x + \dot{x}\mathbf{d}) + (y + \dot{y}\mathbf{d}) = x + y + (\dot{x} + \dot{y})\mathbf{d}$

   (c) $-(x + \dot{x}\mathbf{d}) = -x - \dot{x}\mathbf{d}$

   (d) $\frac{1}{x + \dot{x}d} = \frac{1}{x} - \frac{\dot{x}}{x^2}d$

4. *For $f(x + \dot{x}\mathbf{d})$ the Taylor series expansion is:*

$$f(x + \dot{x}\mathbf{d}) = f(x) + \frac{f'(x)}{1!}\dot{x}\mathbf{d} + \ldots 0 \tag{5.54}$$

   *The* immediate *and important result is that all higher-order terms ($n >= 2$) **disappear** which provides closed-form mathematical expression that represents a function and its derivative.*

**SOL-126**  **CH.SOL- 5.27.**

*The answers are as follows:*

*1.*

$$\sin(x + \dot{x}\mathbf{d}) = \sin(x) + \cos(x)\dot{x}\mathbf{d} \tag{5.55}$$

*2. If we traverse the graph 5.9 from left to right we drive the following simple function:*

$$g(x) = 3 * x + 2 \tag{5.56}$$

*3. We know that:*

$$g(x) = 3 * x + 2 \tag{5.57}$$

$$g'(x) = 3 \tag{5.58}$$

*Now if we expand the function using DN:*

$$g(x + \dot{x}\mathbf{d}) = 3 * (x + \dot{x}\mathbf{d}) + 2 = \tag{5.59}$$
$$3 * x + 3 * (\dot{x}\mathbf{d}) + 2 \tag{5.60}$$

*Rearranging:*

$$3 * x + 2 + 3 * (\dot{x}\mathbf{d}) \tag{5.61}$$

*But since $g(x) = 3 * x + 2$ then:*

$$g(x + \dot{x}\mathbf{d}) = g(x) + g'(x)\dot{x}\mathbf{d} \tag{5.62}$$

*4. Evaluating the function $g(x)$ at $x = 2$ using DN we get:*

$$g(x = 2) = (3 * 2 + 2) + (3)\dot{x}\mathbf{d} = \tag{5.63}$$
$$8 + (3)\dot{x}\mathbf{d} \tag{5.64}$$

*5. The code snippet in 5.20 implements the function using Autograd.*

```
1  import autograd.numpy as np
2  from autograd import grad
3  x = np.array([2.0], dtype=float)
4  def f1(x):
5    return 3*x + 2
6  grad_f1 = grad(f1)
7  print(f1(x))  # > 8.0
8  print(grad_f1(x))  # > 3.0
```

FIGURE 5.20: Autograd

**SOL-127**  **CH.SOL- 5.28.** *The answers are as follows:*

1. *If we traverse the graph 5.9 from left to right we drive the following function:*

$$g(x) = 5 * x^2 + 4 * x + 1 \tag{5.65}$$

2. *We know that:*

$$g(x_1) = 5 * x^2 + 4 * x + 1 \tag{5.66}$$

$$g'(x_1) = 10 * x_1 + 4 \tag{5.67}$$

*Now if we expand the function using DN we get:*

$$g(x + \dot{x}\mathbf{d}) = 5 * (x + \dot{x}\mathbf{d})^2 + 4 * (x + \dot{x}\mathbf{d}) + 1 = \tag{5.68}$$

$$5 * (x^2 + 2 * x + \dot{x}\mathbf{d} + (\dot{x}\mathbf{d})^2) + 4 * x + 4 * (\dot{x}\mathbf{d}) + 1 \tag{5.69}$$

*However by definition* $(\mathbf{d}^2) = 0$ *and therefore that term vanishes. Rearranging the terms:*

$$(5 * x^2 + 4 * x + 1) + (10 * x + 4)\dot{x}\mathbf{d} \tag{5.70}$$

*But since* $g(x) = (5 * x^2 + 4 * x + 1)$ *then:*

$$g(x + \dot{x}\mathbf{d}) = g(x) + g'(x)\dot{x}\mathbf{d} \tag{5.71}$$

3. *Evaluating the function* $g(x)$ *at* $x = 5$ *using DN we get:*

$$g(x = 4) = (5 * 5^2 + 4 * 5 + 1) + (10 * 5 + 4)\dot{x}\mathbf{d} =$$
$$146 + (54)\dot{x}\mathbf{d} \tag{5.72}$$

4. *The code snippet in 5.21 implements the function using Autograd.*

```python
import autograd.numpy as np
from autograd import grad
x = np.array([5.0], dtype=float)
def f1(x):
 return 5*x**2 + 4*x +1
grad_f1 = grad(f1)
print(f1(x)) # > 146.0
print(grad_f1(x)) # > 54.0
```

FIGURE 5.21: Autograd

5.3.14  Forward mode AD

**SOL-128**  **CH.SOL- 5.29.**
*The answers are as follows:*

1. *The function $g(x)$ represented by the expression graph in 5.11 is:*

$$g(x) = A + B * ln(C) \tag{5.73}$$

2. *For a logarithmic function:*

$$\frac{d}{dx} \ln(x) = \frac{1}{x} \tag{5.74}$$

*Therefore, the partial derivatives for the function $g(x)$ are:*

$$\frac{\partial f}{\partial A} = 1$$
$$\frac{\partial f}{\partial B} = ln(C) \tag{5.75}$$
$$\frac{\partial f}{\partial C} = B * \frac{1}{C}$$

---

**SOL-129**  **CH.SOL- 5.30.**    *The answers are as follows:*

1. *True. Both directions yield the exact same results.*

2. *True. Reverse mode is more efficient than forward mode AD (why?).*

3. *True.*

4. *True.*

---

**SOL-130**  **CH.SOL- 5.31.**
*The answers are as follows:*

1. *The function is*

$$f(x_1, x_2) = x_1 x_2 + ln(x_1) \tag{5.76}$$

2. *The graph associated with the forward mode AD is as follows:*

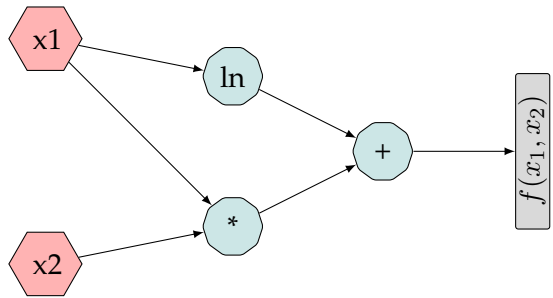

FIGURE 5.22: A Computation graph for $g(x_1, x_2)$ in 5.1

3. *The partial derivatives are:*

$$\frac{\partial f}{\partial x_1} = x_2 - \frac{1}{(x_1)}$$
$$\frac{\partial f}{\partial x_2} = x_1 \tag{5.77}$$

■

5.3.15   Forward mode AD table construction

**SOL-131** ✏ **CH.SOL- 5.32.**
  *The answers are as follows:*

1. *The graph with the intermediate values is depicted in (5.23)*

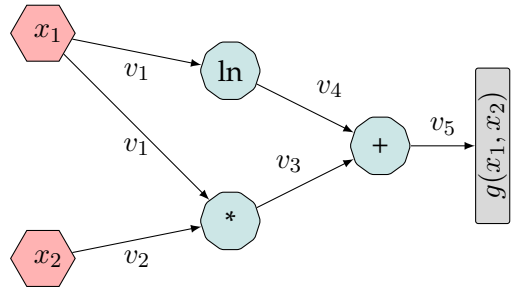

FIGURE 5.23: A derivative graph for $g(x_1, x_2)$ in 5.1

2. *Forward mode AD for $g(x_1, x_2) = \ln(x_1) + x_1 x_2$ evaluated at $(x_1, x_2) = (e^2, \pi)$.*

| Forward-mode function evaluation | | |
|---|---|---|
| $v_{-1} = x_1$ | $= e^2$ | |
| $v_0 \ = x_2$ | $= \pi$ | |
| $v_1 \ = \ln v_{-1}$ | $= \ln(e^2) = 2$ | |
| $v_2 \ = v_{-1} \times v_0$ | $= e^2 \times \pi = 23.2134$ | |
| $v_3 \ = v_1 + v_2$ | $2 + 23.2134 = 25.2134$ | |
| $f \ = v_3$ | $= \approx 25.2134$ | |

TABLE 5.1: Forward-mode AD table for $y = g(x_1, x_2) = \ln(x_1) + x_1 x_2$ evaluated at $(x_1, x_2) = (e^2; \pi)$ and setting $\dot{x}_1 = 1$ to compute $\frac{\partial y}{\partial x_1}$.

3. *The following Python code (5.24) proves that the numerical results are correct:*

```
1  import math
2  print (math.log(math.e*math.e) + math.e*math.e*math.pi)
3  > 25.2134^^I
```

FIGURE 5.24: Python code- AD of the function $g(x_1, x_2)$

4. *Seed values indicate the values by which the dependent and independent variables are initialized to before being propagated in a computation graph. For instance:*

$$\dot{v}_1 = \frac{\partial x_1}{\partial x_1} = 1$$
$$\dot{v}_2 = \frac{\partial x_2}{\partial x_1} = 0$$

*Therefore we set $\dot{x}_1 = 1$ to compute $\frac{\partial y}{\partial x_1}$.*

5. *Here we construct a table for the forward-mode AD for the derivative of $f(x_1, x_2) = \ln(x_1) + x_1 x_2$ evaluated at $(x_1, x_2) = (e^2, \pi)$ while setting $\dot{x}_1 = 1$ to compute $\frac{\partial y}{\partial x_1}$.. In forward-mode AD a derivative is called a **tangent**.*

   *In the derivation that follows, note that mathematically using manual differentiation:*

$$\frac{\mathrm{d}}{\mathrm{d}x_1} \left[\ln(x) + x_2 x\right]$$
$$= \frac{\mathrm{d}}{\mathrm{d}x_1}[\ln(x_1)] + x_2 \cdot \frac{\mathrm{d}}{\mathrm{d}x_1}[x_1]$$
$$= \frac{1}{x_1} + x_2 \cdot 1$$
$$= \frac{1}{x_1} + x_2$$

*and also since $\frac{d}{dx} \ln(x) = \frac{1}{x}$ then $\dot{v}_1 = \frac{1}{v_{-1}} * \dot{v}_{-1} = \dot{v}_{-1}/v_{-1} = \frac{1}{e^2} * 1 = 1/e^2$.*

Forward-mode AD derivative evaluation

$$v_{-1} = x_1 = e^2$$
$$v_0 \;\; = x_2 = \pi$$
$$\dot{v}_{-1} = \dot{x}_1 = 1$$
$$\dot{v}_0 \;\; = \dot{x}_2 = 0$$

---

$$\dot{v}_1 \;\; = \dot{v}_{-1}/v_{-1} = 1/e^2$$
$$\dot{v}_2 \;\; = \dot{v}_{-1} \times v_0 + \dot{v}_0 \times v_{-1} = 1 \times \pi + 0 \times e^2 = \pi$$
$$\dot{v}_4 \;\; = \dot{v}_1 + \dot{v}_2 = 1/e^2 + \pi$$

---

$$\dot{f} \;\; = \; \dot{v}_4 \; = \; 1/e^2 + \pi = \approx 3.2769$$

TABLE 5.3: Forward-mode AD table for $y = g(x_1, x_2) = \ln(x_1) + x_1 x_2$ evaluated at $(x_1, x_2) = (e^2; \pi)$ and setting $\dot{x}_1 = 1$ (seed values are mentioned here: 3) to compute $\frac{\partial y}{\partial x_1}$.

6. *The following Python code (5.25) proves that the numerical results are correct:*

```
1  import autograd.numpy as np
2  from autograd import grad
3  import math
4
5  x1 = math.e* math.e
6  x2 = math.pi
7
8  def f1(x1,x2):
9  return (np.log(x1) + x1*x2)
10
11 grad_f1 = grad(f1)
12
13 print(f1(x1,x2)) # > 25.2134
14 print(grad_f1(x1,x2)) # > 3.2769
```

FIGURE 5.25: Python code- AD of the function $g(x_1, x_2)$

### 5.3.16  Symbolic differentiation

### 5.3.17  Simple differentiation

**SOL-132**  **CH.SOL- 5.33.**

*The answers are as follows:*

1. *Approximate methods such as numerical differentiation suffer from numerical instability and truncation errors.*

2. *In symbolic differentiation, a symbolic expression for the derivative of a function is calculated. This approach is quite slow and requires symbols parsing and manipulation. For example, the number $\sqrt{2}$ is represented in SymPy as the object Pow(2,1/2). Since SymPy employees exact representations Pow(2,1/2)\*Pow(2,1/2) will always equal 2.*

**SOL-133**  **CH.SOL- 5.34.**

*1. First:*

```
import sympy
sympy.init_printing()
from sympy import Symbol
from sympy import diff, exp, sin, sqrt
y = Symbol('y')
y = sympy.Symbol("y")
sigmoid = 1/(1+sympy.exp(-y))^^I
```

FIGURE 5.26: Sigmoid in SymPy

*2. Second:*

```
sig_der=sym.diff(sigmoid, y)
```

FIGURE 5.27: Sigmoid gradient in SymPy

*3. Third:*

```
sig_der.evalf(subs={y:0})
> 0.25
```

FIGURE 5.28: Sigmoid gradient in SymPy

4. *The plot is depicted in 5.29.*

```
1 p = sym.plot(sig_der);
```

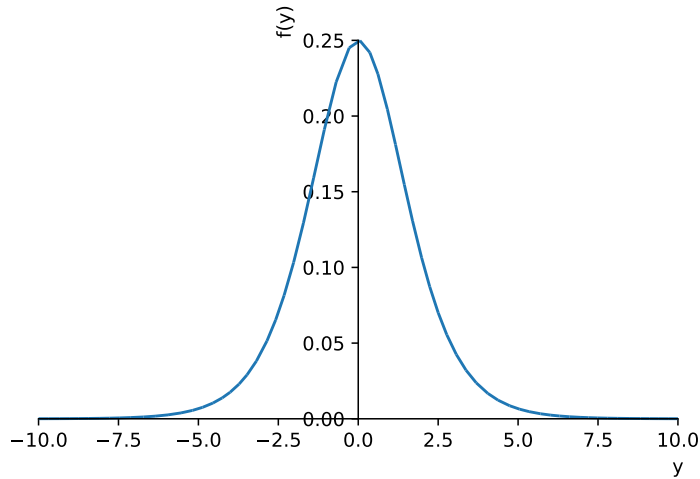

FIGURE 5.29: SymPy gradient of the Sigmoid() function

### 5.3.18  The Beta-Binomial model

**SOL-134**  **CH.SOL- 5.35.**
  *To correctly render the generated LaTeX in this problem, we import and configure several libraries as depicted in 5.30.*

```
import numpy as np
import scipy.stats as st
import matplotlib.pyplot as plt
import sympy as sp
sp.interactive.printing.
init_printing(use_latex=True)
from IPython.display import display, Math, Latex
maths = lambda s: display(Math(s))
latex = lambda s: display(Latex(s))^^I
```

FIGURE 5.30: SymPy imports

1. *The Likelihood function can be created as follows. Note the specific details of generating the Factorial function in SymPy.*

```
1 n = sp.Symbol('n', integer=True, positive=True)
2 r = sp.Symbol('r', integer=True, positive=True)
3 theta = sp.Symbol('theta')
4 # Create the function symbolically
5 from sympy import factorial
6 cNkSym= (factorial(n))/ (factorial(r) *factorial(n-r))
7 cNkSym.evalf()
8 binomSym= cNkSym*((theta **r)*(1-theta)**(n-r))
9 binomSym.evalf()
10 #Convert it to a Numpy-callable function
11 binomLambda = sp.Lambda((theta,r,n), binomSym)
12 maths(r"\operatorname{Bin}(r|n,\theta) = ")
13 display (binomLambda.expr)
14 #Evaluating the SymPy version results in:
15 > binomSym.subs({theta:0.5,r:50,n:100})
16 #Evaluating the pure Numpy version results in:
17 > binomLambda(0.5,50,100)= 0.07958923
```

FIGURE 5.31: Likelihood function using SymPy

*The Symbolic representation results in the following LaTeX:*

$$\mathrm{Bin}(r|n,\theta) = \frac{\theta^r \left(-\theta + 1\right)^{n-r} n!}{r! \left(n - r\right)!} \tag{5.78}$$

2. *The Beta distribution can be created as follows.*

```
1  a = sp.Symbol('a', integer=False, positive=True)
2  b = sp.Symbol('b', integer=False, positive=True)
3  #mu = sp.Symbol('mu', integer=False, positive=True)
4  # Create the function symbolically
5  G = sp.gamma
6  # The normalisation factor
7  BetaNormSym = G(a + b)/(G(a)*G(b))
8  # The functional form
9  BetaFSym = theta**(a-1) * (1-theta)**(b-1)
10 BetaSym=BetaNormSym * BetaFSym
11 BetaSym.evalf()  # this works
12 # Turn Beta into a function
13 BetaLambda = sp.Lambda((theta,a,b), BetaNormSym * BetaFSym)
14 maths(r"\operatorname{Beta}(\theta|a,b) = ")
15 display(BetaSym)
16 #Evaluating the SymPy version results in:
17 > BetaLambda(0.5,2,7)=0.4375
18 #Evaluating the pure Numpy version results in:
19 > BetaSym.subs({theta:0.5,a:2,b:7})=0.4375
```

FIGURE 5.32: Beta distribution using SymPy

*The result is:*

$$\text{Beta}(\theta|a,b) = \frac{\theta^{a-1}\varGamma(a+b)}{\varGamma(a)\varGamma(b)}\,(-\theta+1)^{b-1} \tag{5.79}$$

3. *The plot is depicted in 5.33.*

```python
%pylab inline
mus = arange(0,1,.01)
# Plot for various values of a and b
for ab in [(.1,.1),(.5,.5),(2,20),(2,3), (1,1)]:
    plot(mus, vectorize(BetaLambda)(mus,*ab), label="a=%s b=%s" % ab)
legend(loc=0)
xlabel(r"$\theta$", size=22)
```

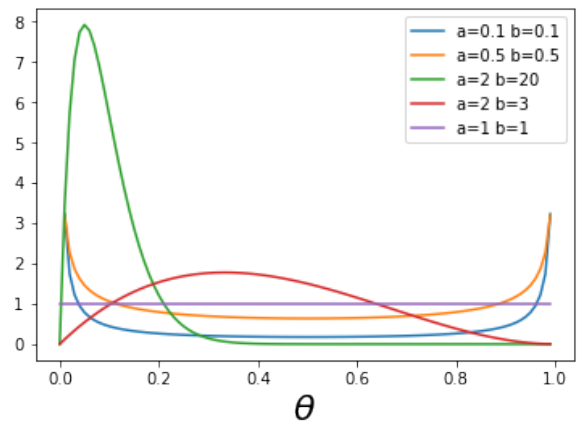

FIGURE 5.33: A plot of the Beta distribution

4. *We can find the posterior distribution by multiplying our Beta prior by the Binomial Likelihood.*

```
1  a = sp.Symbol('a', integer=False, positive=True)
2  b = sp.Symbol('b', integer=False, positive=True)
3  BetaBinSym=BetaSym * binomSym
4  # Turn Beta-bin into a function
5  BetaBinLambda = sp.Lambda((theta,a,b,n,r), BetaBinSym)
6  BetaBinSym=BetaBinSym.powsimp()
7  display(BetaBinSym)
8  maths(r"\operatorname{Beta}(\theta|a,b) \times
   ↪  \operatorname{Bin}(r|n,\theta) \propto %s" %
   ↪  sp.latex(BetaBinSym))
9  > BetaBinSym.subs({theta:0.5,a:2,b:7,n:10,r:3})= 0.051269
10 > BetaBinLambda (0.5,2,7, 10,3)= 0.051269
```

FIGURE 5.34: A plot of the Beta distribution

*The result is:*

$$\mathrm{Beta}(\theta|a,b) \times \mathrm{Bin}(r|n,\theta) \propto$$

$$\frac{\theta^{a+r-1}\left(-\theta+1\right)^{b+n-r-1} n!}{r!\,(n-r)!\,\Gamma(a)\,\Gamma(b)}\,\Gamma(a+b)$$

*So the posterior distribution has the same functional dependence on $\theta$ as the prior, it is just another Beta distribution.*

5. *Mathematically, the relationship is as follows:*

$$\textbf{Prior}:$$
$$\text{Beta}(\theta|a=2,b=7)$$
$$= 56\theta\left(-\theta+1\right)^{6}$$
$$\textbf{Likelihood}:$$
$$\text{Bin}(r=3|n=6,\theta) = 19600\theta^{3}\left(-\theta+1\right)^{47}$$
$$\textbf{Posterior(normalised)}:$$
$$\text{Beta}(\theta|2,7) \times \text{Bin}(3|50,\theta) = 1097600\theta^{4}\left(-\theta+1\right)^{53}$$

(5.80)

```
1  prior = BetaLambda(theta,2,7)
2  maths("\mathbf{Prior}:\operatorname{Beta}(\theta|a=2,b=7) = %s" %
   ↪  sp.latex(prior))
3  likelihood = binomLambda(theta,3,50)  # = binomLambda(0.5,3,10)
4  maths("\mathbf{Likelihood}: \operatorname{Bin}(r=3|n=6,\theta) =
   ↪  %s" % sp.latex(likelihood))
5  posterior = prior * likelihood
6  posterior=posterior.powsimp()
7  maths(r"\mathbf{Posterior
   ↪  (normalised)}:\operatorname{Beta}(\theta|2,7) \times
   ↪  \operatorname{Bin}(3|50,\theta)=%s"
8  posterior.subs({theta:0.5})
9  plt.plot(mus, (sp.lambdify(theta,posterior))(mus), 'r')
10 xlabel("$\\theta$", size=22)
```

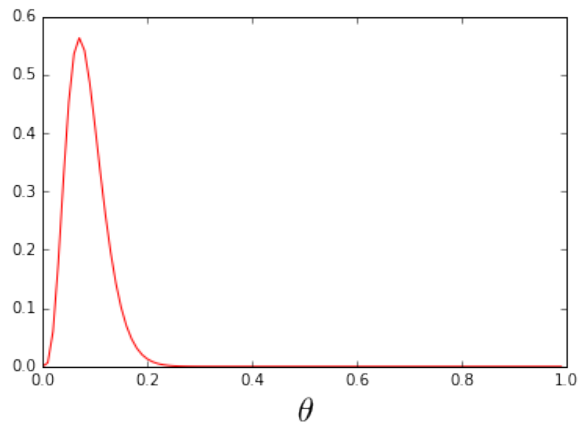

FIGURE 5.35: A plot of the Posterior with the provided data samples.

# PART IV

# BACHELORS

# CHAPTER

*The saddest aspect of life right now is that gathers knowledge faster than society gathers wisdom.*

— Isaac Asimov

# Contents

## 6.1   Introduction

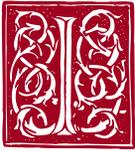Ntuition and practice demonstrate that a poor or an inferior choice may be altogether prevented merely by motivating a group (or an ensemble) of people with diverse perspectives to make a mutually acceptable choice. Likewise, in many cases, neural network ensembles significantly improve the generalization ability of single-model based AI systems [5, 11]. Shortly following the foundation of Kaggle, research in the field had started blooming; not only because researchers are advocating and using advanced ensembling approaches in almost every competition, but also by the **empirical** success of the top winning models. Though the whole process of training ensembles typically involves the utilization of dozens of GPUs and prolonged training periods, ensembling approaches enhance the predictive power of a single model. Though ensembling obviously has a significant impact on the performance of AI systems in general, research shows its effect is particularly dramatic in the field of neural networks [**Russakovsky_2015**, 1, 4, 7, 13]. Therefore, while we could examine combinations of any type of learning algorithms, the focus of this chapter is the combination of neural networks.

## 6.2   Problems

### 6.2.1   Bagging, Boosting and Stacking

---

**PRB-135 ❷ CH.PRB- 6.1.**
  *Mark all the approaches which can be utilized to boost a single model performance:*

  (i) *Majority Voting*

 (ii) *Using K-identical base-learning algorithms*

(iii) *Using K-different base-learning algorithms*

(iv) *Using K-different data-folds*

 (v) *Using K-different random number seeds*

(vi) *A combination of all the above approaches*

**PRB-136 ❷ CH.PRB- 6.2.**

*An argument erupts between two senior data-scientists regarding the choice of an approach for training of a very small medical corpus. One suggest that bagging is superior while the other suggests stacking. Which technique, bagging or stacking, in your opinion is **superior**? Explain in detail.*

(i) *Stacking since each classier is trained on all of the available data.*

(ii) *Bagging since we can combine as many classifiers as we want by training each on a different sub-set of the training corpus.*

**PRB-137 ❷ CH.PRB- 6.3.**

*Complete the sentence: A random forest is a type of a decision tree which utilizes **[bagging/boosting]***

**PRB-138 ❷ CH.PRB- 6.4.**

*The algorithm depicted in Fig. 6.1 was found in an old book about ensembling. Name the algorithm.*

---

**Algorithm 1:** Algo 1

---

**Data:** A set of training data, $Q$ with $N$ elements has been established

**while** *K times* **do**

    Create a random subset of $N'$ data by sampling from $Q$ containing the $N$ samples;

    $N' < N$;

    Execute algorithm Algo 2;

    Return all $N'$ back to $Q$

---

**Algorithm 2:** Algo 2

---

Choose a learner $h_m$;

**while** *K times* **do**

    Pick a training set and train with $h_m$;

---

FIGURE 6.1: A specific ensembling approach

**PRB-139 ❷ CH.PRB- 6.5.**

*Fig. 6.2 depicts a part of a specific ensembling approach applied to the models $x_1, x_2...x_k$. In your opinion, which approach is being utilized?*

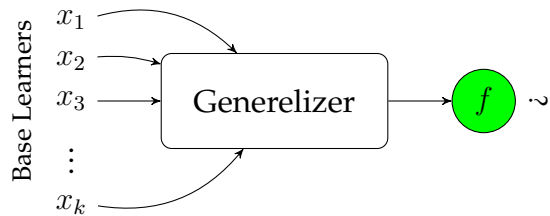

FIGURE 6.2: A specific ensembling approach

*(i) Bootstrap aggregation*

*(ii) Snapshot ensembling*

*(iii) Stacking*

*(iv) Classical committee machines*

---

### PRB-140 ❷ CH.PRB- 6.6.

*Consider training corpus consisting of balls which are glued together as triangles, each of which has either 1, 3, 6, 10, 15, 21, 28, 36, or 45 balls.*

1. *We draw several samples from this corpus as presented in Fig. 6.3 wherein each sample is equiprobable. What type of sampling approach is being utilized here?*

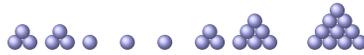

FIGURE 6.3: Sampling approaches

   (i) *Sampling without replacement*

   (ii) *Sampling with replacement*

2. *Two samples are drawn one after the other. In which of the following cases is the covariance between the two samples equals zero?*

   (i) *Sampling without replacement*

   (ii) *Sampling with replacement*

3. *During training, the corpus sampled with replacement and is divided into several folds as presented in Fig. 6.4.*

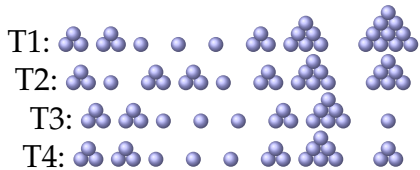

FIGURE 6.4: Sampling approaches

*If 10 balls glued together is a sample event that we know is hard to correctly classify, then it is impossible that we are using:*

   *(i) Bagging*

   *(ii) Boosting*

### 6.2.2 Approaches for Combining Predictors

**PRB-141 ❷ CH.PRB- 6.7.**
   *There are several methods by which the outputs of base classifiers can be combined to yield a single prediction. Fig. 6.5 depicts part of a specific ensembling approach applied to several CNN model predictions for a labelled data-set. Which approach is being utilized?*

   *(i) Majority voting for binary classification*

   *(ii) Weighted majority voting for binary classification*

   *(iii) Majority voting for class probabilities*

   *(iv) Weighted majority class probabilities*

   *(v) An algebraic weighted average for class probabilities*

   *(vi) An adaptive weighted majority voting for combining multiple classifiers*

```
1  l = []
2  for i,f in enumerate(filelist):
3    temp =  pd.read_csv(f)
4    l.append(temp)
5  arr = np.stack(l,axis=-1)
6  avg_results = pd.DataFrame(arr[:,:-1,:].mean(axis=2))
7  avg_results['image'] = l[0]['image']
8  avg_results.columns = l[0].columns
```

FIGURE 6.5: PyTorch code snippet for an ensemble

**PRB-142**  **CH.PRB- 6.8.**
*Read the paper **Neural Network Ensembles** [3] and then **complete the sentence**: If the average error rate for a specific instance in the corpus is less than [...]% and the respective classifiers in the ensemble produce independent [...], then when the number of classifiers combined approaches infinity, the expected error can be diminished to zero.*

**PRB-143**  **CH.PRB- 6.9.**
*True or false: A perfect ensemble comprises of highly correct classifiers that differ as much as possible.*

**PRB-144**  **CH.PRB- 6.10.**
*True or false: In bagging, we re-sample the training corpus with replacement and therefore this may lead to some instances being represented numerous times while other instances not to be represented at all.*

### 6.2.3    Monolithic and Heterogeneous Ensembling

**PRB-145**  **CH.PRB- 6.11.**

1. **True or false:** *Training an ensemble of a single monolithic architecture results in lower model diversity and possibly decreased model prediction accuracy.*

2. **True or false:** *The generalization accuracy of an ensemble increases with the number of well-trained models it consists of.*

3. **True or false:** *Bootstrap aggregation (or bagging), refers to a process wherein a CNN ensemble is being trained using a random subset of the training corpus.*

4. **True or false:** *Bagging assumes that if the single predictors have **independent errors**, then a majority vote of their outputs should be better than the individual predictions.*

---

### PRB-146 ❷ CH.PRB- 6.12.

*Refer to the papers:* **Dropout as a Bayesian Approximation** *[2] and* **Can You Trust Your Model's Uncertainty?** *[12] and answer the following question: Do deep ensembles achieve a better performance on out-of-distribution uncertainty benchmarks compared with Monte-Carlo (MC)-dropout?*

---

### PRB-147 ❷ CH.PRB- 6.13.

1. *In a transfer-learning experiment conducted by a researcher, a number of ImageNet-pretrained CNN classifiers, selected from Table 6.1 are trained on five different folds drawn from the same corpus. Their outputs are fused together producing a composite machine. Ensembles of these convolutional neural networks architectures have been extensively studies an evaluated in various ensembling approaches [4, 9]. Is it likely that the composite machine will produce a prediction with higher accuracy than that of any individual classifier? Explain why.*

| CNN Model | Classes | Image Size | Top-1 accuracy |
|-----------|---------|------------|----------------|
| ResNet152 | 1000 | 224 | 78.428 |
| DPN98 | 1000 | 224 | 79.224 |
| SeNet154 | 1000 | 224 | 81.304 |
| SeResneXT101 | 1000 | 224 | 80.236 |
| DenseNet161 | 1000 | 224 | 77.560 |
| InceptionV4 | 1000 | 299 | 80.062 |

TABLE 6.1: ImageNet-pretrained CNNs. Ensembles of these CNN architectures have been extensively studies and evaluated in various ensembling approaches.

2. **True or False**: *In a classification task, the result of ensembling is always superior.*

3. **True or False**: *In an ensemble, we want differently trained models converge to different local minima.*

---

**PRB-148 ❷ CH.PRB- 6.14.**

*In committee machines, mark all the combiners that do not make direct use of the input:*

(i) *A mixture of experts*

(ii) *Bagging*

(iii) *Ensemble averaging*

(iv) *Boosting*

---

**PRB-149 ❷ CH.PRB- 6.15.**

**True or False**: *Considering a binary classification problem ($y = 0$ or $y = 1$), ensemble averaging, wherein the outputs of individual models are linearly combined to produce a fused output is a form of a **static** committee machine.*

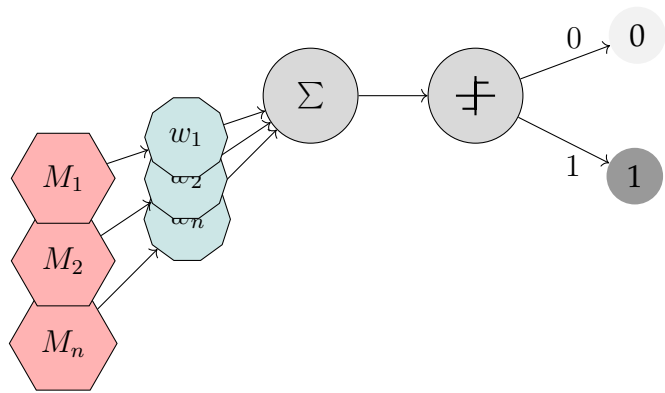

FIGURE 6.6: A typical binary classification problem.

**PRB-150 ❷ CH.PRB- 6.16.**

*True or false:* When using a single model, the risk of overfitting the data increases when the number of adjustable parameters is large compared to cardinality (i.e., size of the set) of the training corpus.

**PRB-151 ❷ CH.PRB- 6.17.**

*True or false:* If we have a committee of $K$ trained models and the errors are uncorrelated, then by averaging them the average error of a model is reduced by a factor of $K$.

### 6.2.4 Ensemble Learning

**PRB-152 ❷ CH.PRB- 6.18.**

1. *Define ensemble learning in the context of machine learning.*

2. *Provide examples of ensemble methods in classical machine-learning.*

3. ***True or false:*** *Ensemble methods usually have stronger generalization ability.*

4. *Complete the sentence: Bagging is **variance/bias** reduction scheme while boosting reduced **variance/bias**.*

6.2.5    Snapshot Ensembling

**PRB-153 ❷ CH.PRB- 6.19.**
   *Your colleague, a well-known expert in ensembling methods, writes the following pseudo code in Python shown in Fig. 6.7 for the training of a neural network. This runs inside a standard loop in each training and validation step.*

```python
import torchvision.models as models
...
models = ['resnext']
    for m in models:
        train ...
        compute VAL loss ...
        amend LR ...
        if (val_acc > 90.0):
            saveModel()
```

FIGURE 6.7: PyTorch code snippet for an ensemble

1. *What type of ensembling can be used with this approach? Explain in detail.*

2. *What is the main advantage of snapshot ensembling? What are the disadvantages, if any?*

**PRB-154 ❷ CH.PRB- 6.20.**
   *Assume further that your colleague amends the code as follows in Fig. 6.8.*

```
1  import torchvision.models as models
2  import random
3  import np
4  ...
5  models = ['resnext']
6      for m in models:
7          train ...
8          compute loss ...
9          amend LR ...
10         manualSeed= draw a new random number
11         random.seed(manualSeed)
12         np.random.seed(manualSeed)
13         torch.manual_seed(manualSeed)
14         if (val_acc > 90.0):
15             saveModel()
```

FIGURE 6.8: PyTorch code snippet for an ensemble

*Explain in detail what would be the possible effects of adding lines 10-13.*

### 6.2.6 Multi-model Ensembling

**PRB-155 ❷ CH.PRB- 6.21.**

1. *Assume your colleague, a veteran in DL and an expert in ensembling methods writes the following Pseudo code shown in Fig. 6.9 for the training of several neural networks. This code snippet is executed inside a standard loop in each and every training/validation epoch.*

```
1  import torchvision.models as models
2  ...
3  models = ['resnext','vgg','dense']
4     for m in models:
5        train ...
6        compute loss/acc ...
7        if (val_acc > 90.0):
8           saveModel()
```

FIGURE 6.9: PyTorch code snippet for an ensemble

*What type of ensembling is being utilized in this approach? Explain in detail.*

2. *Name one method by which NN models may be combined to yield a single prediction.*

### 6.2.7    Learning-rate Schedules in Ensembling

**PRB-156 ❷ CH.PRB- 6.22.**

1. *Referring to Fig. (6.10) which depicts a specific learning rate schedule, describe the basic notion behind its mechanism.*

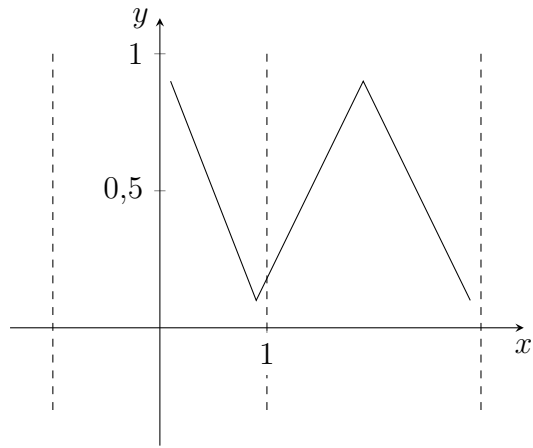

FIGURE 6.10: A learning rate schedule.

2. *Explain how cyclic learning rates [10] can be effective for the training of convolutional neural networks such as the ones in the code snippet of Fig. 6.10.*

3. *Explain how a cyclic cosine annealing schedule as proposed by Loshchilov [10] and [13] is used to converge to multiple local minima.*

## 6.3 Solutions

### 6.3.1 Bagging, Boosting and Stacking

**SOL-135 ✎ CH.SOL- 6.1.**
    *All the presented options are correct.* ■

**SOL-136 ✎ CH.SOL- 6.2.**
    *The correct choice would be stacking. In cases where the given corpus is small, we would most likely prefer training our models on the **full** data-set.* ■

**SOL-137 ✎ CH.SOL- 6.3.**
    *A random forest is a type of a decision tree which utilizes bagging.* ■

**SOL-138**  **CH.SOL- 6.4.**

*The presented algorithm is a classic bagging.*

**SOL-139**  **CH.SOL- 6.5.**

*The approach which is depicted is the **first phase** of stacking. In stacking, we first (phase 0) predict using several base learners and then use a generalizer (phase 1) that learns on top of the base learners predictions.*

**SOL-140**  **CH.SOL- 6.6.**

1. *Sampling with replacement*

2. *Sampling without replacement*

3. *This may be mostly a result of bagging, since in boosting we would have expected misscorrectly classified observations to repeatedly appear in subsequent samples.*

### 6.3.2    Approaches for Combining Predictors

**SOL-141**  **CH.SOL- 6.7.**

*An Algebraic weighted average for class probabilities.*

**SOL-142**  **CH.SOL- 6.8.**

*This is true, [3] provides a mathematical proof.*

**SOL-143**  **CH.SOL- 6.9.**

*This is true. For extension, see instance [8].*

**SOL-144**  **CH.SOL- 6.10.**

*This is true. In a bagging approach, we first randomly draw (with replacement), K ex-*

*amples where K is the size of the original training corpus therefore leading to an imbalanced representation of the instances.* ■

### 6.3.3 Monolithic and Heterogeneous Ensembling

**SOL-145**  **CH.SOL- 6.11.**

1. **True** *Due to their lack of diversity, an ensemble of monolithic architectures tends to perform worse than an heterogeneous ensemble.*

2. **True** *This has be consistently demonstrated in [11, 5].*

3. **True** *In [6] there is a discussion about both using the whole corpus and a subset much like in bagging.*

4. **True** *The total error decreases with the addition of predictors to the ensemble.*

■

**SOL-146**  **CH.SOL- 6.12.**
*Yes, they do.* ■

**SOL-147**  **CH.SOL- 6.13.**

1. *Yes, it is very likely, especially if their errors are independent.*

2. **True** *It may be proven that ensembles of models perform at least as good as each of the ensemble members it consists of.*

3. **True** *Different local minima add to the diversification of the models.*

■

**SOL-148**  **CH.SOL- 6.14.**
*Boosting is the only one that does not.* ■

**SOL-149**  **CH.SOL- 6.15.**

*False By definition, static committee machines use* **only** *the output of the single predictors.* ∎

**SOL-150**  **CH.SOL- 6.16.**

*True* ∎

**SOL-151**  **CH.SOL- 6.17.**

*False Though this may be theoretically true, in practice the errors are rarely uncorrelated and therefore the actual error can not be reduced by a factor of $K$.* ∎

### 6.3.4    Ensemble Learning

**SOL-152**  **CH.SOL- 6.18.**

1. *Ensemble learning is an excellent machine learning idea which displays noticeable benefits in many applications, one such notable example is the widespread use of ensembles in Kaggle competitions. In an ensemble several individual models (for instance ResNet18 and VGG16) which were trained on the same corpus, work in tandem and during inference, their predictions are fused by a pre-defined strategy to yield a single prediction.*

2. *In classical machine learning Ensemble methods usually refer to bagging, boosting and the linear combination of regression or classification models.*

3. ***True** The stronger generalization ability stems from the voting power of diverse models which are joined together.*

4. *Bagging is* **variance** *reduction scheme while boosting reduced* **bias**.

∎

### 6.3.5    Snapshot Ensembling

**SOL-153**  **CH.SOL- 6.19.**

1. *Since only a single model ie being utilized, this type of ensembling is known as snapshot ensembling. Using this approach, during the training of a neural network and in each epoch, a snapshot, e.g. the weights of a trained instance of a model (a PTH file in PyTorch nomenclature) are persisted into permanent storage whenever a certain performance metrics, such as accuracy or loss is being surpassed. Therefore the name "snapshot"; weights of the neural network are being snapshot at specific instances in time. After several such epochs the top-5 performing Snapshots which converged to local minima [4] are combined as part of an ensemble to yield a single prediction.*

2. *Advantages: during a single training cycle, many model instances may be collected. Disadvantages: inherent lack of diversity by virtue of the fact that the same models is being repeatedly used.*

**SOL-154**  **CH.SOL- 6.20.**

*Changing the random seed at each iteration/epoch, helps in introducing variation which may contribute to diversifying the trained neural network models.*

### 6.3.6 Multi-model Ensembling

**SOL-155**  **CH.SOL- 6.21.**

1. *Multi-model ensembling.*

2. *Both averaging and majority voting.*

### 6.3.7 Learning-rate Schedules in Ensembling

**SOL-156**  **CH.SOL- 6.22.**

1. *Capturing the best model of each training cycle allows to obtain multiple models settled on various local optima from cycle to cycle at the cost of training a single mode*

2. *The approach is based on the non-convex nature of neural networks and the ability to converge and escape from local minima using a specific schedule to adjust the learning rate during training.*

3. *Instead of monotonically decreasing the learning rate, this method lets the learning rate cyclically vary between reasonable boundary values.*

## Contents

## 7.1   Introduction

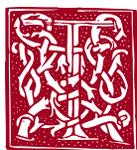

HE extraction of an n-dimensional feature vector (FV) or an *embedding* from one (or more) layers of a pre-trained CNN, is termed *feature extraction* (FE). Usually, FE works by first removing the last fully connected (FC) layer from a CNN and then treating the remaining layers of the CNN as a fixed FE. As exemplified in Fig. (7.1) and Fig. (7.2), applying this method to the ResNet34 architecture, the resulting FV consists of 512 floating point values. Likewise, applying the same logic on the ResNet152 architecture, the resulting FV has 2048 floating point elements.

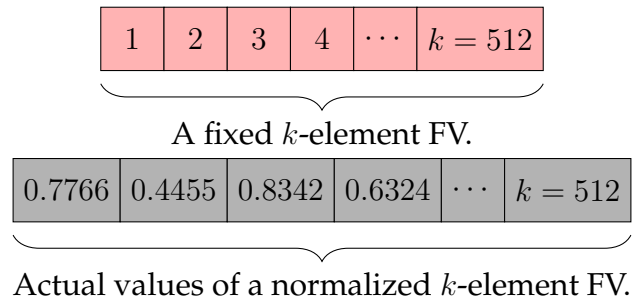

A fixed $k$-element FV.

Actual values of a normalized $k$-element FV.

FIGURE 7.1: A one-dimensional 512-element embedding for a single image from the Res-Net34 architecture. While any neural network can be used for FE, depicted is the ResNet CNN architecture with 34 layers.

```
1  import torchvision.models as models
2  ...
3  res_model = models.resnet34(pretrained=True)
```

FIGURE 7.2: PyTorch decleration for a pre-trained ResNet34 CNN (simplified).

The premise behind FE is that CNNs which were originally trained on the ImageNet Large Scale Visual Recognition Competition [7], can be adapted and used (for instance in a classification task) on a completely different (target) domain **without any additional training** of the CNN layers. The power of a CNN to do so lies in its ability to generalize well beyond the original data-set it was trained on, therefore FE on a new target data-set involves no training and requires only inference.

## 7.2   Problems

### 7.2.1   CNN as Fixed Feature Extractor

Before attempting the problems in this chapter you are highly encouraged to read the following papers [1, 3, 7]. In many DL job interviews, you will be presented with a paper you have never seen before and subsequently be asked questions about it; so reading these references would be an excellent simulation of this real-life task.

**PRB-157 ❷ CH.PRB- 7.1.**

*True or False*: While AlexNet [4] used $11 \times 11$ sized filters, the main novelty presented in the VGG [8] architecture was utilizing filters with much smaller spatial extent, sized $3 \times 3$.

**PRB-158 ❷ CH.PRB- 7.2.**

*True or False*: Unlike CNN architectures such as AlexNet or VGG, ResNet **does not** have any hidden FC layers.

**PRB-159 ❷ CH.PRB- 7.3.**

Assuming the VGG-Net has $138, 357, 544$ *floating point parameters, what is the physical size in Mega-Bytes (MB) required for persisting a trained instance of VGG-Net on permanent storage?*

**PRB-160 ❷ CH.PRB- 7.4.**

*True or False*: Most attempts at researching image representation using FE, focused solely on reusing the activations obtained from layers close to the output of the CNN, and more specifically the fully-connected layers.

**PRB-161 ❷ CH.PRB- 7.5.**

*True or False*: FE in the context of deep learning is particularly useful when the target problem does not include enough labeled data to successfully train CNN that generalizes well.

**PRB-162 ❷ CH.PRB- 7.6.**

Why is a CNN trained on the ImageNet dataset [7] a good candidate for a source problem?

**PRB-163 ❷ CH.PRB- 7.7.**

*Complete the missing parts regarding the VGG19 CNN architecture:*

1. *The VGG19 CNN consists of [...] layers.*

2. *It consists of [...] convolutional and 3 [...] layers.*

3. *The input image size is [...].*

4. *The number of input channels is [...].*

5. *Every image has it's mean RGB value [subtracted / added].*

6. *Each convolutional layer has a [small/large] kernel sized [...].*

7. *The number of pixels for padding and stride is [...].*

8. *There are 5 [...] layers having a kernel size of [...] and a stride of [...] pixels.*

9. *For non-linearity a [rectified linear unit (ReLU [5])/sigmoid] is used.*

10. *The [...] FC layers are part of the linear classifier.*

11. *The first two FC layers consist of [...] features.*

12. *The last FC layer has only [...] features.*

13. *The last FC layer is terminated by a [...] activation layer.*

14. *Dropout [is / is not] being used between the FC layers.*

---

**PRB-164** ❷ **CH.PRB- 7.8.**

 *The following question discusses the method of fixed feature extraction from layers of the VGG19 architecture [8] for the classification of pancreatic cancer. It depicts FE principles which are applicable with minor modifications to other CNNs as well. Therefore, if you happen to encounter a similar question in a job interview, you are likely be able to cope with it by utilizing the same logic. In Fig. (9.7) three different classes of pancreatic cancer are displayed: A, B and C, curated from a dataset of 4K Whole Slide Images (WSI) labeled by a board certified pathologist. Your task is to use FE to correctly classify the images in the dataset.*

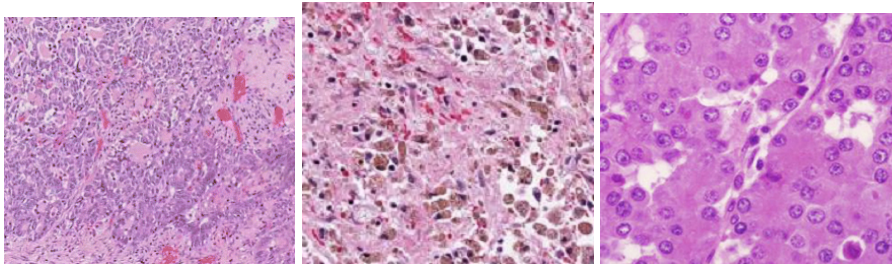

FIGURE 7.3: A dataset of 4K histopathology WSI from three severity classes: A, B and C.

*Table (9.3) presents an incomplete listing of the of the VGG19 architecture [8]. As depicted, for each layer the number of filters (i.e., neurons with unique set of parameters), learnable parameters (weights,biases), and FV size are presented.*

| Layer name | #Filters | #Parameters | # Features |
|---|---|---|---|
| conv4_3 | 512 | 2.3M | 512 |
| fc6 | 4,096 | 103M | 4,096 |
| fc7 | 4,096 | 17M | 4,096 |
| output | 1,000 | 4M | - |
| **Total** | **13,416** | **138M** | **12,416** |

TABLE 7.1: Incomplete listing of the VGG19 architecture

1. *Describe how the VGG19 CNN may be used as fixed FE for a classification task. In your answer be as detailed as possible regarding the stages of FE and the method used for classification.*

2. *Referring to Table (9.3), suggest **three** different ways in which features can be extracted from a trained VGG19 CNN model. In each case, state the extracted feature layer name and the size of the resulting FE.*

3. *After successfully extracting the features for the 4K images from the dataset, how can you now **classify** the images into their respective categories?*

**PRB-165 ❷ CH.PRB- 7.9.**

*Still referring to Table (9.3), a data scientist suggests using the output layer of the VGG19 CNN as a fixed FE. What is the main advantage of using this layer over using for instance, the $fc7$ layer? (Hint: think about an ensemble of feature extractors)*

**PRB-166 ❷ CH.PRB- 7.10.**

*Still referring to Table (9.3) and also to the code snippet in Fig. (7.4), which represents a new CNN derived from the VGG19 CNN:*

```python
import torchvision.models as models
...
class VGG19FE(torch.nn.Module):
 def __init__(self):
  super(VGG19FE, self).__init__()
  original_model = models.VGG19(pretrained=[???])
  self.real_name = (((type(original_model).__name__)))
  self.real_name = "vgg19"

  self.features = [???]
  self.classifier = torch.nn.Sequential([???])
  self.num_feats = [???]

 def forward(self, x):
  f = self.features(x)
  f = f.view(f.size(0), -1)
  f = [???]
  print (f.data.size())
  return f
```

FIGURE 7.4: PyTorch code snippet for extracting the $fc7$ layer from a pre-trained VGG19 CNN model.

1. *Complete line 6; what should be the value of* $\boxed{pretrained}$ *?*

2. *Complete line 10; what should be the value of* $\boxed{self.features}$ *?*

3. *Complete line 12; what should be the value of* $\boxed{self.num\_feats}$ *?*

4. *Complete line 17; what should be the value of* $\boxed{f}$ *?*

---

**PRB-167 ❷ CH.PRB- 7.11.**

*We are still referring to Table (9.3) and using the skeleton code provided in Fig. (7.5) to derive a new CNN entitled **ResNetBottom** from the ResNet34 CNN, to extract a 512-dimensional FV for a given input image. Complete the code as follows:*

1. *The value of* $\boxed{self.features}$ *in line 7.*

2. *The* $\boxed{forward}$ *method in line 11.*

```python
import torchvision.models as models
res_model = models.resnet34(pretrained=True)
class ResNetBottom(torch.nn.Module):
 def __init__(self, original_model):
   super(ResNetBottom, self).__init__()
   self.features = [???]

 def forward(self, x):
  x = [???]
  x = x.view(x.size(0), -1)
  return x
```

FIGURE 7.5: PyTorch code skeleton for extracting a 512-dimensional FV from a pre-trained ResNet34 CNN model.

**PRB-168  CH.PRB- 7.12.**

*Still referring to Table (9.3), the PyTorch based pseudo code snippet in Fig. (7.6) returns the 512-dimensional FV from the modified ResNet34 CNN, given a 3-channel RGB image as an input.*

```python
import torchvision.models as models
from torchvision import transforms
...

test_trans = transforms.Compose([
 transforms.Resize(imgnet_size),
 transforms.ToTensor(),
 transforms.Normalize([0.485, 0.456, 0.406],
 [0.229, 0.224, 0.225])])

def ResNet34FE(image, model):
 f=None
 image = test_trans(image)
 image = Variable(image, requires_grad=False).cuda()
 image= image.cuda()
 f = model(image)
 f = f.view(f.size(1), -1)
 print ("Size : {}".format(f.shape))
 f = f.view(f.size(1),-1)
 print ("Size : {}".format(f.shape))
 f =f.cpu().detach().numpy()[0]
 print ("Size : {}".format(f.shape))
 return f
```

FIGURE 7.6: PyTorch code skeleton for extracting a 512-dimensional FV from a pre-trained ResNet34 CNN model.

*Answer the following questions regarding the code in Fig. (7.6):*

1. What is the purpose of $\boxed{test\_trans}$ in line 5?

2. Why is the parameter $\boxed{requires\_grad}$ set to False in line 14?

3. What is the purpose of $\boxed{f.cpu()}$ in line 23?

4. What is the purpose of $\boxed{detach()}$ in line 23?

5. What is the purpose of $\boxed{numpy()[0]}$ in line 23?

### 7.2.2    Fine-tuning CNNs

**PRB-169 ❷ CH.PRB- 7.13.**

  Define the term **fine-tuning (FT) of an ImageNet pre-trained CNN**.

**PRB-170 ❷ CH.PRB- 7.14.**

  Describe three different methods by which one can fine-tune an ImageNet pre-trained CNN.

**PRB-171 ❷ CH.PRB- 7.15.**

  Melanoma is a lethal form of malignant skin cancer, frequently misdiagnosed as a benign skin lesion or even left completely undiagnosed.

  In the United States alone, melanoma accounts for an estimated $6,750$ deaths per annum [6]. With a 5-year survival rate of $98\%$, early diagnosis and treatment is now more likely and possibly the most suitable means for melanoma related death reduction. Dermoscopy images, shown in Fig. (7.7) are widely used in the detection and diagnosis of skin lesions. Dermatologists, relying on personal experience, are involved in a laborious task of manually searching dermoscopy images for lesions.

  Therefore, there is a very real need for automated analysis tools, providing assistance to clinicians screening for skin metastases. In this question, you are tasked with addressing some of the fundamental issues DL researchers face when building deep learning pipelines. As suggested in [3], you are going to use ImageNet pre-trained CNN to resolve a classification task.

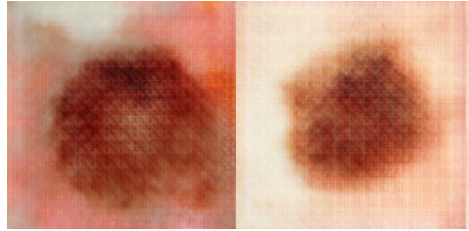

FIGURE 7.7: **Skin lesion categories.** An exemplary visualization of melanoma.

1. *Given that the skin lesions fall into seven distinct categories, and you are training using cross-entropy loss, how should the classes be represented so that a typical PyTorch training loop will successfully converge?*

2. *Suggest several data augmentation techniques to augment the data.*

3. *Write a code snippet in PyTorch to adapt the CNN so that it can predict 7 classes instead of the original source size of* 1000.

4. *In order to fine tune our CNN, the (original) output layer with* 1000 *classes was removed and the CNN was adjusted so that the (new) classification layer comprised seven softmax neurons emitting posterior probabilities of class membership for each lesion type.*

### 7.2.3 Neural style transfer, NST

Before attempting the problems in the section, you are strongly recommended to read the paper: "*A Neural Algorithm of Artistic Style*" [2].

**PRB-172 ❷ CH.PRB- 7.16.**
 *Briefly describe how neural style transfer (NST) [2] works.*

**PRB-173 ❷ CH.PRB- 7.17.**
 **Complete the sentence**: *When using the VGG-19 CNN [8] for neural-style transfer, there different images are involved. Namely they are: [...], [...] and [...].*

**PRB-174 ❷ CH.PRB- 7.18.**

*Refer to Fig. 7.8 and answer the following questions:*

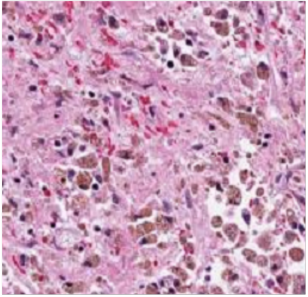

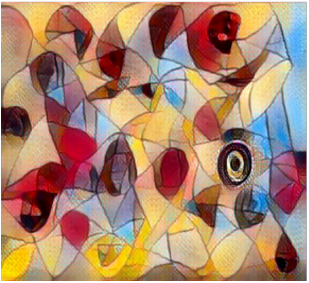

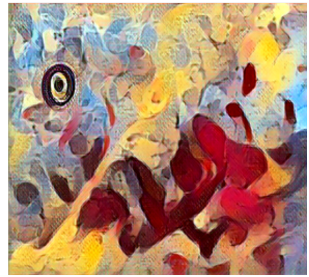

FIGURE 7.8: Artistic style transfer using the style of Francis Picabia's Udnie painting.

1. *Which loss is being utilized during the training process?*

2. *Briefly describe the use of activations in the training process.*

**PRB-175 ❷ CH.PRB- 7.19.**

*Still referring to Fig. 7.8:*

1. *How are the activations utilized in comparing the content of the content image to the content of the combined image?.*

2. *How are the activations utilized in comparing the style of the content image to the*

*style of the combined image?*.

---

**PRB-176 ❷ CH.PRB- 7.20.**

*Still referring to Fig. 7.8. For a new style transfer algorithm, a data scientist extracts a feature vector from an image using a pre-trained ResNet34 CNN (7.9).*

```
import torchvision.models as models
...
res_model = models.resnet34(pretrained=True)
```

FIGURE 7.9: PyTorch declaration for a pre-trained ResNet34 CNN.

*He then defines the cosine similarity between two vectors:*
$u = \{u_1, u_2, \ldots, u_N\}$ *and :*
$v = \{v_1, v_2, \ldots, v_N\}$
*as:*

$$\text{sim}(u, v) = \frac{u \cdot v}{|u||v|} = \frac{\sum_{i=1}^{N} u_i v_i}{\sqrt{\left(\sum_{i=1}^{N} u_i^2\right)\left(\sum_{i=1}^{N} v_i^2\right)}}$$

*Thus, the cosine similarity between two vectors measures the **cosine of the angle** between the vectors irrespective of their magnitude. It is calculated as the dot product of two numeric vectors, and is normalized by the product of the length of the vectors.*

*Answer the following questions:*

1. *Define the term* Gram matrix.

2. *Explain in detail how vector similarity is utilised in the calculation of the* Gram matrix *during the training of NST.*

## 7.3 Solutions

### 7.3.1 CNN as Fixed Feature Extractor

**SOL-157**  **CH.SOL- 7.1.**

*True. The increased depth in VGG-Net was made possible using smaller filters without substantially increasing the number of learnable parameters. Albeit an unwanted side effect of the usage of smaller filters is the increase in the number of filters per-layer.* ■

**SOL-158**  **CH.SOL- 7.2.**

*True. The ResNet architecture terminates with a global average pooling layer followed by a K-way FC layer with a softmax activation function, where K is the number of classes (ImageNet has 1000 classes). Therefore, the ResNet has no hidden FC layers.* ■

**SOL-159**  **CH.SOL- 7.3.** *Note that* $1 bit = 0.000000125$ *MB, therefore:*

$$138,357544 \times 32 = 4427441408 bits = 553.430176 \, MB. \tag{7.1}$$

■

**SOL-160**  **CH.SOL- 7.4.**

*True. There are dozens of published papers supporting this claim. You are encouraged to search them on Arxiv or Google Scholar.* ■

**SOL-161**  **CH.SOL- 7.5.**

*True. One of the major hurdles of training a medical AI system is the lack of annotated data. Therefore, extensive research is conducted to exploit ways for FE and transfer learning, e.g., in the application of ImageNet trained CNNs, to target datasets in which labeled data is scarce.* ■

**SOL-162**  **CH.SOL- 7.6.**

*There are two main reasons why this is possible:*

1. *The huge number of images inside the ImageNet dataset ensures a CNN model that generalizes to additional domains, like the histopathology domain, which is substantially different from the original domain the model was trained one (e.g., cats and dogs).*

2. *A massive array of disparate visual patterns is produced by an ImageNet trained CNN, since it consists of $1,000$ different groups.*

■

## SOL-163  CH.SOL- 7.7.

*Complete the missing parts regarding the VGG19 CNN architecture:*

1. *The VGG19 CNN consists of* 19 *layers.*

2. *It consists of* 5 *convolutional and 3* FC *layers.*

3. *The input image size is* 244*, the default size most ImageNet trained CNNs work on.*

4. *The number of input channels is* 3*.*

5. *Every image has its mean RGB value* subtracted*. (why?)*

6. *Each convolutional layer has a* small *kernel sized* $3 \times 3$*. (why?)*

7. *The number of pixels for padding and stride is the same and equals* 1*.*

8. *There are 5* **convolutional** *layers having a kernel size of* $2 \times 2$ *and a stride of* 2 *pixels.*

9. *For non-linearity a* rectified linear unit (ReLU [5]) *is used.*

10. *The* 3 *FC layers are part of the linear classifier.*

11. *The first two FC layers consist of* 4096 *features.*

12. *The last FC layer has only* 1000 *features.*

13. *The last FC layer is terminated by a* softmax *activation layer.*

14. *Dropout* is *being used between the FC layers.*

■

## SOL-164  CH.SOL- 7.8.

1. *One or more layers of the VGG19 CNN are selected for extraction and a new CNN is designed on top of it. Thus, during inference our target layers are extracted and not the original softmax layer. Subsequently, we iterate and run **inference** over all the images in our pancreatic cancer data-set, extract the features, and persist them to permanent storage such as a solid-state drive (SSD) device. Ultimately, each image has a corresponding FV.*

2. *Regarding the VGG19 CNN, there are numerous ways of extracting and combining features from different layers. Of course, these different layers, e.g., the FC, conv4_3, and fc7 layer may be combined together to form a larger feature vector. To determine which method works best, you shall have to experiment on your data-set; there is no way of a-priory determining the optimal combination of layers. Here are several examples:*

   (a) *Accessing the* $\boxed{\textit{last FC layer}}$ *resulting in a* 1000-D FV. *The output is the score for each of the* 1000 *classes of the ImageNet data-set.*

   (b) *Removing the* $\boxed{\textit{last FC layer}}$ *leaves the fc7 layer, resulting in a* 4096-D FV.

   (c) *Directly accessing the* $\boxed{\textit{conv4\_3 layer}}$ *results in a* 512-D FV.

3. *Once the FVs are extracted, we can train any linear classifier such as an SVM or softmax classifier on the FV data-set, and not on the original images.*

■

---

**SOL-165**  **CH.SOL- 7.9.**

One benefit of using the FC layer is that other ImageNet CNNs can be used in tandem with the VGG19 to create an ensemble since they all produce the same 1000-D sized FV.    ■

---

**SOL-166**  **CH.SOL- 7.10.** *The full code is presented in Fig. (7.10).*

```python
import torchvision.models as models
...
class VGG19FE(torch.nn.Module):
 def __init__(self):
  super(VGG19FE, self).__init__()
  original_model = models.VGG19(pretrained=True)
  self.real_name = (((type(original_model).__name__)))
  self.real_name = "vgg19"

  self.features = original_model.features
  self.classifier = torch.nn.Sequential(
  (*list(original_model.classifier.
  children())[:-1]))
  self.num_feats = 4096

 def forward(self, x):
  f = self.features(x)
  f = f.view(f.size(0), -1)  # (1, 4096) -> (4096,)
  f = self.classifier(f)
  print (f.data.size())
  return f
```

FIGURE 7.10: PyTorch code snippet for extracting the $fc7$ layer from a pre-trained VGG19 CNN model.

1. *The value of the parameter* $\boxed{pretrained}$ *should be True in order to instruct PyTorch to load an ImageNet trained weights.*

2. *The value of* $\boxed{self.features}$ *should be* $\boxed{original\_model.features}$. *This is because we like to retain the layers of the original classifier (original_model).*

3. *The value of* $\boxed{self.num\_feats}$ *should be* $\boxed{4096}$. *(Why?)*

4. *The value of* $\boxed{f}$ *should be* $\boxed{self.classifier(f)}$ *since our newly created CNN has to be invoked to generate the FV.*

**SOL-167**  **CH.SOL- 7.11.**

1. *Line number 7 in Fig. (7.11) takes care of extracting the the correct 512-D FV.*

2. *Line number 11 in Fig. (7.11) extracts the correct 512-D FV by creating a sequential module on top of the existing features.*

```python
import torchvision.models as models
res_model = models.resnet34(pretrained=True)
class ResNetBottom(torch.nn.Module):
def __init__(self, original_model):
 super(ResNetBottom, self).__init__()
 self.features = [???]
def forward(self, x):
 x = [???]
 x = x.view(x.size(0), -1)
 return x
```

FIGURE 7.11: PyTorch code snippet for extracting the $fc7$ layer from a pre-trained VGG19 CNN model.

**SOL-168**  **CH.SOL- 7.12.**

1. *Transforms are incorporated into deep learning pipelines in order to apply one or more operations on images which are represented as tensors. Different transforms are usually utilized during training and inference. For instance, during training we can use a transform to augment our data-set, while during inference our transform may be limited only to normalizing an image. PyTorch allows the use of transforms either during training or inference. The purpose of* test_trans *in line 5 is to normalize the data.*

2. The parameter requires_grad is set to False in line 14 since during inference the computation of gradients is obsolete.

3. The purpose of f.cpu() in line 11 is to move a tensor that was allocated on the GPU to the CPU. This may be required if we want to apply a CPU-based method from the Python numpy package on a Tensor that does not live in the CPU.

4. detach() in line 23 returns a newly created tensor without affecting the current tensor. It also detaches the output from the current computational graph, hence no gradient is backpropagated for this specific variable.

5. The purpose of numpy()[0] in line 23 is to convert the variable (an array) to a numpy compatible variable and also to retrieve the first element of the array.

■

### 7.3.2   Fine-tuning CNNs

**SOL-169 ☑ CH.SOL- 7.13.**
   The term fine-tuning (FT) of an ImageNet pre-trained CNN refers to the method by which one or more of the weights of the CNN are re-trained on a new target data-set, which may or may-not have similarities with the ImageNet data-set.   ■

**SOL-170 ☑ CH.SOL- 7.14.** *The three methods are as follows:*

1. Replacing and re-training only the classifier (usually the FC layer) of the ImageNet pre-trained CNN, on a target data-set.

2. FT all of the layers of the ImageNet pre-trained CNN, on a target data-set.

3. FT part of the layers of the ImageNet pre-trained CNN, on a target data-set.

■

**SOL-171 ☑ CH.SOL- 7.15.**

1. *The categories have to be represented numerically. One such option is presented in Code (7.1).*

```
'MEL': 0, 'NV': 1, 'BCC': 2, 'AKIEC': 3, 'BKL': 4, 'DF': 5,
↪  'VASC': 6
```

CODE 7.1: The seven categories of skin lesions.

2. *Several possible augmentations are presented in Code (7.2). It is usually, that by trial and error one finds the best possible augmentation for a target data-set. However, methods such as AutoAugment may render the manual selection of augmentations obsolete.*

```
self.transforms = []
if rotate:
 self.transforms.append(RandomRotate())
if flip:
 self.transforms.append(RandomFlip())
if brightness != 0:
 self.transforms.append(PILBrightness())
if contrast != 0:
 self.transforms.append(PILContrast())
if colorbalance != 0:
 self.transforms.append(PILColorBalance())
if sharpness != 0:
 self.transforms.append(PILSharpness())
```

CODE 7.2: Pseudeo code for augmentations.

3. *In contrast to the ResNet CNN which ends by an FC layer, the ImageNet pre-trained DPN CNN family, in this case the pretrainedmodels.dpn107, terminated by a Conv2d* *layer and hence must be adapted accordingly if one wishes to change the number fo classes from the 1000 (ImageNet) classes to our skin lession classification problem (7 classes). Line 7 in Code (7.3) demonstrated this idiom.*

```python
import torch
class Dpn107Finetune(nn.Module):
 def __init__(self, num_classes: int, net_kwards):
  super().__init__()
  self.net = pretrainedmodels.dpn107(**net_kwards)
  self.net.__name__= str (self.net)
  self.net.classifier = torch.nn.Conv2d(2688,
   ↪ num_classes,kernel_size=1)
  print(self.net)
```

CODE 7.3: Change between 1000 classes to 7 classes for the ImageNet pre-trained DPN CNN family.

### 7.3.3 Neural style transfer

**SOL-172**  **CH.SOL- 7.16.**

*The images are: a content image, a style image and lastly a combined image.*

**SOL-173**  **CH.SOL- 7.17.**

*The algorithm presented in the paper suggests how to combine the content a first image with the style of a second image to generate a third, stylized image using CNNs.*

**SOL-174**  **CH.SOL- 7.18.**

*The answers are as follows:*

1. *The training pipeline uses a combined loss which consists of a weighted average of the style loss and the content loss.*

2. *Different CNN layers at different levels are utilized to capture both fine-grained stylistic details as well as larger stylistic features.*

**SOL-175**  **CH.SOL- 7.19.**

1. *The content loss is the mean square error (MSE) calculated as the difference between the CNN activations of the last convolutional layer of both the content image and the style images.*

2. *The style loss amalgamates the losses of several layers together. For each layer, the gram matrix (see 7.2) for the activations at that layer is obtained for both the style and the combined images. Then, just like in the content loss, the MSE of the Gram matrices is calculated.*

**SOL-176**  **CH.SOL- 7.20.**

*For each feature map, a feature vector is extracted. The gram matrix captures the correlation between these feature vectors which is then being used in the loss function. Provided a list of feature vectors extracted from the images, $u_1, \ldots, u_k \in \mathbb{R}^n$, the Gram matrix is defined as:*

$$\begin{pmatrix} u_1 \cdot u_1 & \ldots & u_1 \cdot u_k \\ \vdots & \ldots & \vdots \\ u_k \cdot u_1 & \ldots & u_k \cdot u_k \end{pmatrix} \tag{7.2}$$

*The Gram matrix*

# CHAPTER

# 8

<span style="color:#A00000">DEEP LEARNING</span>

*It is the weight, not numbers of experiments that is to be regarded.*

— Isaac Newton.

## Contents

## 8.1    Introduction

I T was Alex Krizhevsky who first demonstrated that a convolutional neural network (CNN) can be effectively trained on the ImageNet large scale visual recognition challenge. A CNN automatically provides some degree of translation and assumes that we wish to learn *filters*, in a data-driven fashion, as a means to extract features describing the inputs. CNNs are applied to numerous computer vision, imaging, and computer graphics tasks as in [24], [23], [15], [5]. Furthermore, they have become extremely popular, and novel architectures and algorithms are continually popping up overnight.

## 8.2    Problems

### 8.2.1    Cross Validation

On the significance of cross validation and stratification in particular, refer to "*A study of cross-validation and bootstrap for accuracy estimation and model selection*" [17].

---

CV approaches

**PRB-177 ❷ CH.PRB- 8.1.**
   *Fig (8.1) depicts two different cross-validation approaches. Name them.*

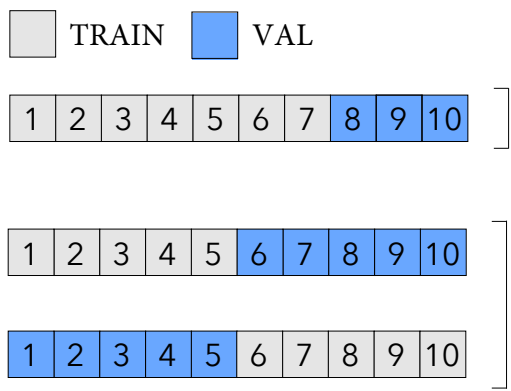

FIGURE 8.1: Two CV approaches

**PRB-178 ❷ CH.PRB- 8.2.**

*1. What is the purpose of the following Python code snippet 8.2 ?*

```
skf = StratifiedKFold(y, n_folds=5, random_state=989,
    shuffle=True)
```

FIGURE 8.2: Stratified K-fold

*2. Explain the benefits of using the K-fold cross validation approach.*

*3. Explain the benefits of using the Stratified K-fold cross validation approach.*

*4. State the difference between K-fold cross validation and stratified cross validation.*

*5. Explain in your own words what is meant by "We adopted a 5-fold cross-validation approach to estimate the testing error of the model".*

K-Fold CV

**PRB-179 ❷ CH.PRB- 8.3.**
 **True or False**: *In a K-fold CV approach, the testing set is completely excluded from the process and only the training and validation sets are involved in this approach.*

**PRB-180 ❷ CH.PRB- 8.4.**
 **True or False**: *In a K-fold CV approach, the final test error is:*

$$CV_{(k)} = \frac{1}{k} \sum_{i=1}^{k} \text{MSE}_i \tag{8.1}$$

**PRB-181 ❷ CH.PRB- 8.5.**

    *Mark all the correct choices regarding a cross-validation approach:*

  (i)  *A 5-fold cross-validation approach results in 5-different model instances being fitted.*

 (ii)  *A 5-fold cross-validation approach results in 1 model instance being fitted over and over again 5 times.*

(iii)  *A 5-fold cross-validation approach results in 5-different model instances being fitted over and over again 5 times.*

(iv)  *Uses K-different data-folds.*

---

**PRB-182 ❷ CH.PRB- 8.6.**

    *Mark all the correct choices regarding the approach that should be taken to compute the performance of K-fold cross-validation:*

  (i)  *We compute the cross-validation performance as the arithmetic mean over the K performance estimates from the validation sets.*

 (ii)  *We compute the cross-validation performance as the best one over the K performance estimates from the validation sets.*

<div align="center">Stratification</div>

---

**PRB-183 ❷ CH.PRB- 8.7.**

    *A data-scientist who is interested in classifying cross sections of histopathology image slices (8.3) decides to adopt a cross-validation approach he once read about in a book. Name the approach from the following options:*

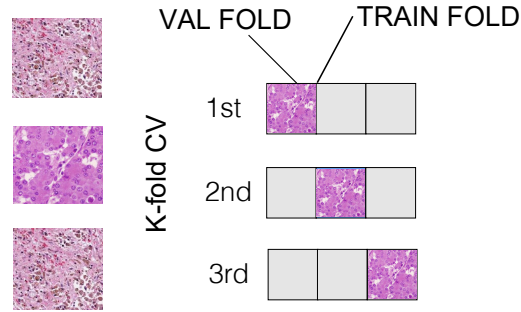

FIGURE 8.3: A specific CV approach

(i) *3-fold CV*

(ii) *3-fold CV with stratification*

(iii) *A (repeated) 3-fold CV*

LOOCV

**PRB-184 ❷ CH.PRB- 8.8.**

1. **True or false***: The leave-one-out cross-validation (LOOCV) approach is a sub-case of k-fold cross-validation wherein $K$ equals $N$, the sample size.*

2. **True or false***: It is always possible to find an optimal value $n$, $K = n$ in K-fold cross-validation.*

### 8.2.2 Convolution and correlation

The convolution operator

**PRB-185 ❷ CH.PRB- 8.9.**

*Equation 8.2 is commonly used in image processing:*

$$(f * g)(t) = \int_{-\infty}^{\infty} f(\tau)g(t - \tau)d\tau \tag{8.2}$$

1. *What does equation 8.2 represent?*

2. *What does $g(t)$ represent?*

---

**PRB-186 ❷ CH.PRB- 8.10.**

  *A data-scientist assumes that:*

 i *A convolution operation is both linear and shift invariant.*

 ii *A convolution operation is just like correlation, except that we flip over the filter before applying the correlation operator.*

 iii *The convolution operation reaches a maximum, only in cases where the filter is mostly similar to a specific section of the input signal.*

*Is he right in assuming so? Explain in detail the meaning of these statements.*

---

The correlation operator

---

**PRB-187 ❷ CH.PRB- 8.11.**

  *Mark the correct choice(s):*

1. *The cross-correlation operator is used to find the location where two different signals are most similar.*

2. *The autocorrelation operator is used to find when a signal is similar to a delayed version of itself.*

---

**PRB-188 ❷ CH.PRB- 8.12.**

  *A data-scientist provides you with a formulae for a discrete 2D convolution operation (8.3):*

$$f(x,y) * h(x,y) = \sum_{m=0}^{M-1} \sum_{n=0}^{N-1} f(m,n)h(x-m,y-n) \qquad (8.3)$$

*Using only (8.3), write the equivalent 2D correlation operation.*

<div align="center">Padding and stride</div>

Recommended reading : "*A guide to convolution arithmetic for deep learning*" by Vincent Dumoulin and Francesco Visin (2016) [22].

---

**PRB-189 ❷ CH.PRB- 8.13.**

*When designing a convolutional neural network layer, one must also define how the filter or kernel slides through the input signal. This is controlled by what is known as the stride and padding parameters or modes. The two most commonly used padding approaches in convolutions are the* VALID *and the* SAME *modes. Given an input stride of* 1:

1. *Define SAME*

2. *Define VALID*

---

**PRB-190 ❷ CH.PRB- 8.14.**

***True or False***: *A valid convolution is a type of convolution operation that does not use any padding on the input.*

---

**PRB-191 ❷ CH.PRB- 8.15.**

*You are provided with a $K \times K$ input signal and a $\theta \times \theta$ filter. The signal is subjected to the **valid** padding mode convolution. What are the resulting dimensions?*

$$arr = \begin{bmatrix} 0 & ... & 0 \\ 0 & ... & 0 \\ 0 & ... & 0 \end{bmatrix} \tag{8.4}$$

---

**PRB-192 ❷ CH.PRB- 8.16.**

*As depicted in (8.4), a filter is applied to a $\times 3$ input signal. Identify the correct choice given a stride of 1 and **Same** padding mode.*

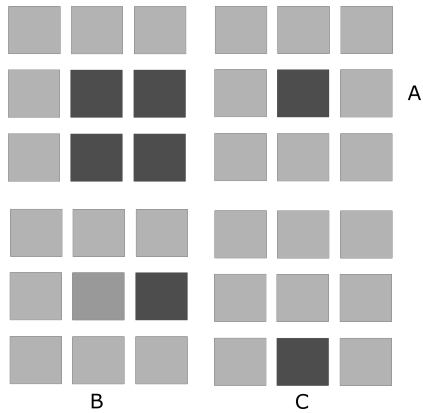

FIGURE 8.4: A padding approach

---

**PRB-193 ❷ CH.PRB- 8.17.**

As depicted in in (8.5), a filter is applied to a $3 \times 3$ input signal, mark the correct choices given a stride of 1.

(i) A represents a VALID convolution and B represents a SAME convolution

(ii) A represents a SAME convolution and B represents a VALID convolution

(iii) Both A and B represent a VALID convolution

(iv) Both A and B represent a SAME convolution

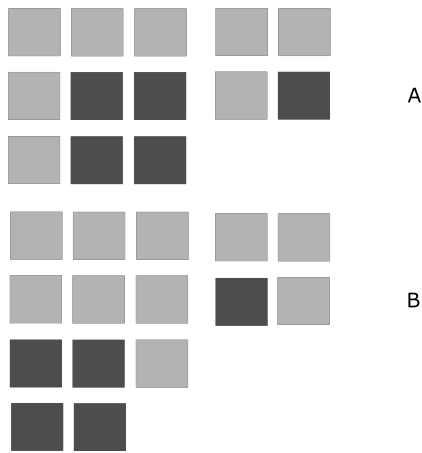

FIGURE 8.5: A padding approach

**PRB-194 ❷ CH.PRB- 8.18.**

In this question we discuss the two most commonly used padding approaches in convolutions; $\boxed{VALID}$ and $\boxed{SAME}$. Fig.8.6 presents python code for generating an input signal $arr001$ and a convolution kernel $filter001$. The input signal, $arr001$ is first initialized to all zeros as follows:

$$arr001 = \begin{bmatrix} 0 & 0 & 0 & 0 & 0 & 0 \\ 0 & 0 & 0 & 0 & 0 & 0 \\ 0 & 0 & 0 & 0 & 0 & 0 \\ 0 & 0 & 0 & 0 & 0 & 0 \\ 0 & 0 & 0 & 0 & 0 & 0 \\ 0 & 0 & 0 & 0 & 0 & 0 \end{bmatrix} \tag{8.5}$$

1. Without actually executing the code, determine what would be the resulting shape of the $convolve2d()$ operation.

2. Manually compute the result of convolving the input signal with the provided filter.

3. Elaborate why the size of the resulting convolutions is **smaller** than the size of the input signal.

```
1  import numpy
2  import scipy.signal
3
4  arr01 = numpy.zeros((6, 6),dtype=float)
5  print (arr01)
6  arr01[:,:3] = 3.0
7  arr01[:,3:] = 1.0
8
9  filter001 = numpy.zeros((3, 3), dtype=float)
10 filter001[:,0] = 2.0
11 filter001[:,2] = -2.0
12
13 output = scipy.signal.convolve2d(arr01, filter, mode='valid')
```

FIGURE 8.6: Convolution and correlation in python

Kernels and filters

**PRB-195 ❷ CH.PRB- 8.19.**

*Equation 8.6 is the discrete equivalent of equation 8.2 which is frequently used in image processing:*

$$(y * k)[i, j] = \sum_n \sum_m y[i - n, j - m]k[n, m] \qquad (8.6)$$

1. *Given the following discrete kernel in the X direction, what would be the equivalent Y direction?*

$$k = \frac{1}{2} \begin{bmatrix} -1 & 1 \\ -1 & 1 \end{bmatrix} \qquad (8.7)$$

2. *Identify the discrete convolution kernel presented in (8.7).*

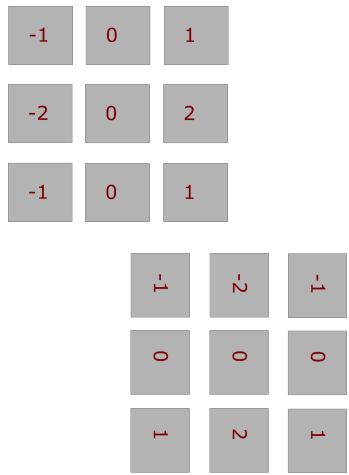

FIGURE 8.7: A 3 by 3 convolution kernel

---

**PRB-196 ❷ CH.PRB- 8.20.**

*Given an image of size $w \times h$, and a kernel with width $K$, how many multiplications and additions are required to convolve the image?*

Convolution and correlation in python

---

**PRB-197 ❷ CH.PRB- 8.21.**

*Fig.8.8 presents two built-in Python functions for the convolution and correlation operators.*

```python
import nympy as np
np.convolve(A,B,"full") # for convolution
np.correlate(A,B,"full") # for cross correlation
```

FIGURE 8.8: Convolution and correlation in python

*1. Implement the convolution operation from scratch in Python. Compare it with the*

*built-in numpy equivalent.*

2. *Implement the correlation operation using the implementation of the convolution operation. Compare it with the built-in numpy equivalent.*

Separable convolutions

**PRB-198 ❷ CH.PRB- 8.22.**
   *The Gaussian distribution in the 1D and 2D is shown in Equations 8.8 and 8.9.*

$$G(x) = \frac{1}{\sqrt{2\pi}\sigma} e^{-\frac{x^2}{2\sigma^2}} \tag{8.8}$$

$$G(x, y) = \frac{1}{2\pi\sigma^2} e^{-\frac{x^2+y^2}{2\sigma^2}} \tag{8.9}$$

   *The Gaussian filter, is an operator that is used to blur images and remove detail and noise while acting like a low-pass filter. This is similar to the way a mean filter works, but the Gaussian filter uses a different kernel. This kernel is represented with a Gaussian bell shaped bump.*
   *Answer the following questions:*

1. *Can 8.8 be used directly on a 2D image?*

2. *Can 8.9 be used directly on a 2D image?*

3. *Is the Gaussian filter separable? if so, what are the advantages of separable filters.*

### 8.2.3    Similarity measures

Image, text similarity

**PRB-199 ❷ CH.PRB- 8.23.**
   *A data scientist extracts a feature vector from an image using a pre-trained ResNet34 CNN (9.5).*

```
1  import torchvision.models as models
2  ...
3  res_model = models.resnet34(pretrained=True)
```

FIGURE 8.9: PyTorch declaration for a pre-trained ResNet34 CNN (simplified).

*He then applies the following algorithm, entitled xxx on the image (9.2).*

```
1   void xxx(std::vector<float>& arr) {
2   float mod = 0.0;
3    for (float i : arr) {
4     mod += i * i;
5    }
6   float mag = std::sqrt(mod);
7   for (float & i : arr) {
8     i /= mag;
9    }
10  }
```

An unknown algorithm in C++11

FIGURE 8.10: listing

*Which results in this vector (8.11):*

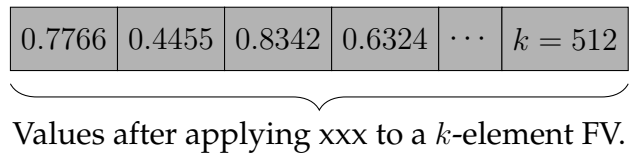

Values after applying xxx to a $k$-element FV.

FIGURE 8.11: A one-dimensional 512-element embedding for a single image from the ResNet34 architecture.

*Name the algorithm that he used and explain in detail why he used it.*

**PRB-200 ❷ CH.PRB- 8.24.**

*Further to the above, the scientist then applies the following algorithm:*

---

**Algorithm 3:** Algo 1

---

**Data:** Two vectors $v1$ and $v2$ are provided
Apply algorithm xxx on the two vectors
Run algorithm 2

---

**Algorithm 4:** Algo 2

```cpp
float algo2(const std::vector<float>& v1, const
   std::vector<float>& v2){
 double mul = 0;
 for (size_t i = 0; i < v1.size(); ++i){
   mul += v1[i] * v2[i];
 }
 if (mul < 0) {
   return 0;
 }
return mul;
}
```

FIGURE 8.12: An unknown algorithm

1. *Name the algorithm algo2 that he used and explain in detail what he used it for.*

2. *Write the mathematical formulae behind it.*

3. *What are the minimum and maximum values it can return?*

4. *An alternative similarity measures between two vectors is:*

$$sim_{euc}(v_1, v_2) = -||v_1 - v_2||. \qquad (8.10)$$

*Name the measure.*

Jacard similarity

**PRB-201 ❷ CH.PRB- 8.25.**

1. *What is the formulae for the Jaccard similarity [12] of two sets?:*

2. *Explain the formulae in plain words.*

3. *Find the Jaccard similarity given the sets depicted in (8.13)*

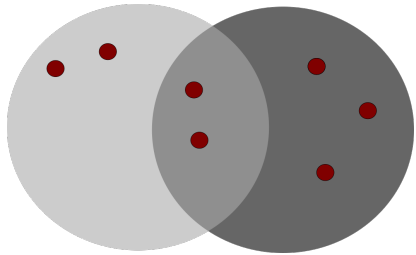

FIGURE 8.13: Jaccard similarity.

4. *Compute the Jaccard similarity of each pair of the following sets:*

   *i 12, 14, 16, 18.*

   *ii 11, 12, 13, 14, 15.*

   *iii 11, 16, 17.*

The Kullback-Leibler Distance

**PRB-202 ❷ CH.PRB- 8.26.**

  *In this problem, you have to actually read 4 different papers, so you will probably not encounter such a question during an interview, however reading academic papers is an excellent skill to master for becoming a DL researcher.*

  *Read the following papers which discuss aspects of the Kullback-Leibler divergence:*

 *i Bennet [2]*

*ii  Ziv [29]*

*iii  Bigi [3]*

*iv  Jensen [1]*

The Kullback-Leibler **divergence**, which was discussed thoroughly in chap 4 is a measure of how different two probability distribution are. As noted, the KL divergence of the probability distributions $P$, $Q$ on a set $X$ is defined as shown in Equation 8.11.

$$D_{KL}(P||Q) = \sum_{x \in X} P(x) log \frac{P(x)}{Q(x)} \tag{8.11}$$

Note however that since KL divergence is a non-symmetric information theoretical measure of distance of $P$ from $Q$, then it is not strictly a distance metric. During the past years, various KL based distance measures (rather than divergence based) have been introduced in the literature generalizing this measure.

Name each of the following KL based distances:

$$D_{KLD1}(P||Q) = D_{KL}(P||Q) + D_{KL}(Q||P) \tag{8.12}$$

$$D_{KLD2}(P||Q) = \sum_{x \in X} (P(x) - Q(x)) log \frac{P(x)}{Q(x)} \tag{8.13}$$

$$D_{KLD3}(P||Q) = \frac{1}{2} \left[ D_{KL}\left(P||\frac{P+Q}{2}\right) + D_{KL}\left(Q||\frac{P+Q}{2}\right) \right] \tag{8.14}$$

$$D_{KLD4}(P||Q) = max\left(D_{KL}(P||Q) + D_{KL}(Q||P)\right) \tag{8.15}$$

MinHash

Read the paper entitled *Detecting near-duplicates for web crawling* [12] and answer the following questions.

**PRB-203 ❷ CH.PRB- 8.27.**

What is the goal of hashing? Draw a simple HashMap of keys and values. Explain what is a collision and the notion of buckets. Explain what is the goal of MinHash.

**PRB-204 ❷ CH.PRB- 8.28.**
*What is Locality Sensitive Hashing or LSH?*

**PRB-205 ❷ CH.PRB- 8.29.**
   ***Complete the sentence****: LSH main goal is to [...] the probability of a colliding, for similar items in a corpus.*

### 8.2.4   Perceptrons

The Single Layer Perceptron

**PRB-206 ❷ CH.PRB- 8.30.**

1. **complete the sentence***: In a single-layer feed-forward NN, there are [...] input(s) and [...]. output layer(s) and no [...] connections at all.*

**PRB-207 ❷ CH.PRB- 8.31.**
   *In its simplest form, a* perceptron *(8.16) accepts only a binary input and emits a binary output. The output, can be evaluated as follows:*

$$output = \begin{cases} 0, & \text{if } \sum_j w_j x_j + b \leq 0, \\ 1, & \text{if } \sum_j w_j x_j + b > 0 \end{cases} . \tag{8.16}$$

*Where* weights *are denoted by* $w_j$ *and* biases *are denoted by* $b$*. Answer the following questions:*

1. **True or False***: If such a perceptron is trained using a labelled corpus, for each participating neuron the values* $w_j$ *and* $b$ *are learned automatically.*

2. **True or False***: If we instead use a new perceptron (sigmoidial) defined as follows:*

$$\sigma(wx + b) \tag{8.17}$$

*where $\sigma$ is the* sigmoid function*:*

$$\sigma(z) = \frac{1}{1 + e^{-z}}. \qquad (8.18)$$

*Then the new perceptron can process inputs ranging between 0 and 1 and emit output ranging between 0 and 1.*

3. *Write the cost function associated with the sigmoidial neuron.*

4. *If we want to train the perceptron in order to obtain the best possible weights and biases, which mathematical equation do we have to solve?*

5. *Complete the sentence: To solve this mathematical equation, we have to apply [...]*

6. *What does the following equation stands for?*

$$\nabla C = \frac{1}{n} \sum_x \nabla C_x \qquad (8.19)$$

*Where:*

$$C_x = \frac{1}{2} \|y(x) - a(x, w, b)\|^2 \qquad (8.20)$$

7. *Complete the sentence: Due to the time-consuming nature of computing gradients for each entry in the training corpus, modern DL libraries utilize a technique that gauges the gradient by first randomly sampling a subset from the training corpus, and then averaging only this subset in every epoch. This approach is known as [...]. The actual number of randomly chosen samples in each epoch is termed [...]. The gradient itself is obtained by an algorithm known as [...].*

The Multi Layer Perceptron

**PRB-208 ❷ CH.PRB- 8.32.**

*The following questions refer to the MLP depicted in (9.1).The inputs to the MLP in (9.1) are $x_1 = 0.9$ and $x_2 = 0.7$ respectively, and the weights $w_1 = -0.3$ and $w_2 = 0.15$ respectively. There is a single hidden node, $H_1$. The bias term, $B1$ equals $0.001$.*

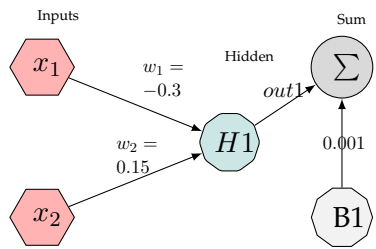

FIGURE 8.14: Several nodes in a MLP.

1. *We examine the mechanism of a single hidden node, $H_1$. The inputs and weights go through a linear transformation. What is the value of the output ($out1$) observed at the $sum$ node?*

2. *What is the value resulting from the application the $sum$ operator?*

3. *Verify the correctness of your results using PyTorch.*

Activation functions in perceptrons

**PRB-209 ❓ CH.PRB- 8.33.**

*The following questions refer to the MLP depicted in (8.15).*

1. *Further to the above, the ReLU non-linear activation function $g(z) = \max\{0, z\}$ is applied (8.15) to the output of the linear transformation. What is the value of the output ($out2$) now?*

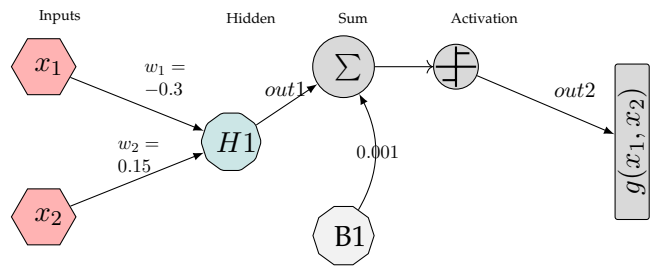

FIGURE 8.15: Several nodes in a MLP.

*2. Confirm your manual calculation using PyTorch tensors.*

Back-propagation in perceptrons

**PRB-210 ❷ CH.PRB- 8.34.**

*Your co-worker, an postgraduate student at M.I.T, suggests using the following activation functions in a MLP. Which ones can never be back-propagated and why?*

*i*

$$f(x) = |x| \tag{8.21}$$

*ii*

$$f(x) = x \tag{8.22}$$

*iii*

$$f(x) = \begin{cases} x\sin(1/x) & \text{if } x \neq 0 \\ 0 & \text{if } x = 0 \end{cases} \tag{8.23}$$

*iv*

$$f(x) = \begin{cases} x^2 & x > 0 \\ -x & x < 0 \\ 0 & x = 0 \end{cases} \tag{8.24}$$

**PRB-211 ❷ CH.PRB- 8.35.**

*You are provided with the following MLP as depicted in 8.16.*

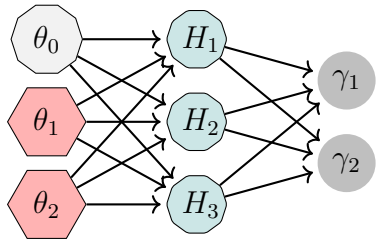

FIGURE 8.16: A basic MLP

*The ReLU non-linear activation function $g(z) = \max\{0, z\}$ is applied to the hidden layers $H_1...H_3$ and the bias term equals $0.001$.*

*At a certain point in time it has the following values 8.17 all of which are belong to the type $torch.FloatTensor$:*

```python
import torch
x= torch.tensor([0.9,0.7]) # Input
w= torch.tensor([
[-0.3,0.15],
[0.32,-0.91],
[0.37,0.47],
]) # Weights
B= torch.tensor([0.002]) # Bias
```

FIGURE 8.17: MLP operations.

1. *Using Python, calculate the output of the MLP at the hidden layers $H_1...H_3$.*

2. *Further to the above, you discover that at a certain point in time that the weights between the hidden layers and the output layers $\gamma_1$ have the following values:*

```python
w1= torch.tensor([
[0.15,-0.46,0.59],
[0.10,0.32,-0.79],
)
```

*What is the value observed at the output nodes $\gamma_1..\gamma_2$?*

3. *Assume now that a Softmax activation is applied to the output. What are the resulting values?*

4. *Assume now that a cross-entropy loss is applied to the output of the Softmax.*

$$L = -\sum_i \hat{y}_i \log(y_i) \tag{8.25}$$

*What are the resulting values?*

The theory of perceptrons

**PRB-212 ❷ CH.PRB- 8.36.**   *If someone is quoted saying:*

*MLP networks are universal function approximators.*

*What does he mean?*

**PRB-213 ❷ CH.PRB- 8.37.**
**True or False**: *the output of a perceptron is 0 or 1.*

**PRB-214 ❷ CH.PRB- 8.38.**
**True or False**: *A multi-layer perceptron falls under the category of supervised machine learning.*

**PRB-215 ❷ CH.PRB- 8.39.**
**True or False**: *The accuracy of a perceptron is calculated as the number of correctly classified samples divided by the total number of incorrectly classified samples.*

Learning logical gates

**PRB-216 ❷ CH.PRB- 8.40.**

*The following questions refer to the SLP depicted in (8.18). The weights in the SLP are $w_1 = 1$ and $w_2 = 1$ respectively. There is a single hidden node, $H_1$. The bias term, $B1$ equals $-2.5$.*

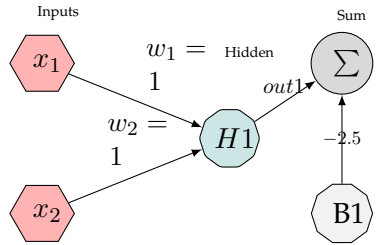

FIGURE 8.18: A single layer perceptron.

1. *Assuming the inputs to the SLP in (8.18) are*

    *i $x_1 = 0.0$ and $x_2 = 0.0$*

    *ii $x_1 = 0.0$ and $x_2 = 1.0$*

    *iii $x_1 = 1.0$ and $x_2 = 0.0$*

    *iv $x_1 = 1.0$ and $x_2 = 1.0$*

   *What is the value resulting from the application the $sum$ operator?*

2. *Repeat the above, assuming now that the bias term $B1$ was amended and equals $-0.25$.*

3. *Define what is the perceptron learning rule.*

4. *What was the most crucial difference between Rosenblatt's original algorithm and Hinton's fundamental papers of 1986:*
   "Learning representations by back-propagating errors" *[22]*
   *and 2012:*
   "ImageNet Classification with Deep Convolutional Neural Networks" *[18]?*

5. *The AND logic gate [7] is defined by the following table (8.19):*

| $x_1$ | $x_2$ | $y$ |
|---|---|---|
| 1 | 1 | 1 |
| 1 | 0 | 0 |
| 0 | 1 | 0 |
| 0 | 0 | 0 |

FIGURE 8.19: Logical AND gate

*Can a perceptron with only two inputs and a single output function as an AND logic gate? If so, find the weights and the threshold and demonstrate the correctness of your answer using a truth table.*

### 8.2.5   Activation functions (rectification)

We concentrate only on the most commonly used activation functions, those which the reader is more likely to encounter or use during his daily work.

**PRB-217 ❷ CH.PRB- 8.41.**

*The Sigmoid $s_c(x) = \frac{1}{1+e^{-cx}}$, also commonly known as the logistic function (Fig. 8.20), is widely used in binary classification and as a neuron activation function in artificial neural networks. Typically, during the training of an ANN, a Sigmoid layer applies the Sigmoid function to elements in the forward pass, while in the backward pass the chain rule is being utilized as part of the backpropagation algorithm. In 8.20 the constant $c$ was selected arbitrarily as 2 and 5 respectively.*

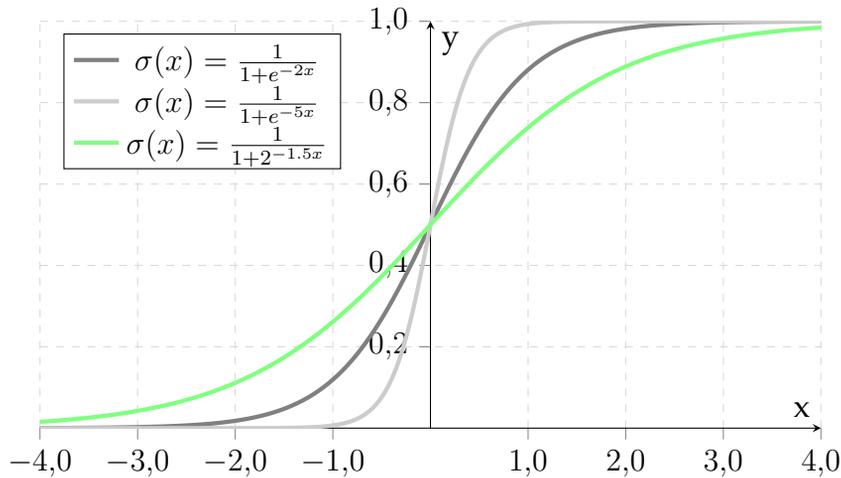

FIGURE 8.20: Examples of two sigmoid functions and an approximation.

*Digital **hardware** implementations of the sigmoid function do exist but they are expensive to compute and therefore several approximation methods were introduced by the research community. The method by [10] uses the following formulas to approximate the exponential function:*

$$e^x \approx Ex(x) \approx 2^{1.44x} \tag{8.26}$$

*Based on this formulation, one can calculate the sigmoid function as:*

$$Sigmoid\,(x) \approx \frac{1}{1 + 2^{-1.44x}} \approx \frac{1}{1 + 2^{-1.5x}} \tag{8.27}$$

1. *Code snippet 8.21 provides a pure C++ based (e.g. not using Autograd) implementation of the forward pass for the Sigmoid function. Implement the backward pass that directly computes the analytical gradients in C++ using Libtorch [19] style tensors.*

```
1  #include <torch/script.h>
2  #include <vector>
3
4  torch::Tensor sigmoid001(const torch::Tensor & x ){
5    torch::Tensor sig = 1.0 / (1.0 + torch::exp(( -x)));
6    return sig;
7  }
```

FIGURE 8.21: Forward pass for the Sigmoid function using Libtorch

2. *Code snippet 8.22 provides a skeleton for printing the values of the sigmoid and its derivative for a range of values contained in the vector $v$. Complete the code (lines 7-8) so that the values are printed.*

```
1  #include <torch/script.h>
2  #include <vector>
3  int main() {
4      std::vector<float> v{0.0, 0.1, 0.2, 0.3,
         ↪  0.4,0.5,0.6,0.7,0.8,0.9,0.99};
5      for (auto it = v.begin(); it != v.end(); ++it) {
6          torch::Tensor t0 = torch::tensor((*it));
7          ...
8          ...
9      }
10 }
```

.

FIGURE 8.22: Evaluation of the sigmoid and its derivative using Libtorch

3. *Manually derive the derivative of eq. 8.27, e.g:*

$$\frac{\mathrm{d}}{\mathrm{d}x}\left[\frac{1}{1+2^{-1.5x}}\right] \tag{8.28}$$

4. *Implement both the forward pass for the Sigmoid function approximation eq. 8.27 that directly computes the analytical gradients in C++ using Libtorch [19].*

5. *Print the values of the Sigmoid function and the Sigmoid function approximation eq. 8.27 for the following vector:*

$$v = [0.0, 0.1, 0.2, 0.3, 0.4, 0.5, 0.6, 0.7, 0.8, 0.9, 0.99] \tag{8.29}$$

Tanh

**PRB-218 ❷ CH.PRB- 8.42.**

*The Hyperbolic tangent nonlinearity, or the tanh function (Fig. 8.23), is a widely used neuron activation function in artificial neural networks:*

$$f_{tanh}(x) = \frac{\sinh(x)}{\cosh(x)} = \frac{e^x - e^{-x}}{e^x + e^{-x}} \tag{8.30}$$

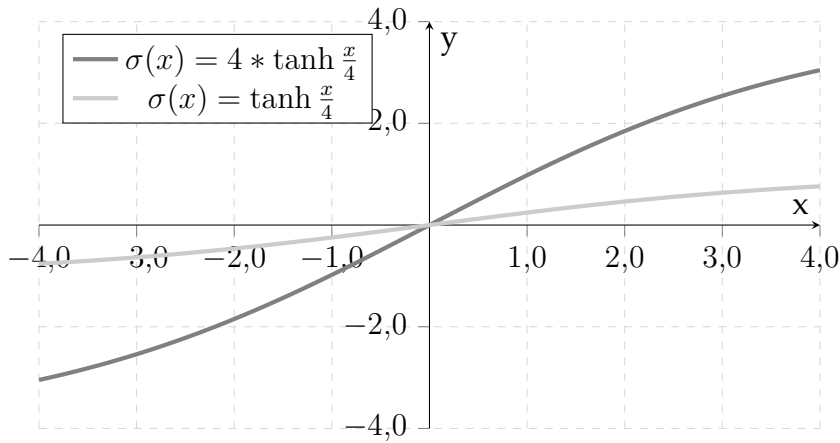

FIGURE 8.23: Examples of two tanh functions.

1. *Manually derive the derivative of the tanh function.*

2. *Use this numpy array as an input* $[[0.37, 0.192, 0.571]]$ *and evaluate the result using pure Python.*

3. *Use the PyTorch based* $torch.autograd.Function$ *class to write a custom Function that implements the forward and backward passes for the tanh function in Python.*

4. *Name the class TanhFunction, and using the gradcheck method from torch.autograd, verify that your numerical values equate the analytical values calculated by gradcheck. Remember you must implement a method entitled* $.apply(x)$ *so that the function can be invoked by Autograd.*

**PRB-219 ❷ CH.PRB- 8.43.**
    *The code snippet in 8.24 makes use of the tanh function.*

```
1  import torch
2
3  nn001 = nn.Sequential(
4    nn.Linear(200, 512),
5    nn.Tanh(),
6    nn.Linear(512, 512),
7    nn.Tanh(),
8    nn.Linear(512, 10),
9    nn.LogSoftmax(dim=1)
10 )
```

FIGURE 8.24: A simple NN based on tanh in PyTorch.

1. *What type of a neural network does nn001 in 8.24 represent?*

2. *How many hidden layers does the layer entitles nn001 have?*

**PRB-220 ❷ CH.PRB- 8.44.**

*Your friend, a veteran of the DL community claims that MLPs based on tanh activation function, have a symmetry around 0 and consequently cannot be saturated. Saturation, so he claims is a phenomenon typical of the top hidden layers in sigmoid based MLPs. Is he right or wrong?*

**PRB-221 ❷ CH.PRB- 8.45.**

*If we initialize the weights of a tanh based NN, which of the following approaches will lead to the **vanishing gradients problem**?.*

  i *Using the normal distribution, with parameter initialization method as suggested by Kaiming [14].*

 ii *Using the uniform distribution, with parameter initialization method as suggested by Xavier Glorot [9].*

iii *Initialize all parameters to a constant zero value.*

**PRB-222 ❷ CH.PRB- 8.46.**

*You friend, who is experimenting with the tanh activation function designed a small CNN with only one hidden layer and a linear output (8.25):*

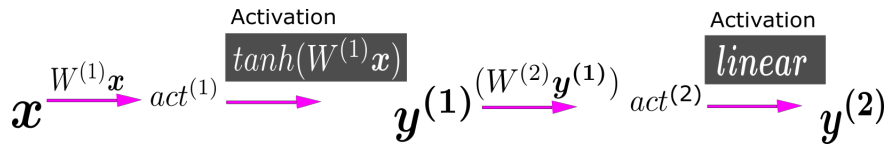

FIGURE 8.25: A small CNN composed of tanh blocks.

*He initialized all the weights and biases (biases not shown for brevity) to zero. What is the most significant design flaw in his architecture?*
*Hint: think about back-propagation.*

ReLU

**PRB-223 ❷ CH.PRB- 8.47.**

*The rectified linear unit, or ReLU $g(z) = \max\{0, z\}$ is the default for many CNN architectures. It is defined by the following function:*

$$f_{\text{ReLU}}(x) = \max(0, x) \tag{8.31}$$

*Or:*

$$f_{\text{ReLU}}(x) = \begin{cases} 1 & \text{if } x > 0 \\ 0 & \text{if } x \leq 0 \end{cases} \tag{8.32}$$

1. *In what sense is the ReLU better than traditional sigmoidal activation functions?*

---

**PRB-224 ❷ CH.PRB- 8.48.**

*You are experimenting with the ReLU activation function, and you design a small CNN (8.26) which accepts an RGB image as an input. Each CNN kernel is denoted by $w$.*

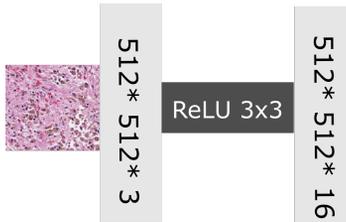

FIGURE 8.26: A small CNN composed of ReLU blocks.

*What is the shape of the resulting tensor $W$?*

---

**PRB-225 ❷ CH.PRB- 8.49.**

*Name the following activation function where $a \in (0, 1)$:*

$$f(x) = \begin{cases} x & \text{if } x > 0 \\ ax & \text{otherwise} \end{cases} \tag{8.33}$$

Swish

**PRB-226 ❷ CH.PRB- 8.50.**

*In many interviews, you will be given a paper that you have never encountered before, and be required to read and subsequently discuss it. Please read* Searching for Activation Functions *[21] before attempting the questions in this question.*

1. *In [21], researchers employed an automatic pipeline for searching what exactly?*

2. *What types of functions did the researchers include in their search space?*

3. *What were the main findings of their research and why were the results surprising?*

4. *Write the formulae for the Swish activation function.*

5. *Plot the Swish activation function.*

### 8.2.6   Performance Metrics

Comparing different machine learning models, tuning hyper parameters and learning rates, finding optimal augmentations, are all important steps in ML research. Typically our goal is to find the *best* model with the lowest errors on both the training and validation sets. To do so we need to be able to measure the *performance* of each approach/model/parameter setting etc. and compare those measures. For valuable reference, read: "*Evaluating Learning Algorithms: A Classification Perspective*" [22]

<div align="center">Confusion matrix, precision, recall</div>

**PRB-227 ❷ CH.PRB- 8.51.**

*You design a binary classifier for detecting the presence of malfunctioning temperature sensors. Non-malfunctioning (N) devices are the majority class in the training corpus. While running inference on an unseen test-set, you discover that the Confusion Metrics (CM) has the following values 8.27:*

|        |   | Predicted |      |
|--------|---|-----------|------|
|        |   | P         | N    |
| Actual | P | 12        | 7    |
|        | N | 24        | 1009 |

FIGURE 8.27: A confusion metrics for functioning (N) temperature sensors. P stands for malfunctioning devices.

1. *Find: TP, TN, FP, FN and correctly label the numbers in table 8.27.*

2. *What is the accuracy of the model?*

3. *What is the precision of the model?*

4. *What is the recall of the model?*

### ROC-AUC

The area under the receiver operating characteristic (ROC) curve, 8.73 known as the AUC, is currently considered to be the standard method to assess the accuracy of predictive distribution models.

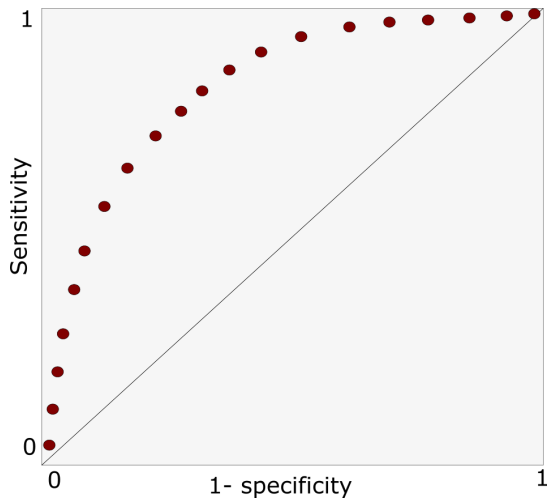

FIGURE 8.28: Receiver Operating Characteristic curve.

**PRB-228 ❷ CH.PRB- 8.52.**

*Complete the following sentences:*

1. *Receiver Operating Characteristics of a classifier shows its performance as a trade off between [...] and [...].*

2. *It is a plot of [...] vs. the [...]. In place of [...], one could also use [...] which are essentially {1 - 'true negatives'}.*

3. *A typical ROC curve has a concave shape with [...] as the beginning and [...] as the end point*

4. *The ROC curve of a 'random guess classifier', when the classifier is completely confused and cannot at all distinguish between the two classes, has an AUC of [...] which is the [...] line in an ROC curve plot.*

---

**PRB-229 ❷ CH.PRB- 8.53.**

*The code 8.30 and Figure 8.29 are the output from running XGBOOST for a binary classification task.*

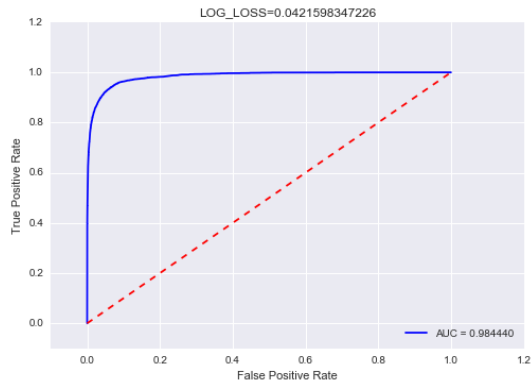

FIGURE 8.29: RUC AUC

```
1  XGBClassifier(base_score=0.5, colsample_bylevel=1,
   ↪   colsample_bytree=0.5,
2  gamma=0.017, learning_rate=0.15, max_delta_step=0, max_depth=9,
3  min_child_weight=3, missing=None, n_estimators=1000, nthread=-1,
4  objective='binary:logistic', reg_alpha=0, reg_lambda=1,
5  scale_pos_weight=1, seed=0, silent=1,
   ↪   subsample=0.9)shape:(316200, 6)
6
7  >ROC AUC:0.984439608912
8  >LOG LOSS:0.0421598347226
```

FIGURE 8.30: XGBOOST for binary classification.

*How would you describe the results of the classification?.*

### 8.2.7    NN Layers, topologies, blocks

CNN arithmetics

**PRB-230 ❷ CH.PRB- 8.54.**
   *Given an input of size of $n \times n$, filters of size $f \times f$ and a stride of $s$ with padding of $p$, what is the output dimension?*

**PRB-231 ❷ CH.PRB- 8.55.**
   *Referring the code snippet in Fig. (8.31), answer the following questions regarding the VGG11 architecture [25]:*

```
1  import torchvision
2  import torch
3  def main():
4   vgg11 = torchvision.models.vgg11(pretrained=True)
5   vgg_layers = vgg11.features
6   for param in vgg_layers.parameters():
7    param.requires_grad = False
8
9   example = [torch.rand(1, 3, 224, 224),
10  torch.rand(1, 3, 512, 512),
11  torch.rand(1, 3, 704, 1024)]
12  vgg11.eval()
13  for e in example:
14   out=vgg_layers(e)
15   print(out.shape)
16 if __name__ == "__main__":
17  main()^^I^^I
```

FIGURE 8.31: CNN arithmetics on the VGG11 CNN model.

1. *In each case for the input variable* example, *determine the dimensions of the tensor which is the output of applying the VGG11 CNN to the respective input.*

2. *Choose the correct option. The last layer of the VGG11 architecture is:*

   i *Conv2d*
   ii *MaxPool2d*
   iii *ReLU*

**PRB-232 ❷ CH.PRB- 8.56.**
*Still referring the code snippet in Fig. (8.31), and specifically to line 7, the code is amended so that the line is replaced by the line:*
*.*

1. *What type of block is now represented by the new line? Print it using PyTorch.*

2. *In each case for the input variable* $\boxed{example}$*, determine the dimensions of the tensor which is the output of applying the block:* $\boxed{vgg\_layers = vgg11.features[:3]}$ *to the respective input.*

**PRB-233 ❷ CH.PRB- 8.57.**

*Table* (8.1) *presents an incomplete listing of the of the VGG11 architecture* [25]. *As depicted, for each layer the number of filters (i. e., neurons with unique set of parameters) are presented.*

| Layer | #Filters |
|-------|----------|
| conv4_3 | 512 |
| fc6 | 4,096 |
| fc7 | 4,096 |
| **output** | 1,000 |

TABLE 8.1: Incomplete listing of the VGG11 architecture.

*Complete the missing parts regarding the dimensions and arithmetics of the VGG11 CNN architecture:*

1. *The VGG11 architecture consists of [...] convolutional layers.*

2. *Each convolutional layer is followed by a [...] activation function, and five [...] operations thus reducing the preceding feature map size by a factor of [...].*

3. *All convolutional layers have a [...] kernel.*

4. *The first convolutional layer produces [...] channels.*

5. *Subsequently as the network deepens, the number of channels [...] after each [...] operation until it reaches [...].*

**PRB-234 ❷ CH.PRB- 8.58.**

*A Dropout layer [26] (Fig. 8.32) is commonly used to regularize a neural network model by randomly equating several outputs (the crossed-out **hidden** node H) to 0.*

Dropout

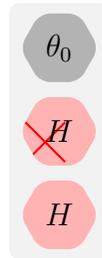

FIGURE 8.32: A Dropout layer (simplified form).

*For instance, in PyTorch [20], a Dropout layer is declared as follows (8.2):*

```python
import torch
import torch.nn as nn
nn.Dropout(0.2)
```

CODE 8.2: Dropout in PyTorch

*Where $nn.Dropout(0.2)$ (Line #3 in 8.2) indicates that the probability of zeroing an element is $0.2$.*

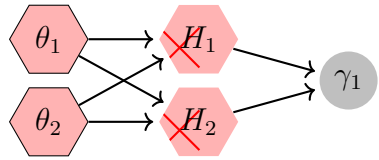

FIGURE 8.33: A Bayesian Neural Network Model

*A new data scientist in your team suggests the following procedure for a Dropout layer which is based on Bayesian principles. Each of the neurons $\theta_n$ in the neural network in (Fig. 8.33) may drop (or not) independently of each other exactly like a Bernoulli trial.*

*During the training of a neural network, the Dropout layer randomly drops out outputs of the previous layer, as indicated in (Fig. 8.32). Here, for illustration purposes, all four neurons are dropped as depicted by the crossed-out **hidden** nodes $H_n$.*

1. *You are interested in the proportion $\theta$ of dropped-out neurons. Assume that the chance of drop-out, $\theta$, is **the same** for each neuron (e.g. a **uniform prior** for $\theta$). Compute the **posterior** of $\theta$.*

2. *Describe the similarities of dropout to bagging.*

---

**PRB-235 ❷ CH.PRB- 8.59.**

*A co-worker claims he discovered an equivalence theorem where, two consecutive Dropout layers [26] can be replaced and represented by a single Dropout layer 8.34.*

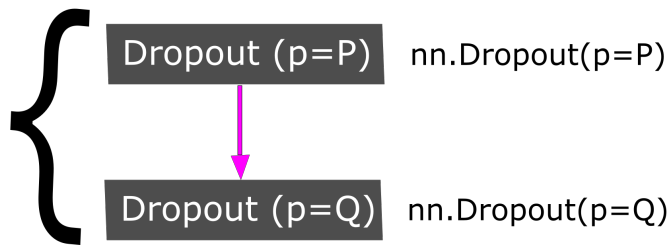

FIGURE 8.34: Two consecutive Dropout layers

*Hi realized two consecutive layers in PyTorch [20], declared as follows (8.3):*

```
1  import torch
2  import torch.nn as nn
3  nn.Sequential(
4    nn.Conv2d(1024, 32),
5    nn.ReLU(),
6    nn.Dropout(p=P, inplace=True),
7    nn.Dropout(p=Q, inplace=True)
8  )
```

CODE 8.3: Consecutive dropout in PyTorch

Where $nn.Dropout(0.1)$ *(Line #6 in [8.3])* indicates that the probability of zeroing an element is $0.1$.

1. *What do you think about his idea, is he right or wrong?*

2. *Either prove that he is right or provide a single example that refutes his theorem.*

### Convolutional Layer

The convolution layer is probably one of the most important layers in the theory and practice of modern deep learning and computer vision in particular.

To study the optimal number of convolutional layers for the classification of two different types of the Ebola virus, a researcher designs a binary classification pipeline using a small CNN with only a few layers ([8.35]):

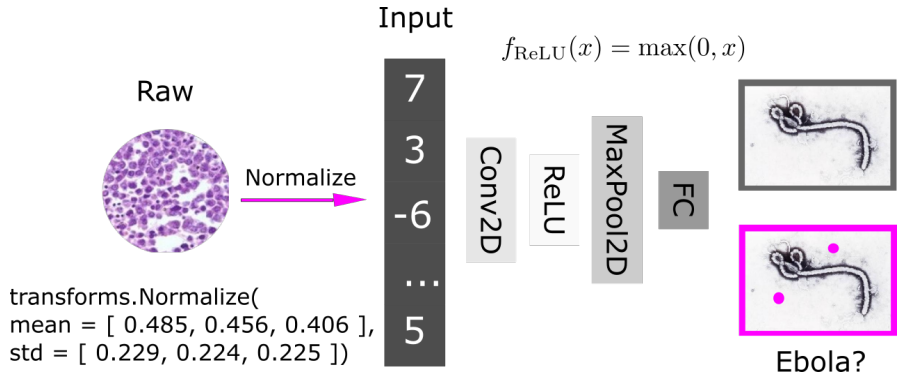

FIGURE 8.35: A CNN based classification system.

Answer the following questions while referring to (8.35):

---

**PRB-236 ❷ CH.PRB- 8.60.**

   *If he uses the following filter for the convolutional operation, what would be the resulting tensor after the application of the convolutional layer?*

$$* \quad \boxed{\begin{array}{c} 3 \\ 1 \end{array}}$$

FIGURE 8.36: A small filter for a CNN

---

**PRB-237 ❷ CH.PRB- 8.61.**

   *What would be the resulting tensor after the application of the ReLU layer (8.37)?*

Input

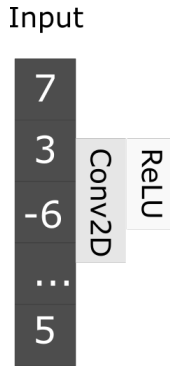

FIGURE 8.37: The result of applying the filter.

**PRB-238 ❷ CH.PRB- 8.62.**
*What would be the resulting tensor after the application of the MaxPool layer (8.78)?*

Pooling Layers

A pooling layer transforms the output of a convolutional layer, and neurons in a pooling layer accept the outputs of a number of adjacent feature maps and merge their outputs into a single number.

MaxPooling

**PRB-239 ❷ CH.PRB- 8.63.**
*The following input 8.38 is subjected to a MaxPool2D(2,2) operation having $2 \times 2$ maxpooling filter with a stride of 2 and no padding at all.*

| | | | |
|---|---|---|---|
| -1 | 0 | 11 | -1 |
| -1 | 7 | 1 | -1 |
| -1 | 0 | 1 | -1 |
| -1 | 0 | 1 | -1 |

FIGURE 8.38: Input to MaxPool2d operation.

*Answer the following questions:*

1. *What is the most common use of max-pooling layers?*

2. *What is the result of applying the MaxPool2d operation on the input?*

---

**PRB-240 ❷ CH.PRB- 8.64.**

*While reading a paper about the MaxPool operation, you encounter the following code snippet  9.1 of a PyTorch module that the authors implemented. You download their pretrained model, and evaluate its behaviour during inference:*

```python
import torch
from torch import nn
class MaxPool001(nn.Module):
 def __init__(self):
  super(MaxPool001, self).__init__()
  self.math = torch.nn.Sequential(
    torch.nn.Conv2d(3, 32, kernel_size=7, padding=2),
    torch.nn.BatchNorm2d(32),
    torch.nn.MaxPool2d(2, 2),
    torch.nn.MaxPool2d(2, 2),
  )
 def forward(self, x):
  print (x.data.shape)
  x = self.math(x)
  print (x.data.shape)
  x = x.view(x.size(0), -1)
  print ("Final shape:{}",x.data.shape)
  return x
model = MaxPool001()
model.eval()
x = torch.rand(1, 3, 224, 224)
out=model.forward(x)
```

CODE 8.4: A CNN in PyTorch

*The architecture is presented in 9.2:*

torch.rand(1, 3, 224, 224)

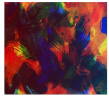

nn.Conv2d(3,32)

nn.BatchNorm2d(32)

nn.MaxPool2d(2,2)

nn.MaxPool2d(2,2)

FIGURE 8.39: Two consecutive MaxPool layers.

*Please run the code and answer the following questions:*

1. *In MaxPool2D(2,2), what are the parameters used for?*

2. *After running line 8, what is the resulting tensor shape?*

3. *Why does line 20 exist at all?*

4. *In line 9, there is a MaxPool2D(2,2) operation, followed by yet a second MaxPool2D(2,2). What is the resulting tensor shape after running line 9? and line 10?*

5. *A friend who saw the PyTorch implementation, suggests that lines 9 and 10 may be replaced by a single MaxPool2D(4,4,) operation while producing the exact same results. Do you agree with him? Amend the code and test your assertion.*

Batch normalization, Gaussian PDF

Recommended readings for this topic are "*Batch Normalization: Accelerating Deep Network Training by Reducing Internal Covariate Shift*" [16] and "*Delving deep into rectifiers: Surpassing human-level performance on imagenet classification*" [14].

A discussion of batch normalization (BN) would not be complete without a discussion of the Gaussian normal distribution. Though it would be instructive to develop the forward and backwards functions for a BN operation from scratch, it would also be quite complex. As an alternative we discuss several aspects of the BN operation while expanding on the Gaussian distribution.

The Gaussian distribution

**PRB-241 ❷ CH.PRB- 8.65.**

1. *What is batch normalization?*

2. *The normal distribution is defined as follows:*

$$P(x) = \frac{1}{\sigma\sqrt{2\pi}} e^{-(x-\mu)^2/2\sigma^2} \qquad (8.34)$$

   *Generally i.i.d. $X \sim \mathcal{N}(\mu, \sigma^2)$ however BN uses the standard normal distribution. What mean and variance does the standard normal distribution have?*

3. *What is the mathematical process of normalization?*

4. *Describe, how normalization works in BN.*

**PRB-242 ❷ CH.PRB- 8.66.**

In python, the probability density function for a normal distribution is given by *8.40*:

```
import scipy
scipy.stats.norm.pdf(x, mu, sigma)
```

FIGURE 8.40: Normal distribution in Python.

1. *Without using Scipy, implement the normal distribution from scratch in Python.*

2. *Assume, you want to back propagate on the normal distribution, and therefore you need the derivative. Using Scipy write a function for the derivative.*

BN

**PRB-243 ❷ CH.PRB- 8.67.**

*Your friend, a novice data scientist, uses an RGB image (8.41) which he then subjects to BN as part of training a CNN.*

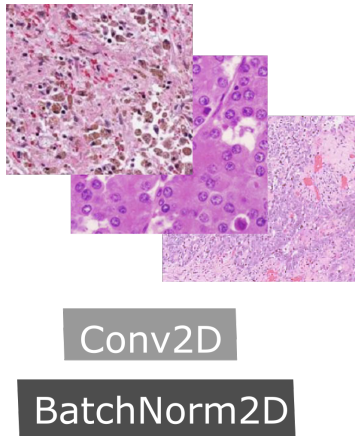

FIGURE 8.41: A convolution and BN applied to an RGB image.

1. *Help him understand, during BN, is the normalization applied pixel-wise or per colour channel?*

2. *In the PyTorch implementation, he made a silly mistake 8.42, help him identify it:*

```
1  import torch
2  from torch import nn
3  class BNl001(nn.Module):
4   def __init__(self):
5    super(BNl001, self).__init__()
6    self.cnn = torch.nn.Sequential(
7      torch.nn.Conv2d(3, 64, kernel_size=3, padding=2),
8   )
9   self.math= torch.nn.Sequential(
10     torch.nn.BatchNorm2d(32),
11     torch.nn.PReLU(),
12     torch.nn.Dropout2d(0.05)
13  )
14  def forward(self, x):
15    ...
```

FIGURE 8.42: A mistake in a CNN

Theory of CNN design

**PRB-244 ❷ CH.PRB- 8.68.**
*True or false:* An activation function applied after a Dropout, is equivalent to an activation function applied before a dropout.

**PRB-245 ❷ CH.PRB- 8.69.**
*Which of the following core building blocks may be used to construct CNNs? Choose all the options that apply:*

  *i  Pooling layers*

 *ii  Convolutional layers*

*iii  Normalization layers*

*iv  Non-linear activation function*

*v  Linear activation function*

---

**PRB-246 ❷ CH.PRB- 8.70.**

*You are designing a CNN which has a single BN layer. Which of the following core CNN designs are valid? Choose all the options that apply:*

  *i  CONV → act → BN → Dropout → . . .*

 *ii  CONV → act → Dropout → BN → . . .*

*iii  CONV → BN → act → Dropout → . . .*

*iv  BN → CONV → act → Dropout → . . .*

  *v  CONV → Dropout → BN → act → . . .*

*vi  Dropout → CONV → BN → act → . . .*

---

**PRB-247 ❷ CH.PRB- 8.71.**

*The following operator is known as the Hadamard product:*

$$OUT = A \odot B \tag{8.35}$$

*Where:*

$$(A \odot B)_{i,j} := (A)_{i,j}(B)_{i,j} \tag{8.36}$$

*A scientist, constructs a Dropout layer using the following algorithm:*

  *i  Assign a probability of $p$ for zeroing the output of any neuron.*

 *ii  Accept an input tensor $T$, having a shape $S$*

*iii  Generate a new tensor $T^` \in \{0,1\}^S$*

*iv  Assign each element in $T^`$ a randomly and independently sampled value from a Bernoulli distribution:*

$$T^`_i \sim B(1, p) \tag{8.37}$$

*v Calculate the OUT tensor as follows:*

$$OUT = T` \odot T \qquad (8.38)$$

*You are surprised to find out that his last step is to multiply the output of a dropout layer with:*

$$\frac{1}{1-p} \qquad (8.39)$$

*Explain what is the purpose of multiplying by the term $\frac{1}{1-p}$.*

---

**PRB-248 ❷ CH.PRB- 8.72.**
*Visualized in (8.43) from a high-level view, is an MLP which implements a well-known idiom in DL.*

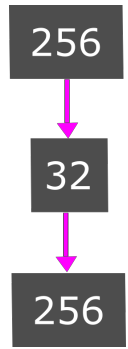

FIGURE 8.43: A CNN block

1. *Name the idiom.*

2. *What can this type of layer learn?*

3. *A fellow data scientist suggests amending the architecture as follows (8.44)*

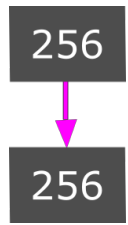

FIGURE 8.44: A CNN block

*Name one disadvantage of this new architecture.*

4. *Name one CNN architecture where the input equals the output.*

CNN residual blocks

**PRB-249 ❷ CH.PRB- 8.73.**
  *Answer the following questions regarding residual networks ([13]).*

1. *Mathematically, the residual block may be represented by:*

$$\mathbf{y} = \mathbf{x} + \mathcal{F}(\mathbf{x}) \tag{8.40}$$

  *What is the function $\mathcal{F}$?*

2. *In one sentence, what was the main idea behind deep residual networks (ResNets) as introduced in the original paper ([13])?*

**PRB-250 ❷ CH.PRB- 8.74.**
  *Your friend was thinking about ResNet blocks, and tried to visualize them in (8.45).*

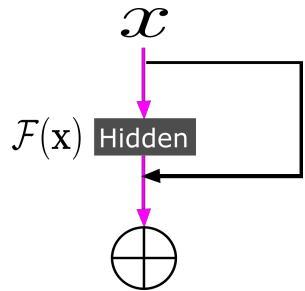

FIGURE 8.45: A resnet CNN block

1. *Assuming a residual of the form* $\mathbf{y} = \mathbf{x} + \mathcal{F}(\mathbf{x})$, *complete the missing parts in Fig.* *(8.45)*.

2. *What does the symbol* $\oplus$ *denotes?*

3. *A fellow data scientist, who had coffee with you said that residual blocks may compute the **identity function**. Explain what he meant by that.*

### 8.2.8 Training, hyperparameters

**PRB-251 ❷ CH.PRB- 8.75.**

*A certain training pipeline for the classification of large images (1024 x 1024) uses the following Hyperparameters (8.46):*

| Hyperparameter | Value |
|---|---|
| Initial learning rate | 0.1 |
| Weight decay | 0.0001 |
| Momentum | 0.9 |
| Batch size | 1024 |

```
1  optimizer = optim.SGD(model.parameters(), lr=0.1,
2  momentum=0.9,
3  weight_decay=0.0001)
4  ...
5  trainLoader = torch.utils.data.DataLoader(
6  datasets.LARGE('../data', train=True, download=True,
7  transform=transforms.Compose([
8  transforms.ToTensor(),
9  ])),
10 batch\_size=1024, shuffle=True)
```

FIGURE 8.46: Hyperparameters.

*In your opinion, what could possibly go wrong with this training pipeline?*

**PRB-252**  **CH.PRB- 8.76.**

*A junior data scientist in your team who is interested in Hyperparameter tuning, wrote the following code (8.5) for spiting his corpus into two distinct sets and fitting an LR model:*

```
1 from sklearn.model_selection import train_test_split
2 dataset = datasets.load_iris()
3 X_train, X_test, y_train, y_test =
4 train_test_split(dataset.data, dataset.target, test_size=0.2)
5 clf = LogisticRegression(data_norm=12)
6 clf.fit(X_train, y_train)
```

CODE 8.5: Train and Validation split.

*He then evaluated the performance of the trained model on the $X_{test}$ set.*

1. *Explain why his methodology is far from perfect.*

2. *Help him resolve the problem by utilizing a difference splitting methodology.*

3. *Your friend now amends the code an uses:*

```
1 clf = GridSearchCV(method, params, scoring='roc_auc', cv=5)
2 clf.fit(train_X, train_y)
```

*Explain why his new approach may work better.*

**PRB-253 ❷ CH.PRB- 8.77.**

   *In the context of Hyperparameter optimization, explain the difference between grid search and random search.*

Labelling and bias

Recommended reading:
"*Added value of double reading in diagnostic radiology,a systematic review*" [8].

**PRB-254 ❷ CH.PRB- 8.78.**

Non-invasive methods that forecast the existence of lung nodules (*8.47*), is a precursor to lung cancer. Yet, in spite of acquisition standardization attempts, the manual detection of lung nodules still remains predisposed to inter mechanical and observer variability. What is more, it is a highly laborious task.

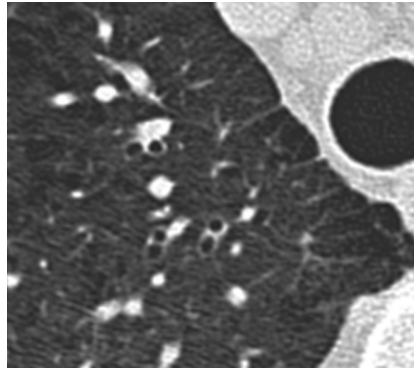

FIGURE 8.47: Pulmonary nodules.

In the majority of cases, the training data is manually labelled by radiologists who make mistakes. Imagine you are working on a classification problem and hire two radiologists for lung cancer screening based on low-dose CT (LDCT). You ask them to label the data, the first radiologist labels only the training set and the second the validation set. Then you hire a third radiologist to label the test set.

1. Do you think there is a design flow in the curation of the data sets?

2. A friend suggests that all there radiologists read all the scans and label them independently thus creating a majority vote. What do you think about this idea?

Validation curve ACC

**PRB-255 ❷ CH.PRB- 8.79.**
Answer the following questions regarding the validation curve visualized in (*8.48*):

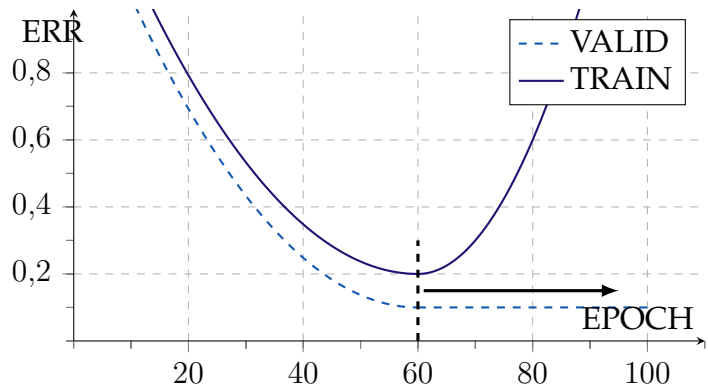

FIGURE 8.48: A validation curve.

1. *Describe in one sentence, what is a validation curve.*

2. *Which hyperparameter is being used in the curve?*

3. *Which well-known metric is being used in the curve? Which other metric is commonly used?*

4. *Which positive phenomena happens when we train a NN longer?*

5. *Which negative phenomena happens when we train a NN longer than we should?*

6. *How this negative phenomena is reflected in 8.48?*

Validation curve Loss

**PRB-256 ❷ CH.PRB- 8.80.**
    *Refer to the validation log-loss curve visualized in (8.49) and answer the following questions:*

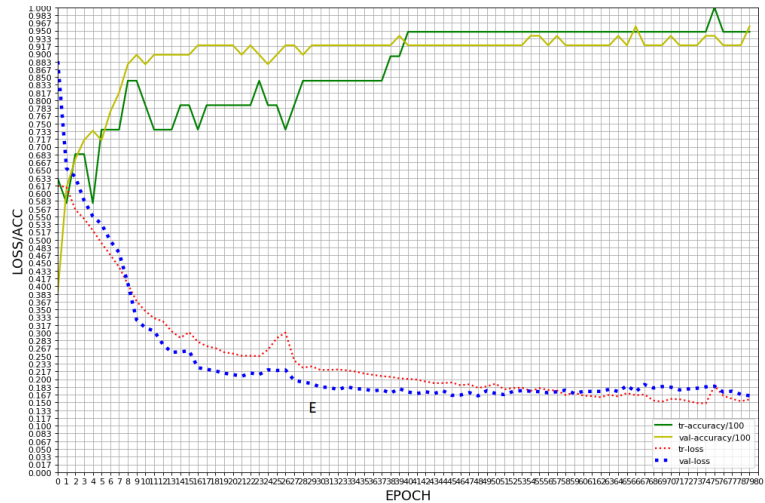

FIGURE 8.49: Log-loss function curve.

1. *Name the phenomena that starts happening right after the marking by the letter E and describe why it is happening.*

2. *Name three different weight initialization methods.*

3. *What is the main idea behind these methods?*

4. *Describe several ways how this phenomena can be alleviated.*

5. *Your friend, a fellow data-scientist, inspects the code and sees the following Hyperparameters are being used:*

| Hyperparameter | Value |
|---|---|
| *Initial LR* | *0.00001* |
| *Momentum* | *0.9* |
| *Batch size* | *1024* |

*He then tells you that the learning rate (LR) is constant and suggests amending the training pipeline by adding the following code (8.50):*

```
1  scheduler = optim.lr_scheduler.ReduceLROnPlateau(opt)
```

FIGURE 8.50: A problem with the log-loss curve.

*What do you think about his idea?*

6. *Provide one reason against the use of the log-loss curve.*

Inference

### PRB-257 ❷ CH.PRB- 8.81.

*You finished training a face recognition algorithm, which uses a feature vector of 128 elements. During inference, you notice that the performance is not that good. A friend tells you that in computer vision faces are gathered in various poses and perspectives. He therefore suggests that during inference you would augment the incoming face five times, run inference on each augmented image and then fuse the output probability distributions by averaging.*

1. *Name the method he is suggesting.*

2. *Provide several examples of augmentation that you might use during inference.*

### PRB-258 ❷ CH.PRB- 8.82.

*Complete the sentence: If the training loss is insignificant while the test loss is significantly higher, the network has almost certainly learned features which are not present in an [...] set. This phenomena is referred to as [...]*

### 8.2.9   Optimization, Loss

Stochastic gradient descent, SGD

### PRB-259 ❷ CH.PRB- 8.83.

*What does the term stochastic in SGD actually mean? Does it use any random number*

*generator?*

---

**PRB-260 ❷ CH.PRB- 8.84.**

*Explain why in SGD, the number of epochs required to surpass a certain loss threshold* increases *as the batch size decreases?*

---

Momentum

---

**PRB-261 ❷ CH.PRB- 8.85.**

*How does momentum work? Explain the role of exponential decay in the gradient descent update rule.*

---

**PRB-262 ❷ CH.PRB- 8.86.**

*In your training loop, you are using SGD and a logistic activation function which is known to suffer from the phenomenon of saturated units.*

1. *Explain the phenomenon.*

2. *You switch to using the tanh activation instead of the logistic activation, in your opinion does the phenomenon still exists?*

3. *In your opinion, is using the tanh function makes the SGD operation to converge better?*

---

**PRB-263 ❷ CH.PRB- 8.87.**

*Which of the following statements holds true?*

 i *In stochastic gradient descent we first calculate the gradient and only then adjust weights for each data point in the training set.*

 ii *In stochastic gradient descent, the gradient for a single sample is not so different from the actual gradient, so this gives a more stable value, and converges faster.*

 iii *SGD usually avoids the trap of poor local minima.*

*iv SGD usually requires more memory.*

<div align="center">Norms, L1, L2</div>

**PRB-264 ❷ CH.PRB- 8.88.**

*Answer the following questions regarding norms.*

1. *Which norm does the following equation represent?*

$$|x1 - x2| + |y1 - y2| \tag{8.41}$$

2. *Which formulae does the following equation represent?*

$$\sqrt{\sum_{i=1}^{n} (x_i - y_i)^2} \tag{8.42}$$

3. *When your read that someone penalized the L2 norm, was the euclidean or the Manhattan distance involved?*

4. *Compute both the Euclidean and Manhattan distance of the vectors:*
   $x1 = [6, 1, 4, 5]$ *and* $x2 = [2, 8, 3, -1]$.

**PRB-265 ❷ CH.PRB- 8.89.**

*You are provided with a pure Python code implementation of the Manhattan distance function (8.51):*

```python
from scipy import spatial
x1=[6,1,4,5]
x2=[2,8,3,-1]
cityblock = spatial.distance.cityblock(x1, x2)
print("Manhattan:", cityblock)
```

FIGURE 8.51: Manhattan distance function.

*In many cases, and for large vectors in particular, it is better to use a GPU for implementing numerical computations. PyTorch has full support for GPU's (and its my favourite DL library ... ), use it to implement the Manhattan distance function on a GPU.*

---

**PRB-266**  **CH.PRB- 8.90.**

*Your friend is training a logistic regression model for a binary classification problem using the L2 loss for optimization. Explain to him why this is a bad choice and which loss he should be using instead.*

---

## 8.3    Solutions

### 8.3.1    Cross Validation

On the significance of cross validation and stratification in particular, refer to "*A study of cross-validation and bootstrap for accuracy estimation and model selection*" [17].

CV approaches

---

**SOL-177**  **CH.SOL- 8.1.**

*The first approach is a leave-one-out CV (**LOOCV**) and the second is a **K-fold** cross-validation approach.*                                                                          ■

---

**SOL-178**  **CH.SOL- 8.2.**

*Cross Validation is a cornerstone in machine learning, allowing data scientists to take full gain of restricted training data. In classification, effective cross validation is essential to making the learning task efficient and more accurate. A frequently used form of the technique is identified as K-fold cross validation. Using this approach, the full data set is divided into $K$ randomly selected folds, occasionally* stratified, ***meaning that each fold has roughly the same class distribution as the overall data set***. *Subsequently, for each fold, all the other $(K-1)$ folds are used for training, while the present fold is used for testing. This process guarantees that sets used for testing, are not used by a classifier that also saw it during training.*                                                                          ■

K-Fold CV

**SOL-179**  **CH.SOL- 8.3.**

*True. We never utilize the test set during a K-fold CV process.* ■

**SOL-180**  **CH.SOL- 8.4.**

*True. This is the average of the individual errors of K estimates of the test error:*

$$\text{MSE}_1, \dots, \text{MSE}_k \tag{8.43}$$

■

**SOL-181**  **CH.SOL- 8.5.**

*The correct answer is: A 5-fold cross-validation approach results in 5-different model instances being fitted. It is a common misconception to think that in a K-fold approach the same model instance is repeatedly used. We must create a new model instance in each fold.* ■

**SOL-182**  **CH.SOL- 8.6.**

*The correct answer is: we compute the cross-validation performance as the **arithmetic mean** over the K performance estimates from the validation sets.* ■

<div align="center">Stratification</div>

**SOL-183**  **CH.SOL- 8.7.**

*The correct answer is: 3-fold CV. A k-fold cross-validation is a special case of cross-validation where we iterate over a dataset set $k$ times. In each round, we split the dataset into k parts: one part is used for validation, and the remaining $k - 1$ parts are merged into a training subset for model evaluation. Stratification is used to balance the classes in the training and validation splits in cases where the corpus is imbalanced.* ■

<div align="center">LOOCV</div>

**SOL-184**  **CH.SOL- 8.8.**

1. **True***: In (LOOCV) $K = N$ the full sample size.*

2. **False***: There is no way of a-priori finding an optimal value for K, and the relationship*

*between the actual sample size and the resulting accuracy is unknown.*

### 8.3.2  Convolution and correlation

The convolution operator

**SOL-185**  **CH.SOL- 8.9.**

1. *This is the definition of a convolution operation on the two signals $f$ and $g$.*

2. *In image processing, the term $g(t)$ represents a **filtering** kernel.*

**SOL-186**  **CH.SOL- 8.10.**

1. *True. These operations have two key features: they are shift invariant, and they are linear. Shift invariance means that we perform the same operation at every point in the image. Linearity means that this operation is linear, that is, we replace every pixel with a linear combination of its neighbours*

2. *True. See for instance Eq. (8.3).*

3. *True.*

The correlation operator

**SOL-187**  **CH.SOL- 8.11.**

1. *True.*

2. *True.*

**SOL-188**  **CH.SOL- 8.12.**

A convolution operation is just like correlation, except that we flip over the filter both horizontally and vertically before correlating.

$$f(x, y) \otimes h(x, y) = \sum_{m=0}^{M-1} \sum_{n=0}^{N-1} f^*(m, n) h(x + m, y + n) \tag{8.44}$$

Padding and stride

Recommended reading : "*A guide to convolution arithmetic for deep learning by Vincent Dumoulin and Francesco Visin (2016)*" [22].

**SOL-189**  **CH.SOL- 8.13.**

1. *The Valid padding only uses values from the original input; however, when the data resolution is not a multiple of the stride, some boundary values are ignored entirely in the feature calculation.*

2. *The Same padding ensures that every input value is included, but also adds zeros near the boundary which are not in the original input.*

**SOL-190**  **CH.SOL- 8.14.**

*True. Contrast this with the two other types of convolution operations.*

**SOL-191**  **CH.SOL- 8.15.**

$$\left\lfloor \frac{K - \theta}{\theta} \right\rfloor + 1 \times \left\lfloor \frac{n - \theta}{\theta} \right\rfloor + 1 \tag{8.45}$$

**SOL-192**  **CH.SOL- 8.16.**

*A is the correct choice.*                                                ■

**SOL-193**  **CH.SOL- 8.17.**

*A represents the VALID mode while B represents the SAME mode.*            ■

**SOL-194**  **CH.SOL- 8.18.**

1. *The resulting output has a shape of $4 \times 4$.*

2. *Convolution operation*

```
[[3. 3. 3. 1. 1. 1.]
 [3. 3. 3. 1. 1. 1.]
 [3. 3. 3. 1. 1. 1.]
 [3. 3. 3. 1. 1. 1.]
 [3. 3. 3. 1. 1. 1.]
 [3. 3. 3. 1. 1. 1.]]

[[ 2.  0. -2.]
 [ 2.  0. -2.]
 [ 2.  0. -2.]]
```

3. *By definition, convolutions in the **valid** mode, reduce the size of the resulting input tensor.*

```
[[  0. -12. -12.   0.]
 [  0. -12. -12.   0.]
 [  0. -12. -12.   0.]
 [  0. -12. -12.   0.]]
```

                                                                          ■

Kernels and filters

**SOL-195**  **CH.SOL- 8.19.**

1. *Flipping by 180 degrees we get:*

$$k = \frac{1}{2} \begin{bmatrix} -1 & -1 \\ 1 & 1 \end{bmatrix} \tag{8.46}$$

2. *The Sobel filter which is being frequently used for edge detection in classical computer vision.*

■

**SOL-196**  **CH.SOL- 8.20.**

*The resulting complexity is given by:*

$$K^2 wh \tag{8.47}$$

■

Convolution and correlation in python

**SOL-197**  **CH.SOL- 8.21.**

1. *Convolution operation:*

```
1  import numpy as np
2  def convolution(A,B):
3   l_A = np.size(A)
4   l_B = np.size(B)
5   C = np.zeros(l_A + l_B -1)
6
7   for m in np.arange(l_A):
8    for n in np.arange(l_B):
9     C[m+n] = C[m+n] + A[m]*B[n]
10
11   return C
```

FIGURE 8.52: Convolution and correlation in python

2. *Correlation operation:*

```
1  def crosscorrelation(A,B):
2   return convolution(np.conj(A),B[::-1])
```

FIGURE 8.53: Convolution and correlation in python

Separable convolutions

## SOL-198  CH.SOL- 8.22.

1. *No.Since images are usually stored as discrete pixel values one would have to use a discrete approximation of the Gaussian function on the filtering mask before performing the convolution.*

2. *No.*

3. *Yes it is separable, a factor that has great implications. For instance, separability means that a 2D convolution can be reduced to two consequent 1D convolutions reducing the computational runtime from* O $(n^2 \, m^2)$ *to* O $(n^2 \, m)$.

### 8.3.3 Similarity measures

Image, text similarity

**SOL-199**  **CH.SOL- 8.23.**

*The algorithm presented in (8.12) normalizes the input vector. This is usually done prior to applying any other method to the vector or before persisting a vector to a database of FVs.*

**SOL-200**  **CH.SOL- 8.24.**

1. *The algorithm presented in (8.1) is one of the most commonly used image similarity measures and is entitled **cosine similarity**. It can be applied to any pair of images.*

2. *The mathematical formulae behind it is:*

   *The cosine similarity between two vectors:*
   $u = \{u_1, u_2, \ldots, u_N\}$ *and* $v = \{v_1, v_2, \ldots, v_N\}$ *is defined as:*

   $$\text{sim}(u, v) = \frac{u \cdot v}{|u||v|} = \frac{\sum_{i=1}^{N} u_i v_i}{\sqrt{\left(\sum_{i=1}^{N} u_i^2\right)\left(\sum_{i=1}^{N} v_i^2\right)}}$$

   *Thus, the cosine similarity between two vectors measures the **cosine of the angle** between the vectors irrespective of their magnitude. It is calculated as the dot product of two numeric vectors, and is normalized by the product of the length of the vectors.*

3. *The minimum and maximum values it can return are 0 and 1 respectively. Thus, a cosine similarity value which is close to 1 indicated a very high similarity while that close to 0 indicates a very low similarity.*

4. *It represents the negative distance in Euclidean space between the vectors.*

**SOL-201**  **CH.SOL- 8.25.**

1. *The general formulae for the Jaccard similarity of two sets is given as follows:*

$$J(A, B) = \frac{|A \cap B|}{|A \cup B|}$$

2. *That is, the ratio of the size of the **intersection** of A and B to the size of their **union**.*

3. *The Jaccard similarity equals:*

$$\frac{2}{7}$$

4. *Given (8.13)*

*For the three combinations of pairs above, we have*

$$J(\{11, 16, 17\}, \{12, 14, 16, 18\}) = \frac{1}{6}$$
$$J(\{11, 12, 13, 14, 15\}, \{11, 16, 17\}) = \frac{1}{7}$$
$$J(\{11, 12, 13, 14, 15\}, \{12, 14, 16, 18\}) = \frac{2}{7}$$

**SOL-202**  **CH.SOL- 8.26.**

*Each KLD corresponds to the definition of:*

i *Jensen [1]*

ii Bennet [2]

iii Bigi [3]

iv Ziv [29]

■

<div align="center">MinHash</div>

Read the paper entitled *Detecting near-duplicates for web crawling* [12] and answer the following questions.

**SOL-203** ✅ **CH.SOL- 8.27.**

*A Hashing function (8.54) maps a value into a constant length string that can be compared with other hashed values.*

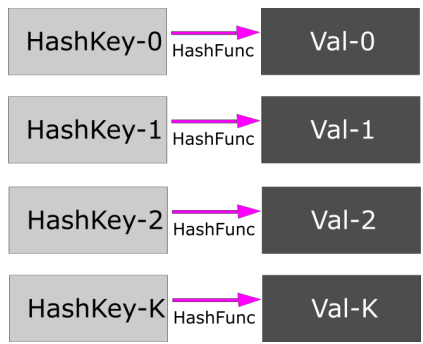

<div align="center">FIGURE 8.54: The idea of hashing</div>

*The idea behind hashing is that items are hashed into buckets, such that **similar items** will have a higher probability of hashing into the **same buckets**.*

*The goal of MinHash is to compute the Jaccard similarity without actually computing the intersection and union of the sets, which would be slower. The main idea behind MinHash is to devise a signature scheme such that the probability that there is a match between the signatures of two sets, $S_1$ and $S_2$, is equal to the Jaccard measure [12].*

■

**SOL-204**  **CH.SOL- 8.28.**

*Locality-Sensitive Hashing (LSH) is a method which is used for determining which items in a given set are similar. Rather than using the naive approach of comparing all pairs of items within a set, items are hashed into buckets, such that similar items will be more likely to hash into the same buckets.*

**SOL-205**  **CH.SOL- 8.29.**

*Maximise.*

### 8.3.4    Perceptrons

The Single Layer Perceptron

**SOL-206**  **CH.SOL- 8.30.**

*Answer: one, one, feedback.*

**SOL-207**  **CH.SOL- 8.31.**

1. *True.*

2. *True.*

3.

$$C(w, b) = \frac{1}{2n} \sum_x \|y(x) - a(x, w, b)\|^2 \qquad (8.48)$$

*where $w$ denotes the collection of all weights in the network, $b$ all the biases, $n$ is the total number of training inputs and $a(x, w, b)$ is the vector of outputs from the network which has weights $w$, biases $b$ and the input $x$.*

4.

$$\arg \min_{w,b} C(w, b). \qquad (8.49)$$

5. *Gradient descent.*

*6. The gradient.*

*7. Stochastic gradient descent. Batch size. Back-propagation.*

■

The Multi Layer Perceptron

**SOL-208**  **CH.SOL- 8.32.**

1. *This operation is a dot product with the given weights. Therefore:*

$$out = x_1 * w_1 + x_2 * w_2 + b_1 =$$
$$0.9 * (-0.3) + 0.7 * 0.15 = -0.164$$
(8.50)

2. *This operation (sum) is a dot product with the given weights **and with the given bias added**. Therefore:*

$$out1 = x_1 * w_1 + x_2 * w_2 + b_1 =$$
$$0.9 * (-0.3) + 0.7 * 0.15 + 0.001 = -0.165$$
(8.51)

3. *Code snippet 8.55 provides a pure PyTorch-based implementation of the MLP operation.*

```
import torch
# .type(torch.FloatTensor)
x= torch.tensor([0.9,0.7])
w= torch.tensor([-0.3,0.15])
B= torch.tensor([0.001])
print (torch.sum(x*w))
print (torch.sum(x*w) + B)
```

FIGURE 8.55: MLP operations.

■

Activation functions in perceptrons

**SOL-209**  **CH.SOL- 8.33.**

1. *Since by definition:*

$$f_{\text{ReLU}}(x) = \begin{cases} 1 & \text{if } x > 0 \\ 0 & \text{if } x \leq 0 \end{cases} \quad (8.52)$$

   *And the output of the linear sum operation was $-0.164$ then, the output $out2 = 0$.*

2. *Code snippet 8.56 provides a pure PyTorch-based implementation of the MLP operation.*

```python
import torch
x= torch.tensor([0.9,0.7])
w= torch.tensor([-0.3,0.15])
B= torch.tensor([0.001])
print (torch.sum(x*w))
print (torch.sum(x*w) + B)
print (torch.relu(torch.sum(x*w + B)))
```

FIGURE 8.56: MLP operations.

Back-propagation in perceptrons

**SOL-210**  **CH.SOL- 8.34.**   *The answers are as follows:*

1. *Non-differentiable at 0.*

2. *Non-differentiable at 0.*

3. *Even though for $x \neq 0$:*

$$f'(x) = \sin\frac{1}{x} - \frac{1}{x}\cos\frac{1}{x},\qquad(8.53)$$

   *the function is still non-differentiable at 0.*

4. *Non-differentiable at 0.*

∎

**SOL-211**  **CH.SOL- 8.35.**

1. *Fig 8.57 uses a loop (inefficient but easy to understand) to print the values:*

```
for i in range(0,w.size(0)):
    print (torch.relu(torch.sum(x*w[i]) + B))
> tensor([0.])
> tensor([0.])
> tensor([0.6630])
```

FIGURE 8.57: MLP operations- values.

2. *The values at each hidden layer are depicted in 8.58*

Output

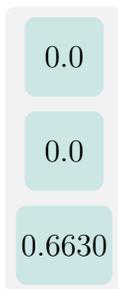

FIGURE 8.58: Hidden layer values, simple MLP.

*3. Fig 8.59 uses a loop (inefficient but easy to understand) to print the values:*

```
x1= torch.tensor([0.0,0.0,0.6630])# Input
w1= torch.tensor([
[0.15,-0.46,0.59],
[0.10,0.32,-0.79],
]).type(torch.FloatTensor) # Weights
for i in range(0,w1.size(0)):
 print (torch.sum(x1*w1[i]))
> tensor(0.3912)
> tensor(-0.5238)
```

FIGURE 8.59: MLP operations- values at the output.

*4. We can apply the Softmax function like so 8.60:*

```
1  x1= torch.tensor([0.0,0.0,0.6630]) # Input
2  w1= torch.tensor([
3  [0.15,-0.46,0.59],
4  [0.10,0.32,-0.79],
5  ]).type(torch.FloatTensor) # Weights
6  out1 = torch.tensor([[torch.sum(x1*w1[0]).item()],
7  [torch.sum(x1*w1[1]).item()]])
8  print (out1)
9  yhat = torch.softmax(out1, dim=0)
10 print (yhat)
11 > tensor([[ 0.3912],
12 [-0.5238]])
13 > tensor([[0.7140],
14 [0.2860]])
```

FIGURE 8.60: MLP operations- Softmax.

5. *For the cross-entropy loss, we use the Softmax values and calculate the result as follows:*

$$-1.0 * \log(0.7140) - 0.0 * \log(0.2860) = 1.31 \qquad (8.54)$$

■

The theory of perceptrons

**SOL-212**  **CH.SOL- 8.36.**

*He means that theoretically [6], a non-linear layer followed by a linear layer, can approximate any non-linear function with arbitrary accuracy, provided that there are enough non-linear neurons*

■

**SOL-213**  **CH.SOL- 8.37.** *True*

■

**SOL-214**  **CH.SOL- 8.38.** *True*

■

**SOL-215**  **CH.SOL- 8.39.**

*False. Divided by the training samples, not the number of incorrectly classified samples.* ▪

Learning logical gates

**SOL-216**  **CH.SOL- 8.40.**

1. *The values are presented in the following table (8.61):*

| Bias = −2.5 | | |
|---|---|---|
| Input | Weighted sum | Output |
| (0,0) | -2.5 | 0 |
| (0,1) | -1.5 | 0 |
| (1,0) | -1.5 | 0 |
| (1,1) | -0.5 | 0 |

FIGURE 8.61: Logical AND: B=-2.5

2. *The values are presented in the following table (8.62):*

| Bias = −0.25 | | |
|---|---|---|
| Input | Weighted sum | Output |
| (0,0) | -0.25 | 0 |
| (0,1) | -0.75 | 0 |
| (1,0) | -0.75 | 0 |
| (1,1) | 1.75 | 1 |

FIGURE 8.62: Logical AND: B=-0.25

3. *The perceptron learning rule is an algorithm that can automatically compute optimal weights for the perceptron.*

4. *The main addition by [22] and [18] was the introduction of a differentiable activation function.*

5. *if we select w1 = 1;w2 = 1 and threshold=1. We get:*

$$x_1 = 1, x_2 = 1 :$$
$$n = 1 \times 1 + 1 \times 1 = 2, thus, y = 1$$
$$x_1 = 1, x_2 = -1 :$$
$$n = 1 \times 1 + 1 \times (-1) = 0, thus, y = -1$$
$$x_1 = -1, x_2 = 1 :$$
$$n = 1 \times (-1) + 1 \times 1 = 0, thus, y = -1$$
$$x_1 = -1, x_2 = -1 :$$
$$n = 1 \times (-1) + 1 \times (-1) = -2, thus, y = -1$$

(8.55)

*Or summarized in a table (8.63):*

| AND gate | | |
|---|---|---|
| $in_1$ | $in_2$ | out |
| 0 | 0 | 0 |
| 0 | 1 | 0 |
| 1 | 0 | 0 |
| 1 | 1 | 1 |

FIGURE 8.63: Logical AND gate

### 8.3.5 Activation functions (rectification)

We concentrate only on the most commonly used activation functions, those which the reader is more likely to encounter or use during his daily work.

Sigmoid

**SOL-217**  **CH.SOL- 8.41.**

1. *Remember that the analytical derivative is of the sigmoid:*

$$\frac{d}{dx}s(x) = \frac{d}{dx}((1 + e^{-x})^{-1}) \tag{8.56}$$

$$\frac{d}{dx}s(x) = -1((1 + e^{-x})^{(-1-1)})\frac{d}{dx}(1 + e^{-x}) \tag{8.57}$$

$$\frac{d}{dx}s(x) = -1((1 + e^{-x})^{(-2)})(\frac{d}{dx}(1) + \frac{d}{dx}(e^{-x})) \tag{8.58}$$

$$\frac{d}{dx}s(x) = -1((1 + e^{-x})^{(-2)})(0 + e^{-x}(\frac{d}{dx}(-x))) \tag{8.59}$$

$$\frac{d}{dx}s(x) = -1((1 + e^{-x})^{(-2)})(e^{-x})(-1) \tag{8.60}$$

$$\frac{d}{dx}s(x) = ((1 + e^{-x})^{(-2)})(e^{-x}) \tag{8.61}$$

$$\frac{d}{dx}s(x) = \frac{1}{(1 + e^{-x})^2}(e^{-x}) \tag{8.62}$$

$$\frac{d}{dx}s(x) = \frac{(e^{-x})}{(1 + e^{-x})^2} \tag{8.63}$$

*Code snippet 8.64 provides a pure C++ based implementation of the backward pass that directly computes the analytical gradients in C++.*

```cpp
#include <torch/script.h>
#include <vector>

torch::Tensor sigmoid001_d(torch::Tensor & x) {
  torch::Tensor s = sigmoid001(x);
  return (1 - s) * s;
}
```

FIGURE 8.64: Backward pass for the Sigmoid function using Libtorch.

2. *Code snippet 8.65 depicts one way of printing the values.*

```cpp
#include <torch/script.h>
#include <vector>
int main() {
    std::vector<float> v{0.0, 0.1, 0.2, 0.3,
    ↪ 0.4,0.5,0.6,0.7,0.8,0.9,0.99};
    for (auto it = v.begin(); it != v.end(); ++it) {
      torch::Tensor t0 = torch::tensor((*it));
      std::cout << (*it) << "," <<
      ↪ sigmoid001(t0).data().detach().item()
      .toFloat()<< ","
      << sigmoid001_d (t0).data().detach().item().toFloat()
      << '\n';
    }
}
```

FIGURE 8.65: Evaluation of the sigmoid and its derivative in C++ using Libtorch.

3. *The manual derivative of eq. 8.27 is:*

$$3\ln(2) \times \left[ \frac{2^{-1.5x}}{\left(2^{-1.5x}+1\right)^2} \right] \qquad (8.64)$$

4. *The forward pass for the Sigmoid function approximation eq. 8.27 is presented in code snippet 8.66:*

```
1 #include <torch/script.h>
2 #include <vector>
3 torch::Tensor sig_approx(const torch::Tensor & x ){
4   torch::Tensor sig = 1.0 / (1.0 +  torch::pow(2,( −1.5*x)));
5   return sig;
6 }
```

FIGURE 8.66: Forward pass for the Sigmoid function approximation in C++ using Libtorch.

5. *The values are 8.67: :*

```
1 #include <torch/script.h>
2 #include <vector>
3 int main() {
4  std::vector<float> v{0.0, 0.1, 0.2, 0.3,
     ↪  0.4,0.5,0.6,0.7,0.8,0.9,0.99};
5  for (auto it = v.begin(); it != v.end(); ++it) {
6     torch::Tensor t0 = torch::tensor((*it));
7     std::cout << (*it) << "," <<
       ↪  sigmoid001(t0).data().detach().item()
8     .toFloat()<< ","<< sig_approx (t0).data().detach().item().
9     toFloat()<<'\n';
10  }
```

FIGURE 8.67: Printing the values for Sigmoid and Sigmoid function approximation in C++ using Libtorch.

*An the values are presented in Table 8.2:*

| Value | Sig | Approx |
|-------|-----|--------|
| 0 | 0.5 | 0.5 |
| 0.1 | 0.524979 | 0.52597 |
| 0.2 | 0.549834 | 0.5518 |
| 0.3 | 0.574443 | 0.577353 |
| 0.4 | 0.598688 | 0.602499 |
| 0.5 | 0.622459 | 0.627115 |
| 0.6 | 0.645656 | 0.65109 |
| 0.7 | 0.668188 | 0.674323 |
| 0.8 | 0.689974 | 0.69673 |
| 0.9 | 0.710949 | 0.71824 |
| 0.99 | 0.729088 | 0.736785 |
| | | |

TABLE 8.2: Computed values for the Sigmoid and the Sigmoid approximation.

Tanh

**SOL-218**  **CH.SOL- 8.42.**
*The answers are as follows:*

1. *The derivative is:*

$$f_{\tanh}(x) = 1 - f_{\tanh}(x)^2 \tag{8.65}$$

2. *Code snippet 8.68 implements the forward pass using pure Python.*

```
1  import numpy as np
2  xT =
   ↪  torch.abs(torch.tensor([[0.37,0.192,0.571]],requires_grad=True))
3  .type(torch.DoubleTensor)
4  xT_np=xT.detach().cpu().numpy()
5  print ("Input: \n",xT_np)
6  tanh_values = np.tanh(xT_np)
7  print ("Numpy:", tanh_values)
8  > Numpy: [[0.35399172 0.18967498 0.51609329]]
```

FIGURE 8.68: Forward pass for tanh using pure Python.

3. *In order to implement a PyTorch based $torch.autograd.Function$ function such as tanh, we must provide both the forward and backward passes implementation. The mechanism behind this idiom in PyTorch is via the use of a **context**, abbreviated ctx which is like a state manager for automatic differentiation. The implementation is depicted in 8.69:*

```python
import torch

class TanhFunction(torch.autograd.Function):
 @staticmethod
 def forward(ctx, x):
  ctx.save_for_backward( x )
  y = x.tanh()
  return y

 @staticmethod
 def backward(ctx, grad_output):
  input, = ctx.saved_tensors
  dy_dx = 1 / (input.cosh() ** 2)
  out = grad_output * dy_dx
  print ("backward:{}".format(out))
  return out
```

FIGURE 8.69: Tanh in PyTorch.

4. *Code snippet 8.70 verifies the correctness of the implementation using gradcheck.*

```
1  import numpy as np
2  import numpy as np
3  xT = torch.abs(torch.tensor([[0.37,0.192,0.571]],
4  requires_grad=True))
5  .type(torch.DoubleTensor)
6  xT_np=xT.detach().cpu().numpy()
7  tanh_values = np.tanh(xT_np)
8  tanh_values_torch = tanhPyTorch(xT)
9  print ("Torch:", tanh_values_torch)
10 from torch.autograd import gradcheck, Variable
11 f = TanhFunction.apply
12 test=gradcheck(lambda t: f(t), xT)
13 print(test)
14 > PyTorch version: 1.7.0
15 > Torch: tensor([[0.3540, 0.1897, 0.5161]], dtype=torch.float64)
16 > backward:tensor([[0.8747, 0.9640, 0.7336]],dtype=torch.float64)
```

FIGURE 8.70: Invoking gradcheck on tanh.

**SOL-219**  **CH.SOL- 8.43.**

1.  *The type of NN is a MultiLayer Perceptron or MLP.*

2.  *There are two hidden layers.*

**SOL-220**  **CH.SOL- 8.44.**

   *He is partially correct , see for example **Understanding the difficulty of training deep feedforward neural networks** [9].*

**SOL-221**  **CH.SOL- 8.45.**

*Initialize all parameters to a constant zero value. When we apply the tanh function to an input which is very large, the output which is almost zero, will be propagated to the remaining partial derivatives leading to the well known phenomenon.*

■

**SOL-222**  **CH.SOL- 8.46.**

*During the back-propagation process, derivatives are calculated with respect to $(W^{(1)})$ and also $(W^{(2)})$. The design flaw:*

  i *Your friend initialized all weights and biases to zero.*

 ii *Therefore any gradient with respect to $(W^{(2)})$ would also be zero.*

iii *Subsequently, $(W^{(2)})$ will never be updated.*

 iv *This would inadvertently cause the derivative with respect to $(W^{(1)})$ to be always zero.*

  v *Finally, would also never be updated $(W^{(1)})$.*

■

ReLU

**SOL-223**  **CH.SOL- 8.47.**

*The ReLU function has the benefit of not saturating for positive inputs since its derivative is one for any positive value.*

■

**SOL-224**  **CH.SOL- 8.48.**

*The shape is:*

$$3 \times 3 \times 3 \times 16$$

■

**SOL-225**  **CH.SOL- 8.49.**

*The activation function is a leaky ReLU which in some occasions may outperform the*

*ReLU activation function.*                                               ■

<div align="center">Swish</div>

**SOL-226** ✏️ **CH.SOL- 8.50.**

1. *They intended to find new better-performing activation functions.*

2. *They had a list of basic mathematical functions to choose from, for instance the exponential families exp(), sin(), min and max.*

3. *Previous research found several activation function properties which were considered very useful. For instance, gradient preservation and non-monotonicity. However the surprising discovery was that the swish function violates both of these previously deemed useful properties.*

4. *The equation is:*

$$f(x) = x \cdot \sigma(x) \tag{8.66}$$

5. *The plot is <span style="color:blue">8.71</span>*

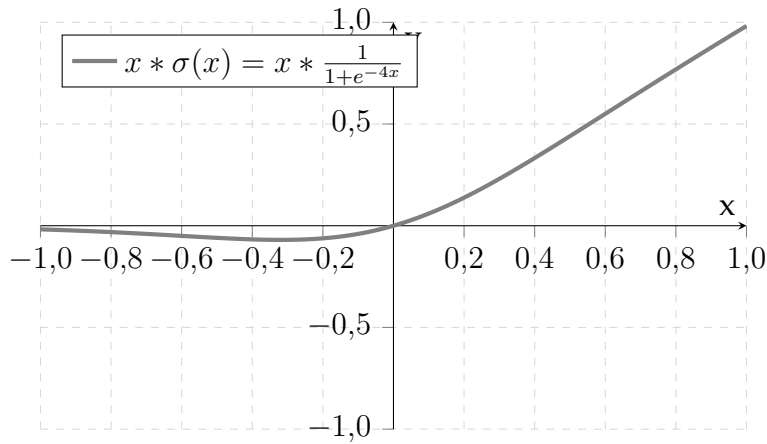

FIGURE 8.71: A plot of the Swish activation function.

### 8.3.6 Performance Metrics

Confusion matrix, precision, recall

**SOL-227**  **CH.SOL- 8.51.**

1. *The values are labelled inside 8.27:*

|  |  | Predicted | |
| --- | --- | --- | --- |
|  |  | P | N |
| Truth | P | TP=12 | FN=7 |
|  | N | FP=24 | TN=1009 |

FIGURE 8.72: TP, TN, FP, FN.

2.
$$acc = \frac{12 + 1009}{12 + 7 + 24 + 1009} = 0.97 \qquad (8.67)$$

3.
$$prec = \frac{12}{12 + 24} = 0.333 \qquad (8.68)$$

4.
$$recall = \frac{12}{12 + 7} = 0.631 \qquad (8.69)$$

∎

#### ROC-AUC

The area under the receiver operating characteristic (ROC) curve, 8.73 known as the AUC, is currently considered to be the standard method to assess the accuracy of predictive distribution models.

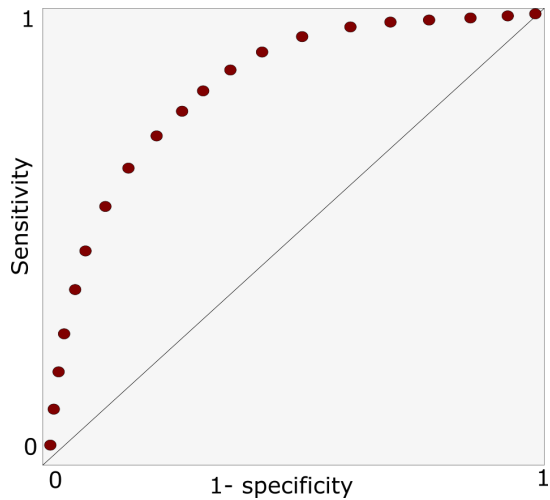

FIGURE 8.73: Receiver Operating Characteristic curve.

**SOL-228**  **CH.SOL- 8.52.**

*ROC allows to attest the relationship between sensitivity and specificity of a binary classifier. Sensitivity or true positive rate measures the proportion of positives correctly classified; specificity or true negative rate measures the proportion of negatives correctly classified. Conventionally, the true positive rate tpr is plotted against the false positive rate fpr, which is one minus true negative rate.*

1. *Receiver Operating Characteristics of a classifier shows its performance as a trade off between* selectivity *and* sensitivity.

2. *It is a plot of* 'true positives' *vs. the* 'true negatives'. *In place of* 'true negatives', *one could also use* 'false positives' *which are essentially 1 -* 'true negatives'.

3. *A typical ROC curve has a concave shape with* (0,0) *as the beginning and* (1,1) *as the end point*

4. *The ROC curve of a 'random guess classifier', when the classifier is completely confused and cannot at all distinguish between the two classes, has an AUC of 0.5, the* 'x = y' *line in an ROC curve plot.*

**SOL-229 ✎ CH.SOL- 8.53.**

*The ROC curve of an ideal classifier (100% accuracy) has an AUC of 1, with 0.0 'false positives' and 1.0 'true positives'. The ROC curve in our case, is almost ideal, which may indicate over-fitting of the XGBOOST classifier to the training corpus.*

### 8.3.7 NN Layers, topologies, blocks

CNN arithmetics

**SOL-230 ✎ CH.SOL- 8.54.**

*Output dimension: $L \times L \times M$ where $L = \frac{n-f+2p}{s} + 1$*

**SOL-231 ✎ CH.SOL- 8.55.**

*The answers are as follows:*

1. *Output dimensions:*

   i *torch.Size([1, 512, 7, 7])*

   ii *torch.Size([1, 512, 16, 16])*

   iii *torch.Size([1, 512, 22, 40])*

2. *The layer is MaxPool2d.*

**SOL-232 ✎ CH.SOL- 8.56.**

*The answers are as follows:*

1. *A convolutional block 8.74.*

```
1 Sequential(
2 (0): Conv2d(3, 64, kernel_size=(3, 3), stride=(1, 1), padding=(1,
  ↪ 1))
3 (1): ReLU(inplace=True)
4 (2): MaxPool2d(kernel_size=2, stride=2, padding=0, dilation=1,
  ↪ ceil_mode=False
5 )
```

FIGURE 8.74: Convolutional block from the VGG11 architecture.

2. *The shapes are as follows:*

  i *torch.Size([1, 64, 112, 112])*

  ii *torch.Size([1, 64, 256, 256])*

  iii *torch.Size([1, 64, 352, 512])*

**SOL-233**  **CH.SOL- 8.57.**

*The VGG11 architecture contains **seven** convolutional layers, each followed by a **ReLU** activation function, and five **max-polling operations**, each reducing the respective feature map by a factor of **2**. All convolutional layers have a $3 \times 3$ kernel. The first convolutional layer produces **64** channels and subsequently, as the network deepens, the number of channels **doubles** after each **max-pooling** operation until it reaches **512**.*

<div align="center">Dropout</div>

**SOL-234**  **CH.SOL- 8.58.**

1. *The observed data, e.g the dropped neurons are distributed according to:*

$$(x_1, \ldots, x_n)|\theta \overset{iid}{\sim} \mathrm{Bern}(\theta) \tag{8.70}$$

*Denoting $s$ and $f$ as success and failure respectively, we know that the likelihood is:*

$$p\left(x_1, \ldots, x_n | \theta\right) = \theta^s (1-\theta)^f \tag{8.71}$$

*With the following parameters $\alpha = \beta = 1$ the beta distribution acts like Uniform prior:*

$$\theta \sim \text{Beta}(\alpha, \beta), \text{ given } \alpha = \beta = 1 \tag{8.72}$$

*Hence, the **prior** density is:*

$$p(\theta) = \frac{1}{B(\alpha, \beta)} \theta^{\alpha-1} (1-\theta)^{\beta-1} \tag{8.73}$$

*Therefore the **posterior** is:*

$$\begin{aligned}
p\left(\theta | x_1, \ldots, x_n\right) &\propto p\left(x_1, \ldots, x_n | \theta\right) p(\theta) \\
&\propto \theta^S (1-\theta)^f \theta^{\alpha-1} (1-\theta)^{\beta-1} \\
&= \theta^{\alpha+s-1} (1-\theta)^{\beta+f-1}
\end{aligned} \tag{8.74}$$

2. *In dropout, in every training epoch, neurons are randomly pruned with probability $P = p$ sampled from a Bernoulli distribution. During inference, all the neurons are used but their output is multiplied by the a-priory probability $P$. This approach resembles to some degree the model averaging approach of bagging.*

■

## SOL-235  CH.SOL- 8.59.
*The answers are as follows:*

1. *The idea is true and a solid one.*

2. *The idiom may be exemplified as follows 8.75:*

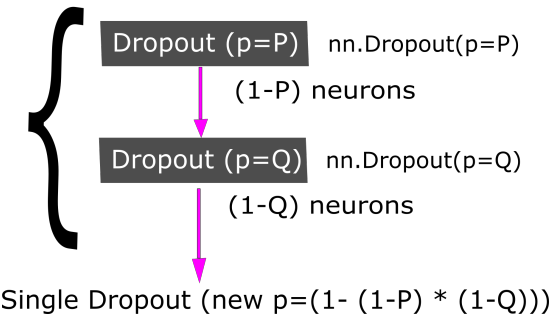

FIGURE 8.75: Equivalence of two consecutive dropout layers

*The probabilities add up by multiplication at each layer, resulting in a single dropout layer with probability:*

$$1 - (1 - p)(1 - q) \tag{8.75}$$

■

Convolutional Layer

**SOL-236** ✎ **CH.SOL- 8.60.**
  *The result is (8.76):*

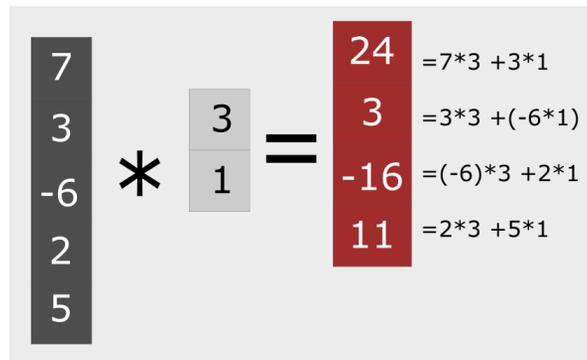

FIGURE 8.76: The result of applying the filter.

■

**SOL-237** ✏ **CH.SOL- 8.61.**

*The result is (8.77):*

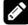

FIGURE 8.77: The result of applying a ReLU activation.

**SOL-238** ✏ **CH.SOL- 8.62.**

*The result is (8.78):*

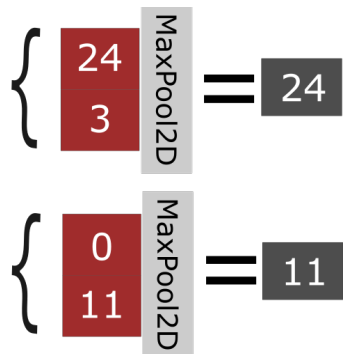

FIGURE 8.78: The result of applying a MaxPool layer.

Pooling Layers
MaxPooling

**SOL-239**  **CH.SOL- 8.63.**

*The answers are as follows:*

1. *A max-pooling layer is most commonly used after a convolutional layer in order to reduce the spatial size of CNN feature maps.*

2. *The result is 8.79:*

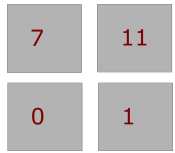

FIGURE 8.79: Output of the MaxPool2d operation.

■

**SOL-240**  **CH.SOL- 8.64.**

1. *In MaxPool2D(2,2), the first parameter is the size of the pooling operation and the second is the stride of the pooling operation.*

2. *The BatchNorm2D operation does not change the shape of the tensor from the previous layer and therefore it is:*
   $torch.Size([1, 32, 222, 222])$.

3. *During the training of a CNN we use model.train() so that Dropout layers are fired. However, in order to run inference, we would like to turn this firing mechanism off, and this is accomplished by model.eval() instructing the PyTorch computation graph not to activate dropout layers.*

4. *The resulting tensor shape is:*
   $torch.Size([1, 32, 55, 55])$
   *If we reshape the tensor like in line 17 using:*
   $x = x.view(x.size(0), -1)$

*Then the tensor shape becomes:*
$torch.Size([1, 96800])$

5. *Yes, you should agree with him, as depicted by the following plot* 8.80:

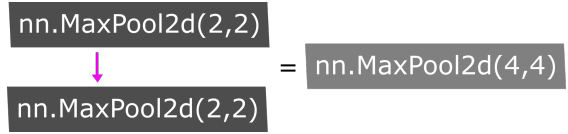

FIGURE 8.80: A single MaxPool layer.

■

Batch normalization, Gaussian PDF
The Gaussian distribution

## SOL-241 ✐ CH.SOL- 8.65.

*The answers are as follows:*

1. *BN is a method that normalizes the mean and variance of each of the elements during training.*

2. $X \sim \mathcal{N}(0, 1)$ *a mean of zero and a variance of one. The standard normal distribution occurs when* $(\sigma)^2 = 1$ *and* $\mu = 0$.

3. *In order to normalize we:*

   i *Step one is to subtract the mean to shift the distribution.*

   ii *Divide all the shifted values by their standard deviation (the square root of the variance).*

4. *In BN, the normalization is applied on an element by element basis. During training at each epoch, every element in the batch has to be shifted and scaled so that it has a zero mean and unit variance within the batch.*

■

**SOL-242**  **CH.SOL- 8.66.**

1. *One possible realization is as follows 8.81:*

```python
from math import sqrt
import math
def normDist(x, mu, sigSqrt):
  return (1 / sqrt(2 * math.pi * sigSqrt)) * math.e ** ((-0.5) *
    ↪ (x - mu) ** 2 / sigSqrt)
```

FIGURE 8.81: Normal distribution in Python: from scratch.

2. *The derivative is given by 8.82:*

```python
scipy.stats.norm.pdf(x, mu, sigma)*(mu - x)/sigma**2
```

FIGURE 8.82: The derivative of a Normal distribution in Python.

■

BN

**SOL-243**  **CH.SOL- 8.67.**

1. *During training of a CNN, when a convolution is being followed by a BN layer, for each of the three RGB channels a single separate mean and variance is being computed.*

2. *The mistake he made is using a BN with a batch size of 32, while the output from the convolutional layer is 64.*

■

Theory of CNN design

**SOL-244**  **CH.SOL- 8.68.**
*True.*

∎

**SOL-245**  **CH.SOL- 8.69.**
*All the options may be used to build a CNN.* ∎

**SOL-246**  **CH.SOL- 8.70.** *While the original paper ([16]) suggests that BN layers be used **before** an activation function, it is also possible to use BN after the activation function. In some cases, it actually leads to better results ([4]).*

∎

**SOL-247**  **CH.SOL- 8.71.**
*When dropout is enabled during the training process, in order to keep the expected output at the same value, the output of a dropout layer must be multiplied with this term. Of course, during inference no dropout is taking place at all.* ∎

**SOL-248**  **CH.SOL- 8.72.**

1. *The idiom is a bottleneck layer ([27]), which may act much like an autoencoder.*

2. *Reducing and then increasing the activations, may force the MLP to learn a more compressed representation.*

3. *The new architecture has far more connections and therefore it would be prone to overfitting.*

4. *Once such architecture is an autoencoder ([28]).*

∎

CNN residual blocks

**SOL-249**  **CH.SOL- 8.73.**

1. *The function $\mathcal{F}$ is the residual function.*

2. *The main idea was to add an **identity** connection which skips two layers all together.*

---

**SOL-250**  **CH.SOL- 8.74.**

1. *The missing parts are visualized in (8.83).*

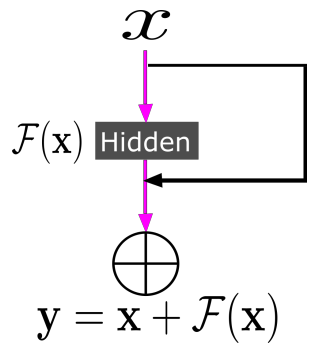

FIGURE 8.83: A resnet CNN block

2. *The symbol represents the addition operator.*

3. *Whenever $\mathcal{F}$ returns a zero, then the input $X$ will reach the output without being modified. Therefore, the term identity function.*

---

8.3.8   Training, hyperparameters

Hyperparameter optimization

**SOL-251**  **CH.SOL- 8.75.**

*The question states that image size is quite large, and the batch size is 1024, therefore it may fail to allocate memory on the GPU with an Out Of Memory (OOM) error message. This*

*is one of the most commonly faced errors when junior data-scientist start training models.*

**SOL-252**  **CH.SOL- 8.76.**

1. *Since hs is tuning his Hyperparameters on the validation set, he would most probably overfit to the validation set which he also used for evaluating the performance of the model.*

2. *One way would be to amend the splitting, is by first keeping a fraction of the training set aside, for instance* 0.1, *and then split the remaining .90 into a training and a validation set, for instance 0.8 and 0.1.*

3. *His new approach uses GridSearchCV with 5-fold cross-validation to tune his Hyperparameters. Since he is using cross validation with five folds, his local CV metrics would better reflect the performance on an unseen data set.*

**SOL-253**  **CH.SOL- 8.77.**

   *In* **grid search**, *a set of pre-determined values is selected by a user for each dimension in his search space, and then thoroughly attempting each and every combination. Naturally, with such a large search space the number of the required combinations that need to be evaluated scale exponentially in the number of dimensions in the grid search.*

   *In* **random search** *the main difference is that the algorithm samples completely random points for each of the dimensions in the search space. Random search is usually faster and may even produce better results.*

<div align="center">Labelling and bias</div>

Recommended reading:
"*Added value of double reading in diagnostic radiology,a systematic review*" [8].

**SOL-254**  **CH.SOL- 8.78.**

   *There is a potential for bias in certain settings such as this. If the whole training set is labelled only by a single radiologist, it may be possible that his professional history would*

*inadvertently generate bias into the corpus. Even if we use the form of radiology report reading known as double reading it would not be necessarily true that the annotated scans would be devoid of bias or that the quality would be better [8].*

Validation curve ACC

**SOL-255**  **CH.SOL- 8.79.**

   *The answers are as follows:*

1. *A validation curve displays on a single graph a chosen hyperparameter on the horizontal axis and a chosen metric on the vertical axis.*

2. *The hyperparameter is the number of epochs*

3. *The quality metric is the error (1 -accuracy). Accuracy, error = (1`accuracy) or loss are typical quality metrics.*

4. *The longer the network is trained, the better it gets on the training set.*

5. *At some point the network is fit too well to the training data and loses its capability to generalize. While the classifier is still improving on the training set, it gets worse on the validation and the test set.*

6. *At this point the quality curve of the training set and the validation set diverge.*

Validation curve Loss

**SOL-256**  **CH.SOL- 8.80.**

   *The answers are as follows:*

1. *What we are witnessing is phenomena entitled a plateau. This may happen when the optimization protocol can not improve the loss for several epochs.*

2. *There possible methods are:*

     *i Constant*

     *ii Xavier/Glorot uniform*

  *iii Xavier/Glorot normal*

3. *Good initialization would optimally generate activations that produce initial gradients that are larger than zero. One idea is that the training process would converge faster if unit variance is achieved ([16]). Moreover, weights should be selected carefully so that:*

  *i They are large enough thus preventing gradients from decaying to zero.*

  *ii They are not too large causing activation functions to over saturate.*

4. *There are several ways to reduce the problem of plateaus:*

  *i Add some type of regularization.*

  *ii In cases wherein the plateau happens right at the beginning, amend the way weights are initialized.*

  *iii Amending the optimization algorithm altogether, for instance using SGD instead of Adam and vice versa.*

5. *Since the initial LR is already very low, his suggestion may worsen the situation since the optimiser would not be able to jump off and escape the plateau.*

6. *In contrast to accuracy, Log loss has no upper bounds and therefore at times may be more difficult to understand and to explain.*

■

Inference

**SOL-257**  **CH.SOL- 8.81.**

1. *Usually data augmentation, is a technique that is heavily used during training, especially for increasing the number of instances of minority classes. In this case, augmentations are using during inference and this method is entitled Test Time Augmentation (TTA).*

2. *Here are several image augmentation methods for TTA, with two augmentations shown also in PyTorch.*

Horizontal flip

Vertical flip

Rotation

Scaling

Crops

```
1 transforms.HorizolntalFlip(p=1)(image)
2 transforms.VerticalFlip(p=1)(image)
```

FIGURE 8.84: Several image augmentation methods for TTA.

**SOL-258**  **CH.SOL- 8.82.**

*i Unseen*

*ii Overfitting*

### 8.3.9   Optimization, Loss

Stochastic gradient descent, SGD

**SOL-259**  **CH.SOL- 8.83.**

*There is no relation to random number generation, the true meaning is the use of batches during the training process.*

**SOL-260**  **CH.SOL- 8.84.**

*A larger batch size decreases the variance of the gradient estimation of SGD. Therefore, if your training loop uses larger batches, the model will converge faster. On the other hand, smal-*

*ler batch sizes increase the variance, leading to the opposite phenomena; longer convergence times.*

<div align="center">Momentum</div>

**SOL-261**  **CH.SOL- 8.85.**

*Momentum introduces an extra term which comprises a moving average which is used in gradient descent update rule to exponentially decay the historical gradients Using such term has been demonstrated to accelerate the training process ([11]) requiring less epochs to converge.*

**SOL-262**  **CH.SOL- 8.86.**

*The answers are as follows:*

1. *The derivative of the logistic activation function is extremely small for either negtive or positive large inputs.*

2. *The use of the tanh function does not alleviate the problem since we can scale and translate the sigmoid function to represent the tanh function:*

$$\tanh(z) = 2\sigma(2z) - 1 \tag{8.76}$$

*While the sigmoid function is centred around 0.5, the tanh activation is centred around zero. Similar to the application of BN, centring the activations may aid the optimizer converge faster. Note: there is no relation to SGD; the issue exists when using other optimization functions as well.*

**SOL-263**  **CH.SOL- 8.87.**

*The answers are as follows:*

i *True.*

ii *False. In stochastic gradient descent, the gradient for a single sample is **quite** different*

*from the actual gradient, so this gives a more **noisy** value, and converges **slower***

*iii  True.*

*iv  False. SGD requires less memory.*

■

<div align="center">Norms, L1, L2</div>

**SOL-264**  **CH.SOL- 8.88.**

1. *The L2 norm.*

2. *The Euclidean distance which is calculated as the square root of the sum of differences between each point in a set of two points.*

3. *The Manhattan distance is an L1 norm (introduced by Hermann Minkowski) while the Euclidean distance is an L2 norm.*

4. *The Manhattan distance is:*

$$|6 - 2| \ + \ |1 - 8| + |4 - 3| \ + \ |5 - (-1)| $$
$$= 4 + 7 + 1 + 6 = 18 \tag{8.77}$$

5. *The Euclidean distance is:*

$$\sqrt{(6 - 2)^2 + (1 - 8)^2 + (4 - 3)^2 + (5 - (-1))^2}$$
$$= \sqrt{102} \tag{8.78}$$

■

**SOL-265**  **CH.SOL- 8.89.**

*The PyTorch implementation is in (8.85). Note that we are allocating tensors on a GPU but first they are created on a CPU using numpy. This is also always the interplay between the CPU and the GPU when training NN models. Note that this only work if you have GPU available; in case there is no GPU detected, the code has a fallback to the CPU.*

```
1  %reset -f
2  import torch
3  import numpy
4
5  use_cuda = torch.cuda.is_available()
6  device = torch.device("cuda" if use_cuda else "cpu")
7  print (device)
8  x1np=numpy.array([6,1,4,5])
9  x2np=numpy.array([2,8,3,-1])
10 x1t=torch.FloatTensor(x1np).to(device)  # Move to GPU if available
11 x2t=torch.FloatTensor(x2np).to(device)
12 dist = torch.sqrt (torch.pow(x1t - x2t, 2).sum())
13 dist
14 >cuda
15 >tensor(10.0995, device='cuda:0')
```

FIGURE 8.85: Manhattan distance function in PyTorch.

**SOL-266**  **CH.SOL- 8.90.**

*The L2 loss is suitable for a target, or a response variable that is continuous. On the other hand, in a binary classification problem using LR we would like the output to match either zero or one and a natural candidate for a loss function is the binary cross-entropy loss.*

# PART V

# PRACTICE EXAM

# CHAPTER

<div style="color: #A6093D;">9</div>

<span style="color: #A6093D;">JOB INTERVIEW MOCK EXAM</span>

*A man who dares to waste **one hour** of time has not discovered the value of life.*

— Charles Darwin

## Contents

Stressful events, such as a job interview, prompt concern and anxiety (as they do for virtually every person), but it's the lack of preparation that fuels unnecessary nervousness. Many perceive the interview as a potentially threatening event. Testing your knowledge in AI using a mock exam, is an effective way to not only identifying your weaknesses and to pinpointing the concepts and topics that need brushing up, but also to becoming more relaxed in similar situations. Remember that at the heart of job interview confidence is feeling relaxed.

Doing this test early enough, gives you a head-start before the actual interview, so that you can target areas that require perfection. The exam includes questions from a wide variety of topics in AI, so that these areas are recognised and it would then be a case of solving all the problems in this book over a period of few months to be properly prepared. Do not worry even if you can not solve any of the problems in the exam as some of them are quite difficult.

# DEEP LEARNING JOB INTERVIEW MOCK EXAM

EXAM INSTRUCTIONS:

YOU SHOULD NOT SEARCH FOR SOLUTIONS ON THE WEB. MORE GENERALLY, YOU ARE URGED TO TRY AND SOLVE THE PROBLEMS WITHOUT CONSULTING ANY REFERENCE MATERIAL, AS WOULD BE THE CASE IN A REAL JOB INTERVIEW.

### 9.0.1   Rules

**REMARK**: In order to receive credits, you must:

  i Show all work neatly.

 ii A sheet of formulas and calculators are permitted but not notes or texts.

iii Read the problems **CAREFULLY**

iv Do not get **STUCK** at any problem (or in local minima ...) for too much time!

 v After completing all problems, a double check is **STRONGLY** advised.

vi You have **three hours** to complete all questions.

<h1 style="text-align:center">9.1   Problems</h1>

<h3 style="text-align:center">9.1.1   Perceptrons</h3>

**PRB-267 ❷ CH.PRB- 9.1.** *[PERCEPTRONS]*

*The following questions refer to the MLP depicted in (9.1).The inputs to the MLP in (9.1) are $x_1 = 0.9$ and $x_2 = 0.7$ respectively, and the weights $w_1 = -0.3$ and $w_2 = 0.15$ respectively. There is a single hidden node, $H_1$. The bias term, $B1$ equals 0.001.*

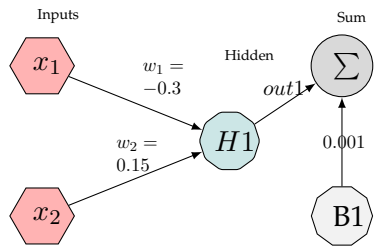

FIGURE 9.1: Several nodes in a MLP.

1. *We examine the mechanism of a single hidden node, $H_1$. The inputs and weights go through a linear transformation. What is the value of the output ($out1$) observed at the $sum$ node?*

2. *What is the resulting value from the application of the $sum$ operator?*

3. *Using PyTorch tensors, verify the correctness of your answers.*

<h3 style="text-align:center">9.1.2   CNN layers</h3>

**PRB-268 ❷ CH.PRB- 9.2.** *[CNN LAYERS]*

*While reading a paper about the MaxPool operation, you encounter the following code snippet 9.1 of a PyTorch module that the authors implemented. You download their pre-trained model, and examine its behaviour during inference:*

```python
import torch
from torch import nn
class MaxPool001(nn.Module):
 def __init__(self):
  super(MaxPool001, self).__init__()
  self.math = torch.nn.Sequential(
    torch.nn.Conv2d(3, 32, kernel_size=7, padding=2),
    torch.nn.BatchNorm2d(32),
    torch.nn.MaxPool2d(2, 2),
    torch.nn.MaxPool2d(2, 2),
  )
 def forward(self, x):
  print (x.data.shape)
  x = self.math(x)
  print (x.data.shape)
  x = x.view(x.size(0), -1)
  print ("Final shape:{}",x.data.shape)
  return x
model = MaxPool001()
model.eval()
x = torch.rand(1, 3, 224, 224)
out=model.forward(x)
```

CODE 9.1: A CNN in PyTorch

*The architecture is presented in 9.2:*

torch.rand(1, 3, 224, 224)

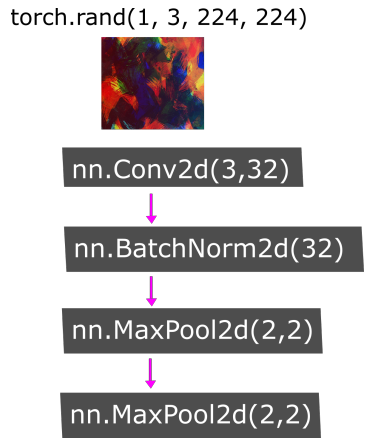

nn.Conv2d(3,32)

nn.BatchNorm2d(32)

nn.MaxPool2d(2,2)

nn.MaxPool2d(2,2)

FIGURE 9.2: Two consecutive MaxPool layers.

*Please run the code and answer the following questions:*

1. *In MaxPool2D(2,2), what are the parameters used for?*

2. *After running line 8, what is the resulting tensor shape?*

3. *Why does line 20 exist at all?*

4. *In line 9, there is a MaxPool2D(2,2) operation, followed by yet a second MaxPool2D(2,2). What is the resulting tensor shape after running line 9? and line 10?*

5. *A friend who saw the PyTorch implementation, suggests that lines 9 and 10 may be replaced by a single MaxPool2D(4,4,) operation while producing the exact same results. Do you agree with him? Amend the code and test your assertion.*

### 9.1.3    Classification, Logistic regression

**PRB-269 ❷ CH.PRB- 9.3.  *[CLASSIFICATION, LR]***

*To study factors that affect the survivability of humans infected with COVID19 using logistic regression, a researcher considers the link between lung cancer and COVID19 as a*

*plausible risk factor. The predictor variable is a count of **removed** pulmonary nodules (Fig. 9.3) in the lungs.*

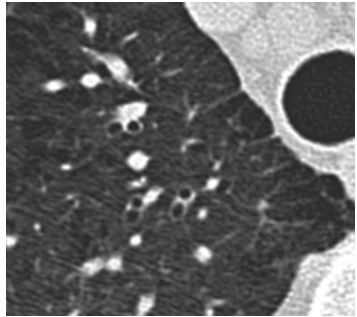

FIGURE 9.3: Pulmonary nodules.

*The response variable Y measures whether the patient shows any remission (as in the manifestations of a disease, e. g. yes=1, no=0) when the pulmonary nodules count shifts up or down. The output from training a logistic regression classifier is as follows:*

```
                                 Standard
         Parameter   DF  Estimate    Error
         Intercept    1   -4.8792   1.0732
 Pulmonary nodules    1    0.0258   0.0194
```

1. *Estimate the probability of improvement when the count of removed pulmonary nodules of a patient is 33.*

2. *Find out the removed pulmonary nodules count at which the estimated probability of improvement is 0.5.*

3. *Find out the estimated odds ratio of improvement for an increase of **1**, in the total removed pulmonary nodule count.*

4. *Obtain a 99% confidence interval for the true odds ratio of improvement increase of **1** in the total removed pulmonary nodule count. Remember that The most common confidence levels are* 90%, 95%, 99%, *and* 99.9%.

| Confidence Level | $z$ |
|:---:|:---:|
| 90% | 1.645 |
| 95% | 1.960 |
| 99% | **2.576** |
| 99.9% | 3.291 |

TABLE 9.1: Common confidence levels

*Table 9.1 lists the $z$ values for these levels.*

<h3 style="text-align:center">9.1.4   Information theory</h3>

**PRB-270 ❷ CH.PRB- 9.4.**  *[INFORMATION THEORY]*

   *This question discusses the link between binary classification, information gain and decision trees. Recent research suggests that the co-existence of **influenza** (Fig. 9.4) and COVID19 virus may decrease the survivability of humans infected with the COVID 19 virus. The data (Table 9.2) comprises a training set of feature vectors with corresponding class labels which a researcher intents classifying using a decision tree.*

   *To study factors affecting **COVID19 eradication**, the deep-learning researcher collects data regrading two independent binary variables; $\theta_1$ (T/F) indicating whether the patient is a female, and $\theta_2$ (T/F) indicating whether the human tested positive for the influenza virus. The binary response variable, $\gamma$, indicates whether eradication was observed (e.g. eradication=+, no eradication=-).*

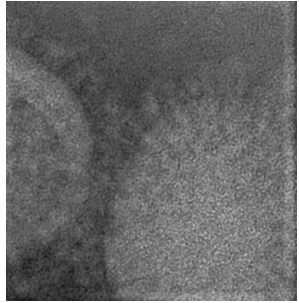

FIGURE 9.4: The influenza virus.

*Referring to Table (9.2), each row indicates the observed values, columns ($\theta_i$) denote features and rows ($<\theta_i, \gamma_i>$) denote labelled instances while class label ($\gamma$) denotes whether eradication was observed.*

| $\gamma$ | $\theta_1$ | $\theta_2$ |
|---|---|---|
| + | T | T |
| - | T | F |
| + | T | F |
| + | T | T |
| - | F | T |

TABLE 9.2: Decision trees and the COVID19 virus.

1. *Describe what is meant by **information gain**.*

2. *Describe in your own words how does a decision tree work.*

3. *Using $log_2$, and the provided dataset, calculate the sample entropy $H(\gamma)$.*

4. *What is the information gain $IG(X_1) \equiv H(\gamma) - H(|\theta_1)$ for the provided training corpus?*

**PRB-271 ❷ CH.PRB- 9.5.**

What is the entropy of a **biased coin**? Suppose a coin is biased such that the probability of 'heads' is $p(x_h) = 0.98$.

1. **Complete the sentence:** We can predict 'heads' for each flip with an accuracy of [___]%.

2. **Complete the sentence:** If the result of the coin toss is 'heads', the amount of Shannon information gained is [___] bits.

3. **Complete the sentence:** If the result of the coin toss is 'tails', the amount of Shannon information gained is [___] bits.

4. **Complete the sentence:** It is always true that the more information is associated with an outcome, the [**more/less**] surprising it is.

5. Provided that the ratio of tosses resulting in 'heads' is $p(x_h)$, and the ratio of tosses resulting in 'tails' is $p(x_t)$, and also provided that $p(x_h) + p(x_t) = 1$, what is the formula for the **average surprise**?

6. What is the value of the **average surprise** in bits?

**PRB-272 ❷ CH.PRB- 9.6.**

**Complete the sentence:** The relative entropy $D(p||q)$ is the measure of (a) [___] between two distributions. It can also be expressed as a measure of the (b)[___] of assuming that the distribution is $q$ when the (c)[___] distribution is $p$.

### 9.1.5    Feature extraction

**PRB-273 ❷ CH.PRB- 9.7.  *[FEATURE EXTRACTION]***

A data scientist extracts a feature vector from an image using a pre-trained ResNet34 CNN (*9.5*).

```
1  import torchvision.models as models
2  ...
3  res_model = models.resnet34(pretrained=True)
```

FIGURE 9.5: PyTorch declaration for a pre-trained ResNet34 CNN (simplified).

*He then applies the following algorithm, entitled xxx on the image (9.2).*

CODE 9.2: An unknown algorithm in C++11

```
1   void xxx(std::vector<float>& arr){
2     float mod = 0.0;
3     for (float i : arr) {
4       mod += i * i;
5     }
6     float mag = std::sqrt(mod);
7     for (float & i : arr) {
8       i /= mag;
9     }
10  }
```

*Which results in this vector (9.6):*

| 0.7766 | 0.4455 | 0.8342 | 0.6324 | $\cdots$ | $k = 512$ |

Values after applying xxx to a $k$-element FV.

FIGURE 9.6: A one-dimensional 512-element embedding for a single image from the ResNet34 architecture.

*Name the algorithm that he used and explain in detail why he used it.*

**PRB-274  CH.PRB- 9.8.**
    *[FEATURE EXTRACTION]*

*The following question discusses the method of fixed feature extraction from layers of the VGG19 architecture for the classification of the COVID19 pathogen. It depicts FE principles which are applicable with minor modifications to other CNNs as well. Therefore, if you happen to encounter a similar question in a job interview, you are likely be able to cope with it by utilizing the same logic.*

*In (Fig. 9.7), 2 different classes of human cells are displayed; infected and not-infected, which were curated from a dataset of $4K$ images labelled by a majority vote of two expert virologists. Your task is to use FE to correctly classify the images in the dataset.*

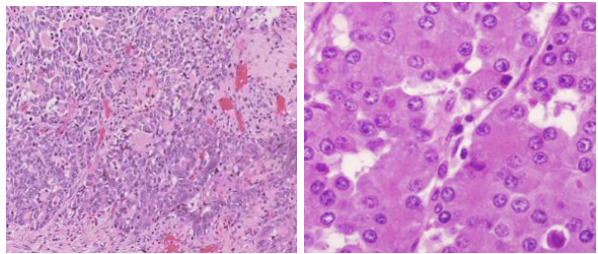

FIGURE 9.7: A dataset of human cells infected by the COVID19 pathogen.

*Table (9.3) presents an incomplete listing of the of the VGG19 architecture. As depicted, for each layer the number of filters (i.e. neurons with unique set of parameters), learnable parameters (e.g. weights and biases), and FV size are presented.*

| Layer name | #Filters | #Parameters | # Features |
|------------|----------|-------------|------------|
| conv4_3 | 512 | 2.3M | 512 |
| fc6 | 4,096 | 103M | 4,096 |
| fc7 | 4,096 | 17M | 4,096 |
| output | 1,000 | 4M | - |
| **Total** | **13,416** | **138M** | **12,416** |

TABLE 9.3: Incomplete listing of the of the VGG19 architecture

1. *Describe how the VGG19 CNN may be used as fixed FE for a classification task. In your answer be as detailed as possible regarding the stages of FE and the method used for classification.*

2. *Referring to Table (9.3), suggest **three** different ways in which features can be extracted from a trained VGG19 CNN model. In each case, state the extracted feature layer name and the size of the resulting FE.*

3. *After successfully extracting the features for the 4k images from the dataset, how can you now **classify** the images into their respective categories?*

<div style="text-align:center">9.1.6    Bayesian deep learning</div>

**PRB-275 ❷ CH.PRB- 9.9.** *[BAYESIAN DEEP LEARNING]*

*A recently published paper presents a new layer for Bayesian neural networks (BNNs). The layer behaves as follows. During the feed-forward operation, each of the hidden neurons $H_n$ , $n \in \{1, 2, \}$ in the neural network in (Fig. 9.8) **may, or may not** fire, independently of each other, according to a known prior distribution.*

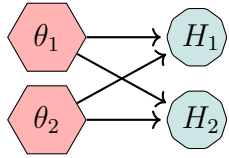

FIGURE 9.8: Likelihood in a BNN model.

*The chance of firing, $\gamma$, is the same for each hidden neuron. Using the formal definition, calculate the **likelihood function** of each of the following cases:*

1. *The hidden neuron is distributed according to $X \sim \mathrm{B}(n, \gamma)$ random variable and fires with a probability of $\gamma$. There are 100 neurons and only 20 are **fired**.*

2. *The hidden neuron is distributed according to $X \sim U(0, \gamma)$ random variable and fires with a probability of $\gamma$.*

---

**PRB-276 ❷ CH.PRB- 9.10.**

*During pregnancy, the Placenta Chorion Test is commonly used for the diagnosis of hereditary diseases (Fig. 9.9).*

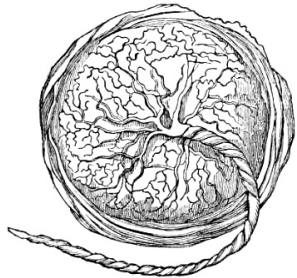

FIGURE 9.9: Foetal surface of the placenta

*Assume, that a new test entitled **the Placenta COVID19 Test** has the exact same properties as the Placenta Chorion Test. The test has a probability of 0.95 of being correct **whether or not** a COVID19 pathogen is present. It is known that 1/100 of pregnancies result in*

*COVID19 virus being passed to foetal cells. Calculate the probability of a test indicating that a COVID19 virus is present.*

---

### PRB-277 ❷ CH.PRB- 9.11.

*A person who was unknowingly infected with the COVID19 pathogen takes a walk in a park crowded with people. Let $y$ be the number of successful infections in $5$ independent social interactions or infection attempts (trials), where the probability of "success" (infecting someone else) is $\theta$ in each trial. Suppose your prior distribution for $\theta$ is as follows: $P(\theta = 1/2) = 0.25$, $P(\theta = 1/6) = 0.5$, and $P(\theta = 1/4) = 0.25$.*

1. *Derive the **posterior distribution** $p(\theta|y)$.*

2. *Derive the **prior predictive distribution** for $y$.*

---

### PRB-278 ❷ CH.PRB- 9.12.

*The 2014 west African Ebola (Fig. 9.10) epidemic has become the largest and fastest-spreading outbreak of the disease in modern history with a death tool far exceeding all past outbreaks combined. Ebola (named after the Ebola River in Zaire) first emerged in 1976 in Sudan and Zaire and infected over 284 people with a mortality rate of 53%.*

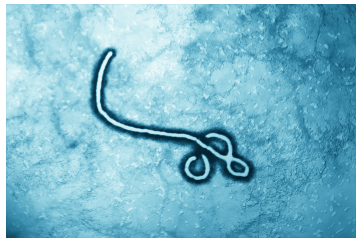

FIGURE 9.10: The Ebola virus.

*This rare outbreak, underlined the challenge medical teams are facing in containing epidemics. A junior data scientist at the centre for disease control (CDC) models the possible spread and containment of the Ebola virus using a numerical simulation. He knows that out of a population of $k$ humans (the number of trials), $x$ are carriers of the virus (success in*

*statistical jargon). He believes the sample likelihood of the virus in the population, follows a Binomial distribution:*

$$L(\gamma) = \begin{pmatrix} n \\ y \end{pmatrix} \gamma^y (1-\gamma)^{n-y},$$

$$\gamma \in [0,1], \quad y = 1, 2, \ldots, n \quad , \tag{9.1}$$

*where:*

$$\begin{pmatrix} n \\ y \end{pmatrix} = \frac{n!}{(n-y)!y!}. \tag{9.2}$$

*As the senior researcher in the team, you guide him that his parameter of interest is $\gamma$, the proportion of infected humans in the entire population.*

*The expectation and variance of the binomial are:*

$$E(y|\gamma, n) = n\gamma, \ , \ V(y|\gamma, n) = n\gamma(1-\gamma). \tag{9.3}$$

*Answer the following:*

1. *For the likelihood function of the form $l_x(\gamma) = \log L_x(\gamma)$ what is the log-likelihood function?*

2. *Find the log-likelihood function $ln\,(L(\gamma))$*

3. *Find the gradient vector $g(\gamma)$*

4. *Find the Hessian matrix $H(\gamma)$*

5. *Find the Fisher information $I(\gamma)$*

6. *In a population spanning 10,000 individuals, 300 were infected by Ebola. Find the MLE for $\gamma$ and the standard error associated with it.*

# PART VI
## VOLUME TWO

# CHAPTER

VOLUME TWO - PLAN

*Nothing exists until it is measured.*

— Niels Bohr, 1985

## Contents

## 10.1    Introduction

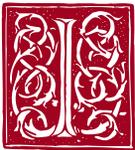 T is important at the outset to understand we could not possibly include everything we wanted to include in the first VOLUME of this series. While the first volume is meant to introduce many of the core subjects in AI, the second volume takes another step down that road and includes numerous, more advanced subjects. This is a short glimpse into the plan for VOLUME-2 of this series. This second volume focuses on more advanced topics in AI

10.2    AI system design

10.3    Advanced CNN topologies

10.4    1D CNN's

10.5    3D CNN's

10.6    Data augmentations

10.7    Object detection

10.8    Object segmentation

10.9    Semantic segmentation

10.10    Instance segmentation

10.11    Image classification

10.12    Image captioning

10.13    NLP

# List of Tables

# List of Figures

# Alphabetical Index

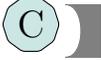

# C

## G

## H

## I

## J

## K

## L

# M

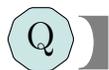

## R

## S

## U

## V

## W

## X